\documentclass{article}

 \PassOptionsToPackage{numbers, compress}{natbib}
 \usepackage[preprint]{neurips/neurips_2026}

\usepackage[utf8]{inputenc}
\usepackage[T1]{fontenc}
\usepackage{xcolor}

\definecolor{linkblue}{HTML}{14315D}
\usepackage[
    colorlinks=true,
    linkcolor=linkblue,
    citecolor=linkblue,
    urlcolor=linkblue,
    filecolor=linkblue
]{hyperref}

\usepackage{url}
\usepackage{booktabs}
\usepackage{amsfonts}
\usepackage{nicefrac}
\usepackage{microtype}

\usepackage{graphicx}
\usepackage{subcaption}

\usepackage{wrapfig}

\usepackage{pifont}
\newcommand{\cmark}{\ding{51}}
\newcommand{\xmark}{\ding{55}}
\newcommand{\pmark}{\raisebox{0.15ex}{\scriptsize$\triangle$}}

\usepackage{amsmath}
\usepackage{mathtools}
\usepackage{amssymb}
\usepackage{amsthm}

\usepackage{multirow}
\usepackage{siunitx}
\theoremstyle{plain}
\newtheorem{theorem}{Theorem}[section]

\theoremstyle{definition}
\newtheorem{definition}[theorem]{Definition}

\theoremstyle{remark}

\usepackage{longtable}
\usepackage{array}
\usepackage{makecell}
\usepackage{paracol}
\usepackage{enumitem}
\usepackage{xspace}
\newcommand{\name}{\textsc{MT-InfoSeek}\xspace}

\usepackage{titlesec}
\titlespacing*{\paragraph}{0pt}{0pt}{0.5em}

\usepackage[most]{tcolorbox}

\newtcolorbox{examplebox}[1]{
  enhanced,
  breakable,
  colback=gray!2,
  colframe=gray!45,
  colbacktitle=gray!20,
  coltitle=black,
  boxrule=0.4pt,
  arc=2pt,
  left=5pt,
  right=5pt,
  top=4pt,
  bottom=4pt,
  before skip=5pt,
  after skip=5pt,
  fonttitle=\bfseries\footnotesize,
  title={#1},
}

\usepackage{ragged2e}

\newtcolorbox{promptbox}[1]{
  colback=gray!5!white,
  colframe=gray!75!black,
  title={#1},
  fonttitle=\bfseries,
  fontupper=\ttfamily\small,
  boxrule=0.5pt,
  sharp corners,
  breakable,
  before upper={\RaggedRight\sloppy},
  after upper={\par}
}
\newcommand{\blankline}{\par\vspace{\baselineskip}}

\usepackage{fontawesome5}

\definecolor{TakeawayBlue}{RGB}{235,242,252}

\definecolor{TakeawayAccent}{RGB}{80,120,200}

\definecolor{ImpColor}{RGB}{170,70,35}

\newcommand{\takeawaytag}[1]{%
  \begingroup
  \setlength{\fboxsep}{0pt}%
  \colorbox{TakeawayBlue}{%
    \strut\textcolor{TakeawayAccent}{\faLightbulb}\,
    \textbf{Takeaway~#1}%
  }%
  \endgroup
}

\newcommand{\takeaway}[2]{%
  \noindent\takeawaytag{#1}\hspace{0.6em}%
  {\color{black}\emph{#2}}\par
}

\usepackage{neurips/algorithm}
\usepackage{neurips/algorithmic}

\title{
Do LLMs Know What to Ask and When? Evaluating Multi-Turn Information Seeking
}

\author{%
\begin{tabular}{@{}c@{}}
\small
Yepeng Huang\textsuperscript{\dag,}\thanks{Equal contribution.} \quad
Jiawen Zhang\textsuperscript{\dag,}\footnotemark[1] \quad
Michelle Dai\textsuperscript{\dag} \quad
Xiaorui Su\textsuperscript{\dag}
\\[-0.1em]
\small
Shanghua Gao\textsuperscript{\dag} \quad
Zi Wang\textsuperscript{\dag,}\textsuperscript{\ddag} \quad
Marinka Zitnik\textsuperscript{\dag}
\\[0.25em]
{\normalfont\footnotesize
\textsuperscript{\dag}Harvard University \quad
\textsuperscript{\ddag}Google DeepMind}
\\[0.4em]
{
\normalfont\footnotesize
\href{https://github.com/mims-harvard/MT-InfoSeek}{\faCode\; Code}
\hspace{1.2em}
\href{https://zitniklab.hms.harvard.edu/MT-InfoSeek/}{\faGlobe\; Project page}
\hspace{1.2em}
\href{https://github.com/mims-harvard/MT-InfoSeek/tree/main/data}{\faDatabase\; Dataset}
}
\end{tabular}%
}

\begin{document}

\maketitle

\begin{abstract}
When a user question is underspecified, a capable model should recognize that its context is insufficient, identify the missing information, ask for it, and respond only once that information determines a unique answer. We formalize multi-turn information seeking as solving a $k$-underspecified constraint satisfaction problem, where $k$ is the number of variables jointly required to determine the target and therefore measures the degree of missing information. We instantiate the formulation in \name, a controlled evaluation suite of 5,251 problems and 9,006 task instances spanning mathematics, logic, biology, medicine, and general knowledge. We evaluate models along three axes: what they ask, when they ask it, and how the acquired information affects the final answer. Performance degrades across models and domains as underspecification increases. Models recognize that additional information is needed but underestimate how much, and in logical problems at $k=2$ they under-predict the degree of missing information about four times as often as they over-predict it. They also fail to identify a minimal sufficient set of queries, improve only marginally when given the true $k$, and often stop before acquiring sufficient information. In tasks with ordered dependencies, an incorrect query order reduces final accuracy even when the model eventually acquires all necessary information. We measure information seeking directly through final sufficiency, which records whether the acquired information determines the target independent of answer generation. This separation shows differences between models that final accuracy alone does not capture, and indicates that the ability to seek information over multiple turns is distinct from the ability to generate answers and is not measured by current LLM evaluations.

\end{abstract}

\section{Introduction} 

Large language models (LLMs) often operate under incomplete information in multi-turn interactions, because a user task may be underspecified~\footnote{We study underspecification, not ambiguity. \cite{li2025questbench} draws the distinction, and Appendix~\ref{app:underspecification} discusses it further.}. A capable model should recognize that its context is insufficient, identify and acquire the missing information, and respond only once that information determines a unique answer~\citep{Amershi2019hai,Bansal2024challenges}. Interactive settings depend on this capability, but direct and systematic evaluation remains difficult.

Final task success does not indicate whether a model acquired the necessary information. 
Existing multi-turn benchmarks measure task success, dialogue quality, or user-facing outcomes, without specifying which information the model should request or whether it stopped only after the target became identifiable~\citep{li2024mediq,johri2025patientinteract,kobalczyk2025activebed,zhou2025arbench}. Model priors and the ability to generate answers therefore confound the evaluation. In MediQ~\citep{li2024mediq}, for example, a 30B model answers correctly in roughly 60--70\% of cases from the initial incomplete information alone (Appendix~\ref{appendix:pre_study_high_acc}), so final accuracy overstates the ability to seek information. Simulated users and free-form responses make query-level evaluation ambiguous, because a failure may follow from a wrong question, an uninformative response, or a response that does not map onto task variables.

\begin{figure*}[t]
    \centering
    \vspace{-20pt}
    \includegraphics[width=\textwidth]{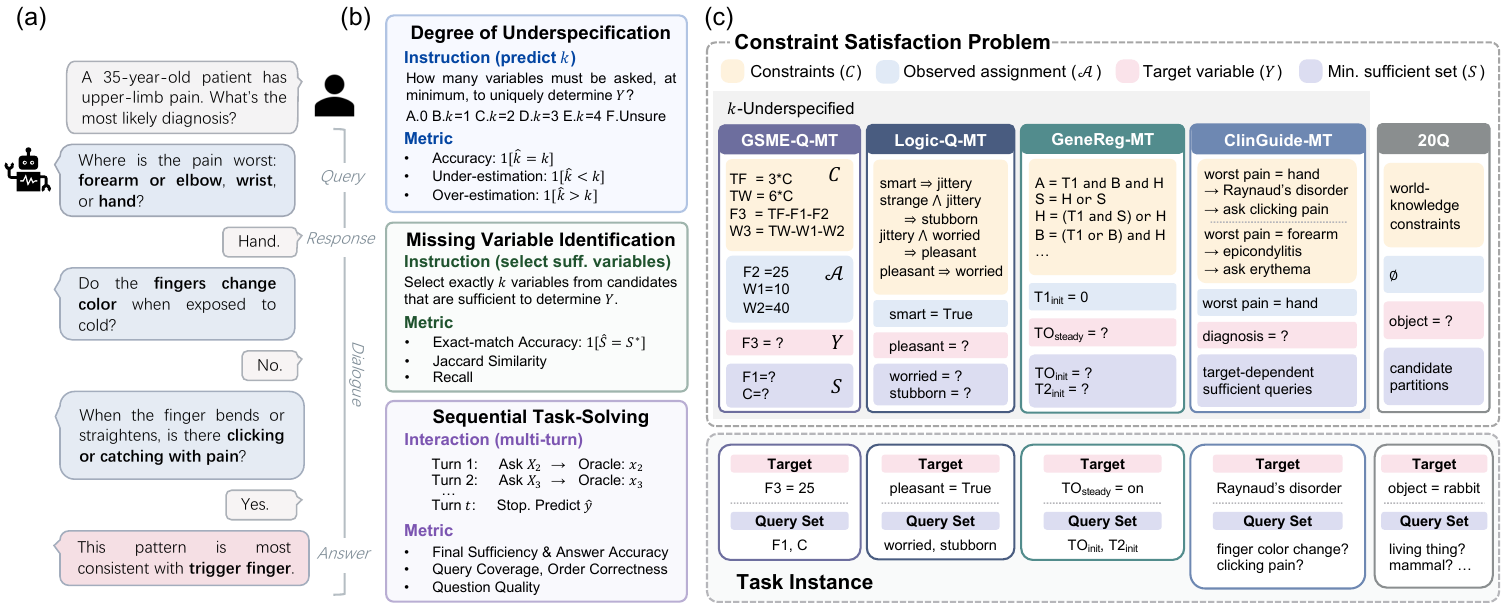}
    \captionsetup{font=small}
    \caption{
    Overview of the \name evaluation framework.
    \textbf{(a) Information-seeking interaction}: the model queries an oracle over multiple turns, acquires the missing information, and identifies a hidden target.
    \textbf{(b) Evaluation protocols}: we evaluate three abilities, predicting the degree of underspecification $k$, identifying a minimal sufficient set of variables, and solving the task through sequential interaction.
    \textbf{(c) Task construction}: each problem specifies constraints, an observed assignment, a target variable, and its minimal sufficient sets, and induces task instances that share this information and differ in the hidden target value. One example problem is shown for each domain.
    }
    \label{fig:overview}
    \vspace{-24pt}
\end{figure*}

Representing an underspecified task as a constraint satisfaction problem (CSP) removes these sources of ambiguity, because the constraints determine which variables the model needs before it can answer, and each question the model asks either names one of those variables or does not. QuestBench formalizes underspecified reasoning problems as CSPs and provides the ground-truth missing variable that a model should query~\citep{li2025questbench}. It covers the special case where a single missing variable is sufficient to determine the answer, so one question ends the interaction. Multi-turn information seeking differs in structure, because several variables may be jointly necessary. The model queries one variable per turn, receives its value, and updates the set of assignments that remain feasible, so it must choose a variable at every turn and judge when the acquired information is sufficient to stop. Evaluating this requires representing joint sufficiency and tracking how each queried value changes the feasible space across turns. Neither single-turn query selection nor final accuracy captures these dimensions.

\paragraph{Present work.} We formalize multi-turn information seeking as solving a $k$-underspecified CSP, where \(k\) variables are jointly required to determine the target variable. The formulation provides ground truth for which variables the model needs, allows controlled variation in the degree of underspecification \(k\), and defines when the model should stop asking and answer. We instantiate the formulation in \name, an evaluation suite spanning mathematics, logic, biology, medicine, and general knowledge, with 5,251 problems and 9,006 task instances (Fig.~\ref{fig:overview}). Each problem induces a family of task instances that share the same initial information and differ in the target value, so a model cannot succeed by guessing the most likely target.

We study three components of multi-turn information seeking, which variables the model queries, when it stops, and how the acquired information affects the final answer. Models underestimate the degree of underspecification, perform worse as the number of missing variables grows, and stop before they acquire sufficient information. 
A wrong first query does not determine the outcome, since models that continue to search for target-relevant variables reach sufficiency at close to the rate of models that query correctly on the first turn, and the number of turns predicts success more strongly as $k$ grows.
In tasks with ordered dependencies, an incorrect query order lowers final accuracy even when the model acquires all necessary information.
Question quality alone does not account for success either, because models given the same completed interaction can still differ in whether they correctly answer. Separating information seeking from answer generation therefore shows differences between models that final accuracy alone does not capture.
We make the following contributions:
\begin{itemize}[leftmargin=20pt, topsep=0pt, itemsep=1pt, parsep=0pt, partopsep=0pt]
\item A formalization of multi-turn information seeking as solving $k$-underspecified CSPs, with ground truth for the variables the model needs and for when it has enough of them.
\item \name, an evaluation suite spanning mathematics, logic, biology, medicine, and general knowledge.
\item An evaluation of closed and open-weight LLMs under this formulation, which measures information seeking through final sufficiency and separates it from final accuracy.
\end{itemize}

\section{Related Work} 

\paragraph{Multi-turn information seeking.}
Recent work studies LLMs in interactive settings, where models acquire information, respond to feedback, and adapt over successive turns.
These include RL-based interaction~\citep{shani2024multiturnrlhf,qian2025userrl}, collaborative environments~\citep{wu2025collabllm}, decision-oriented games~\citep{zhou2025arbench,grand2026shoot,zhou2025multiturnagent}, and medical dialogue~\citep{li2024mediq,johri2025patientinteract,wang2025healthq}.
A related line of work studies clarification under ambiguity or underspecification, including  generation of clarification questions~\citep{yu2020clarify,pyatkin2023clarify,zhang2025clarify}, Bayesian experimental design for question selection~\citep{kobalczyk2025activebed,choudhury2025bedllm}, and proactive clarification in retrieval-augmented systems~\citep{li2026gps}.
These studies establish settings and mechanisms for interaction, feedback, and ambiguity resolution.
We treat multi-turn interaction as controlled information acquisition, where the model identifies missing variables that determine the target, queries them in as few turns as possible, and stops once the acquired information is sufficient to produce an answer.

\paragraph{Benchmarking under incomplete information.}
Existing studies evaluate information seeking across clinical, mathematical, logical, and common-knowledge domains~\citep{li2024mediq,johri2025patientinteract,kobalczyk2025activebed,zhou2025arbench,li2025questbench,yue2026interactive}, but they primarily measure performance through final task success, dialogue quality, or judge-mediated outcomes.
QuestBench~\citep{li2025questbench} is closest to our setting, since it gives ground-truth missing variables through an underspecified CSP formulation, and we extend this formulation to the multi-turn setting where the model acquires information sequentially.
\citet{laban2025llms_get_lost} show that LLMs perform worse when fully specified instructions are split into shards and revealed over turns.
Our work is complementary. We give the model an underspecified task and test whether it identifies the missing variables, queries them over successive turns, and stops once the acquired information determines the target.
Appendix Table~\ref{tab:benchmark_comparison} compares \name with existing benchmarks in detail.

\section{\name Evaluation Framework}

We formalize multi-turn information seeking as solving an underspecified CSP (\S\ref{sec:problem_formulation}). We then define the protocols and metrics that separate information seeking from answer generation (\S\ref{sec:evaluation_protocol}), and construct \name, an evaluation suite that instantiates the formulation in mathematics, logic, biology, medicine, and general knowledge (\S\ref{sec:benchmark_construction}).

\subsection{Problem Formulation}
\label{sec:problem_formulation}

We define a problem as a tuple $P = \langle \mathcal{X}, \mathcal{D}, \mathcal{C}, \mathcal{A}, Y \rangle$, where $\mathcal{X}$ is a set of variables with domains $\mathcal{D}$, $\mathcal{C}$ is a set of logical constraints over these variables, $\mathcal{A}$ is an observed partial assignment, and $Y$ is the target variable. 
Let $\Omega(P)$ denote the feasible space, the set of all full assignments consistent with both $\mathcal{C}$ and $\mathcal{A}$. Intuitively, $\Omega(P)$ contains the underlying states that remain possible given the available information (Fig.~\ref{fig:overview}a).

\begin{examplebox}{Example: 20 Questions animal-guessing game (fixed-trait view)}
\footnotesize
\textbf{Variables} \(\mathcal X\) include the target variable \(Y\) and latent traits (e.g., \textit{can\_swim}). 
\textbf{Constraints} \(\mathcal C\) define logical relations (e.g., $Y=\textit{penguin} \Rightarrow \textit{can\_swim}=\textit{True}$).
 Given \textbf{observed assignment} \(\mathcal A=\{\textit{can\_swim}=\texttt{True}\}\), 
the feasible space \(\Omega(P)\) contains all identities and traits consistent with this clue.
\textbf{Interaction:} Each question aims to assign a value to a trait, progressively restricting
\(\Omega(P)\) until all remaining assignments agree on \(Y\).
\end{examplebox}

To formalize this interactive process, we write $\mathcal{U}(P) = \mathcal{X} \setminus (\operatorname{dom}(\mathcal{A}) \cup \{Y\})$ for the set of queryable variables, those that are unassigned and are not the target, and define the following properties.
\begin{itemize}[leftmargin=5pt, topsep=0pt, itemsep=1pt, parsep=0pt, partopsep=0pt]
    \item Known Target \& Underspecification: $Y$ is \emph{known under $\mathcal A$} if all feasible assignments in $\Omega(P)$ share the same value for $Y$. If multiple values remain possible (e.g., knowing only \textit{can\_swim} leaves both $Y=\textit{penguin}$ and $Y=\textit{dolphin}$ feasible), the problem is \emph{underspecified}.
    \item Sufficient Set: A subset of variables $S \subseteq \mathcal U(P)$ is \emph{sufficient} if, for every valid assignment $s$ to $S$, the target variable $Y$ becomes known under the updated assignment $\mathcal A \cup \{S=s\}$.
    \item Minimal $k$-Sufficient Set ($k$-MSS): A set $S \subseteq \mathcal U(P)$ is a $k$-MSS if it is sufficient, has a size of $k$, and no strictly smaller subset of $\mathcal U(P)$ is sufficient. This value $k$ defines the \emph{problem-level degree of underspecification}.
\end{itemize}

We distinguish a \emph{problem} from its induced \emph{task instances} $T_y=(P,y)$, where $y \in \{Y(\omega) : \omega \in \Omega(P)\}$ is the hidden target value and $Y(\omega)$ is the value of $Y$ under a full assignment $\omega$. The task instances induced by a single problem $P$ form its \emph{task family}, and they share $\mathcal{C}$ and $\mathcal{A}$ and differ only in $y$. The \emph{task-level} degree of underspecification, the minimum number of variables needed to determine a specific $y$, can be strictly smaller than the problem-level degree $k$. The asymmetry follows because $k$ is the worst-case number of queries needed across all feasible target values, while a specific $y$ can be separated from the rest of the feasible space in fewer than $k$ queries.

\begin{examplebox}{Example of task-level asymmetry}
\footnotesize
Suppose the observed assignment $\mathcal{A}=\{\textit{can\_swim}=\textit{True}\}$ leaves four feasible target values, $Y \in \{\textit{shark}, \textit{dolphin}, \textit{penguin}, \textit{turtle}\}$. Distinguishing every target value in this set requires a 3-MSS, for example $S=\{\textit{has\_gills}, \textit{lays\_eggs}, \textit{has\_feathers}\}$, so the problem-level degree is $k=3$. 

If the task is $T_{\textit{shark}}$, then querying $\textit{has\_gills}$ alone is sufficient, since the answer $\textit{True}$ eliminates the other three candidates, and the task-level degree is $1$. Solving $T_{\textit{penguin}}$ requires all three variables to exclude the fish, the mammal, and the reptile, so its task-level degree is $3$.
\end{examplebox}

Prior single-turn studies represent the special case of $k=1$~\citep{li2025questbench}. For $k>1$, solving a task instance $T_y$ requires making a series of decisions, which are studied here. At each turn $t$, the model queries a variable $X_t \in \mathcal{U}(P)$. The oracle returns its value $x_t$, and the partial assignment becomes $\mathcal{A}_{t+1} = \mathcal{A}_t \cup \{X_t = x_t\}$, starting from $\mathcal{A}_1=\mathcal{A}$. The interaction continues until the model commits to an answer (Appendix~\ref{app:csp_formulation}).

\subsection{Multi-Turn Evaluation Protocol}
\label{sec:evaluation_protocol}

We define three evaluations for each problem $P$ and its task family, which separate the components of information seeking that a single interaction would otherwise combine. The first two operate at the problem level. They present the model with $P$ alone and ask it to report how much information is missing and which variables would supply it, so a failure at this stage reflects the model's assessment of the problem and not its conduct of a multi-turn interaction. The third operates at the task level. The model receives $P$, queries one variable per turn, observes the returned value, and decides at every turn whether to query again or to answer, so its performance depends on the sequence of queries it selects and on when it stops.
\begin{itemize}[leftmargin=5pt, topsep=0pt, itemsep=1pt, parsep=0pt, partopsep=0pt]
\item \textbf{Degree of underspecification prediction ($k$-prediction):} Given problem $P$, the model predicts the degree of underspecification $\hat{k}$. This setting tests whether the model recognizes how much information is missing.
\item \textbf{Missing variables identification ($k$-MSS identification):} Given problem $P$, and in one condition the true $k$, the model selects a minimal sufficient set from $\mathcal{U}(P)$. This setting tests whether the model plans a minimal query strategy without executing it.
\item \textbf{Sequential task-solving (multi-turn interaction):} Given a task instance $T_y=(P,y)$, the model observes only $P$. At each turn, it queries a variable $X_t \in \mathcal{U}(P)$ from an oracle or predicts $\hat{y}$. This setting tests the full interaction, where the model acquires information, updates its state, and answers once it judges the information sufficient.
\end{itemize}

To control for priors over targets, we evaluate models on the complete task family of each problem $P$. Because every instance in a family shares the same observed assignment $\mathcal{A}$ and differs only in the hidden target $y$, a model cannot succeed by guessing the most likely target and must instead narrow the feasible space through its queries. Appendix~\ref{appendix:multiturn_protocol} provides further details on our multi-turn evaluation protocol.

\paragraph{Metrics.}
We use different metrics for each setting, with formal definitions in Appendix~\ref{appendix:metrics}. For \textbf{$k$-prediction}, we report accuracy together with the rates of overestimation and underestimation, the fraction of problems where $\hat{k} > k$ and where $\hat{k} < k$. For \textbf{MSS identification}, we measure exact-set accuracy and Jaccard similarity against the ground truth MSS. For \textbf{sequential task solving}, we report \emph{final accuracy}, whether $\hat{y} = y$, and \emph{final sufficiency}, whether the acquired information determines $y$ uniquely. Final sufficiency differs from exact-set accuracy in MSS identification, because it is measured after a multi-turn interaction and reflects the variables the model actually queried. We also report behavioral metrics, including the total number of turns and the coverage, correctness, and ordering of queries. From reasoning traces, we compute the MSS relative mention rate, the fraction of the variables the model considers querying that belong to the MSS, and awareness of underspecification (Appendix~\ref{app:turn1-cot-recovery}).

\subsection{Datasets}
\label{sec:benchmark_construction}

\name instantiates multi-turn information seeking in mathematics, logic, biology, medicine, and general knowledge. Four domains are structured, in the sense that the queryable variables $\mathcal{U}(P)$ are predefined and finite and the model selects one of them at each turn. In these domains, we generate $k$-underspecified problems for $k \in \{1, 2, 3, 4\}$ where feasible, by masking variables from fully specified states and verifying that each masked set is a valid $k$-MSS. 
We then evaluate models on the task instances $T_y$ that each problem induces. The fifth domain, 20Q, is unstructured, since the model composes its own questions instead of selecting from a predefined $\mathcal{U}(P)$, and each question induces its own partition of the candidate set. 
The problem-level degree $k$ and the exact MSS are therefore not defined in advance, and 20Q tests information seeking under open-ended exploration. Appendix~\ref{appendix:benchmark_construction} reports dataset statistics and construction details.

\paragraph{Logic-Q-MT.}
\label{problem:logicq}
We extend the logical CSPs from \citet{li2025questbench} from single missing-variable settings to multi-turn interactions. Starting from rule-based reasoning environments, we recursively construct $k$-MSSs, requiring the model to sequentially identify and query the missing facts necessary to prove or disprove a target proposition. This domain explicitly evaluates information seeking and sufficiency judgment under strict formal logic.
Details are provided in Appendix~\ref{appendix:logicq_multi}.

\paragraph{GSME-Q-MT and GSME-Q-MT-Ext.}
GSME-Q contains an equation-based version of the grade-school math problems with human-annotated missing conditions required to compute an answer~\citep{li2025questbench}. Because the original dataset masks only a single variable ($k=1$), we extend it to a multi-turn setting (\emph{GSME-Q-MT}) by systematically masking multiple quantities. However, we found that GSME-Q-MT is nearly saturated; modern LLMs solve it reliably due to shallow constraints and weakly entangled variables. Consequently, we introduce \emph{GSME-Q-MT-Ext}, a rigorously enriched variant featuring deeper dependency structures, larger variable sets, and strictly validated $k$-MSSs. This extension provides a more diagnostic testbed for evaluating dependency tracking and step-wise mathematical information acquisition. Details are provided in Appendix~\ref{appendix:gsme_multi}.

\paragraph{GeneReg-MT.}
We construct tasks from Boolean models of gene regulation. The variables $\mathcal{X}$ are gene expression states, the constraints $\mathcal{C}$ are Boolean update rules, and the target $Y$ is either a steady state or the value of a marker gene at convergence. We refer to the two settings as steady state identification and marker identification. We mask initial gene expression values that determine $Y$, so the model must acquire the genes that fix the outcome under cyclic regulatory dependencies. Appendix~\ref{appendix:genereg_multi} gives the construction details.

\begin{wrapfigure}[13]{r}{0.5\textwidth}
    \vspace{-18pt}
    \centering
    \includegraphics[width=\linewidth]{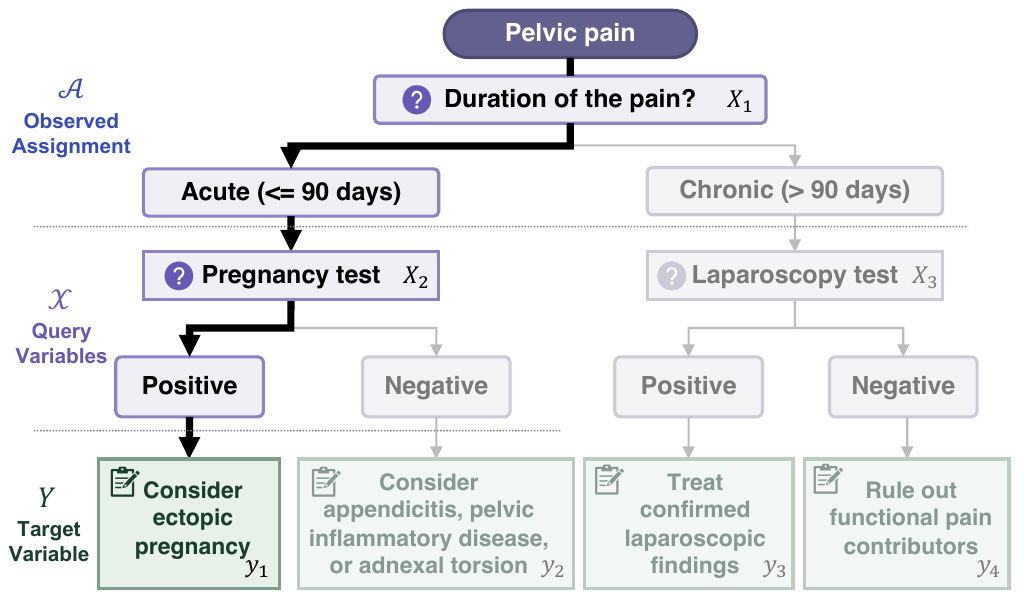}
    \captionsetup{font=small}
    \caption{Example ClinGuide-MT diagnostic pathway for pelvic pain \citep{patient_history, nice2021pelvicfloor, scott2014symptom}.
    }
    \vspace{-8pt}
    \label{fig:ClinGuideSchematic}
\end{wrapfigure}

\paragraph{ClinGuide-MT.}
Clinical reasoning proceeds by acquiring evidence in sequence before a diagnosis or management decision~\citep{ledley1959reasoning,mehandru2025erreason,sikora2026dontknow}. We construct tasks from diagnostic decision trees in clinical guidelines and textbooks (Fig.~\ref{fig:ClinGuideSchematic}). Internal nodes are the clinical variables $\mathcal{X}$, including symptoms, risk factors, and test results, and leaf nodes are the target outcome $Y$, a diagnosis, treatment, triage decision, or follow-up recommendation. We create $k$-underspecified problems by withholding the values of the last $k$ variables on a diagnostic pathway from the root to a leaf. Because a pathway encodes conditional dependencies, where an earlier finding determines which later variables are relevant, ClinGuide-MT tests both which variables the model queries and in which order. Appendix~\ref{appendix:clinguide} gives the construction details.

\paragraph{20Q.}
We use the 20 Questions game as an open-domain information-seeking setting. The hidden target $Y$ is sampled from a finite candidate set, and the model must deduce it by generating natural language yes-or-no questions. Each question partitions the feasible space $\Omega(P)$ into subsets (\texttt{yes}, \texttt{no}, or \texttt{pass}, where \texttt{pass} indicates ambiguity). Unlike the structured domains above, 20Q does not provide a predefined, finite $\mathcal U(P)$. Instead, the model dynamically generates constraints over the candidate set. 
20Q thus serves as an open-ended extension of our framework: claims involving $k$ and MSS identification are grounded in the four structured domains, while 20Q tests free-form question generation and integration of natural-language evidence. 
We evaluate models on two subsets: \textsc{Common} (animals, places, food) and \textsc{Thing} (general objects). 
Details on 20Q are provided in Appendix~\ref{appendix:20q}.

\section{Results}

We organize our experiments around three research questions.
\emph{RQ1: Which variables to query?} 
We evaluate whether models recognize the missing information and identify the variables that determine the target, through $k$-prediction, MSS identification, and sequential task solving. Our primary metric is \emph{final sufficiency}, which records whether the acquired information determines the target, independent of the answer the model gives. The separation matters because models answer Logic-Q-MT correctly when we provide the values of all MSS variables (Table~\ref{app_tab:logic_q_all_conditions}), so failures under underspecification follow from insufficient information seeking and not from an inability to reason to the answer.
\emph{RQ2: In which order to query them?} 
We evaluate whether models acquire information in a valid order when the task imposes dependencies among variables. ClinGuide-MT provides this structure, since an earlier finding determines which variables are relevant next.
\emph{RQ3: How does the model use the acquired information?} 
We analyze whether the acquired information improves target identification. We report final accuracy together with the reduction in uncertainty at each turn and the ambiguity of the questions, which separates the informativeness of a question from the use the model makes of the answer.
Complete benchmarking results are in Appendix~\ref{app:benchmarking_results}, with additional analyses in Appendix~\ref{app:add_results}.

\paragraph{Models.}
We evaluate closed and open-weight LLMs, including GPT-5, GPT-5-mini, Gemini-3-Flash, Qwen3-4B-Thinking, Qwen3-30B-A3B-Thinking-FP8, and gpt-oss-20B. 
Compute constraints limit Gemini-3.1-Pro, gpt-oss-120B, Qwen3-Next-80B-A3B-Thinking-FP8, and the Qwen3.5 family to a subset of settings; the corresponding appendix tables list the models evaluated in each setting.

\paragraph{General experimental setup.}
Unless otherwise specified, models are not given an explicit turn budget, but interactions are capped at 10 turns for consistency; this cap does not significantly affect final sufficiency (Appendix Fig.~\ref{app_fig:logicq_multi_multiturn_budget20_vs_budget10}). 20Q sets a 20-turn budget per the game design.  
Oracles vary by dataset: we use an oracle algorithm for GSME-Q-MT, Logic-Q-MT, and GeneReg-MT, and an LLM-based oracle for ClinGuide-MT and 20 Questions (Appendix~\ref{app:info_source}).
Qualitative conclusions hold across oracle response policies (Appendix~\ref{app:oracle_robustness}), and the LLM-based oracles are validated for self-consistency, cross-oracle agreement, and agreement with human annotation (Appendix~\ref{app:oracle_reliability}).
In Logic-Q-MT, alternative MSSs are forbidden as queries to ensure fair comparison, and our conclusions are insensitive to the designated MSS (Appendix~\ref{app:alt_mss}).
Implementation details and dataset-specific prompts are provided in Appendix~\ref{app:implementation_detail}.
Prompts were fixed before the model comparison and shared across models (Appendix~\ref{app:prompt_sensitivity}).

\begin{wrapfigure}[14]{r}{0.5\textwidth}
    \vspace{-16pt}
    \centering
    \includegraphics[width=\linewidth]{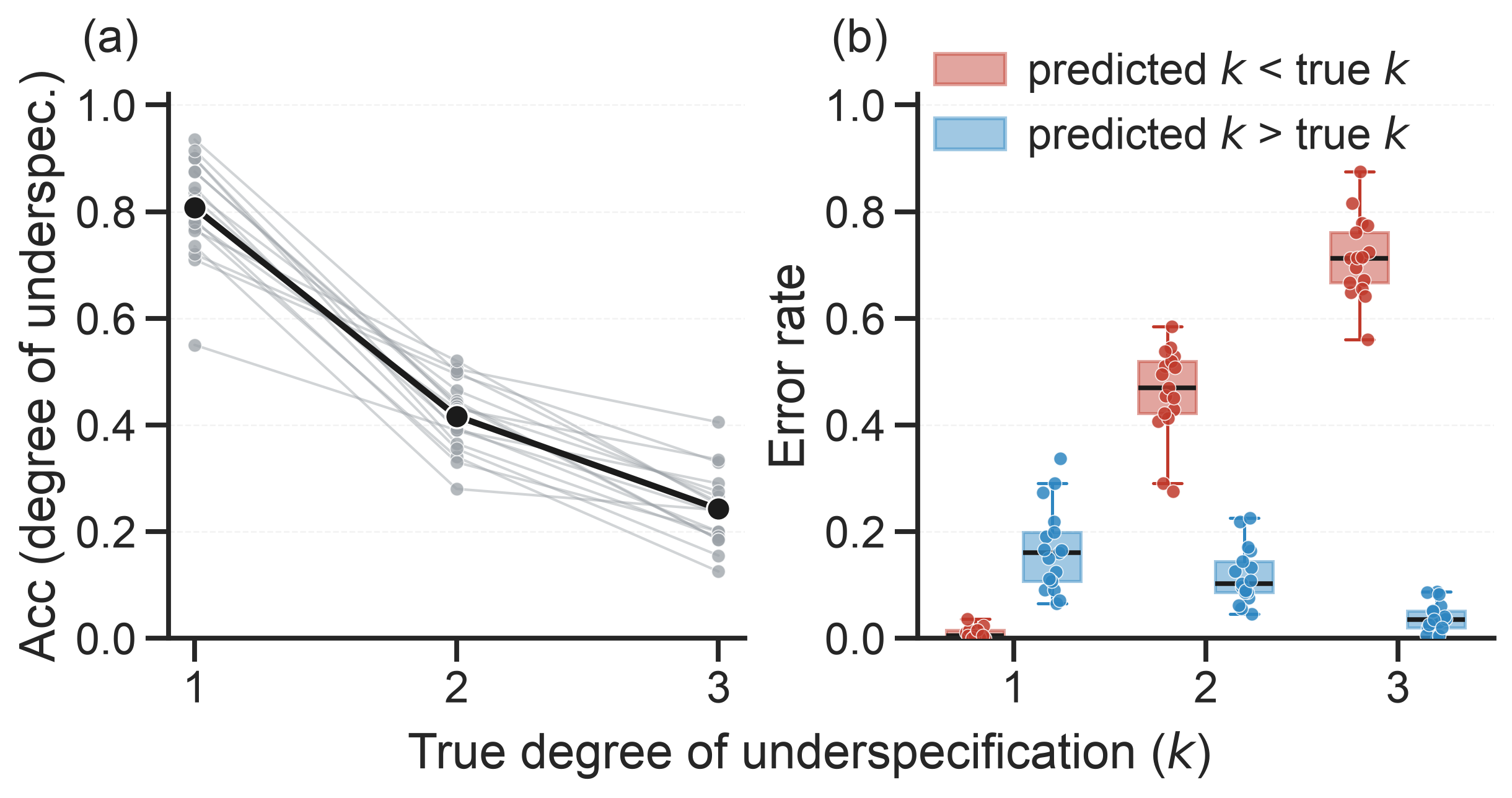}
    \vspace{-16pt}
    \captionsetup{font=small}
    \caption{
    (a) Accuracy of predicting the true $k$. Light gray lines denote individual models and the black line their mean. (b) Rate at which models predict a value smaller (red) or larger (blue) than the true $k$; dots are individual models.
    }
    \label{fig:logicq_multi_level_underspec}
    \vspace{-8pt}
\end{wrapfigure}

\takeaway{1 (RQ1)}
{Models detect that their context is insufficient to determine the target, but they underestimate how many additional variables they need before the target becomes identifiable.}
We first test whether LLMs can quantify missing information via multiple-choice \(k\)-prediction, selecting from $\{0,1,2,3,4,\text{not sure}\}$, in Logic-Q-MT.
We found that models rarely classify underspecified problems as fully specified (only predicting $0.6{\pm}0.8\%$ of problems as $k{=}0$), but they struggle to calibrate how much information is missing (Fig.~\ref{fig:logicq_multi_level_underspec}). Accuracy drops sharply as $k$ increases, falling below $0.5$ for $k \geq 2$ (per-model results in Appendix Table~\ref{app_tab:logic_q_k_prediction}). Errors are consistently biased toward underestimation: for $k=2$, where choices are balanced above and below the true value, models predict a smaller $k$ more than $4\times$ as often as a larger one on average.

\takeaway{2 (RQ1)}
{Detecting that information is missing does not imply identifying which information. Even when given the degree of underspecification $k$, models often fail to select a minimal set of variables that determines the target.}
\begin{wrapfigure}[16]{r}{0.5\textwidth}
    \vspace{0pt}
    \centering
    \includegraphics[width=0.5\textwidth]{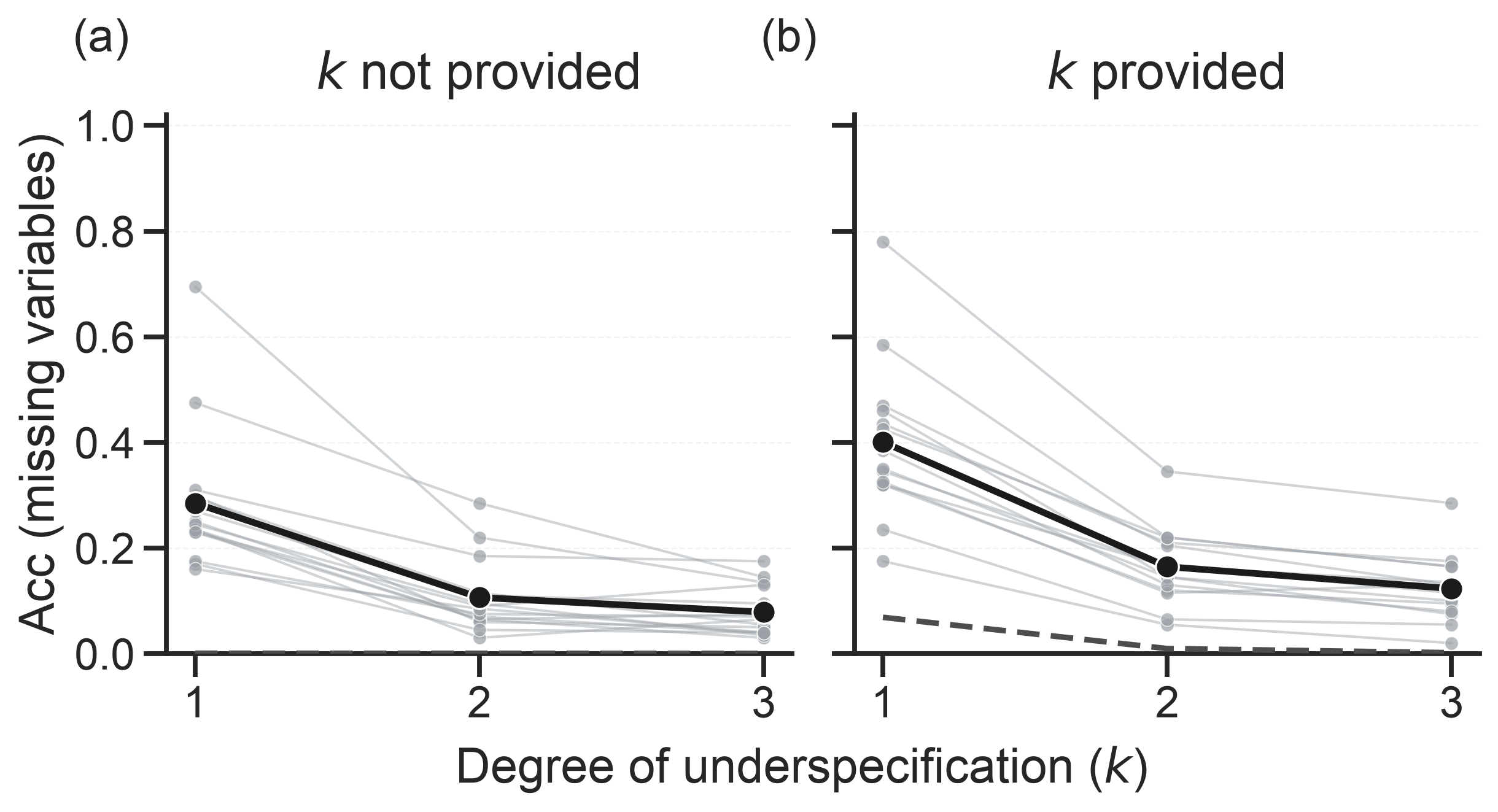}
    \vspace{-16pt}
    \captionsetup{font=small}
    \caption{
    Models predict the MSS of each problem in Logic-Q-MT, either (a) without knowing $k$ or (b) given the true $k$ and asked to select exactly $k$ variables. 
    We report exact-set accuracy against the true MSS. 
    Gray lines denote individual models, black lines the mean, and dashed lines a uniform-random baseline.
    }
    \vspace{-8pt}
    \label{fig:logicq_multi_mc}
\end{wrapfigure}
We next ask whether LLMs know what missing variables to ask for to solve underspecified problems. 
When models freely choose up to four variables ($k$ not provided), their exact-set accuracy is low and drops sharply as $k$ increases: no model exceeds $0.4$ accuracy at $k{=}2$ or $0.2$ at $k{=}3$ (Fig.~\ref{fig:logicq_multi_mc}a). 
Providing the true $k$ improves performance, but only modestly; both exact-set accuracy and Jaccard similarity remain low for $k{\geq}2$ (Fig.~\ref{fig:logicq_multi_mc}b). 
Thus, estimating the amount of missing information is helpful but insufficient: the harder challenge is identifying which variables form the target-relevant MSS. 
Jaccard similarity and recall trends are consistent (Appendix Fig.~\ref{app_fig:logicq_multi_mc_jaccard_recall}; per-model results in Appendix Table~\ref{app_tab:logic_q_all_conditions}).

\takeaway{3 (RQ1)}
{Under a fixed query budget, models often acquire more of the required information when queries are spread across turns instead of being issued together in one turn.}
We next study how interaction budgets affect final sufficiency. We prompt LLMs with different budgets of turns (Appendix~\ref{appendix:logicq_multi_prompts}) and number of questions allowed per turn, and measure the final sufficiency. 
In Logic-Q-MT and GeneReg-MT, final sufficiency generally increases as more turns are allowed, and, under the same total query budget, allocating queries across more turns often outperforms allowing more queries per turn (Appendix Figs.~\ref{app_fig:logicq_multi_multiturn_budget_exp},\ref{app_fig:genereg_multi_multiturn_budget_exp}).
Appendix Fig.~\ref{app_fig:logicq_multi_multiturn_sufficient_proportion_vs_turn} traces the proportion of sufficient interactions turn by turn, with the separation between models widening as $k$ increases.
This pattern shows that the benefit of interaction is not captured by the total number of queries alone, and how queries are distributed across turns also matters. 
However, forced multi-turn execution can introduce additional failures when models underestimate $k$. In GSME-Q-MT, constraining models to query only one variable per turn often causes them to stop too early and directly guess the final answer, thereby degrading performance (Table~\ref{tab:gsme_q_mt_summary}, Appendix Fig.~\ref{app_fig:gsme_multi_degrad}).

\begin{wrapfigure}[12]{r}{0.48\columnwidth}
    \vspace{-20pt}
    \centering
    \includegraphics[width=\linewidth]{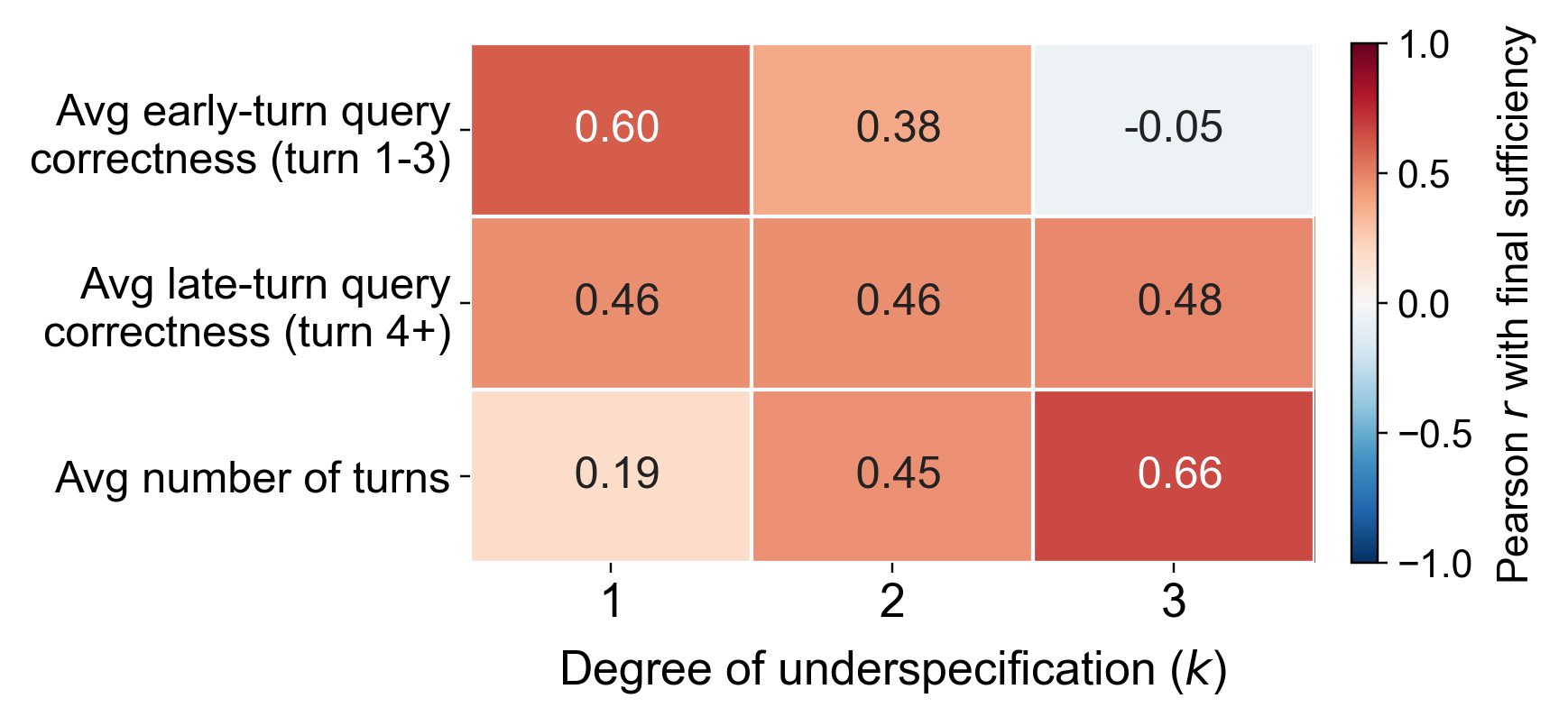}
    \vspace{-16pt}
    \captionsetup{font=small}
    \caption{
    Correlation between per-model behavioral features and final sufficiency in Logic-Q-MT. 
    Each cell reports the Pearson correlation across models between final sufficiency and a behavioral feature, computed separately for each $k$. 
    }
    \vspace{-8pt}
    \label{fig:logicq_multi_multiturn_corr_model_behavior_vs_model_performance_heatmap}
\end{wrapfigure}
\takeaway{4 (RQ1)}
{Success on highly underspecified problems depends less on the first query and more on sustained information acquisition across turns. Early-turn query correctness predicts final sufficiency at small $k$, and at larger $k$, stronger models keep identifying useful variables over longer interactions.}
\label{exp:model_level_corr_behavior_vs_perf}
The preceding results motivate a closer look at which model behaviors might explain final sufficiency.
For each $k$, we compute model-level Pearson correlations between final sufficiency and three behavioral features in Logic-Q-MT: early-turn (turns 1-3) query correctness, late-turn (turns 4+) query correctness, and number of turns. 
Early-turn correctness is strongly correlated with final sufficiency when underspecification is mild ($r{=}0.60$ at $k{=}1$), but its correlation vanishes at $k{=}3$ ($r{=}{-}0.05$; Fig.~\ref{fig:logicq_multi_multiturn_corr_model_behavior_vs_model_performance_heatmap}, Appendix Fig.~\ref{app_fig:logicq_multi_multiturn_corr_model_behavior_vs_performance_scatterplot}). 
On the other hand, late-turn correctness remains consistently correlated with success ($r{=}0.46$--$0.48$), and the correlation with the number of turns increases with $k$ ($r{=}0.19{\rightarrow}0.66$). 
A similar turn-count trend appears in GeneReg-MT, reaching $r{=}0.77$ on steady state identification (Appendix Fig.~\ref{app_fig:genereg_multi_dyn_attr_multiturn_correlation_model_behavior_vs_suff_heatmap}) and $0.97$ on marker identification (Appendix Fig.~\ref{app_fig:genereg_multi_dyn_marker_multiturn_correlation_model_behavior_vs_suff_heatmap}) at $k{=}4$.

\takeaway{5 (RQ1)}
{A first query that misses the minimal sufficient set is often recoverable. Interactions in which the model keeps searching for target-relevant variables reach sufficiency at close to the rate of interactions whose first query is correct.}
Building on Takeaway 4, we ask whether missing an MSS variable on the first turn necessarily prevents a interaction from reaching sufficiency.  
We focus on tasks with $k{\geq}2$ and compare two groups of interactions: those where the first query targets an MSS variable, and those where the first query does not target an MSS variable but the second query does. 

\begin{wrapfigure}[17]{r}{0.45\textwidth}
    \vspace{-4pt}
    \centering
    \includegraphics[width=\linewidth]{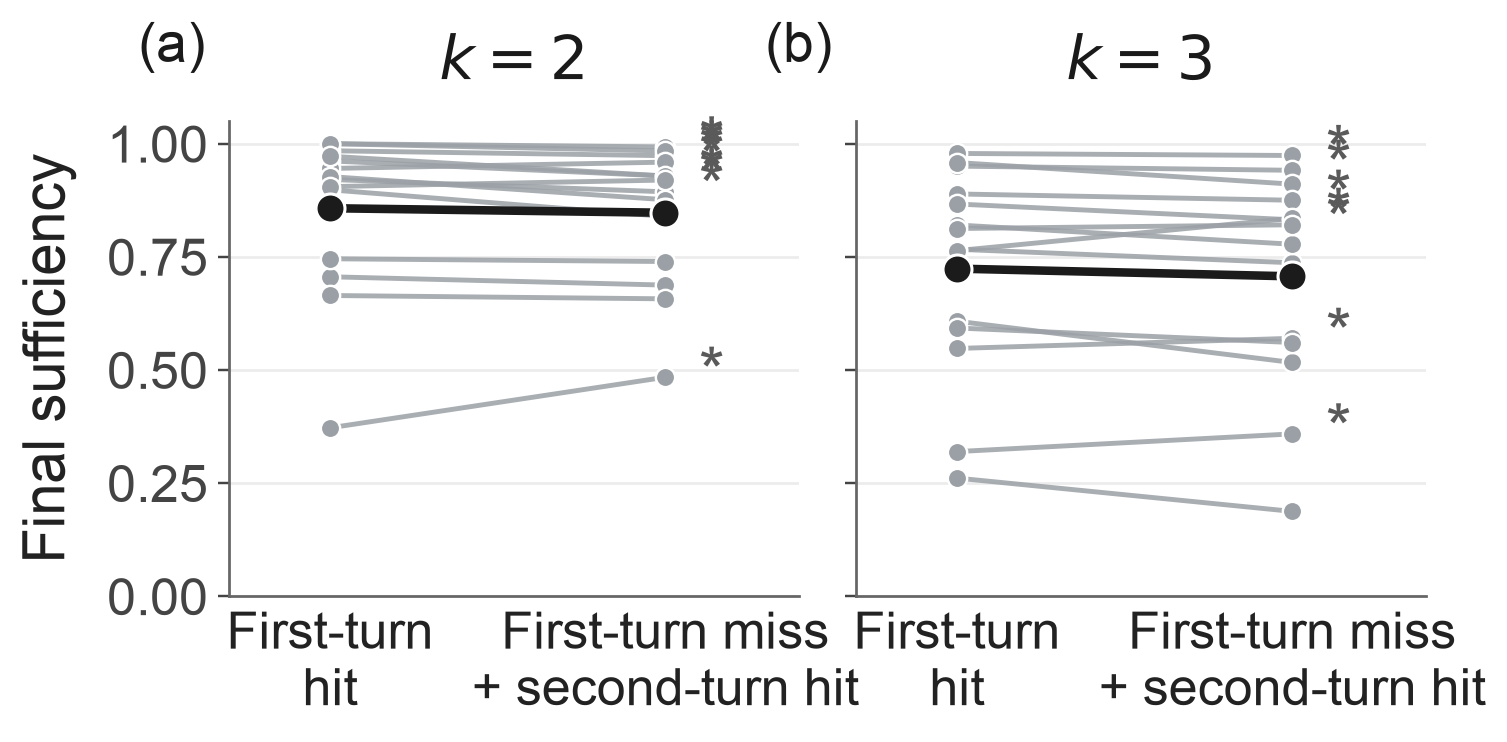}
    \vspace{-16pt}
    \captionsetup{font=small}
    \caption{
    Final sufficiency between interactions where the first query targets an MSS variable and interactions where the first query misses an MSS variable but the second query targets one. 
    Gray lines denote individual models and the black line their mean. Stars mark models for which this comparison passes a one-sided Miettinen--Nurminen non-inferiority test with a 10 percentage-point margin ($p{<}0.05$).
    }
    \vspace{-8pt}
    \label{fig:logicq_multi_multiturn_early_recovery_pairwise_comparison_logicq}
\end{wrapfigure}
In Logic-Q-MT, the latter group achieves final sufficiency close to the first-turn-hit group for both $k{=}2$ and $k{=}3$ (Fig.~\ref{fig:logicq_multi_multiturn_early_recovery_pairwise_comparison_logicq}; per-model results in Appendix Fig.~\ref{fig:logicq_multi_muliturn_early_recovery_allk_gap10pp}).
The gap is small for most models: it is within 5 percentage points for 12 out of 14 models, and within 10 percentage points for all models (Appendix Fig.~\ref{fig:logicq_multi_muliturn_early_recovery_sensitivity_analysis}). 
Using a one-sided Miettinen--Nurminen non-inferiority test with a 10 percentage-point margin, we find statistically supported near-parity for 8/14 models at $k{=}2$ and 7/14 models at $k{=}3$.  
The pattern is qualitatively similar but weaker in GeneReg-MT, where recovery cases are less frequent (per-model results in Appendix Figs.~\ref{fig:genereg_multi_dyn_attr_muliturn_early_recovery_allk_gap10pp},~\ref{fig:genereg_multi_dyn_marker_muliturn_early_recovery_allk_gap10pp}; sensitivity in Appendix Fig.~\ref{fig:genereg_multi_muliturn_early_recovery_sensitivity_analysis}).
These results show that early query misses are not necessarily fatal: what matters is whether the model continues to search and identify target-relevant missing variables.

\emph{First-turn mentions of MSS variables predict later recovery.}
To probe why some first-turn misses are recoverable, we analyze the model's visible first-turn reasoning traces. 
In Logic-Q-MT, the first-turn MSS relative mention rate is positively associated with both second-turn hit after a first-turn miss (log odds ratio $=0.365$, $p{<}$1e-4) and final sufficiency (log odds ratio $=0.528$, $p<$1e-4), even after controlling for CoT length, first-turn query mention, degree of underspecification, model type, and problem-difficulty factors (Appendix Fig.~\ref{app_fig:turn1_cot_covariate_comparison_logicq_multi}). By contrast, awareness of underspecification has near-zero adjusted association. 
These results suggest that some recovery cases are preceded by specific partial recognition of the relevant missing variables, even when the model does not select those variables as its first query. 
MSS relative mention rate remains positive in steady-state and marker identification subsets in GeneReg-MT, though the effects are noisier (Appendix Fig.~\ref{app_fig:turn1_cot_covariate_comparison_grn_multi}).

\takeaway{6 (RQ2)}
{In order-dependent tasks, querying the right variables is not enough if the model queries them in the wrong order. Clinical diagnostic pathways impose conditional dependencies, and violating them lowers final accuracy even when the model eventually acquires every necessary variable.}

In ClinGuide-MT, diagnostic pathways impose ordered dependencies among clinical variables, where earlier findings determine which downstream questions are relevant and how their answers should be interpreted. For example, as illustrated in Fig.~\ref{fig:ClinGuideSchematic}, whether a patient has acute or chronic pelvic pain changes which examination should be performed next. This setting allows us to evaluate not only whether models identify the underspecified information, but also whether they acquire it in a dependency-consistent order.

\begin{figure}[htp]
    \centering
    \vspace{-4pt}
    \includegraphics[width=1.0\linewidth]{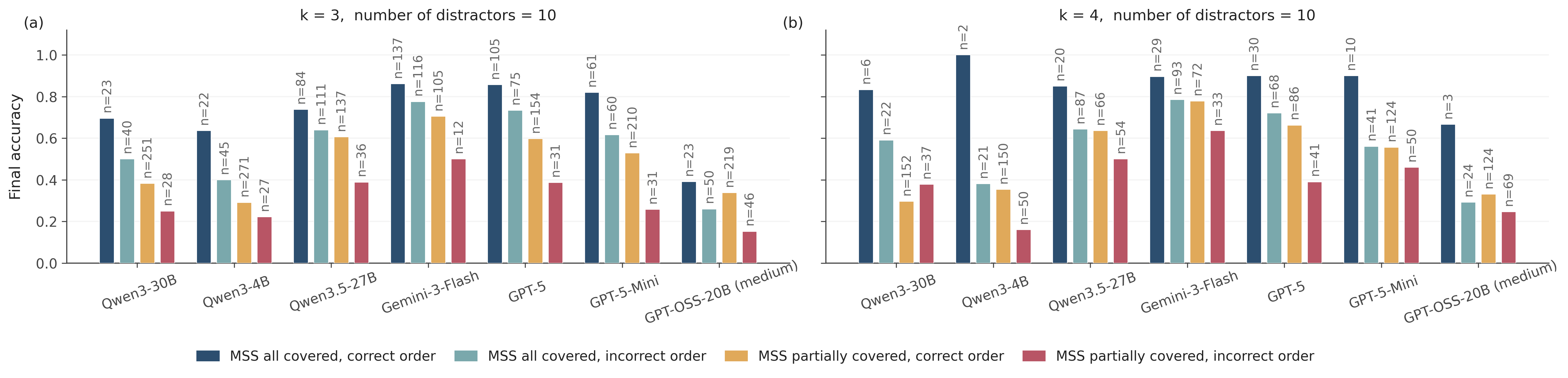}
    \captionsetup{font=small}
    \caption{ClinGuide-MT final accuracy stratified by MSS coverage and query order for $k\in\{3,4\}$ with $10$ distractor variables.}
    \vspace{-16pt}
    \label{fig:clinguide_order}
\end{figure}

We find that following the correct query order consistently improves final accuracy both in MSS fully covered and partially covered interactions (Fig.~\ref{fig:clinguide_order}). This effect is more pronounced in more difficult settings with larger $k$ and more distractors.
We further quantify the effect of query coverage, order correctness, and query correctness to final accuracy using regression analysis, (Appendix Fig.~\ref{app:clinguide_heatmap_full} and Appendix~\ref{app:clinguide_fullresultstext}). Query coverage remains the strongest and most stable positive predictor of final accuracy, while order correctness is generally the second strongest positive predictor. Query correctness (i.e., avoiding off-branch or irrelevant queries) provides additional benefit and becomes more salient in more difficult settings.
These effects persist even in a deliberately favorable open-book setting that provides all source diagnostic algorithms verbatim, largely reducing the task to locating and following the correct pathway; accuracy rises substantially yet remains unsaturated (Appendix~\ref{app:clinguide_openbook}).%

\begin{figure}[!htp]
    \centering
    \vspace{-8pt}
    \begin{subfigure}[t]{0.43\textwidth}
        \centering
        \includegraphics[width=\textwidth]{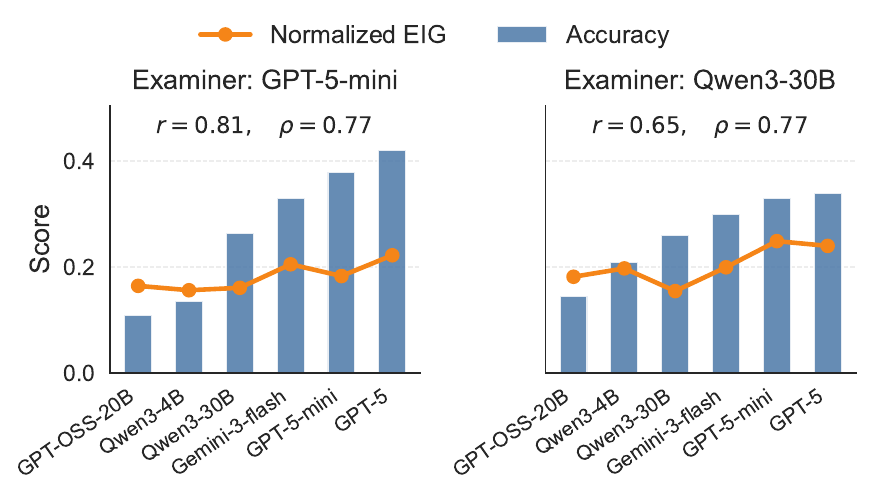}
        \vspace{-16pt}
        \caption{Acc. ($\uparrow$) vs. question informativeness ($\uparrow$).}
        \label{fig:thing_acc_quality}
    \end{subfigure}
    \hfill
    \begin{subfigure}[t]{0.55\textwidth}
        \centering
        \includegraphics[width=\textwidth]{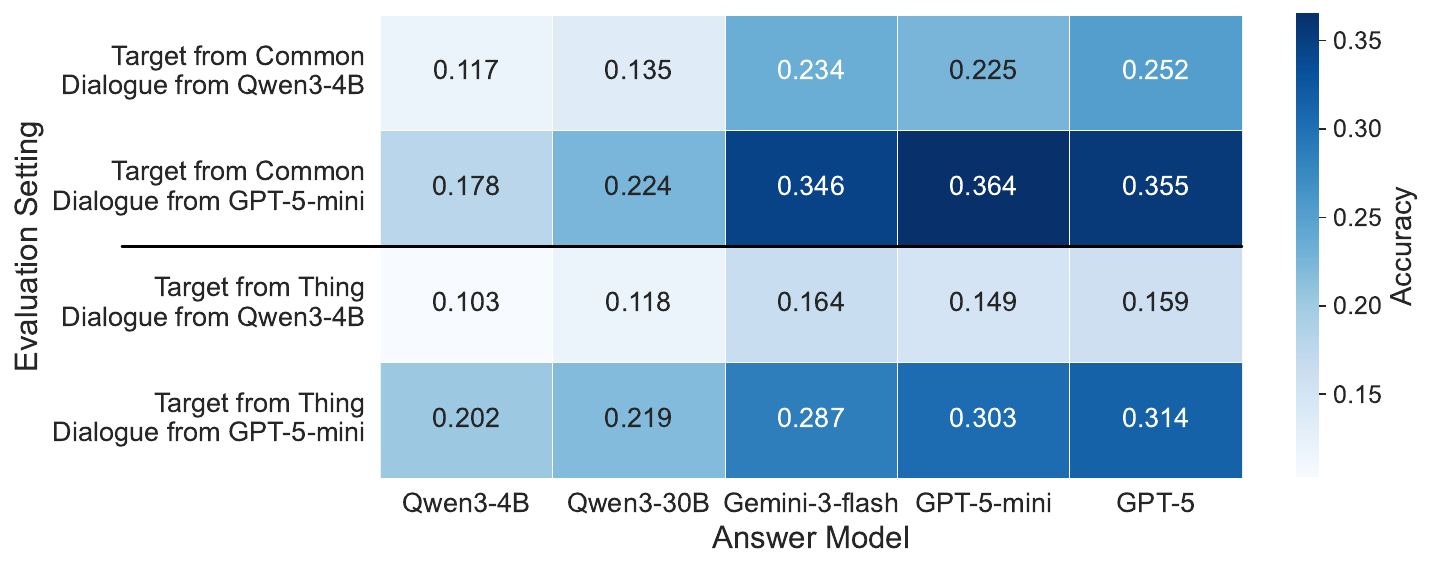}
        \vspace{-16pt}
        \caption{Final target identification from the same QA dialogue.
        }
        \label{fig:trace_transfer_heatmap}
    \end{subfigure}
    \captionsetup{font=small}
    \caption{Interaction-level analysis on 20Q.
    (a) Final accuracy and question informativeness, measured by normalized expected information gain. $r$ denotes Pearson correlation and $\rho$ denotes Spearman rank correlation.
    (b) Target identification accuracy across answer models given the same completed QA dialogue.}
    \vspace{-8pt}
    \label{fig:trace_analysis}
\end{figure}

\takeaway{7 (RQ3)}
{Informative queries are necessary but not sufficient for identifying the target. Models whose queries carry similar expected information gain differ substantially in final accuracy, so success also depends on how the model integrates the acquired evidence into its belief over the remaining candidates.}
We measure question informativeness using the average normalized expected information gain (nEIG) of a model's questions (Appendix~\ref{app:20q_metric}); nEIG matches blinded human preferences in a forced-choice study (76.5\% of 200 pairs, 86.5\% at high annotator confidence; Appendix~\ref{app:nig_human_validation}). As shown in Fig.~\ref{fig:thing_acc_quality}, higher-gain questions generally correlate with higher final target-identification accuracy, suggesting that effective information seeking is important for 20Q. However, models with similar question informativeness can still differ substantially in final accuracy, indicating that success also depends on how well they use the collected answers.
To isolate this effect, Fig.~\ref{fig:trace_transfer_heatmap} performs trace transfer: given the same completed QA dialogue, we replace the original guesser with different answer models and ask each to infer the target. Stronger answer models identify the target more reliably from identical traces, showing that some failures come not from asking uninformative questions, but from failing to integrate the acquired evidence.

\begin{figure}[htp]
    \centering
    \vspace{-8pt}
    \begin{subfigure}[t]{0.42\textwidth}
        \centering
        \includegraphics[width=\textwidth]{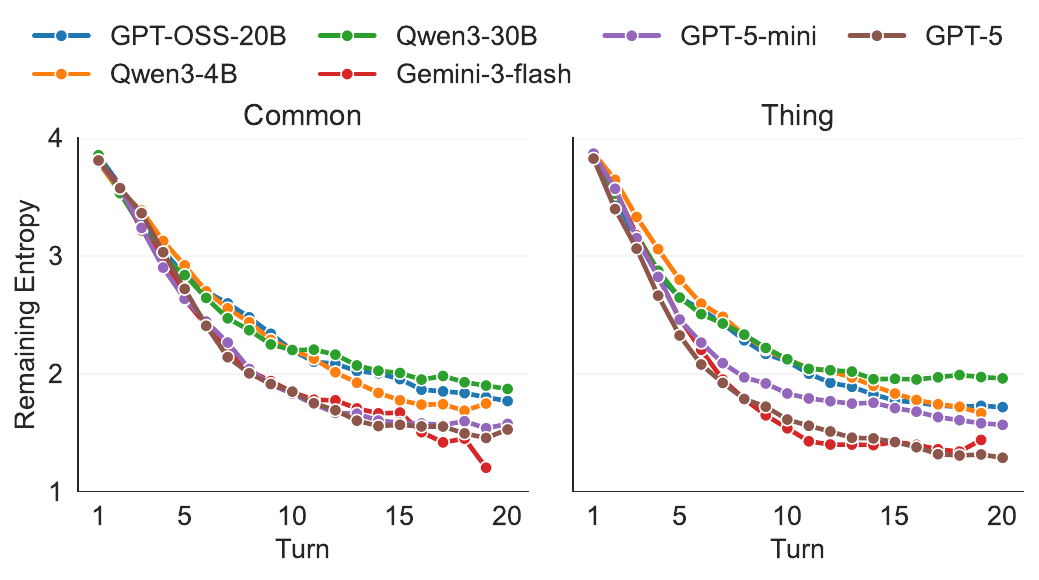}
        \vspace{-8pt}
        \caption{Remaining entropy across turns.}
        \label{fig:remaining_entropy_by_dataset}
    \end{subfigure}
    \hfill
    \begin{subfigure}[t]{0.57\textwidth}
        \centering
        \includegraphics[width=\textwidth]{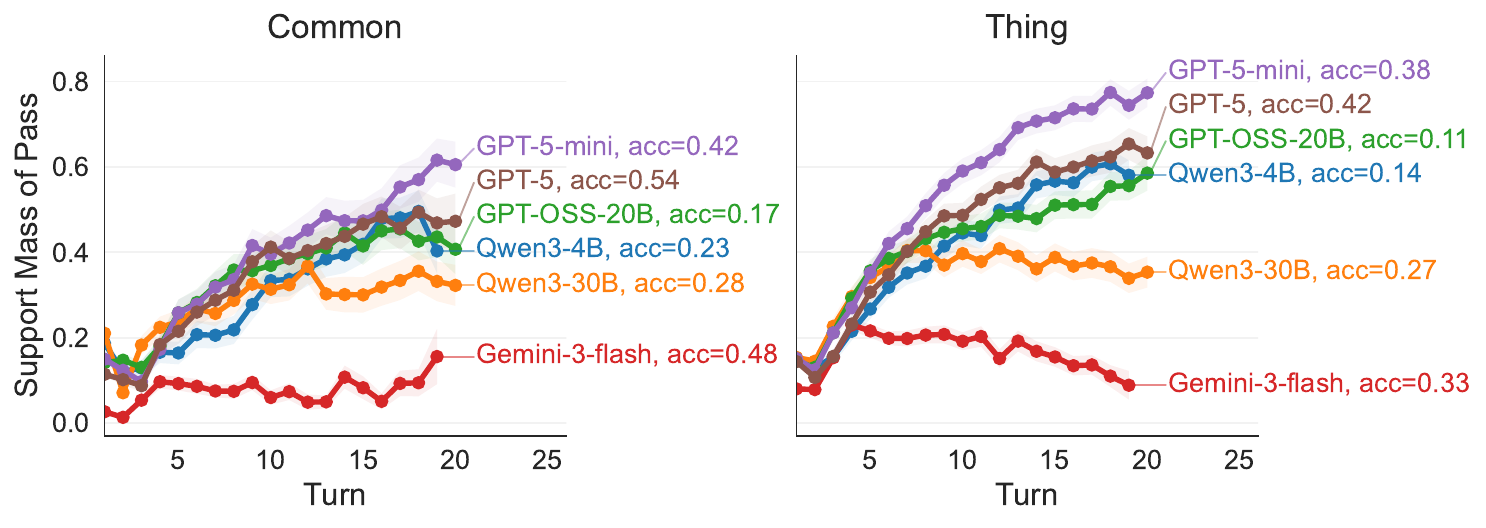}
        \vspace{-8pt}
        \caption{Probability mass of candidates consistent with \textit{pass}.}
        \label{fig:pass_mass_by_dataset}
    \end{subfigure}
    \captionsetup{font=small}
    \caption{Turn-level analysis on 20Q under the examiner GPT-5-mini.}
    \vspace{-8pt}
    \label{fig:entropy_passmass_analysis}
\end{figure}

\takeaway{8 (RQ3)}
{Models narrow the candidate space well and separate the remainder poorly. Early queries eliminate most of the feasible space, and later queries often fail to distinguish closely related target values that remain.}
We measure remaining entropy as the uncertainty of the posterior distribution over candidate targets after each turn. As shown in Fig.~\ref{fig:remaining_entropy_by_dataset}, remaining entropy decreases across turns for nearly all models and settings, indicating that the interaction generally provides useful information for narrowing the candidate set. However, the curves often flatten in later turns, suggesting that models become less effective at further reducing uncertainty once the candidate space has already been narrowed. Stronger models maintain lower entropy throughout the interaction, reflecting more efficient evidence acquisition and use, whereas weaker models plateau earlier and retain more residual uncertainty. These results suggest that the challenge is not only in making initial progress, but also in continuing to distinguish among the remaining plausible targets.
Additional turn-level results, including both examiners, are provided in Appendix~\ref{app:20q_full_results}.

\takeaway{9 (RQ3)}
{A partially informative answer is not a failed turn. Stronger models extract evidence from responses that do not separate the candidates and use it in later turns to narrow the remaining target values.}
Fig.~\ref{fig:pass_mass_by_dataset} reports \textit{pass} support mass at each turn, i.e., how much of the model's current candidate belief falls on targets for which the question cannot be clearly answered with yes or no (Appendix~\ref{app:20q_metric}). Higher pass mass means the question less cleanly partitions the remaining candidates under the binary-answer protocol.
For most models, pass mass increases over the interaction, suggesting that later questions tend to become more borderline or partially applicable as the remaining candidates become harder to distinguish. 
However, high pass mass does not consistently imply lower final accuracy. 
Several stronger models remain competitive despite asking high-pass questions, suggesting that \textit{pass} feedback can still provide useful partial evidence when models continue to update the search space rather than treating it as a failed query. 
We further analyze representative failure modes through qualitative case studies in Appendix~\ref{app:20q_case_study}.

\section{Discussion and Conclusion}
\label{discussion}

We study multi-turn information seeking under underspecification, where the model recognizes the missing information, acquires it through interaction, and uses it to determine the target. Across mathematics, logic, biology, medicine, and general knowledge, current LLMs detect underspecification but underestimate its degree, query incomplete sets of variables, and stop before the acquired information determines the target. Final accuracy alone therefore does not measure interactive competence. Evaluation should also record whether the model recognizes that its context is insufficient, which variables it needs, and whether it acquires them over successive turns. \name supports this evaluation by measuring final sufficiency, which separates the ability to answer from the ability to acquire the information that determines the answer.

\paragraph{Limitations and future work.}
Future work should extend the setting to richer user models, with ambiguous, noisy, changing, or inconsistent responses~\citep{ni2026user_model}, to probabilistic formulations of uncertainty, to longer conversations~\citep{motwani2026longcot}, and to forms of underspecification beyond missing variables, such as missing rules or unknown constraints~\citep{grand2026shoot,yang2026programbench}. 
We treat every query as equally expensive, while acquisition in practice differs in effort, latency, invasiveness, and risk. Unequal costs would turn query selection into a decision under a budget. An LLM serves as the oracle in ClinGuide-MT and 20Q. 
Consistency, cross-model, and human-agreement checks indicate that the oracle does not account for the failures we report (Appendix~\ref{app:oracle_reliability}), and human-audited protocols remain a useful extension. 
Normalized expected information gain measures the expected reduction in uncertainty over a surrogate candidate pool, which we validate against blinded human judgments (Appendix~\ref{app:nig_human_validation}). 
It does not measure question informativeness in a broader sense. ClinGuide-MT uses pathways from published guidelines to evaluate  information seeking, and it does not establish clinical readiness. 
Analyses of reasoning traces and internal states may further clarify when a model tracks the missing information and when apparent recovery reflects shallow exploration~\citep{serapio2025psychometric}.

\ack{
We thank Alexander D'Amour, Yuchang Su, Kexin Chen, Yasha Ektafaie, Kevin Li, and Pengwei Sui for helpful discussions. 
Z.W. provided advisory contributions; this work was partially conducted during a prior lecturing appointment at Harvard University in a Google DeepMind capacity.
Y.H., M.D., X.S., S.G., and M.Z. gratefully acknowledge the support, in part, by NSF CAREER Award 2339524, ARPA-H Biomedical Data Fabric (BDF) Toolbox Program, Amazon Faculty Research, Google Research Scholar Program, AstraZeneca Research, GlaxoSmithKline Award, Roche Alliance with Distinguished Scientists (ROADS) Program, Sanofi iDEA-iTECH Award, Boehringer Ingelheim Award, Merck Award, Optum AI Research Collaboration Award, Pfizer Research, Gates Foundation (INV-079038), Chan Zuckerberg Initiative, Collaborative Center for XDP at Massachusetts General Hospital, John and Virginia Kaneb Fellowship at Harvard Medical School, Biswas Computational Biology Initiative in partnership with the Milken Institute, and the Kempner Institute for the Study of Natural and Artificial Intelligence at Harvard University. This work was delivered as part of the AURORA project supported by the Cancer Grand Challenges partnership funded by Cancer Research UK ([CGCAI1-Mar26/100001]).
Any opinions, findings, conclusions or recommendations expressed in this material are those of the authors and do not necessarily reflect the views of the funders.
}
\clearpage

\bibliography{ref}
\bibliographystyle{unsrtnat}

\newpage
\appendix
\section{Ethics statement and broader impacts}
\label{app:broader_impact}
This work formalizes and measures multi-turn information seeking, supporting evaluation beyond final-answer accuracy: whether models recognize missing information, ask appropriate follow-up questions, and avoid unsupported guesses.
Such evaluation may help develop interactive systems that are more transparent about uncertainty and more cautious under underspecification.
The evaluation suite that instantiates our framework, \name, is constructed from public or synthetically generated sources and does not require collecting private user data.

At the same time, our results are diagnostic measurements on controlled tasks rather than evidence of deployment readiness.
Although \name includes clinical decision procedures and biological networks, these tasks serve only controlled evaluation, not real-world clinical or scientific decision-making.
The suite also does not cover the full diversity of real information-seeking interactions and may reflect cultural, linguistic, or domain-specific assumptions.
We have released code, data, and prompts to support reproducibility, auditing, and further stress-testing.

\clearpage

\section{Additional motivation}
\label{appendix:additional_motivation}

\subsection{Underspecification and the inference gap}
\label{app:underspecification}

\textbf{Underspecification and ambiguity.} Underspecification and ambiguity concern different stages of an interaction. An ambiguous request admits
more than one interpretation, so the uncertainty lies in which problem the user intends
to pose. An underspecified request poses its problem clearly and withholds what
is needed to solve it~\citep{li2025questbench,zhang2023ambiguityres}. What is missing is information about
the case, not the user's intent. 

\textbf{Why underspecification is difficult to detect.} Ambiguity tends to
declare itself, since the competing readings are present in the request.
Underspecification leaves no such trace: users are rarely in a position to know what the model lacks, and a model can produce a confident answer by silently assigning presumptive values to variables it never queried.
Multi-turn
interaction compounds the problem, because an assumption adopted at an early turn
conditions every turn that follows and surfaces, if at all, only in the final
answer. 

\textbf{The inference gap.} 
Clinical medicine has studied this problem under its own name, the
\emph{inference gap}, defined as the distance between what is known at the point
of care and the evidence required to reach a decision for the individual
patient. Bridging it is treated as the leap clinicians make daily in fitting
generalized clinical evidence to a particular case~\citep{iom2007learning}. The gap is closed
by acquiring what is missing, whether by asking the patient, ordering a test, or
running a clinical study. These differ in cost and latency but
share a common structure: each requires recognizing that the target is not yet
determined, and then selecting an action expected to narrow the possibilities
that remain consistent with what is already known. Strong inference describes
the same procedure in scientific practice, where a crucial experiment is chosen
for its capacity to exclude competing hypotheses~\citep{platt1964strong}. Our information-seeking
tasks sit above the act of acquisition: we ask whether a model recognizes that
its context is insufficient and identifies what would resolve it, not
how that information can be obtained. 
This is a simplification: all queries carry the same cost, and every problem admits a sufficient set by construction, whereas in practice acquisition costs vary and some gaps cannot be closed by any available evidence.
What we measure
comes first in any setting: a model that cannot tell what is missing will not
ask for it, order it, or design an experiment to obtain it.

\subsection{LLMs can still achieve high accuracy even with limited information}
\label{appendix:pre_study_high_acc}

\begin{table}[h]
    \centering
    \caption{Performance of Qwen3-30B-A3B-Thinking-FP8 in the full MediQ dataset, with limited or full information provided to the LLM in prompt, and no interaction with user allowed.}
    \begin{tabular}{l c c c}
        \toprule
         & \textbf{CRAFT\_MD} & \textbf{MedQA-dev} & \textbf{MedQA-test} \\
        \midrule
        Limited information & 0.69 & 0.6 & 0.58 \\
        Full information    & 0.91 & 0.88 & 0.86 \\
        Proportion* & 74\% & 62\% & 61\% \\
        \bottomrule
        \multicolumn{4}{p{0.7\linewidth}}{\footnotesize * Proportion of questions where the model selects the correct choice in at least 2/4 repeats (limited information, no interaction).}
    \end{tabular}
\end{table}

Prompts:
\begin{promptbox}{System Prompt}
You are a medical doctor trying to reason through a real-life clinical case. Based on your understanding of basic and clinical science, medical knowledge, and mechanisms underlying health, disease, patient care, and modes of therapy, respond according to the task specified by the user. Base your response on the current and standard practices referenced in medical guidelines.
\end{promptbox}

\begin{promptbox}{User Prompt}
A patient comes into the clinic presenting with a symptom as described in the conversation log below:

PATIENT INFORMATION: \{initial\_or\_full\_info\} \\
QUESTION: \{question\} \\
OPTIONS: \{options\}

YOUR TASK: Given the information so far, you MUST produce a factual conclusion. Pick an option and respond ONLY with the letter choice and NOTHING ELSE. Do NOT ask any questions. Do NOT provide any explanations or rationales.
\end{promptbox}

\begin{table}[htp]
\centering
\caption{\label{tab:benchmark_comparison}
\textbf{Comparison with existing benchmarks for interactive information seeking.}
\cmark, \pmark, and \xmark \ 
denote full, partial or judge-mediated, and no direct support, respectively.
\textit{Multi-turn}: evaluation involves multiple interaction turns.
\textit{Proactive}: the model must actively ask questions or choose information-gathering actions.
\textit{Obj. Query}: objective labels are provided for what information should be queried.
\textit{Timing/Order}: the benchmark evaluates when to ask, when to stop, or the order of acquired information.
\textit{Target Ctrl.}: multiple compatible target instances are evaluated under the same initial information to control target-specific bias.
}

\setlength{\tabcolsep}{4pt}
\resizebox{\textwidth}{!}{
\begin{tabular}{l|cccccl}
\toprule
\textbf{Benchmark} &
\textbf{Multi-turn} &
\textbf{Proactive} &
\textbf{Obj. Query} &
\textbf{Timing/Order} &
\textbf{Target Ctrl.} &
\textbf{Domains} \\
\midrule
QuestBench~\citep{li2025questbench} & \xmark & \cmark & \cmark & \xmark & \xmark & Logic, planning, math \\
AR-Bench\citep{zhou2025arbench} & \cmark & \cmark & \pmark & \pmark & \xmark & Puzzles, numbers \\
Active Task Disamb.\citep{kobalczyk2025activebed} & \cmark & \cmark  & \xmark & \xmark & \xmark & 20Q, code \\
MediQ\citep{li2024mediq} & \cmark & \cmark  & \xmark & \pmark & \xmark & Clinical QA \\
CRAFT-MD\citep{johri2025patientinteract} & \cmark & \cmark  & \xmark & \pmark & \xmark & Clinical diagnosis \\
LLMs Get Lost\citep{laban2025llms_get_lost} & \cmark & \xmark  & \xmark & \pmark & \xmark & Code, DB, math, NLG \\
Interactive Bench.\citep{yue2026interactive} & \cmark & \cmark  & \pmark & \pmark & \xmark & Logic, math, games \\
\midrule
\textbf{\name (Ours)} & \cmark & \cmark  & \cmark & \cmark & \cmark & Math, logic, bio, clinical, 20Q \\
\bottomrule
\end{tabular}
}
\end{table}

\section{Formalization and evaluation details}
\subsection{Formal definitions}
\label{app:csp_formulation}

We define an information-seeking problem as a CSP $ P = \langle \mathcal X, \mathcal D, \mathcal C, \mathcal A, Y\rangle,$
where: (i) \(\mathcal X = \{X_i\}_{i=1}^N\) is a finite set of variables, (ii) \(\mathcal D = \{\mathcal D_i\}_{i=1}^N\) is a family of domains, where \(\mathcal D_i\) is the domain of \(X_i\), (iii) \(\mathcal C = \{c_j(\{X_i\}_{i\in I_j})\}_{j=1}^M\) is a set of constraints, where each \(c_j\) is a Boolean predicate over a subset of variables, (iv) \(\mathcal A = \bigwedge_{i\in I_{\mathsf{known}}}(X_i = x_i)\) is the known partial assignment on a subset \(I_{\mathsf{known}} \subseteq [N]\), with \(x_i \in \mathcal D_i\), (v) \(Y\) is the target variable whose value we wish to determine.

We assume that $(\mathcal C, \mathcal A)$ is satisfiable. Let $\Omega(P)$ denote the set of full assignments $\omega$ over $\mathcal X$ that satisfy both \(\mathcal C\) and \(\mathcal A\).
For $\omega\in\Omega(P)$, we write $Y(\omega)$ for the value of the target variable under~$\omega$.

\begin{definition}[Known]
Let \(\psi\) be a conjunction of variable assignments consistent with \((\mathcal C, \mathcal A)\). We say that the target variable \(Y\) is \emph{known under \(\psi\)} if all feasible assignments satisfying \(\psi\) agree on the value of \(Y\):
\[
\mathrm{Known}(Y \mid \psi)
\iff
\forall \omega_1,\omega_2 \in \Omega(P),\ 
(\omega_1 \vDash \psi \wedge \omega_2 \vDash \psi)
\Rightarrow
Y(\omega_1)=Y(\omega_2).
\]
\end{definition}

We write $\mathrm{dom}(\mathcal A)$ for the set of variables assigned in $\mathcal A$. Let \(
\mathcal U(P) = \mathcal X \setminus (\mathrm{dom}(\mathcal A) \cup \{Y\})
\) denote the queryable variables of \(P\).

\begin{definition}[Sufficient set]
For \(S \subseteq \mathcal U(P)\), we say \(S\) is \textit{sufficient to determine \(Y\) under \(\mathcal A\)} if for every partial assignment \(s\) to \(S\) consistent with \((\mathcal C, \mathcal A)\), we have
\[
\mathrm{Known}(Y \mid \mathcal A \wedge (S = s)).
\]
\end{definition}

\begin{definition}[Underspecification~\citep{li2025questbench}]
A problem $P$ is \textit{underspecified} if the target variable cannot be determined from the current information alone, i.e., \(\neg\text{Known}(Y \mid \mathcal A) \). 
\end{definition} 

We can now formalize multi-turn information-seeking on top of the preliminaries above. 

\begin{definition}[Minimal sufficient set]
Let \(P=\langle \mathcal X, \mathcal D, \mathcal C, \mathcal A, Y\rangle\) be an underspecified problem. A set of variables $S \subseteq \mathcal U(P)$ is a \emph{minimal $k$-sufficient set} for $P$ if:
\begin{enumerate}
    \item \textbf{Cardinality:} $|S| = k$.
    \item \textbf{Sufficiency:} $S$ is sufficient to determine $Y$ under $\mathcal A$.
    \item \textbf{Global minimality}: No set $S' \subseteq \mathcal U(P)$ with \(|S'| < k\) is sufficient to determine \(Y\) under \(\mathcal A\).
\end{enumerate}
\end{definition}

\label{method:distinction_problem_task}
Evidently, $k$ is unique for an underspecified \textit{problem} and characterizes the degree of underspecification of the problem. We term such problem with a degree of underspecification of $k$ a $k$-underspecified problem.
For each compatible value
\(
y \in \{Y(\omega): \omega \in \Omega(P)\},
\)
we define a \emph{task instance} as
\[
T_y=(P,y),
\]
where \(y\) is the (hidden) ground-truth value of the target variable \(Y\). Thus, a single problem \(P\) induces a family of task instances, one for each compatible target value.
Importantly, we also define task-level degree of underspecification for $T_y$.
\begin{definition}[Task-level degree of underspecification]
Let \(P=\langle \mathcal X, \mathcal D, \mathcal C, \mathcal A, Y\rangle\) be an underspecified problem with problem-level degree of underspecification \(k\), and let
\(\mathsf{MSS}(P)\) denote the collection of all minimal \(k\)-sufficient sets for \(P\).

For a task instance \(T_y=(P,y)\), its \emph{task-level degree of underspecification} is
\[
\begin{aligned}
k_{\mathsf{task}}(T_y)
:= \min \Bigl\{
|W| :\;&
\exists S \in \mathsf{MSS}(P),\ W \subseteq S, \forall \omega \in \Omega(P),\\ 
& Y(\omega)=y \Rightarrow \mathrm{Known}\!\left(Y \mid \mathcal A \wedge (W=\omega|_W)\right)
\Bigr\}.
\end{aligned}
\]
\end{definition}
Evidently, $k_{\mathsf{task}}(T_y)$ can be smaller than the problem-level degree of underspecification of the corresponding problem \(P\). For example, consider the problem \(P=\langle \mathcal X,\mathcal D, \mathcal C, \mathcal A, Y\rangle\) with
\(
\mathcal C=\{X_1=1 \Rightarrow Y=y_1,\;\; X_1=0 \land X_2=1 \Rightarrow Y=y_2,\;\; X_1=0 \land X_2=0 \Rightarrow Y=y_3\},
\)
\(\mathcal A=\emptyset\), and target variable \(Y\). This is a 2-underspecified problem. However, for the task instance \(T_{y_1}\), querying only the value of \(X_1\) is sufficient to derive $Y=y_1$.
Note, in certain cases, there can be a set of variables whose cardinality is smaller than $k_{\mathsf{task}}$, but can still solve the task. This is due to the existence of target value-specific shortcuts, but they are not subset of any of the minimal sufficient sets of the problem.

\subsection{Multi-turn interaction formalization}
\label{appendix:multiturn_protocol}
To formalize the multi-turn setup, fix a task instance \(T_y\), a model \(M\), and an information source \(O\). The model is initialized with the problem \(P\) only. Accordingly, the initial interaction history is
\(
\mathcal H_0 = P.
\)
At turn \(t\), the model generates a reasoning trace \(R_t\) and a query \(Q_t\) conditioned on the current history:
\(
(R_t,Q_t)=M(\mathcal H_{t-1}).
\)
Here, \(Q_t\) requests the value of some variable \(X^{(t)} \in \mathcal U(P)\). The information source then returns a response
\(
I_t = O(T_y,Q_t).
\)
The interaction history is updated as
\(\mathcal H_t = \mathcal H_{t-1} \cup \{(R_t,Q_t,I_t)\}.
\)
The interaction terminates when the model stops querying and instead outputs a reasoning trace \(R_{\mathrm{final}}\) and a prediction \(\hat y\) for the target value of \(Y\). The full interaction is therefore
\(
\mathcal H_M(T_y)=\bigl((R_1,Q_1,I_1),\dots,(R_T,Q_T,I_T),(R_{\mathrm{final}},\hat y)\bigr).
\)

\subsection{Evaluation metrics}
\label{appendix:metrics}

We consider the following metrics for each problem setup:
\begin{itemize}[leftmargin=*]
    \item \textbf{Degree of underspecification:} Accuracy (degree of underspecification), predicted $k$ less than ground-truth $k$, predicted $k$ greater than ground-truth $k$.

    \item \textbf{Missing variables:} For each problem $P$, we use a single ground-truth minimal sufficient set $S = \{X^{(1)}, \ldots, X^{(k)}\}$. Let the model-predicted
    variable set be $\hat S$. We define the following metrics:
    (1) Accuracy (missing variables) = $\mathbb I\{\hat S = S\}$,
    (2) Jaccard similarity =
    $\frac{|\hat S \cap S|}{|\hat S \cup S|}$.
    
    \item \textbf{Task-solving:} For a task \(T_y\) induced from a problem \(P\) with minimal sufficient set
    \(
    S = \{ X^{(1)}, X^{(2)}, \ldots, X^{(k)} \},
    \)
    let model prediction be \(\hat y\), and let the ordered variables queried by the model be
    \(
    \hat Q = (\hat X^{(1)}, \ldots, \hat X^{(T)}),
    \)
    where \(T\) is the last turn the model queries. 
    Let the deduplicated query sequence be
    \(
    \hat Q^{\mathrm{uniq}} = \mathrm{Unique}(\hat Q),
    \)
    i.e., the ordered list obtained from \(\hat Q\) by removing repeated queries while preserving first-occurrence order. 
    Let the hit subsequence of \(\hat Q\) be
    \(
    \hat Q^* = (\,\hat X \in \hat Q^{\mathrm{uniq}} : \hat X \in S\,),
    \)
    i.e., the ordered list of queried variables that belong to \(S\).
    For ClinGuide-MT, let the ordered sufficient variables be
    \(
    Q = (X^{(1)}, X^{(2)}, \ldots, X^{(k)}).
    \)
    We then define the following metrics:
    \begin{itemize}[leftmargin=*]
        \item \textbf{Task completion:}
        (1) Accuracy: \(\mathbb I\{\hat y = y\}\),
        (2) Final sufficiency: \( \mathbb I \{\mathrm{Known}(Y \mid  \mathcal H_T)\} \),
        (3) Turn to sufficiency: \( \min_t \{ t: \mathbb I \{ \mathrm{Known}(Y \mid \mathcal H_t)\} = 1 \} \) (for final-sufficient runs only),
        (4) Total turns: \(T\).
    
        \item \textbf{Turn queries (unordered):}
        (1) First-turn hit: \(\mathbb I \{\hat X^{(1)} \in S  \}\),
        (2) Second-turn hit: \(\mathbb I \{\hat X^{(2)} \in S \setminus \{\hat X^{(1)}\} \}\),
        (3) Query correctness: defines the proportion of queried variables that belong to MSS \(\mathrm{QueryCorr}(\hat{Q}) = \frac{1}{T} \sum_{i=1}^{T} \mathbb{I}\{\hat{X}^{(i)} \in S\}.\)
        (4) Query coverage: measures the fraction of MSS variables that have been successfully queried \( \mathrm{QueryCov}(\hat Q)=\frac{|\hat Q^*|}{k} \in [0,1] \).
    
        \item \textbf{Turn queries (ordered, in ClinGuide-MT only):} 
        Query order correctness: we evaluate the correctness of the query order using the Damerau-Levenshtein distance, which measures the minimum number of insertions, deletions, substitutions, and adjacent transpositions required to transform one sequence into another. Here, we compare the sequence of $\hat{Q}^*$ with the ground-truth ordered sequence $Q$ using the normalized distance: \(d_{\mathrm{DL}}(\hat{Q}^*, Q)\). We define order correctness as a normalized similarity score: \( \mathrm{OrdCorr}(\hat{Q}) = 1 - \frac{d_{\mathrm{DL}}(\hat{Q}^*, Q)}{\max(|\hat{Q}^*|, |Q|)} \in [0,1].\)
    \end{itemize}

\end{itemize}

Table~\ref{tab:metrics_glossary} summarizes all evaluation metrics used throughout the paper.

\begingroup
\small
\setlength{\tabcolsep}{4pt}
\renewcommand{\arraystretch}{1.2}

\begin{longtable}{@{}
  >{\raggedright\arraybackslash}p{0.185\linewidth}
  >{\raggedright\arraybackslash}p{0.245\linewidth}
  >{\raggedright\arraybackslash}p{0.40\linewidth}
  >{\raggedright\arraybackslash}p{0.075\linewidth}@{}}
\caption{Glossary of evaluation metrics. \emph{Used in:} \textbf{L}=Logic-Q-MT, \textbf{G}=GeneReg-MT,
\textbf{M}=GSME-Q-MT/-Ext, \textbf{C}=ClinGuide-MT, \textbf{20Q}=20 Questions.
$k/\hat{k}$: true/predicted degree of underspecification;
$S$: ground-truth MSS ($\hat{S}$: predicted);
$\hat{Q}$: variables queried over $T$ turns, $\hat{Q}^{*}$: those in $S$;
$Q$: order-dependent sufficient sequence (ClinGuide-MT);
$\mathcal{H}_t$: history through turn $t$; $\hat{y}/y$: predicted/true target; $p_t$: belief over 20Q candidates.}
\label{tab:metrics_glossary}\\
\toprule
\textbf{Metric} & \textbf{Notation} & \textbf{Description} & \textbf{Used in}\\
\midrule
\endfirsthead

\multicolumn{4}{@{}l}{\small\itshape (Table~\ref{tab:metrics_glossary} continued)}\\
\toprule
\textbf{Metric} & \textbf{Notation} & \textbf{Description} & \textbf{Used in}\\
\midrule
\endhead

\midrule
\multicolumn{4}{r@{}}{\small\itshape continued on next page}\\
\endfoot

\bottomrule
\endlastfoot

\multicolumn{4}{@{}l}{\textbf{\textit{(A) Uncertainty awareness (problem-level)}}}\\
\addlinespace[2pt]
Degree accuracy & $\mathbb{I}\{\hat{k}=k\}$ & Predicted degree of underspecification matches ground truth. & L\\
Under-prediction rate & $\mathbb{I}\{\hat{k}<k\}$ & Predicts fewer missing variables than required. & L\\
Over-prediction rate & $\mathbb{I}\{\hat{k}>k\}$ & Predicts more missing variables than required. & L\\
MSS accuracy & $\mathbb{I}\{\hat{S}=S\}$ & Predicted variable set equals the MSS. & L\\
Jaccard similarity & $|\hat{S}\cap S|/|\hat{S}\cup S|$ & Overlap between predicted set and MSS. & L\\
Recall & $|\hat{S}\cap S|/|S|$ & Fraction of MSS variables recovered. & L\\
\addlinespace[3pt]

\multicolumn{4}{@{}l}{\textbf{\textit{(B) Task completion (sequential task-solving)}}}\\
\addlinespace[2pt]
Final accuracy & $\mathbb{I}\{\hat{y}=y\}$ & Final answer matches the hidden target. & L, G, M, C, 20Q\\
Final sufficiency & $\mathbb{I}\{\mathrm{Known}(Y\mid\mathcal{H}_T)\}$ & Acquired information uniquely determines the target. & L, G, M\\
Turn to sufficiency & $\min_t\{t:\mathrm{Known}(Y\mid\mathcal{H}_t)\}$ & Earliest turn at which information becomes sufficient. & L, G\\
Total turns & $T$ & Interaction turns before committing to an answer. & L, G, M, C, 20Q\\
\addlinespace[3pt]

\multicolumn{4}{@{}l}{\textbf{\textit{(C) Query behavior (turn-level)}}}\\
\addlinespace[2pt]
First-turn hit & $\mathbb{I}\{\hat{X}^{(1)}\in S\}$ & First query is an MSS variable. & L, G\\
Second-turn hit & $\mathbb{I}\{\hat{X}^{(2)}\in S\setminus\{\hat{X}^{(1)}\}\}$ & Second query is a new MSS variable. & L, G\\
Query correctness & $\frac{1}{T}\sum_{i=1}^{T}\mathbb{I}\{\hat{X}^{(i)}\in S\}$ & Proportion of queries targeting sufficient variables. & L, G, C\\
Query coverage & $|\hat{Q}^{*}|/k$ & Fraction of MSS variables queried. & L,G,C\\
Query order correctness & $1-\dfrac{d_{\mathrm{DL}}(\hat{Q}^{*},Q)}{\max(|\hat{Q}^{*}|,|Q|)}$ & Normalized Damerau--Levenshtein similarity to the ground-truth order. & C\\
Over-questioning rate & --- & Keeps querying after information is already sufficient. & L, G, M\\
\addlinespace[3pt]

\multicolumn{4}{@{}l}{\textbf{\textit{(D) Reasoning-trace (CoT) diagnostics}}}\\
\addlinespace[2pt]
MSS relative mention rate & $R=\dfrac{\#\text{MSS vars planned}}{\#\text{queryable vars planned}}$ & Among variables the model considers asking about, the share that belong to the MSS ($R{=}0$ if none). & L, G\\
Awareness of underspecification & --- & Binary flag that more information is needed, without naming a variable. & L, G\\
\addlinespace[3pt]

\multicolumn{4}{@{}l}{\textbf{\textit{(E) 20Q information-theoretic metrics}}}\\
\addlinespace[2pt]
Entropy & $H(p_t)=-\sum_{c}p_t(c)\log p_t(c)$ & Remaining uncertainty over candidate targets. & 20Q\\
Expected information gain (EIG) & $\mathrm{EIG}(q_t)=H(p_t)-\mathbb{E}_{a}\!\left[H(p_{t+1}^{(a)})\right]$ & Expected entropy reduction over answers $a$ (mutual information $I(Y;A_t\mid q_t)$). & 20Q\\
Normalized EIG (nEIG) & $\mathrm{EIG}(q_t)/(H(p_t)+\epsilon)$ & Expected fraction of uncertainty removed; question-informativeness score. & 20Q\\
Pass mass & $\sum_{c:g(c,q_t)=\texttt{pass}}p_t(c)$ & Belief mass a question cannot cleanly split by yes/no. & 20Q\\

\end{longtable}
\endgroup

\subsection{Information source and oracle design}
\label{app:info_source}

To test LLM's information seeking ability with different types of information sources, we design three oracles that respond to queries from LLMs. 
\begin{itemize}[leftmargin=*]
    \item \textbf{Adversarial oracle:} Returns a deterministic value that is consistent with the problem constraints but chosen to maximally delay identification of the target outcome, i.e., preserving as many candidate outcomes as possible.

    \item \textbf{Random oracle:} Samples a value from the set of assignments consistent with the problem constraints. When the queried variable is not uniquely determined, the oracle may respond with “not sure”, reflecting ambiguity.

    \item \textbf{Cooperative oracle:} Returns a deterministic value consistent with the ground-truth assignment that most directly reduces uncertainty about the target outcome.
    
\end{itemize}
Note that: (1) the adversarial and cooperative oracles both act greedily (one-step look ahead). (2) While the cooperative oracle yields strictly easier evaluations, the adversarial oracle does not necessarily require more queries than the random oracle. The random oracle may introduce ambiguity through "not sure" responses, whereas the adversarial oracle must respond with a deterministic answer.

\begin{figure}[htp]
    \centering
    \includegraphics[width=0.8\linewidth]{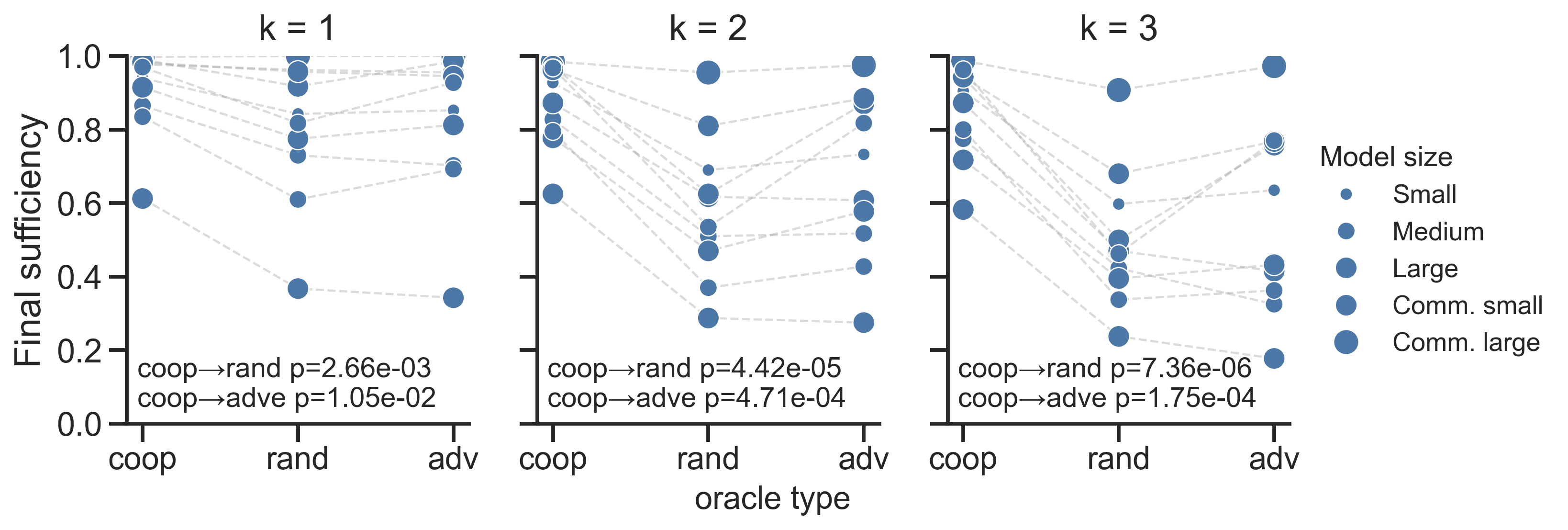}
    \caption{Performance of LLMs in Logic-Q-MT task-solving with different oracle types. coop stands for cooperative, rand stands for random, and adv stands for adversarial. With a budget of 10 turns, 1 query per turn allowed. $p$-values from two-sided paired T-test.}
    \label{app_fig:logicq_multi_multiturn_oracle_type_exp}
\end{figure}

\begin{figure}[htp]
    \centering
    \includegraphics[width=0.8\linewidth]{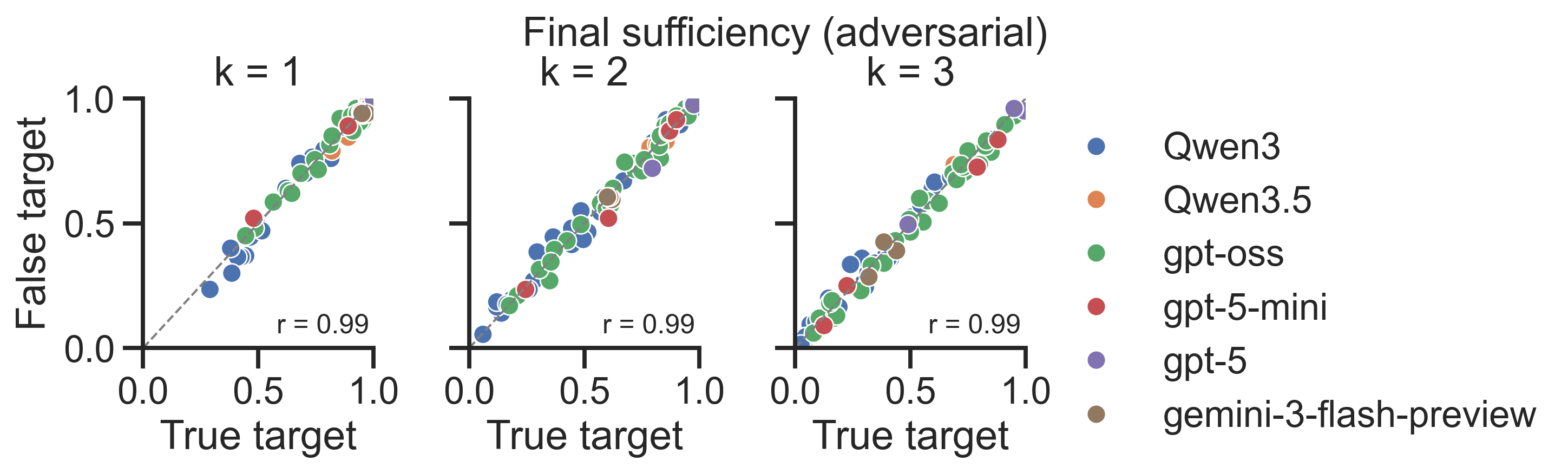}
    \caption{Correlation between performances on tasks with True target variable values and with False target variable values for LLMs in Logic-Q-MT task-solving. Runs are with adversarial oracles and different budget (1 turn x 4 queries, 4 turns x 1 query, 10 turns x 1 query) and alternative minimal sufficient set variable forbiddance statuses (forbid, allow). Due to (realistic) budget constraints, commercial models are only run with forbid. $r$ from Pearson's correlation.}
    \label{app_fig:logicq_multi_multiturn_target_true_false_adversarial}
\end{figure}

\begin{figure}[htp]
    \centering
    \includegraphics[width=0.8\linewidth]{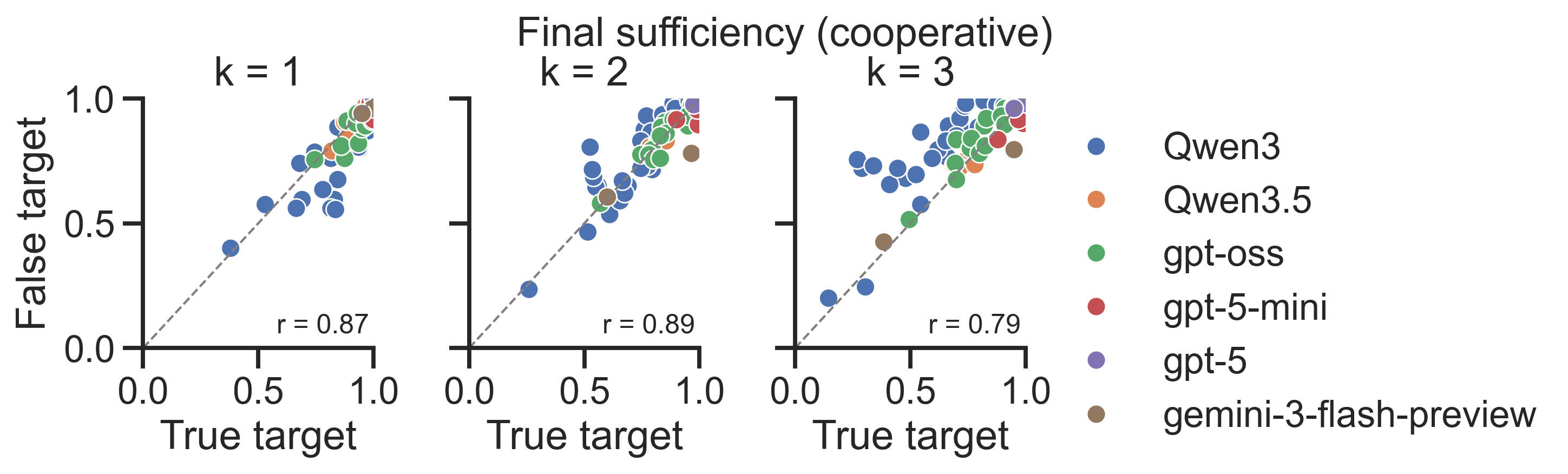}
    \caption{Correlation between performances on tasks with True target variable values and with False target variable values for LLMs in Logic-Q-MT task-solving. Runs are with cooperative oracles and different budget (1 turn x 4 queries, 4 turns x 1 query, 10 turns x 1 query) and alternative minimal sufficient set variable forbiddance statuses (forbid, allow). Due to (realistic) budget constraints, commercial models are only run with forbid. $r$ from Pearson's correlation.
    }
    \label{app_fig:logicq_multi_multiturn_target_true_false_cooperative}
\end{figure}

\begin{figure}[htp]
    \centering
    \includegraphics[width=0.8\linewidth]{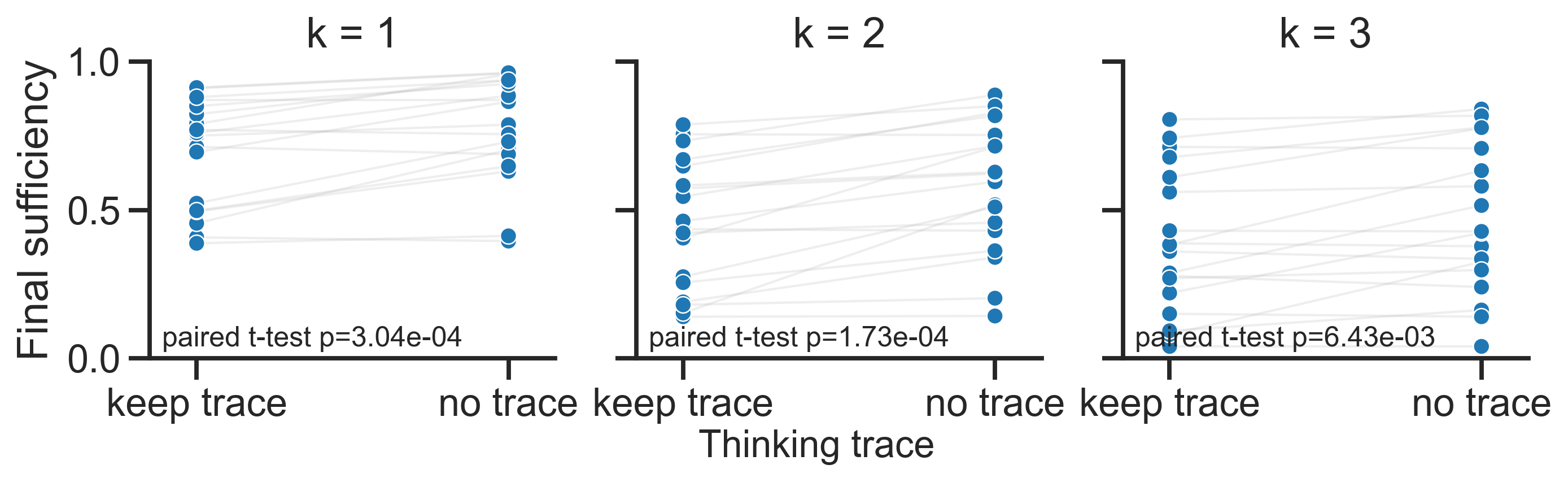}
    \caption{Performance of Qwen3-30B-A3B-Thinking-FP8 in Logic-Q-MT task-solving, with or without thinking traces appended in conversation history to the next turn. Runs are with different budget (1 turn x 4 queries, 4 turns x 1 query, 10 turns x 1 query), oracle type (adversarial, random, cooperative), and alternative minimal sufficient set variable forbiddance (forbid, allow).}
    \label{app_fig:logicq_multi_multiturn_thinking_trace_exp}
\end{figure}

For controlled tasks, we use adversarial oracles to evaluate a worst-case information-seeking setting, where the oracle answers consistently with the hidden target while avoiding unnecessary disclosures. 
This prevents models from benefiting from overly helpful responses and tests whether they can select target-relevant queries under realistic constraints such as privacy and limited observability. 
As sanity checks, models perform worse with random or adversarial oracles than with cooperative oracles (Appendix Fig.~\ref{app_fig:logicq_multi_multiturn_oracle_type_exp}), and adversarial-oracle results are more consistent across True- and False-target variants than cooperative-oracle (Appendix Figs.~\ref{app_fig:logicq_multi_multiturn_target_true_false_adversarial},\ref{app_fig:logicq_multi_multiturn_target_true_false_cooperative}). 
We do not carry models' previous turn reasoning traces to the next turn inference as it could lead to degradation in performance for specific models (Appendix Fig.~\ref{app_fig:logicq_multi_multiturn_thinking_trace_exp}).

\subsection{Robustness to oracle response policies}
\label{app:oracle_robustness}

In the structured domains, every oracle response remains consistent with the underlying CSP regardless of the response policy: the adversarial, random, and cooperative oracles defined in Appendix~\ref{app:info_source} differ only in how they select among valid responses. To test whether the reported findings depend on the adversarial policy used in the main experiments, we reran GeneReg-MT under the random and cooperative oracles for four representative models (Qwen3-4B-Thinking, Qwen3-30B-A3B-Thinking-FP8, GPT-5-mini, and gpt-oss-20B-high). 

First, the decline with the degree of underspecification persists: from $k=1$ to $k=4$, final sufficiency decreased for all four models under both oracles, by 29.7--41.9\,pp under the random oracle and 25.0--39.0\,pp under the cooperative oracle. Second, the turn-budget effect persists: 54 of 64 fitted budget slopes were positive, with most exceptions in the marker-identification task. Third, recovery from a first-turn miss persists: among episodes whose first query missed but whose second query hit, final sufficiency was 0.89--1.05$\times$ that of first-turn-hit episodes under the random oracle and 0.92--1.09$\times$ under the cooperative oracle, and after adjusting for GeneReg-MT task type and $k$, the recovery association was positive for all eight model--oracle pairs and reached $p<0.05$ for six of the eight.

Oracle type does change absolute difficulty as expected: cooperative responses yielded the highest pooled final sufficiency for all four models. However, it does not account for the qualitative conclusions above.

\subsection{Reliability of LLM-based oracles}
\label{app:oracle_reliability}

Logic-Q-MT, GSME-Q-MT, and GeneReg-MT use deterministic algorithmic oracles, so their responses are correct by construction. Only the open-ended ClinGuide-MT and 20Q settings rely on an LLM-based oracle, a design shared with interactive clinical benchmarks \citep{li2024mediq}: in 20Q, an oracle LLM answers free-form questions about the hidden target (yes, no, or pass), and in ClinGuide-MT, an oracle LLM decides whether a model's free-text question matches an information item on the guideline-derived diagnostic pathway \citep{patient_history}. Because LLM outputs can vary with sampling and input formulation \citep{sclar2024quantifying, mizrahi2024state}, we evaluate the reliability of these LLM-mediated information sources along four axes: within-oracle consistency, cross-oracle agreement, human validation, and impact on task outcomes.

\paragraph{Within-oracle consistency.}
For 20Q, we resampled the oracle's response to each logged question three times at temperature 0.6. All three samples agree on 91.9\%--95.6\% of questions ($\kappa \approx 0.91$--$0.94$; Table~\ref{app_tab:twentyq_oracle_consistency}), and most variation involves an uncertainty label rather than conflicting substantive answers. For ClinGuide-MT, we replayed each oracle on identical inputs (system prompt, patient context, and model query), reported separately for each evaluated answerer model. Repeated-response agreement ranges from 94.0\% to 97.7\% (Table~\ref{app_tab:clinguide_oracle_consistency}), with GPT-5-mini---the oracle used in the main ClinGuide-MT experiments---the more stable of the two; the maximum variation is 3.7\,pp and the relative ordering of answerer models is unchanged. Most inconsistencies are conservative shifts to or from ``Not sure'' (80.6\%--95.6\% of inconsistent cases) rather than switches between substantive answers.

\begin{table}[htp]
\centering
\small
\caption{Within-oracle consistency of the 20Q oracles: percentage of questions on which three responses sampled at temperature 0.6 all agree, with confidence intervals in brackets.}
\label{app_tab:twentyq_oracle_consistency}
\begin{tabular}{lcc}
\toprule
\textbf{Oracle} & \textsc{Common} (\%) & \textsc{Thing} (\%) \\
\midrule
Qwen3-30B-A3B-Instruct-FP8 & 95.4 [93.6, 97.1] & 95.6 [93.8, 97.2] \\
GPT-5-mini & 92.7 [90.5, 94.9] & 91.9 [89.5, 94.3] \\
\bottomrule
\end{tabular}
\end{table}

\begin{table}[htp]
\centering
\small
\caption{Within-oracle consistency of the ClinGuide-MT oracles when replayed on identical inputs (system prompt, patient context, and model query), reported separately for each evaluated answerer model, with confidence intervals in brackets. The last column is the share of inconsistent cases involving a shift to or from ``Not sure''.}
\label{app_tab:clinguide_oracle_consistency}
\begin{tabular}{llcc}
\toprule
\textbf{Oracle} & \textbf{Answerer} & \textbf{Agreement (\%)} & \textbf{``Not sure'' share (\%)} \\
\midrule
GPT-5-mini & Qwen3-30B-A3B-Thinking-FP8 & 97.7 [96.6, 98.7] & 86.8 \\
GPT-5-mini & GPT-5 & 97.3 [96.6, 98.4] & 80.6 \\
GPT-5.4 & Qwen3-30B-A3B-Thinking-FP8 & 94.1 [92.7, 96.5] & 95.6 \\
GPT-5.4 & GPT-5 & 94.0 [92.8, 96.2] & 90.8 \\
\bottomrule
\end{tabular}
\end{table}

\paragraph{Cross-oracle agreement.}
For 20Q, we regenerated responses to the same 545 questions using alternative oracle models: GPT-5-mini and GPT-5.4 agree with Qwen3-30B-A3B-Instruct-FP8 on 93.4\% and 94.1\% of responses, respectively. Most disagreements concern the boundary between a decisive answer and ``pass'' rather than reversals between substantive answers; Table~\ref{app_tab:twentyq_cross_oracle_examples} shows representative boundary cases together with an independent human label. For ClinGuide-MT, we compared the two oracle choices (GPT-5-mini and GPT-5.4) on identical inputs, separately for each evaluated answerer model; the two oracles agree on 94.0\% of responses for both answerer models (Table~\ref{app_tab:clinguide_cross_oracle}).

\begin{table}[htp]
\centering
\small
\caption{Representative 20Q cross-oracle disagreements. Oracle responses are listed in the order Qwen3-30B-A3B-Instruct-FP8 / GPT-5-mini / GPT-5.4; the human label comes from an independent annotator. Disagreements concentrate on category boundaries rather than reversals between substantive answers.}
\label{app_tab:twentyq_cross_oracle_examples}
\begin{tabular}{llcc}
\toprule
\textbf{Target} & \textbf{Question} & \textbf{Oracle responses} & \textbf{Human} \\
\midrule
T-Rex & Is T-Rex a living thing? & no / no / yes & yes \\
Christmas tree & Is Christmas tree a living thing? & no / pass / pass & pass \\
Mount Rushmore & Is Mount Rushmore man-made? & yes / pass / pass & pass \\
\bottomrule
\end{tabular}
\end{table}

\begin{table}[htp]
\centering
\small
\caption{Cross-oracle agreement in ClinGuide-MT: percentage of identical inputs on which the two oracle choices (GPT-5-mini and GPT-5.4) return the same response, with confidence intervals in brackets.}
\label{app_tab:clinguide_cross_oracle}
\begin{tabular}{lc}
\toprule
\textbf{Answerer} & \textbf{GPT-5-mini vs.\ GPT-5.4 agreement (\%)} \\
\midrule
Qwen3-30B-A3B-Thinking-FP8 & 94.0 [92.4, 96.4] \\
GPT-5 & 94.0 [93.1, 95.9] \\
\bottomrule
\end{tabular}
\end{table}

\paragraph{Human validation.}
For 20Q, a graduate-level annotator independently labelled 300 questions, evenly split between \textsc{Common} and \textsc{Thing}, without seeing the oracle responses. Human--oracle agreement is high for all three oracle models (Table~\ref{app_tab:twentyq_human_agreement}), and disagreements again concentrate on category boundaries or partially applicable questions. For ClinGuide-MT, we manually audited 200 oracle responses and labelled each as correct, a clear mismatch, or borderline, where borderline means that the model's question has meaningful clinical overlap with a desired question, so that either matching it or returning ``Not sure'' can be reasonable (Table~\ref{app_tab:clinguide_human_audit}).

\begin{table}[htp]
\centering
\small
\caption{Human--oracle agreement on 20Q: fraction of 300 independently annotated questions (150 per dataset) on which the human label matches each oracle's response.}
\label{app_tab:twentyq_human_agreement}
\begin{tabular}{lccc}
\toprule
\textbf{Dataset} & \textbf{Qwen3-30B-A3B-Instruct-FP8} & \textbf{GPT-5-mini} & \textbf{GPT-5.4} \\
\midrule
\textsc{Common} & 0.907 & 0.953 & 0.960 \\
\textsc{Thing} & 0.933 & 0.940 & 0.960 \\
\bottomrule
\end{tabular}
\end{table}

\begin{table}[htp]
\centering
\small
\caption{Manual audit of 200 ClinGuide-MT oracle responses. Borderline denotes questions with meaningful clinical overlap with a desired question, where either matching or returning ``Not sure'' can be reasonable.}
\label{app_tab:clinguide_human_audit}
\begin{tabular}{lccc}
\toprule
\textbf{Answerer} & \textbf{Correct (\%)} & \textbf{Clear mismatch (\%)} & \textbf{Borderline (\%)} \\
\midrule
GPT-5 & 91 & 5 & 4 \\
Qwen3-30B-A3B-Thinking-FP8 & 85 & 9 & 6 \\
\bottomrule
\end{tabular}
\end{table}

\paragraph{Impact on task outcomes.}
Finally, we assess whether residual oracle error could affect the reported conclusions. For 20Q, we audited the 1,669 turns for which two independent reference models (Claude-Opus-4.8 and GPT-5.6-sol) unanimously assigned a definite Yes/No label. GPT-5-mini agreed with this reference panel on 99.9\% of the audited turns, corresponding to an error rate of 0.1\%, and 110 of 111 audited episodes contained no identifiable oracle error, whereas the guesser failed on 72\% of episodes; oracle errors therefore cannot explain the observed 20Q failure rate or the main comparative conclusions. For ClinGuide-MT, the manual audit identified 7\% clear mismatches and 5\% borderline matches pooled across the two answerer models. The dominant error mode is overmatching a related but non-equivalent question, such as mapping a query about hair-loss type to a duration-specific question, while borderline cases involve meaningful clinical overlap without being clear paraphrases; Table~\ref{app_tab:clinguide_mismatch_examples} shows representative cases.

\begin{table}[htp]
\centering
\scriptsize
\setlength{\tabcolsep}{3pt}
\caption{Representative ClinGuide-MT oracle errors from the manual audit of 200 responses.}
\label{app_tab:clinguide_mismatch_examples}
\begin{tabular}{lll}
\toprule
\textbf{Model question} & \textbf{Oracle-matched question} & \textbf{Human verdict} \\
\midrule
Hair-loss type & Hair-loss duration under one year & Mismatch: type $\neq$ duration \\
Ataxia with falling & Cause of gait abnormality & Mismatch: symptom $\neq$ mechanism \\
Specific question about mania & Broader composite psychiatric screen & Borderline: partial overlap, not a clear paraphrase \\
\bottomrule
\end{tabular}
\end{table}

\clearpage

\section{Dataset construction details}
\label{appendix:benchmark_construction}

Dataset statistics can be found in Table~\ref{tab:dataset_stats}.

\begin{table}[htp]
\centering
\caption{
Dataset statistics by degree of underspecification. 
Columns under Problems and Tasks report the total count and the count for each $k$; ``--'' denotes not applicable.
}
\resizebox{\textwidth}{!}{
\begin{tabular}{lrrrrrrrrrrl}
\toprule
\multirow{2}{*}{Dataset} 
& \multicolumn{5}{c}{\# Problems} 
& \multicolumn{5}{c}{\# Tasks} 
& \multirow{2}{*}{Source} \\
\cmidrule(lr){2-6} \cmidrule(lr){7-11}
& Total & $k{=}1$ & $k{=}2$ & $k{=}3$ & $k{=}4$
& Total & $k{=}1$ & $k{=}2$ & $k{=}3$ & $k{=}4$
& \\
\midrule
Logic-Q-MT     & 600  & 200 & 200 & 200 & 0 & 1200 & 400 & 400 & 400 & 0 & \cite{Zhang2023simplelogic} \\
GeneReg-MT     & 800  & -- & -- & -- & -- & 2326 & 451 & 626 & 729 & 520 & 38 GRNs~\cite{Kadelka2024grn} \\
GSME-Q-MT      & 2956 & 1067 & 884 & 709 & 296 & 2956 & 1067 & 884 & 709 & 296 & \cite{li2025questbench} \\
GSME-Q-MT-Ext  & 524  & 134 & 144 & 123 & 123 & 524  & 134 & 144 & 123 & 123 & Synthetic \\
ClinGuide-MT   & 370  & -- & -- & -- & -- & 1595 & 521 & 477 & 370 & 227 & 59 algorithms~\citep{patient_history, scott2014symptom} \\
20Q            & 1    & -- & -- & -- & -- & 405  & -- & -- & -- & -- & \cite{uot} \\
\bottomrule
\end{tabular}
\label{tab:dataset_stats}
}
\end{table}

\subsection{Logic-Q-MT}
\label{appendix:logicq_multi}

In the 1-underspecified problem construction pipeline~\cite{li2025questbench}, each pair of true-target and false-target assignments $(\mathcal A_i^{(y)}, \mathcal A_j^{(\neg y)})$ is examined, and pairs differing only in the assignment of one variable $x_d$ are collected for further validation: after removing $x_d$, the shared context $(\mathcal A_i^{(y)}\setminus x_d) \wedge (\mathcal A_j^{(\neg y)}\setminus x_d)$ should not already determine the target value. However, when extending to larger $k$, merely collecting the variables whose assignments differ between $(\mathcal A_i^{(y)}, \mathcal A_j^{(\neg y)})$ is neither sufficient nor necessary for identifying minimal $k$-sufficient sets. For example, XOR/XNOR-style dependencies at $k=2$ cannot be captured by a single pairwise flip criterion. We therefore design a recursive construction algorithm for candidate $k$-sufficient CSPs (Algorithm~\ref{alg:recursive_logic_q_multi}), followed by a feasibility-aware validator (Algorithm~\ref{alg:final_validation_logic_q_multi}).

For Logic-Q-MT, the SimpleLogic rules are represented as Horn CNF clauses: a rule
\[
(x_1\wedge \cdots \wedge x_m)\Rightarrow y
\]
is stored as
\[
(\neg x_1 \vee \cdots \vee \neg x_m \vee y).
\]
Thus, feasibility and entailment can be decided by Horn-SAT. In our implementation, we use unit propagation with empty-clause detection to test feasibility, and use refutation to test whether a target value is forced.

Concretely, for a partial assignment $\psi$, define
\[
\mathsf{Force}_{\mathcal C}(\psi,Y)=
\begin{cases}
\bot, & \mathcal C\wedge \psi \text{ is infeasible},\\
1, & \mathcal C\wedge \psi \models Y,\\
0, & \mathcal C\wedge \psi \models \neg Y,\\
?, & \text{otherwise}.
\end{cases}
\]
For Horn CNF, this is computed by unit propagation: feasibility is checked by whether unit propagation derives an empty clause; $Y$ is forced true iff $\mathcal C\wedge\psi\wedge \neg Y$ is infeasible, and forced false iff $\mathcal C\wedge\psi\wedge Y$ is infeasible.

Let $\mathsf{Suff}(\mathcal A,S,Y)$ denote the predicate that $S$ is sufficient to determine $Y$ under context $\mathcal A$:
\[
\mathsf{Suff}(\mathcal A,S,Y)
\iff
\exists s \text{ feasible for } S
\;\;\text{and}\;\;
\forall s \text{ feasible for } S,\ 
\mathsf{Force}_{\mathcal C}(\mathcal A\wedge (S=s),Y)\in\{0,1\}.
\]

From a complete SimpleLogic problem, we first obtain candidate minimal $1$-sufficient sets using an improved algorithm based on~\citep{li2025questbench}. Assume that at step $k-1$ we have obtained candidate minimal $(k-1)$-sufficient sets $(\mathcal A_{k-1},\mathcal S_{k-1})$. For each assigned literal $\ell_k=(x_k=a_k)$ in $\mathcal A_{k-1}$, we remove it from the context and add its variable to the query set:
\[
\mathcal A_k = \mathcal A_{k-1}\setminus \{\ell_k\},\qquad
\mathcal S_k = \mathcal S_{k-1}\cup\{x_k\}.
\]
Since $\mathcal A_{k-1}=\mathcal A_k\wedge (x_k=a_k)$, sufficiency of $\mathcal S_{k-1}$ under $\mathcal A_{k-1}$ already verifies the branch $x_k=a_k$. We therefore additionally check the flipped branch:
\[
\mathsf{Suff}\!\left(
\mathcal A_k\wedge (x_k=\mathrm{flip}(a_k)),
\mathcal S_{k-1},Y
\right).
\]
This yields a candidate $k$-sufficient set. We then apply global minimality and feasibility-aware validation over the final query space.

\begin{algorithm}[htp]
  \caption{Recursive construction of candidate $k$-sufficient sets}
  \label{alg:recursive_logic_q_multi}
  \begin{algorithmic}
    \STATE {\bfseries Input:} Horn CNF constraints $\mathcal C$; target $Y$;
    forcing oracle $\mathsf{Force}_{\mathcal C}$; candidate minimal
    $(k{-}1)$-sufficient sets $\mathcal M_{k-1}$.
    \STATE {\bfseries Output:} Candidate $k$-sufficient sets $\mathcal M_k$.

    \FORALL{$(\mathcal A_{k-1},\mathcal S_{k-1}) \in \mathcal M_{k-1}$}
      \FORALL{assigned literals $\ell_k=(x_k=a_k) \in \mathcal A_{k-1}$}
        \STATE $\mathcal A_k \gets \mathcal A_{k-1}\setminus\{\ell_k\}$
        \STATE $\mathcal S_k \gets \mathcal S_{k-1}\cup\{x_k\}$

        \STATE \COMMENT{Non-triviality: the reduced context alone should not determine $Y$.}
        \IF{$\mathsf{Force}_{\mathcal C}(\mathcal A_k,Y)\in\{0,1\}$}
          \STATE \textbf{continue}
        \ENDIF

        \STATE \COMMENT{The branch $x_k=a_k$ is inherited from the $(k-1)$-sufficient item.}
        \STATE \COMMENT{Check the flipped branch.}
        \IF{$\neg\mathsf{Suff}\!\left(\mathcal A_k\wedge(x_k=\mathrm{flip}(a_k)),\mathcal S_{k-1},Y\right)$}
          \STATE \textbf{continue}
        \ENDIF

        \STATE \COMMENT{Efficient candidate-level minimality check.}
        \STATE $ok \gets \texttt{true}$
        \FORALL{$\mathcal U \subseteq \mathcal U(P(\mathcal C, \mathcal A_k))$ with $|\mathcal U|=k-1$}
          \IF{$\mathsf{Suff}(\mathcal A_k,\mathcal U,Y)$}
            \STATE $ok \gets \texttt{false}$
            \STATE \textbf{break}
          \ENDIF
        \ENDFOR

        \IF{$ok$}
          \STATE $\mathcal M_k \gets \mathcal M_k \cup \{(\mathcal A_k,\mathcal S_k)\}$
        \ENDIF
      \ENDFOR
    \ENDFOR
  \end{algorithmic}

  \vspace{0.3em}
  {\footnotesize
  This recursive procedure is an efficient candidate generator. Because infeasible assignments and alternative sufficient sets can create false positives, we apply the feasibility-aware validator in Algorithm~\ref{alg:final_validation_logic_q_multi}.}
\end{algorithm}

\begin{algorithm}[htp]
  \caption{Feasibility-aware validation for Logic-Q-MT}
  \label{alg:final_validation_logic_q_multi}
  \begin{algorithmic}
    \STATE {\bfseries Input:} CSP $P$ with Horn CNF constraints $\mathcal C$ and context $\mathcal A_k$;
    candidate set $\mathcal S_k=\{x_1,\dots,x_k\}$; target $Y$;
    askable universe $\mathcal U(P)$; forcing oracle $\mathsf{Force}_{\mathcal C}$.
    \STATE {\bfseries Output:} \texttt{accept}/\texttt{reject}.

    \STATE $\mathcal T \gets \emptyset$ \COMMENT{A partial truth table over feasible assignments to $\mathcal S_k$.}

    \STATE \COMMENT{Step 1: Feasibility-aware sufficiency.}
    \FORALL{$a \in \{0,1\}^k$}
      \STATE $\phi_a \gets \bigwedge_{i=1}^k (x_i=a_i)$
      \STATE $z \gets \mathsf{Force}_{\mathcal C}(\mathcal A_k\wedge\phi_a,Y)$

      \IF{$z=\bot$}
        \STATE \textbf{continue} \COMMENT{Discard infeasible assignment.}
      \ELSIF{$z=?$}
        \STATE \textbf{return} \texttt{reject} \COMMENT{$\mathcal S_k$ is not sufficient.}
      \ELSE
        \STATE $\mathcal T[a]\gets z$
      \ENDIF
    \ENDFOR

    \IF{$\mathcal T=\emptyset$}
      \STATE \textbf{return} \texttt{reject}
    \ENDIF

    \STATE \COMMENT{Step 2: Essentiality of each variable in $\mathcal S_k$.}
    \FOR{$j=1$ {\bfseries to} $k$}
      \STATE $\texttt{essential}\gets\texttt{false}$
      \FORALL{$b\in\{0,1\}^{k-1}$ assignments to $\mathcal S_k\setminus\{x_j\}$}
        \STATE Let $a^{(0)}=(x_j=0,b)$ and $a^{(1)}=(x_j=1,b)$.
        \IF{$a^{(0)}\in\mathrm{dom}(\mathcal T)$ \AND
             $a^{(1)}\in\mathrm{dom}(\mathcal T)$ \AND
             $\mathcal T[a^{(0)}]\neq \mathcal T[a^{(1)}]$}
          \STATE $\texttt{essential}\gets\texttt{true}$
          \STATE \textbf{break}
        \ENDIF
      \ENDFOR
      \IF{$\texttt{essential}=\texttt{false}$}
        \STATE \textbf{return} \texttt{reject}
      \ENDIF
    \ENDFOR

    \STATE \COMMENT{Step 3: Global minimality against alternative query sets.}
    \FORALL{$\mathcal U\subseteq \mathcal U(P)$ with $|\mathcal U|=k-1$}
      \IF{$\mathsf{Suff}(\mathcal A_k,\mathcal U,Y)$}
        \STATE \textbf{return} \texttt{reject}
      \ENDIF
    \ENDFOR

    \STATE \textbf{return} \texttt{accept}
  \end{algorithmic}

  \vspace{0.3em}
  {\footnotesize
  For Logic-Q-MT, $\mathsf{Force}_{\mathcal C}$ is implemented by Horn-SAT:
  unit propagation detects infeasible branches, and the entailment of $Y$ or $\neg Y$ is checked by refutation. Since sufficiency is monotone in the queried variable set, checking all $(k-1)$-subsets rules out all smaller sufficient sets as well.}
\end{algorithm}

\paragraph{Brute-force essentiality check.} 
We further clarify the need of the essentiality validation.
Consider $\mathcal S_2=\{x_1,x_2\}$ and suppose the feasible assignments over $(x_1,x_2,y)$ are exactly
\[
\Omega=\{(0,0,0),(1,1,1),(1,0,1)\},
\]
where the missing combination $(0,1)$ is infeasible due to contradictions with the constraints.
Then $y$ does \emph{not} depend essentially on $x_2$: there is no pair of feasible assignments that agrees on $x_1$
but flips $x_2$ and changes $y$.
However, a recursive subset-based check can fail to expose this issue when it never explicitly conditions on the infeasible combination involving the removed variable. Such a procedure may incorrectly treat $\mathcal S_2$ as minimal. This motivates our feasibility-aware essentiality validation, which enumerates all assignments to $\mathcal S_k$, removes infeasible rows, and checks essentiality directly on the remaining table.

\paragraph{Recursive coverage is not guaranteed over non-Cartesian feasible domains.}
A general complete coverage using the recursive algorithm is not guaranteed once we allow arbitrary CSP rules (constraints) over the variables (i.e., the feasible set is an arbitrary relation rather than a full Cartesian product). In particular, classical results in the spirit of Salomaa apply to total functions over Cartesian domains, but need not extend to arbitrary constraint relations~\footnote{See, statements that for an $n$-ary function $f:A^n\to B$ depending on all variables, there exists a coordinate whose fixing preserves dependence on the remaining $n-1$ variables; this relies on the full Cartesian domain assumption~\cite{salomaa1964essential} (restated in a convenient form by Couceiro and Lehtonen~\cite{couceiro2007effect}).}.

As a counterexample, consider the following $k=4$ case. Let $a,b,c,d,y\in\{0,1\}$ and $\mathcal A=\top$. Define the CSP extensionally by allowing exactly the following nine assignments (ordered as $(a,b,c,d,y)$):

\begin{align*}
\Omega \;=\;\{
&(0,0,0,0,0),
(0,0,0,1,0),
(0,0,1,0,1),
(0,1,0,0,1),
(1,0,0,1,1),\\
&(1,0,1,0,1),
(1,0,1,1,1),
(1,1,1,0,0),
(1,1,1,1,1)
\}.    
\end{align*}

\emph{(i) $\{a,b,c,d\}$ is globally minimal $4$-sufficient for $y$ under $\mathcal A$.}
First, $y$ is not known under $\mathcal A=\top$ since $\Omega$ contains both $y=0$ and $y=1$ solutions. 
Second, knowing all of $(a,b,c,d)$ determines $y$ because each feasible $(a,b,c,d)$ pattern occurs with a unique $y$ in $\Omega$. Finally, no 3-subset suffices: for each 3-subset there exist two feasible assignments agreeing on those three variables but disagreeing on $y$, e.g.,
\begin{align*}
\{a,b,c\}:~&(1,1,1,0)\mapsto y=0 \;\text{ vs. }\; (1,1,1,1)\mapsto y=1,\\
\{a,b,d\}:~&(0,0,1,0)\mapsto y=1 \;\text{ vs. }\; (0,0,0,0)\mapsto y=0,\\
\{a,c,d\}:~&(0,1,0,0)\mapsto y=1 \;\text{ vs. }\; (0,0,0,0)\mapsto y=0,\\
\{b,c,d\}:~&(0,0,0,1)\mapsto y=0 \;\text{ vs. }\; (1,0,0,1)\mapsto y=1.
\end{align*}
Hence the minimum sufficient size under $\mathcal A$ is $4$, and $\{a,b,c,d\}$ is globally minimal.

\emph{(ii) No single assignment reduces it to a globally minimal $3$-sufficient problem.}
Conditioning on any single variable assignment collapses the instance so that the minimum sufficient size drops below $3$:
\begin{align*}
&\min\text{-size}(\mathcal A\wedge a=0)=2,\;
\min\text{-size}(\mathcal A\wedge a=1)=2,\; \\
&\min\text{-size}(\mathcal A\wedge b=0)=2,\;
\min\text{-size}(\mathcal A\wedge b=1)=2, \\
&\min\text{-size}(\mathcal A\wedge c=0)=2,\;
\min\text{-size}(\mathcal A\wedge c=1)=2,\; \\
&\min\text{-size}(\mathcal A\wedge d=0)=2,\;
\min\text{-size}(\mathcal A\wedge d=1)=1.
\end{align*}

Therefore, although the original instance is globally minimal $4$-sufficient, there exists no single variable/value assignment that produces a context whose globally minimal sufficient size is $3$. This blocks complete coverage by a recursive $k=3\to4$ construction.

\subsection{GSME-Q-MT and GSME-Q-MT-Ext}
\label{appendix:gsme_multi}

This section describes the construction of our GSME-based multi-turn information-seeking datasets. 
We first extend the original GSME-Q benchmark into a multi-turn setting, denoted as GSME-Q-MT, and then introduce GSME-Q-MT-Ext, an enriched variant generated from DAG-structured arithmetic programs. 

\begin{figure}[htp]
    \centering
    \includegraphics[width=1.0\linewidth]{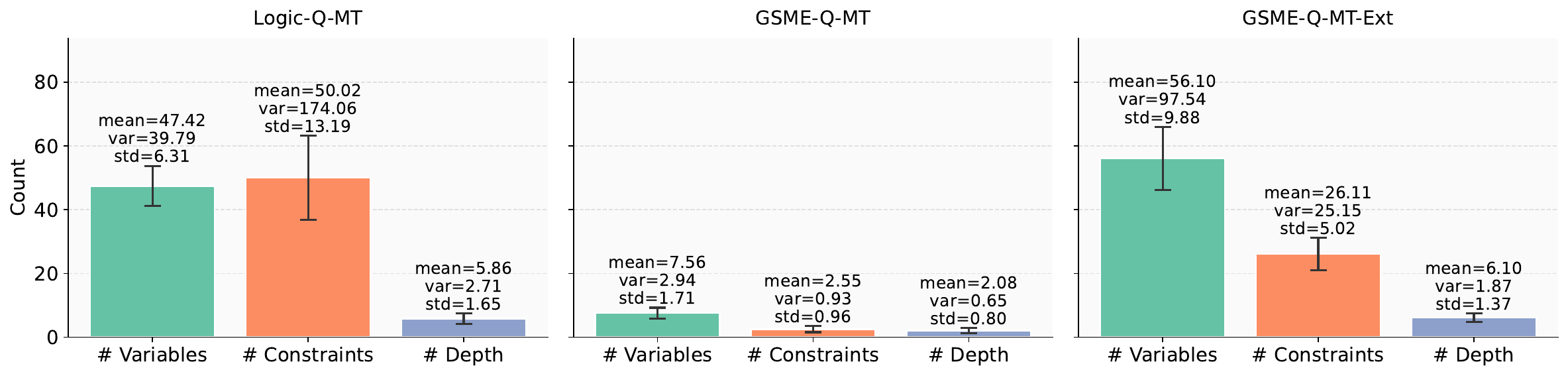}
    \caption{Complexity Across Task Settings.}
    \label{app_fig:gsme_statistics}
\end{figure}

\subsubsection{GSME-Q-MT construction}

GSME-Q-MT is derived from the original GSME-Q equation systems. 
For each problem, we identify base variables that are not defined by other equations, mask a subset of them, and validate whether the masked subset forms a minimal \(k\)-sufficient set for determining the target (Algorithm~\ref{alg:ksuff_gsme_mt}). 
This yields controlled multi-turn tasks where the model must ask for multiple missing quantities before solving the problem.

\paragraph{Base-variable pool.}
For each original GSME-Q instance, we parse the equation system and identify the target variable \(y\), the arithmetic constraints \(\mathcal{C}\), and the set of leaf or base variables. 
A base variable is a variable that is not produced as the output of any equation. 
These variables correspond to primitive quantities that can be directly provided in the problem statement. 
We exclude dataset-provided distractors from the base-variable pool used for constructing true missing conditions. 
We also require each resulting instance to retain a minimum number of revealed base variables, preventing degenerate contexts where the problem contains too little visible information.

\paragraph{Dependency graph and relevant leaves.}
From the equation system, we construct a directed dependency graph. 
An edge points from an input variable to an output variable whenever the input is used in an arithmetic constraint defining the output. 
We compute the ancestor set of the target variable, denoted \(\mathrm{Anc}(y)\). 
Let \(\mathcal{L}\) be the set of leaf variables. 
The goal-relevant leaf set is
\[
\mathcal{L}_{\mathrm{rel}}
=
\mathcal{L}\cap \mathrm{Anc}(y).
\]
Only variables in \(\mathcal{L}_{\mathrm{rel}}\) are considered as true missing variables. 
This avoids treating structurally irrelevant distractors as necessary information.

\paragraph{Minimal \(k\)-sufficient set construction.}
For a target value of \(k\), we sample candidate held-out sets \(S_k\subseteq \mathcal{L}_{\mathrm{rel}}\) with \(|S_k|=k\). 
For each candidate \(S_k\), we reveal all relevant leaves outside \(S_k\):
\[
\mathcal{A}_k
=
\left\{
x=v_x: x\in \mathcal{L}_{\mathrm{rel}}\setminus S_k
\right\}.
\]
We retain \(S_k\) only if it satisfies the following three conditions:
\[
(1) \ \neg \mathrm{Known}(y \mid \mathcal{A}_k), 
\]
\[
(2) \ \mathrm{Known}(y \mid \mathcal{A}_k \cup \{x=v_x: x\in S_k\}), 
\]
\[
(3) \ \forall S'\subset S_k,
\neg \mathrm{Known}(y \mid \mathcal{A}_k \cup \{x=v_x: x\in S'\}).
\]
The first condition enforces underspecification: the visible information alone is insufficient. 
The second condition enforces sufficiency: revealing the full held-out set makes the target uniquely determined. 
The third condition enforces global minimality, or strict necessity: no smaller subset of the held-out variables is enough.

\paragraph{Candidate query set.}
Each instance is accompanied by a candidate set of queryable variables. 
This set contains the true missing variables \(S_k\), dataset-provided distractors when available, and additional decoys. 
The decoys are selected to discourage superficial strategies. 
They may include variables that are derivable from already revealed information, variables that are structurally irrelevant to the target, or plausible quantities that appear in the problem but are not necessary for determining \(y\). 
Thus, the model must distinguish necessary information from irrelevant or redundant information.

\paragraph{Difficulty ranking.}
A single original problem may induce multiple valid held-out sets. 
When more than a fixed number of valid sets are found, we rank them by a difficulty score and keep the top instances. 
The score can incorporate structural and ambiguity-based factors, such as the maximum distance from held-out leaves to the target and the number of possible target values under the visible information. 
This avoids overrepresenting trivially easy masks.

\begin{algorithm}[th]
\caption{Constructing Minimal \(k\)-Sufficient GSME-Q-MT}
\label{alg:ksuff_gsme_mt}
\begin{algorithmic}[1]
\REQUIRE Source GSME problems \(\mathcal{T}\); maximum degree \(k_{\max}\); candidate budget \(N\); keep-top \(M\)
\ENSURE Generated dataset \(\mathcal{T}'\)

\STATE \(\mathcal{T}' \leftarrow \emptyset\)

\FOR{each problem instance \(p \in \mathcal{T}\)}
  \STATE Parse variables \(\mathcal X\), domains \(\mathcal D\), constraints \(\mathcal C\), full assignment \(\omega^\star\), and target variable \(Y\)
  \STATE Construct the dependency graph induced by \(\mathcal C\)
  \STATE Identify queryable variables relevant to \(Y\): \(\mathcal U \leftarrow \mathrm{Leaves}(\mathcal X) \cap \mathrm{Anc}(Y)\)

  \FOR{\(k=1\) to \(k_{\max}\)}
    \IF{\(k > |\mathcal U|\)}
      \STATE \textbf{continue}
    \ENDIF

    \STATE Sample up to \(N\) candidate sets \(S \subseteq \mathcal U\) with \(|S|=k\)
    \STATE \(\mathcal V_k \leftarrow \emptyset\)

    \FOR{each candidate set \(S\)}
      \STATE Define observed assignment
      \[
      \mathcal A_S \leftarrow \{X=\omega^\star(X): X\in \mathcal U\setminus S\}
      \]

      \IF{\(\mathrm{Known}(Y \mid \mathcal A_S)\)}
        \STATE \textbf{continue} \COMMENT{\(Y\) is already determined}
      \ENDIF

      \IF{there exists an assignment \(s\) to \(S\) consistent with \((\mathcal C,\mathcal A_S)\) such that \(\neg \mathrm{Known}(Y \mid \mathcal A_S \wedge s)\)}
        \STATE \textbf{continue} \COMMENT{\(S\) is not sufficient}
      \ENDIF

      \IF{there exists \(S' \subset S\) such that, for every consistent assignment \(s'\) to \(S'\), \(\mathrm{Known}(Y \mid \mathcal A_S \wedge s')\)}
        \STATE \textbf{continue} \COMMENT{\(S\) is not minimal}
      \ENDIF

      \STATE \(\mathcal V_k \leftarrow \mathcal V_k \cup \{S\}\)
    \ENDFOR

    \STATE Rank \(\mathcal V_k\) by difficulty and keep the top \(M\) sets

    \FOR{each kept set \(S \in \mathcal V_k\)}
      \STATE \(y^\star \leftarrow Y(\omega^\star)\)
      \STATE Rewrite \(p\) by masking variables in \(S\)
      \STATE Construct the candidate query set from \(\mathcal U\)
      \STATE Emit sample
      \[
      \tau = (P_S, \mathcal A_S, S, Y, y^\star, \mathcal U)
      \]
      \STATE \(\mathcal T' \leftarrow \mathcal T' \cup \{\tau\}\)
    \ENDFOR
  \ENDFOR
\ENDFOR

\STATE \textbf{return} \(\mathcal T'\)
\end{algorithmic}
\end{algorithm}

\subsubsection{GSME-Q-MT-Ext construction}

Although GSME-Q-MT extends GSME-Q to multiple missing variables, the resulting problems remain limited by the structure of the original dataset. 
Many instances contain few variables, short dependency chains, and weak interactions among missing quantities. 
As a result, identifying the missing variables is often easy once the model parses the equation structure. 
To create a more discriminative testbed, we construct GSME-Q-MT-Ext (Algorithm~\ref{alg:complex_gsme_generation}).

\paragraph{Latent program representation.}
Each GSME-Q-MT-Ext instance is generated from a directed acyclic graph $G=(V,E)$,
where nodes \(V\) correspond to variables and edges \(E\) encode arithmetic dependencies. 
Leaf nodes represent primitive quantities. 
Internal nodes represent derived quantities defined by arithmetic rules. 
One node \(y\in V\) is designated as the goal variable. 
The graph is treated as the canonical latent structure of the problem, while the natural-language or symbolic problem statement is a surface rendering of this latent program.
This representation gives direct control over structural properties, including the number of variables, the number of constraints, reasoning depth, branch-merge patterns, and the separation between goal-relevant and distractor variables.

\paragraph{Controlled structural complexity.}
GSME-Q-MT-Ext controls complexity along four main axes.
Depth is defined as the length of the longest directed path from a goal-relevant leaf variable to the target variable. 
To ensure that depth reflects actual reasoning requirements, we explicitly construct a goal-bearing backbone chain rather than merely adding disconnected equations.
Second, we introduce \emph{branching and merge structures}. 
A merge node is an internal variable whose value depends on multiple parent variables. 
Such nodes create situations where multiple upstream quantities jointly determine a downstream quantity. 
This is important for multi-turn evaluation because the model may need to acquire several missing leaves before the target becomes determined.
Third, we include \emph{distractor subgraphs}. 
A distractor subgraph is a valid arithmetic component that appears in the instance but is not an ancestor of the target. 
Distractor subgraphs increase structural clutter without changing the target value. 
They test whether the model can separate relevant missing information from plausible but unnecessary information.
Fourth, we control the number of \emph{goal-relevant leaves}. 
This is especially important for constructing large-\(k\) instances, because a problem must contain enough relevant primitive quantities to support a minimal \(k\)-sufficient held-out set.

\paragraph{Generation pipeline.}
The GSME-Q-MT-Ext generation pipeline consists of five stages.

\emph{Stage 1: sample structural targets.}
For each instance, we sample or specify target structural parameters, including the number of variables, the number of arithmetic rules, the target reasoning depth, the number of goal-relevant leaves, and the amount of distractor structure.

\emph{Stage 2: build the goal-relevant subgraph.}
We first construct a backbone chain that guarantees the target reasoning depth. 
We then attach additional branches and merge nodes to create richer dependency patterns. 
The construction is biased toward the requested number of relevant leaves, ensuring that the instance can support the desired missing-information level \(k\).

\emph{Stage 3: add distractor subgraphs.}
We optionally add disconnected arithmetic components that do not affect the target variable. 
These components are valid and internally consistent, but none of their variables is an ancestor of \(y\). 
They are included in the full problem representation and the query candidate pool, making the information-seeking task less reducible to surface matching.

\emph{Stage 4: assign values by forward execution.}
We sample values for leaf variables and compute all internal node values by executing the DAG in topological order. 
Whenever division is used, we enforce divisibility constraints so that all generated values remain integer-friendly.

\emph{Stage 5: validate and filter.}
Each generated instance is validated before being retained. 
We check that the graph is acyclic, that all arithmetic rules are numerically consistent, that the requested structural targets are satisfied, that the target has a valid ancestry structure, and that the target is solvable from full information. 
We then construct and validate minimal \(k\)-sufficient held-out sets following the same criterion used in GSME-Q-MT.

\begin{algorithm}[th]
\caption{GSME-Q-MT-Ext generation}
\label{alg:complex_gsme_generation}
\begin{algorithmic}[1]
\REQUIRE Number of samples \(B\); structural parameter ranges \(\Theta\); maximum missing set size \(k_{\max}\); candidate budget \(N\); keep-top \(M\)
\ENSURE Generated dataset \(\mathcal{T}_{\mathrm{Ext}}\)

\STATE \(\mathcal{T}_{\mathrm{Ext}} \leftarrow \emptyset\)

\WHILE{\(|\mathcal{T}_{\mathrm{Ext}}| < B\)}
  \STATE Sample structural targets \(\theta \sim \Theta\)
  \STATE Build a goal-bearing backbone chain with target depth specified by \(\theta\)
  \STATE Add branch and merge nodes to increase structural entanglement
  \STATE Add optional distractor subgraphs disconnected from the target
  \STATE Let \(G=(V,E)\) be the resulting graph and choose target node \(y\)
  \STATE Sample leaf values and compute internal values by topological forward execution
  \IF{\(G\) is not acyclic or any arithmetic constraint is invalid}
    \STATE \textbf{continue}
  \ENDIF
  \IF{\(G\) does not satisfy the requested structural targets}
    \STATE \textbf{continue}
  \ENDIF

  \STATE Compute all leaves \(\mathcal{L}_{\mathrm{all}}\)
  \STATE Compute relevant leaves \(\mathcal{L}_{\mathrm{rel}} = \mathcal{L}_{\mathrm{all}}\cap \mathrm{Anc}(y)\)
  \STATE Construct valid held-out sets

  \IF{no valid held-out set is found}
    \STATE \textbf{continue}
  \ENDIF

  \STATE Render the full symbolic problem, rewritten partial-information problem, candidate query set, and supervision fields
  \STATE Add generated samples to \(\mathcal{T}_{\mathrm{Ext}}\)
\ENDWHILE

\STATE \textbf{return} \(\mathcal{T}_{\mathrm{Ext}}\)
\end{algorithmic}
\end{algorithm}

\subsection{GeneReg-MT}
\label{appendix:genereg_multi}

Gene regulation is fundamental to cellular biological systems. Schematics of gene regulation are usually represented by a network, called Gene Regulatory Networks (GRNs), where nodes represent genes and edges represent how activations of genes affect downstream gene expression. GRNs are often modeled as Boolean networks, which are simple but can yield accurate quantitative results with limited data. Each of the Boolean networks will converge onto a finite set of attractor states, including fixed points (steady states) and limit cycles. 
These networks provide a testbed to study LLM's information seeking ability in the natural science domain. In particular, we curate the dataset from~\citep{Kadelka2024grn}, which contains 122 Boolean GRNs. We further filter the dataset such that there are 6 to 30 nodes in the graph. 
For each GRN, based on the number of its fixed points, we consider two goals: (1) Steady-state identification (steady state): Infer which one steady state a partial initial gene expression value assignment will lead to, provided all steady states for this GRN. (2) Steady gene expression identification (marker): Infer which one gene expression value of a specific marker gene a partial initial gene expression value assignment will lead to, provided all steady states for this GRN. 
A CSP is formalized in this dataset as:
\begin{itemize}
    \item $\mathcal{X}$: Genes in the GRN (values spanning the full Cartesian space)
    \item $\mathcal D$: $\{\{\mathrm{Active}, \mathrm{Inactive} \} \forall x \in \mathcal X\}$ 
    \item $\mathcal{A}$: Partial observation of the initial gene vector
    \item $\mathcal C$: Boolean update rules
    \item $y$: The unique steady state, or a marker gene value at convergence. 
    \item $\mathcal S$: Some additional gene expressions
\end{itemize}

To create $k$-underspecified tasks for the former goal, we first obtain sets of gene expression value assignments that will determine the target and cannot be reduced. We then randomly drop $k$ values and brute-force verify whether they form a minimal $k$-sufficient set. For the latter goal, we exploit its Cartesian variable space and recursively construct $k=1,2,3,4$-sufficient problems, similar to Logic-Q-MT. In total, we construct 800 problems from 38 GRNs, which induce 2,326 tasks.

\textbf{Example: Gene regulatory network as a CSP.}
Consider a Boolean gene regulatory network with variables
\[
\mathcal X=\{I,S,M_{34},Z,M_{200}\},
\]
where \(I\) denotes the external input, \(S\) denotes \textit{Snail}, \(M_{34}\) denotes \(miR\text{-}34\), \(Z\) denotes \textit{Zeb}, and \(M_{200}\) denotes \(miR\text{-}200\). 
Each variable is Boolean, so for every \(X_i\in\mathcal X\),
\[
\mathcal D_i=\{\mathrm{Active},\mathrm{Inactive}\},
\]
or equivalently \(\mathcal D_i=\{1,0\}\).

The constraints \(\mathcal C\) encode the steady-state conditions of the Boolean update rules:
\[
\begin{aligned}
S &\leftarrow I \wedge \neg(S \wedge M_{34}),\\
M_{34} &\leftarrow \neg S \wedge \neg Z,\\
Z &\leftarrow \neg M_{200} \vee (S \wedge Z),\\
M_{200} &\leftarrow \neg S \vee (Z \wedge M_{200}).
\end{aligned}
\]
Here each constraint requires the gene value at convergence to agree with its Boolean regulatory rule.

The partial observation \(\mathcal A\) specifies known values of some variables, such as an observed or intervened input condition. For example,
\[
\mathcal A=\{I=\mathrm{Active}\}
\]
restricts the feasible assignments to steady states consistent with an active input. The feasible set is
\[
\Omega(P)=
\left\{
x\in \prod_{X_i\in\mathcal X}\mathcal D_i:
x \models \mathcal C
\text{ and }
x \models \mathcal A
\right\}.
\]

The target variable \(Y\) may be the full steady-state gene expression vector,
\[
Y=(S,M_{34},Z,M_{200}),
\]
or a marker gene value at convergence, such as \(Y=Z\). Additional queryable gene expressions are represented by a set
\[
\mathcal S\subseteq \mathcal X\setminus \operatorname{dom}(\mathcal A),
\]
whose observed values can further restrict \(\Omega(P)\).

For instance, if \(\mathcal A=\{I=\mathrm{Inactive}\}\), the constraints imply a unique feasible steady state:
\[
\Omega(P)=
\left\{
(I=0,\ S=0,\ M_{34}=1,\ Z=0,\ M_{200}=1)
\right\}.
\]
Thus the full steady state \(Y=(S,M_{34},Z,M_{200})\) is known.

In contrast, if \(\mathcal A=\{I=\mathrm{Active}\}\), the feasible steady states are
\[
\Omega(P)=
\left\{
(I=1,\ S=1,\ M_{34}=0,\ Z=1,\ M_{200}=0),
(I=1,\ S=1,\ M_{34}=0,\ Z=1,\ M_{200}=1)
\right\}.
\]
The full steady-state vector is therefore not uniquely determined, because \(M_{200}\) remains ambiguous. 
However, a marker target such as \(Y=Z\) is known under \(\mathcal A\), since all feasible assignments agree that \(Z=\mathrm{Active}\).

\subsection{ClinGuide-MT}
\label{appendix:clinguide}

\subsubsection{ClinGuide-MT problem formulation and construction}

Clinical guidelines formalize expert knowledge into structured decision procedures for diagnosis and management. They are often organized around presenting symptoms and specify a sequence of conditional decisions that progressively narrow the set of plausible diagnoses, treatments, triage decisions, or follow-up recommendations. This structure makes clinical guidelines a natural source of controlled multi-turn information-seeking tasks: a model must identify which missing clinical evidence is needed, query it in an appropriate order, and stop once the final recommendation is determined.
We curate 59 structured diagnostic algorithms spanning 11 clinical areas, including cardiovascular, gastrointestinal, and musculoskeletal conditions, using established clinical textbooks and guidelines as references. Each algorithm is represented as a decision tree. Internal nodes correspond to clinical variables $X$, such as symptoms, risk factors, medication or substance use, physical-examination findings, laboratory tests, imaging results, or procedure findings. Outgoing edges correspond to possible values of the variable, such as positive or negative test results, acute versus chronic symptom duration, or alternative symptom presentations. Leaf nodes correspond to final outcomes $Y$, including diagnoses, treatments, triage decisions, or recommended follow-up examinations.
Fig.~\ref{fig:ClinGuideSchematic} shows a simplified diagnostic algorithm for pelvic pain. The pathway begins with the presenting symptom and proceeds through conditionally relevant variables, such as duration of pain, pregnancy-test result, and laparoscopy finding, before reaching a terminal recommendation. A complete root-to-leaf path therefore defines one diagnostic pathway: an ordered sequence of clinical variable--value pairs together with a final outcome.

\paragraph{Pathway validation.}
\label{app:clinguide_validation}
The pathways are built upon the diagnostic algorithms in the source textbook~\citep{patient_history}. While constructing each problem, every extracted algorithm was manually proofread against the textbook and the journal articles it cites, with particular attention to preserving decision nodes, branches, and terminal recommendations so that each branch leads to a unique outcome; one clinical expert additionally reviewed ten samples of the extracted algorithms and confirmed their fidelity to the source material.

To construct multi-turn tasks, we first normalize each diagnostic algorithm only to resolve structural ambiguity, yielding a set of unambiguous root-to-leaf paths. For each leaf node, we backtrack to the root and extract the corresponding ordered pathway. For each $k \in \{1,2,3,4\}$, whenever the path length permits, we mask the final $k$ clinical variables on the path. The unmasked prefix is converted into the observed patient context $\mathcal A$, while the masked variables define the minimal sufficient set $S$ whose values are needed to recover the final outcome $Y$. Full dataset statistics are reported in Appendix Table~\ref{tab:dataset_stats}.

Each task also includes a query space. The valid query variables are the clinical variables appearing in the corresponding decision tree. To test whether models can identify the correct variables to ask about, we add distractor variables sampled from other ClinGuide-MT problems, excluding variables that overlap with the current tree. These distractors expand the query space but do not affect the ground-truth final outcome. Each resulting task contains: (i) an observed patient context, (ii) a set of valid and distractor query variables, (iii) the masked order-dependent variables and their ground-truth values, and (iv) the final recommendation as the target outcome. Details of system prompts and oracle prompts are provided in Appendix~\ref{appendix:clinguide-prompt}.

The complete formulation of ClinGuide-MT is specified as follows: 
\begin{itemize} 
    \item $\mathcal X$: the set of clinical variables, including symptoms, risk factors, medication or substance use, laboratory tests, imaging results, procedures, and other clinical findings; 
    \item $\mathcal D$: the domains of the variables in $\mathcal X$, specifying their possible values; 
    \item $\mathcal A$: the observed partial assignment, corresponding to the known patient context provided by the unmasked pathway prefix; 
    \item $\mathcal C$: the underlying clinical decision structure encoded by the guideline-derived decision tree; 
    \item $Y$: the target outcome, corresponding to the final recommendation, such as a diagnosis, treatment, triage decision, or follow-up action; 
    \item $S$: the masked variables whose values must be queried to determine $Y$. 
\end{itemize}

At evaluation time, the model is given the observed context $\mathcal A$ and the query space $\mathcal U(P)$, but not the underlying decision structure $\mathcal C$. Thus, ClinGuide-MT tests whether a model can use its implicit clinical knowledge and the available query options to identify the missing variables needed to determine the final recommendation.

\subsection{20 Questions}
\label{appendix:20q}

\subsubsection{Problem formulation}

Each 20Q instance contains a hidden target \(y\), sampled from a finite target set \(\mathcal{Y}_{\mathrm{target}}\). 
At turn \(t\), the model has observed the interaction history
\[
\mathcal{H}_t
=
\{(q_1,a_1),\ldots,(q_{t-1},a_{t-1})\},
\]
where \(q_i\) is the \(i\)-th question asked by the model and \(a_i\in\{\texttt{yes},\texttt{no},\texttt{pass}\}\) is the examiner's answer. 
Given \(\mathcal{H}_t\), the model either asks a new yes-or-no question \(q_t\) or produces a final guess \(\hat{y}\):
\[
q_t \sim \pi_{\theta}(\cdot \mid \mathcal{H}_t)
\quad \text{or} \quad
\hat{y} \sim \pi_{\theta}(\cdot \mid \mathcal{H}_t).
\]
The goal is to identify the hidden target \(y\) within a limited number of turns.

The examiner answers each question using an oracle-style function
\[
a_t = g(y,q_t),
\qquad
a_t \in \{\texttt{yes},\texttt{no},\texttt{pass}\}.
\]
The answer \texttt{yes} means that the question is true for the target; \texttt{no} means that it is false; and \texttt{pass} means that the question is ambiguous, underspecified, only partially true, or cannot be answered cleanly with yes or no.

\paragraph{Relation to CSPs.}
20Q can be viewed as a generalized underspecified CSP. 
The hidden target \(Y\) is the variable to be determined, and the candidate set provides its finite domain. 
Each question-answer pair \((q_t,a_t)\) adds a new constraint on feasible target values. 
After observing \(\mathcal{H}_t\), the remaining feasible targets are those compatible with all previous answers:
\[
\Omega_t
=
\left\{
c\in \mathcal{Y}_{\mathrm{eval}}
:
g(c,q_i)=a_i \ \text{for all compatible prior turns}
\right\}.
\]
However, unlike our fixed variable-query tasks, the model does not choose from a predefined set of missing variables. 
The query space is open-ended, and each question defines a new partition of the candidate space. 
Therefore, the standard notions of a minimal \(k\)-sufficient set and a fixed degree of underspecification \(k\) are not directly applicable.

\subsubsection{Candidate-space surrogate}

In principle, evaluating a question requires knowing how much it narrows the latent space of all possible target values. 
Because this space is open-ended and difficult to enumerate, we approximate it with a finite candidate set:
\[
\mathcal{Y}_{\mathrm{eval}}
=
\{c_1,c_2,\ldots,c_N\}.
\]
This candidate set serves as a surrogate hypothesis space for computing question-level and trajectory-level information-seeking metrics.
At the beginning of each episode, we initialize a uniform belief over candidates:
\[
p_0(c)=\frac{1}{|\mathcal{Y}_{\mathrm{eval}}|},
\qquad
c\in \mathcal{Y}_{\mathrm{eval}}.
\]
The belief distribution is updated turn by turn according to the observed answers. 
This belief is used only for offline evaluation and is not shown to the model.

\paragraph{Candidate pools.}
We evaluate two main candidate subsets. 
\textsc{Common} combines common animals, places, foods, and objects. 
\textsc{Thing} focuses on object-like targets. 
We also maintain category-level pools for analysis. 
Table~\ref{tab:20q_pool_stats} summarizes the target and evaluation candidate sizes.

\begin{table}[th]
\centering
\small
\caption{Statistics of 20Q candidate pools. 
The target column reports the number of hidden targets used for evaluation, while the candidate column reports the size of the candidate-space surrogate used for offline question-quality estimation.}
\label{tab:20q_pool_stats}
\begin{tabular}{lcc}
\toprule
\textbf{Pool} & \textbf{\# Targets} & $|\mathcal{Y}_{\mathrm{eval}}|$ \\
\midrule
Animals & 27 & 116 \\
Places & 21 & 71 \\
Food & 26 & 92 \\
Objects & 37 & 126 \\
\midrule
\textsc{Common} & 111 & 405 \\
\textsc{Thing} & 200 & 188 \\
\bottomrule
\end{tabular}
\end{table}

\subsubsection{Question-induced partitions}

For a candidate \(c\in\mathcal{Y}_{\mathrm{eval}}\) and a model-generated question \(q_t\), the examiner function assigns one of three labels:
\[
g(c,q_t)\in\{\texttt{yes},\texttt{no},\texttt{pass}\}.
\]
Thus, each question partitions the candidate space into three subsets:
\[
\mathcal{Y}^{\texttt{yes}}(q_t)
=
\{c:g(c,q_t)=\texttt{yes}\},
\]
\[
\mathcal{Y}^{\texttt{no}}(q_t)
=
\{c:g(c,q_t)=\texttt{no}\},
\]
\[
\mathcal{Y}^{\texttt{pass}}(q_t)
=
\{c:g(c,q_t)=\texttt{pass}\}.
\]
An informative question should induce a partition that substantially reduces uncertainty about the hidden target under the current belief. 
For example, broad questions such as whether the target is a living organism may be useful early in the episode, while more specific questions may become useful once the candidate space has been narrowed.

\subsubsection{Belief update}

After the model asks \(q_t\) and receives the observed answer \(a_t\), we update the belief over candidates using a soft likelihood model. 
For each candidate \(c\), we define
\[
P(a_t\mid c,q_t)
=
\begin{cases}
\alpha, & g(c,q_t)=a_t, \\
\beta, & g(c,q_t)=\texttt{pass}, \\
\gamma, & \text{otherwise},
\end{cases}
\]
where we set
\[
\alpha=0.98,\qquad
\beta=0.50,\qquad
\gamma=0.02.
\]
The unnormalized posterior is
\[
\tilde{p}_{t+1}(c)
=
p_t(c)P(a_t\mid c,q_t),
\]
and the normalized posterior is
\[
p_{t+1}(c)
=
\frac{\tilde{p}_{t+1}(c)}
{\sum_{c'\in \mathcal{Y}_{\mathrm{eval}}}\tilde{p}_{t+1}(c')}.
\]
    This update assigns high weight to candidates consistent with the observed answer, low weight to inconsistent candidates, and intermediate weight to candidates for which the question is ambiguous or not cleanly answerable.

We use a soft update rather than hard filtering because examiner answers over open-ended natural-language questions can be noisy. 
The \texttt{pass} response is also not equivalent to contradiction: it indicates ambiguity or insufficient grounding, so candidates in the \texttt{pass} bucket are downweighted less aggressively than candidates with directly inconsistent answers.

\subsubsection{Evaluation metrics for 20Q}
\label{app:20q_metric}

We evaluate 20Q at both the instance level and the turn level. 
Final accuracy measures whether the model identifies the hidden target, while the information-seeking metrics measure whether the questions efficiently reduce uncertainty.

\paragraph{Final Accuracy.}
The primary task-completion metric is whether the final prediction \(\hat{y}\) matches the hidden target \(y\):
\[
\mathrm{Acc}
=
\mathbb{I}[\hat{y}=y].
\]
We additionally report the average number of turns used before the model makes its final guess.

\paragraph{Entropy.}
At each turn, we measure the remaining uncertainty over the candidate-space surrogate using Shannon entropy:
\[
H(p_t)
=
-\sum_{c\in\mathcal{Y}_{\mathrm{eval}}}p_t(c)\log p_t(c).
\]
Lower entropy means that the interaction history has narrowed the candidate space more effectively.

\paragraph{Expected Information Gain.}
We quantify how informative a question \(q_t\) is by its \emph{expected} reduction in entropy under the current belief, evaluated before the answer is observed.
Reusing the soft likelihood \(P(a\mid c,q_t)\) from the belief update, the predictive distribution over the possible answers \(a\in\{\texttt{yes},\texttt{no},\texttt{pass}\}\) is
\[
P(a\mid q_t)
=
\sum_{c\in\mathcal{Y}_{\mathrm{eval}}}p_t(c)\,\bar{P}(a\mid c,q_t),
\qquad
\bar{P}(a\mid c,q_t)
=
\frac{P(a\mid c,q_t)}{\sum_{a'}P(a'\mid c,q_t)},
\]
and the hypothetical posterior under a candidate answer \(a\) is
\[
p_{t+1}^{(a)}(c)
=
\frac{p_t(c)\,\bar{P}(a\mid c,q_t)}{\sum_{c'\in\mathcal{Y}_{\mathrm{eval}}}p_t(c')\,\bar{P}(a\mid c',q_t)}.
\]
The expected information gain is the mutual information between the hidden target \(Y\) and the as-yet-unobserved answer \(A_t\) under the current belief:
\[
\mathrm{EIG}(q_t)
=
H(p_t)-\!\!\sum_{a\in\{\texttt{yes},\texttt{no},\texttt{pass}\}}\!\! P(a\mid q_t)\,H\!\left(p_{t+1}^{(a)}\right)
=
I(Y;A_t\mid q_t).
\]
A larger value indicates that the question is expected to remove more uncertainty about the hidden target.
Unlike the realized reduction \(H(p_t)-H(p_{t+1})\), \(\mathrm{EIG}\) depends only on the current belief and the answer likelihoods induced by the question, and therefore does not require observing the actual answer.

\paragraph{Normalized Expected Information Gain.}
Because later turns naturally have less remaining uncertainty, raw \(\mathrm{EIG}\) is not directly comparable across turns. 
We therefore normalize it by the current entropy:
\[
\mathrm{nEIG}(q_t)
=
\frac{\mathrm{EIG}(q_t)}{H(p_t)+\epsilon}\in[0,1],
\]
where \(\epsilon\) is a small constant for numerical stability. 
This measures the fraction of current uncertainty the question is expected to remove, and is our main per-turn question-informativeness metric.

\paragraph{Remaining Entropy Curve.}
We track \(H(p_{t+1})\) across turns to obtain a remaining entropy curve. 
This curve measures how efficiently the model narrows the candidate space over the interaction. 
A sharper decrease indicates more efficient information seeking.

\paragraph{Pass Mass.}
For each turn \(t\), after the model asks question \(q_t\), we compute the belief mass assigned to candidates for which the question receives a \texttt{pass} label:
\[
\mathrm{PassMass}_t
=
\mathrm{PassMass}(q_t)
=
\sum_{c:g(c,q_t)=\texttt{pass}}p_t(c).
\]
This is a per-turn metric computed under the current belief \(p_t\), before updating the belief with the observed answer \(a_t\). 
Thus, pass mass measures how much of the currently plausible candidate space cannot be cleanly partitioned by the question asked at turn \(t\). 
High pass mass usually indicates that the question is ambiguous, underspecified, difficult to ground, or poorly aligned with the remaining candidate pool. 
This metric is useful for diagnosing questions that sound plausible but do not cleanly divide the current hypotheses into informative \texttt{yes} and \texttt{no} regions.

\subsubsection{Offline question-informativeness evaluation}

The belief update and information-gain metrics are used only for offline analysis and do not affect the interaction protocol seen by the model. 
During the game, the model observes only the interaction history and the examiner's answer to each question. 
After an episode is complete, we replay the trajectory over the candidate-space surrogate and compute per-turn partitions, posterior beliefs, entropy, expected information gain, normalized expected information gain, and pass mass. 
Unless otherwise specified, we use Qwen3-30B-A3B-Instruct-FP8 to support the offline evaluation.

This offline evaluation serves two purposes. 
First, it separates final task success from question informativeness: a model may guess correctly after asking weak questions, or fail despite asking several informative ones. 
Second, it enables fine-grained diagnosis of multi-turn information seeking, such as whether a model asks broad high-gain questions early, avoids ambiguous questions, and progressively narrows the candidate space across turns.

\subsubsection{Human validation of normalized expected information gain}
\label{app:nig_human_validation}

nEIG (Appendix~\ref{app:20q_metric}) measures a narrower property than holistic question informativeness: the reduction of uncertainty under the current belief and the surrogate candidate space. To test whether this quantity tracks human judgments of the same property, we conducted a blinded forced-choice study on 200 interaction states drawn from 111 \textsc{Common} episodes.

\paragraph{Protocol.}
For each sampled state at turn $t \in [2, 8]$, we paired (i) the question generated by Qwen3-30B-A3B-Thinking-FP8 at that turn with (ii) a question generated by GPT-5-mini from the identical truncated history. We excluded identical questions and pairs with $|\Delta\mathrm{nEIG}| < 0.01$, and retained at most four pairs per episode, separated by at least two turns. A graduate-level annotator saw the candidate pool, the preceding question--answer history, and the two candidate questions in randomized order. The annotator was blinded to the hidden target, the answers to the candidate questions, the model identities, and the $\mathrm{nEIG}$ scores, selected which question would better narrow the remaining possibilities, and rated confidence on a scale from 1 to 3.

\paragraph{Results.}
The higher-$\mathrm{nEIG}$ question matched the human preference in 153 of 200 pairs (76.5\%), compared with a 50\% chance rate. Agreement was 86.5\% for judgments made at the highest confidence level (3), compared with 70.6\% and 70.7\% at confidence levels 2 and 1, respectively. Table~\ref{app_tab:nig_human_validation} stratifies agreement by the absolute $\mathrm{nEIG}$ difference between the two candidate questions; agreement is significantly above chance in every bin.

\begin{table}[htp]
\centering
\small
\caption{Human--$\mathrm{nEIG}$ agreement stratified by the absolute difference in normalized expected information gain between the two candidate questions. Each cell reports the fraction of pairs in which the higher-$\mathrm{nEIG}$ question matched the blinded annotator's preference; $p$-values compare each agreement rate against the chance rate of $0.5$.}
\label{app_tab:nig_human_validation}
\begin{tabular}{lcc}
\toprule
$|\Delta\mathrm{nEIG}|$ & Human--$\mathrm{nEIG}$ agreement & $p$-value vs.\ $0.5$ \\
\midrule
$<0.05$        & 72.3\% & $<10^{-4}$ \\
$0.05$--$0.10$ & 86.5\% & $<10^{-7}$ \\
$0.10$--$0.15$ & 81.3\% & $<10^{-4}$ \\
\bottomrule
\end{tabular}
\end{table}

\paragraph{Scope of the validation.}
These results provide direct evidence that normalized expected information gain aligns with human judgments of which question better reduces uncertainty over the remaining candidates, particularly when the human preference is expressed with high confidence. They do not imply that the metric reproduces holistic judgments of question informativeness. In particular, $\mathrm{nEIG}$ is defined relative to the surrogate candidate pool, the belief-update procedure, and the offline evaluator described in Appendix~\ref{app:20q_metric}, and its values inherit the assumptions of these components.

\clearpage

\section{Prompts and implementation details}
\label{app:implementation_detail}

\subsection{Prompt templates}

\paragraph{Logic-Q-MT.}
\label{appendix:logicq_multi_prompts}
\blankline

\begin{promptbox}{System Prompt (missing variables, k not provided, adapted from~\cite{li2025questbench})}
Suppose you know the following rules about Alice:\\
\{rules\_nl\}
\blankline
You are trying to discern whether a statement about Alice is true given some facts. You have a budget to ask about up to \{max\_k\} attributes at once. You must decide whether you have enough information to determine whether the final statement is true. You may respond with one of the following:
\blankline
Instructions:

1. If you already have enough information to determine the truth value of the statement, respond strictly with: "End questioning".

2. Otherwise, you MUST select a set of attributes (at least 1 and at most \{max\_k\}) to query. Choose the best combination that provides the most information regarding the statement.

3. Format the question strictly as: "Question: Is Alice [attribute\_1]? Is Alice [attribute\_2]? ..." (for at least 1 and at most \{max\_k\} attributes).

4. Do not output any other text.
\end{promptbox}

\begin{promptbox}{System Prompt (missing variables, k provided, adapted from~\cite{li2025questbench})}
Suppose you know the following rules about Alice:

\{rules\_nl\}
\blankline
You are trying to discern whether a statement about Alice is true given some facts. You must select exactly \{k\} attributes of Alice to query in order to gain the most information about the final statement.
\blankline

Instructions:

1. You MUST select exactly \{k\} attributes to query. Choose the best combination that provides the most information regarding the statement.

2. Format the question strictly as: "Question: Is Alice [attribute\_1]? Is Alice [attribute\_2]? ..." (for exactly \{k\} attributes).

3. Do not output any other text.
\end{promptbox}

\begin{promptbox}{System Prompt (degree of underspecification, adapted from~\cite{li2025questbench})}
Suppose you know the following rules about Alice:

\{rules\_nl\}
\blankline

You will be given some facts about Alice (some attributes known true/false), and then a target yes/no question of the form "Is Alice X?". Decide whether the target question can be answered with certainty from the given facts and rules.
\blankline

Output in exactly one of these formats:

- If the question is NOT ambiguous (it can be answered definitively as Yes or No): "Answer: No"

- If the question IS ambiguous because some attribute values are missing: output the MINIMUM number of additional attributes that must be known to answer definitively, as an integer 1-4: "Answer: 1", "Answer: 2", "Answer: 3", or "Answer: 4"

- If you are sure it is ambiguous but cannot determine the count: "Answer: Yes, but not sure about how many"

Do not output any other text.
\end{promptbox}

\begin{promptbox}{System Prompt (sequential task-solving, no budget in prompt)}
Suppose you know the following rules about Alice:

\{rules\_nl\}
\blankline

You are trying to discern whether Alice is {goal} given some facts. You must decide whether you have enough information to determine whether the final statement is true. At each turn, you may respond with one of the following:
\blankline

Instructions:

1. If you already have enough information to determine the truth value of the statement, respond strictly with: "Answer: \{goal\}" or "Answer: not \{goal\}". 

2. Otherwise, you MUST select exactly one attribute to query. Choose the single attribute that provides the most information regarding the statement, such that the answer can be determined with least questions being asked. 

3. Format the question strictly as: "Question: Is Alice [attribute]?"

4. Do not output any other text.
\blankline

IMPORTANT: The initial facts provided are INSUFFICIENT. You MUST ask questions to gather the missing information before you can answer correctly. Do not guess - ask questions first.
\end{promptbox}

\begin{promptbox}{System Prompt (sequential task-solving, with budget in prompt)}
Suppose you know the following rules about Alice:

\{rules\_nl\}
\blankline

You are trying to discern whether Alice is \{goal\} given some facts. You can take at most \{max\_turns\} turn(s). In each turn, you may ask about up to \{max\_num\_queries\_per\_turn\} attribute(s). You must decide whether you have enough information to determine whether the final statement is true. At each turn, you may respond with one of the following:
\blankline

Instructions:

1. If you already have enough information to determine the truth value of the statement, respond strictly with: "Answer: \{goal\}" or "Answer: not \{goal\}".

2. Otherwise, you MUST select a set of attributes (between 1 and \{max\_num\_queries\_per\_turn\}) to query. Choose the smallest sufficient combination that provides enough information regarding the statement.

3. Format the question strictly as: "Question: Is Alice [attribute\_1]? Is Alice [attribute\_2]? ..." (for at least 1 attribute).

4. Do not output any other text.
\blankline

IMPORTANT: The initial facts provided are INSUFFICIENT. You MUST ask questions to gather the missing information before you can answer correctly. Do not guess - ask questions first.
\end{promptbox}

\begin{promptbox}{User Prompt}
\{known\_facts\}\\
\{known\_untrue\_facts\}\\
\{invalid\_qs\}\\
Is Alice \{goal\}?
\end{promptbox}

\paragraph{GeneReg-MT.}
\label{appendix:genereg_multi_prompts}
\blankline

\begin{promptbox}{System Prompt (steady state identification, sequential task-solving)}
You are reasoning about a Boolean gene regulatory network under synchronous updates. You will be given a partial observation of the initial state vector at time t=0, and your task is to identify which attractor ID (fixed point or cycle) the system will reach under synchronous updates starting from that initial state.
\blankline

Instructions:

- Genes are binary (0/1).

- The update semantics are synchronous (all genes update at the same discrete time step).

- Any gene values you ask about are values in the initial state (time t=0).

- The final answer refers to the reached attractor (fixed point or cycle).

- Respond at each turn with exactly one of:
  1) Question: [GENE\_NAME]
  2) Answer: [ATTRACTOR\_ID\_INTEGER]
\blankline

Rules:

1. If current information is sufficient, output "Answer: [ATTRACTOR\_ID\_INTEGER]".

2. Otherwise ask for one gene (or up to the configured max per turn) in format "Question: [GENE\_NAME]".

3. Do not output any extra text.
\blankline

IMPORTANT: The current observations are insufficient. Ask questions before answering when uncertain.
\end{promptbox}

\begin{promptbox}{System Prompt (marker identification, sequential task-solving)}
You are reasoning about a Boolean gene regulatory network under synchronous updates. You will be given a partial observation of the initial state vector at time t=0, and your task is to identify the value of marker gene \{marker\_gene\} in the reached attractor (steady behavior under synchronous updates).
\blankline

Instructions:

- Genes are binary (0/1).

- The update semantics are synchronous (all genes update at the same discrete time step).

- Any gene values you ask about are values in the initial state (time t=0).

- Respond at each turn with exactly one of:

  1) Question: [GENE\_NAME]
  
  2) Answer: 0   or   Answer: 1
\blankline

Rules:

1. If current information is sufficient, output "Answer: 0" or "Answer: 1".

2. Otherwise ask for one gene (or up to the configured max per turn) in format "Question: [GENE\_NAME]".

3. Do not output any extra text.
\blankline

IMPORTANT: The current observations are insufficient. Ask questions before answering when uncertain.
\end{promptbox}

\begin{promptbox}{User Prompt (steady state identification, no budget in prompt)}
Gene index order for all bitstrings below (left->right): \{gene\_index\_text\}

Target: determine the attractor ID reached from the true initial state.

Synchronous Boolean update rules:

\{rules\}
\blankline

Attractor catalog (fixed points and cycles) for this model:

\{attractors\_catalog\}
\blankline
            
Catalog format:

- FixedPoint: a single bitstring state.
            
- Cycle: multiple bitstrings joined by " | " (one state after another in the cycle).
\blankline

Currently observed gene values:

\{obs\}
\blankline

You may NOT ask about these genes:

\{banned\}

\end{promptbox}

\begin{promptbox}{User Prompt (marker identification, no budget in prompt)}
Gene index order for all bitstrings below (left->right): \{gene\_index\_text\}

Target: determine converged marker value of \{marker\_gene\} (0/1).

Synchronous Boolean update rules:

\{rules\}
\blankline

Attractor catalog (fixed points and cycles) for this model:

\{attractors\_catalog\}
\blankline
            
Catalog format:

- FixedPoint: a single bitstring state.
            
- Cycle: multiple bitstrings joined by " | " (one state after another in the cycle).
\blankline

Currently observed gene values:

\{obs\}
\blankline

You may NOT ask about these genes:

\{banned\}

\end{promptbox}

\begin{promptbox}{User Prompt (steady state identification, with budget in prompt)}
Gene index order for all bitstrings below (left->right): \{gene\_index\_text\}

Target: determine the attractor ID reached from the true initial state.

You can take up to \{max\_turns\} turns

You may ask about up to \{max\_num\_q\_per\_turn\} genes per turn
\blankline

Synchronous Boolean update rules:

\{rules\}
\blankline

Attractor catalog (fixed points and cycles) for this model:

\{attractors\_catalog\}
\blankline
            
Catalog format:

- FixedPoint: a single bitstring state.
            
- Cycle: multiple bitstrings joined by " | " (one state after another in the cycle).
\blankline

Currently observed gene values:

\{obs\}
\blankline

You may NOT ask about these genes:

\{banned\}

\end{promptbox}

\begin{promptbox}{User Prompt (marker identification, with budget in prompt)}
Gene index order for all bitstrings below (left->right): \{gene\_index\_text\}

Target: determine converged marker value of \{marker\_gene\} (0/1).

You can take up to \{max\_turns\} turns

You may ask about up to \{max\_num\_q\_per\_turn\} genes per turn
\blankline

Synchronous Boolean update rules:

\{rules\}
\blankline

Attractor catalog (fixed points and cycles) for this model:

\{attractors\_catalog\}
\blankline

Catalog format:

- FixedPoint: a single bitstring state.

- Cycle: multiple bitstrings joined by " | " (one state after another in the cycle).
\blankline

Currently observed gene values:

\{obs\}
\blankline

You may NOT ask about these genes:

\{banned\}

\end{promptbox}

\paragraph{GSME-Q-MT and GSME-Q-MT-Ext.}
\label{appendix:gsme_prompt}
\blankline

\begin{promptbox}{System Prompt (sequential task-solving)}
You are solving a math problem. You must decide whether you have enough information to solve this problem.\\
\blankline
Rules:\\
1) You may ask only about leaf variables (given as candidates).\\
2) You must NOT ask about the goal variable.\\
3) You must NOT ask about any non-leaf variable.\\
4) If you ask an invalid variable, the user will refuse and you must ask again.\\
\blankline
When asking, use one of these formats:\\
- "Question: What is VAR?"\\
- "Questions: VAR1, VAR2, ..."\\
\blankline
When answering, use:\\
- "Answer: <number>" (raw number only)
\blankline
Do not output other formats.
\blankline
You may ask at most ONE variable per turn. Use fewer turns when possible.\\
\end{promptbox}

\begin{promptbox}{User Prompt}
Math problem: \{problem\}
\end{promptbox}

\paragraph{ClinGuide-MT.}
\label{appendix:clinguide-prompt}
\blankline

\begin{promptbox}{System Prompt (Model, sequential task-solving)}
You are a clinical decision support system. You may only ask questions from the following.
\blankline
Allowed questions: 

\{allowed questions\}
\blankline
Possible recommendations: 

\{all recommendations\}
\blankline
Instructions:

You will be given an initial patient presentation and may ask up to \{budget\} follow-up questions over multiple turns. Your goal is to select the correct recommendation from possible recommendations. 
\blankline
At each turn:

- If you have enough information to determine a single recommendation, respond strictly with: "Answer: [recommendation]".

- Otherwise, ask exactly one question to obtain the most critical missing information. Respond strictly with: "Question: [question]?"

- You may only ask questions that relate to the allowed questions above. Do not ask about topics not listed.
\blankline
Question selection strategy:

Not all allowed topics may be equally informative for this patient. At each turn, choose the single question whose answer is most likely to distinguish among the possible recommendations above, allowing you to reach the correct recommendation in the fewest total questions.
\blankline
Note: Some of the allowed question topics are drawn from other clinical guidelines and are distractors — they are not relevant to the possible recommendations for this patient. Before asking, critically assess whether each topic can actually distinguish among the recommendations listed above. Skip topics that cannot.
\end{promptbox}

\begin{promptbox}{User Prompt (Model, sequential task-solving)}
You are presented a patient with the following information:

\{Initial patient context + oracle history\}
\blankline
What question/recommendation should you ask/give to the patient?
\end{promptbox}

\begin{promptbox}{System Prompt (Oracle, sequential task-solving)}

You are an oracle that answers clinical questions about a patient based ONLY on provided information.
\blankline
You will receive:

1. A list of desired question-answer pairs about the patient. The desired question might be a summary or concept that is desirable to be asked about, but not an actual question.

2. One or more questions to answer
\blankline
For each question you must:

1. Try to match it with one of the desired questions (allow paraphrasing, e.g., "patient's age" matches "age of the patient", "is the patient pregnant" matches "pregnancy test", etc.)

2. If matched, output the corresponding answer from the desired pairs

3. If no match exists, output "Not sure"
\blankline
Critical rules:

- Only answer based on the provided question-answer pairs

- Do not make up or infer any information

- If a question cannot be matched to a desired question, you must output "Not sure"

- Allow matching for questions that are paraphrased but semantically equivalent

- The past Q\&A history is provided only for context/reference - do not exclude any options based on what was previously answered

- Even if a question was asked before, you should still match and answer it normally

\end{promptbox}

\begin{promptbox}{User Prompt (Oracle, sequential task-solving)}
Desired question-answer pairs about this patient:

\{desired pairs text\}\{past history text\}
\blankline
Questions to answer:

\{questions text\}
\end{promptbox}

\paragraph{20Q.}
\label{appendix:20q_prompt}
\blankline

\begin{promptbox}{System Prompt (Guesser, sequential task-solving)}
Let us play the game of 20 questions. I am impersonating the thing, X.
\blankline
Your goal is to identify X within at most 20 turns.
\blankline
Rules:\\
1. Ask exactly one question only.\\
2. Each question should start with "Is X" whenever possible.\\
3. Do not ask more than one question in a turn.\\
4. Do not add explanation, justification, or extra text.\\
5. If this is the final turn, do not ask a question. Instead, make exactly one final guess.\\
\blankline
Final-turn output rule:\\
- On the final turn, output only:\\
  X is "<your best guess>".
\blankline
Let us begin. Ask me the first question.
\blankline
Reply with exactly one action only.\\
Allowed actions:\\
1. one yes/no question, or\\
2. one direct guess of X.\\
Do not output multiple questions.\\
Do not output both a question and a guess.\\
Keep the reply short.
\end{promptbox}

\begin{promptbox}{System Prompt (Examiner, sequential task-solving)}
Let us play the game of 20 questions. You are the answerer and I am the guesser. X is \{item\}.
\blankline
You must answer each question truthfully based on being X, but you must follow this output rule strictly:
\blankline
Allowed responses:\\
- "Yes"\\
- "No"\\
- "Partially"\\
- "Not applicable"\\
- "You guessed it. X is \{item\}."
\blankline
Important rules:

1. Output exactly one of the allowed responses above.

2. Do not add any explanation, justification, hint, clarification, punctuation, or extra words.

3. Do not reveal any additional information about X beyond the allowed response.

4. If the guess is exactly correct, output: "You guessed it. X is \{item\}."

5. For all other questions, output only one of: "Yes", "No", "Partially", or "Not applicable".
\blankline
Examples of forbidden outputs:\\
- "Yes, because it is a mammal."\\
- "No. It usually lives in Africa."\\
- "Partially, since that depends on context."
\blankline
Examples of valid outputs:\\
- "Yes"\\
- "No"\\
- "Partially"\\
- "Not applicable"\\
\blankline
Let us begin. Here is my first question.
\end{promptbox}

\begin{promptbox}{System Prompt (Offline Evaluator, sequential task-solving)}
You are a strict oracle used inside a 20-questions benchmark.\\
You will be given one yes/no style question about X and a numbered list of candidate entities.\\
For each candidate, answer what the truthful response would be IF X were exactly that candidate.
\blankline
Allowed labels:\\
- Yes\\
- No\\
- Pass\\
\blankline
Use Pass when the question is ambiguous, malformed, underspecified, only partially true, context-dependent, or cannot be answered cleanly with a definite Yes or No.\\
\blankline
Output format requirements:\\
Return exactly one line per candidate.\\
Each line must be exactly:\\
<index>. <Yes|No|Pass>\\
Examples:\\
1. Yes\\
2. No\\
3. Pass\\
Do not include candidate names.\\
Do not include explanations.\\
Do not omit any candidate.\\
Do not add any extra text before or after the answer list.
\end{promptbox}

\begin{promptbox}{User Prompt (Offline Evaluator, sequential task-solving)}
Candidate: \{candidate\}\\
Question: \{question\}
\blankline
Return exactly one token: Yes / No / Pass
\end{promptbox}

\subsection{Experiment compute resources}
\label{appendix:inference_setup}
All open-weight models are served locally with \texttt{vLLM} behind an
OpenAI-compatible endpoint and queried by our multi-turn evaluator
through that endpoint. Each evaluation job launches a dedicated
\texttt{vLLM} server in a containerized environment on a
dynamically-selected free port; the evaluator only begins after both a
liveness check and a model-readiness check succeed. Closed-API models
are queried through their respective provider APIs using the same
evaluator.

Open-weight inference uses H100 GPUs. Tensor parallelism is set per
model size: TP$=4$ for Qwen3.5-122B-A10B-FP8, TP$=2$ for
Qwen3-Next-80B-A3B-Thinking-FP8, and TP$=1$ for all other open-weight
models in our study (Qwen3.5-\{4B, 9B, 27B-FP8, 35B-A3B-FP8\},
Qwen3-4B-Thinking-2507, Qwen3-30B-A3B-Thinking-2507-FP8, and
gpt-oss-\{20B, 120B\}).

Across all servers we use a maximum context length of $131{,}072$
tokens, an FP8 KV cache where supported, chunked prefill, asynchronous scheduling, and disable multimodal inputs and prefix caching for consistency. GPU memory utilization is set to $0.95$ for single-GPU servers and $0.90$ for tensor-parallel servers. Per-model batching parameters (\texttt{max-num-batched-tokens}, \texttt{max-num-seqs}) and evaluator concurrency are tuned to saturate the server without exceeding its sequence cap. For \texttt{gpt-oss} models we set the reasoning effort to \texttt{medium} or \texttt{high} as indicated in the corresponding results; reasoning-parser settings follow the model family defaults.

\clearpage

\section{Benchmarking results}
\label{app:benchmarking_results}

We benchmark LLMs' performance in multi-turn information seeking. Open-weight models are hosted locally on Nvidia H100 GPUs and run with vLLM. Proprietary models are run through APIs with default inference parameters recommended by the respective providers. 

\begin{figure}[htp]
    \vspace{-10pt}
    \centering
    \includegraphics[width=1.0\linewidth]{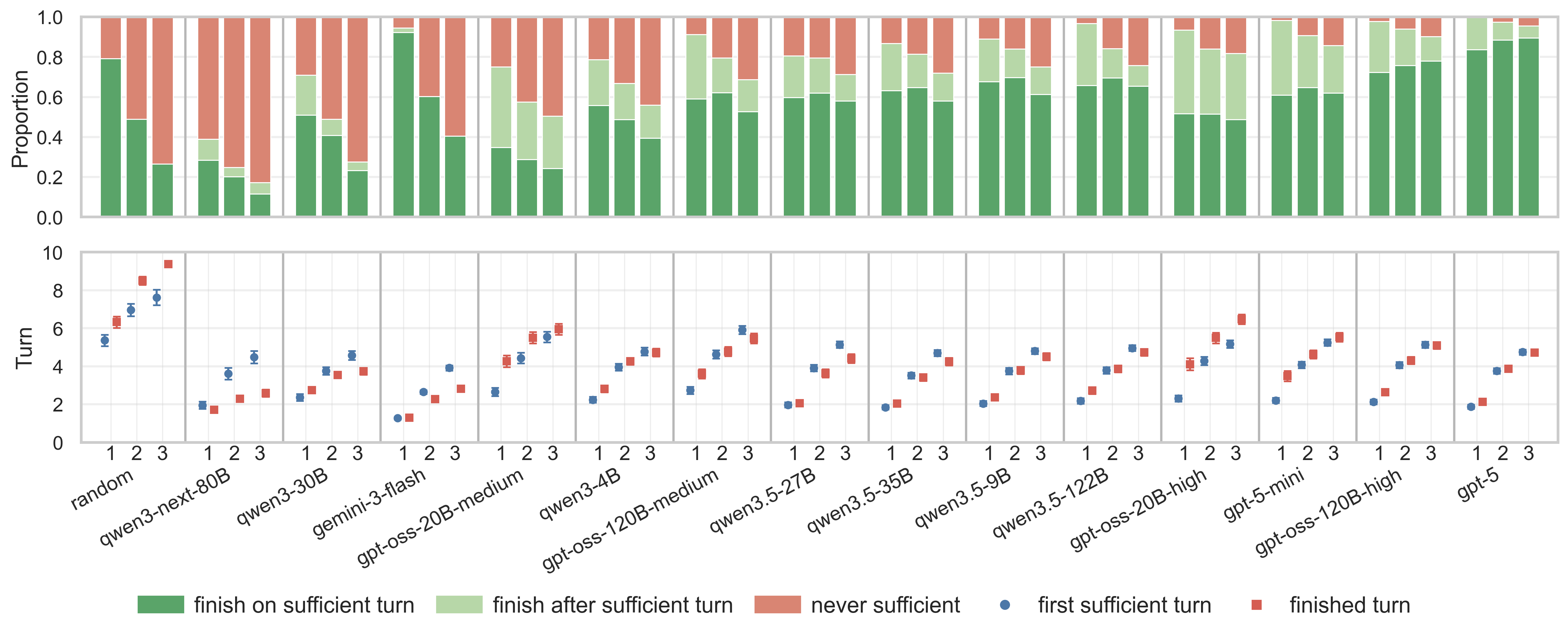}
    \vspace{-16pt}
    \caption{Performance and turns used of LLMs in Logic-Q-MT task-solving. Note that the proportion of interactions with final sufficiency = 1 is equal to the proportions that finish on sufficient turn (green) and that finish after sufficient turn (light green). LLMs are ordered by final sufficiency for $k=3$ tasks. ``Random'' randomly selects a variable that has not been queried from the valid variables per turn. 
    }
    \label{fig:logicq_multi_muliturn_benchmark}
    \vspace{-8pt}
\end{figure}

\begin{figure}[htp]
    \centering
    \includegraphics[width=1.0\linewidth]{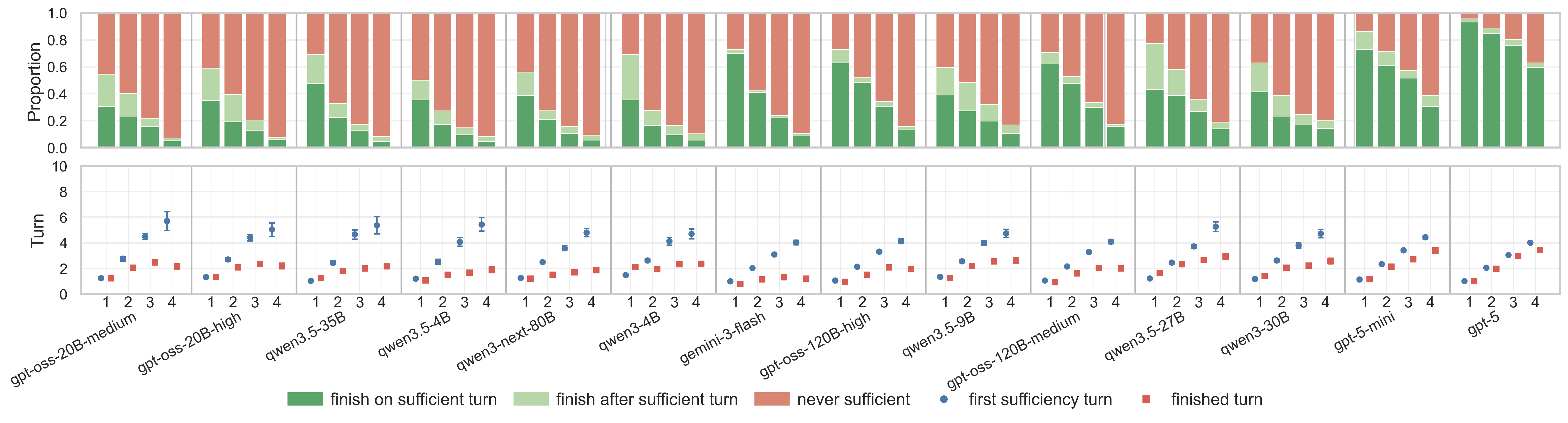}
    \caption{Performance and turns used of LLMs in GeneReg-MT dynamic steady-state task-solving. Similar to Fig.~\ref{fig:logicq_multi_muliturn_benchmark}  the proportion of samples with final sufficiency = 1 is equal to the proportions that finish on sufficient turn (green) and that finish after sufficient turn (light green). The high performance of random baseline is due to the low number of valid variables per task. 
    LLMs are ordered by final sufficiency for $k=4$ tasks. We do not include a random baseline as there are fewer than 10 valid variables in many tasks, which naturally leads to near-perfect performance of random baselines.}
    \label{fig:genereg_multi_dyn_attr_muliturn_benchmark}
\end{figure}

\begin{figure}[htp]
    \centering
    \includegraphics[width=1.0\linewidth]{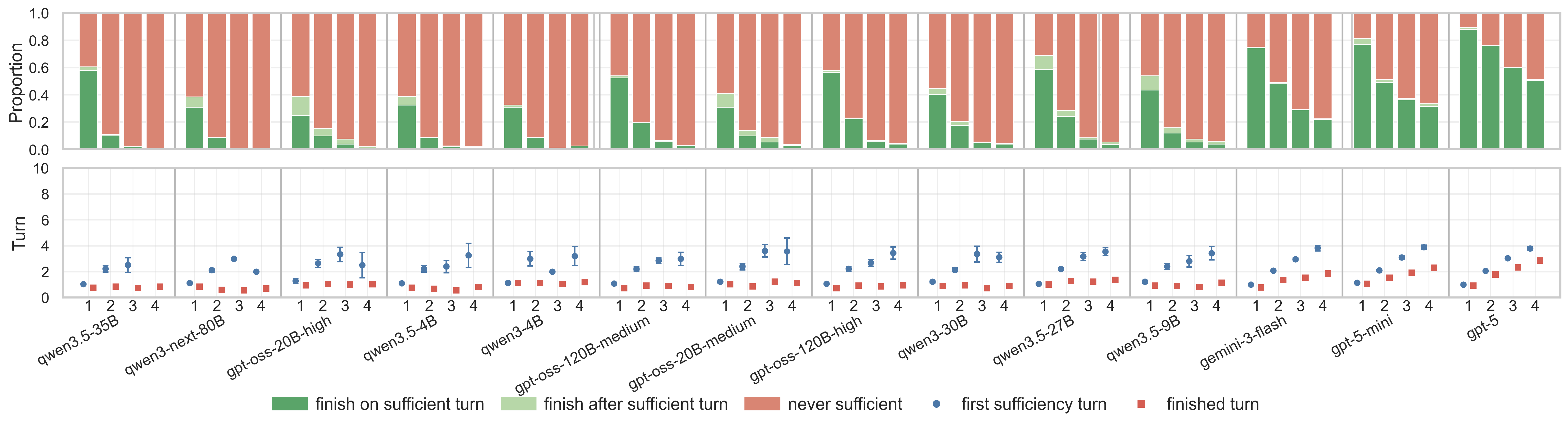}
    \caption{Performance and turns used of LLMs in GeneReg-MT dynamic marker task-solving. Similar to Fig.~\ref{fig:logicq_multi_muliturn_benchmark}  the proportion of samples with final sufficiency = 1 is equal to the proportions that finish on sufficient turn (green) and that finish after sufficient turn (light green). The high performance of random baseline is due to the low number of valid variables per task. More concretely, for $k=1,2,3,4$, there are 86/100, 86/100, 84/100, 84/100 tasks with 10 or fewer valid variables, respectively, to query for. Some ``final sufficiency turn'' values are missing in the lower panel, because there is no final sufficient interaction for that model and that $k$. LLMs are ordered by final sufficiency for $k=4$ tasks. We do not include a random baseline as there are fewer than 10 valid variables in many tasks, which naturally leads to near-perfect performance of random baselines.}
    \label{fig:genereg_multi_dyn_marker_muliturn_benchmark}
\end{figure}

\emph{Logic-Q-MT, GeneReg-MT, GSME-Q-MT, GSME-Q-MT-ext.}
As noted in the main Results, we treat \emph{final sufficiency} as a key metric for evaluating which variables a model queries with these datasets. 
We found that models exhibit varying performances, with some models (such as GPT-5 and GPT-5-mini) capable of querying all required information in almost all cases in Logic-Q-MT and in many cases in GeneReg-MT, while some usually failing to query sufficiently (Figs.~\ref{fig:logicq_multi_muliturn_benchmark},\ref{fig:genereg_multi_dyn_attr_muliturn_benchmark},\ref{fig:genereg_multi_dyn_marker_muliturn_benchmark}; full results in Tables~\ref{tab:logic_q_mt_wide},~\ref{tab:genereg_mt_summary}). 
Again, although we did not include any information about the degree of underspecification in prompt, LLMs generally perform worse in tasks with higher $k$'s.

\emph{ClinGuide-MT.} Results are shown in Table~\ref{tab:clinguide_summary}.

\emph{20-Questions.} Results are shown in Table~\ref{tab:twenty_question_summary}.

\begin{table*}[htbp]
\centering
\scriptsize
\setlength{\tabcolsep}{3pt}
\renewcommand{\arraystretch}{1.05}
\caption{
Logic-Q-MT results across different underspecification levels $k$. 
Suff. denotes final sufficiency, Acc. denotes answer accuracy, Turn denotes the average number of turns, and OverQ denotes over-questioning rate.
Gemini-3.1-Pro is evaluated on a reduced set of conditions and is therefore excluded from the cross-model analyses reported in the main text.
}
\label{tab:logic_q_mt_wide}
\resizebox{\textwidth}{!}{
\begin{tabular}{lcccccccccccc}
\toprule
\multirow{2}{*}{\textbf{Model}}
& \multicolumn{4}{c}{$\boldsymbol{k=1}$}
& \multicolumn{4}{c}{$\boldsymbol{k=2}$}
& \multicolumn{4}{c}{$\boldsymbol{k=3}$} \\
\cmidrule(lr){2-5} \cmidrule(lr){6-9} \cmidrule(lr){10-13}
& Suff. $\uparrow$ & Acc. $\uparrow$ & Turn & OverQ $\downarrow$
& Suff. $\uparrow$ & Acc. $\uparrow$ & Turn & OverQ $\downarrow$
& Suff. $\uparrow$ & Acc. $\uparrow$ & Turn & OverQ $\downarrow$ \\
\midrule

GPT-5
& 1.000 & 1.000 & 2.128 & 0.585
& 0.975 & 0.988 & 3.870 & 0.785
& 0.955 & 0.973 & 4.720 & 0.815 \\

GPT-5-mini
& 0.983 & 0.990 & 3.488 & 0.715
& 0.908 & 0.938 & 4.628 & 0.785
& 0.858 & 0.935 & 5.530 & 0.790 \\

Gemini-3-Flash
& 0.945 & 0.970 & 1.303 & 0.213
& 0.603 & 0.808 & 2.275 & 0.293
& 0.405 & 0.675 & 2.815 & 0.270 \\

Qwen3-4B-Thinking
& 0.788 & 0.858 & 2.813 & 0.555
& 0.668 & 0.815 & 4.265 & 0.578
& 0.560 & 0.710 & 4.725 & 0.470 \\

Qwen3.5-9B
& 0.890 & 0.913 & 2.355 & 0.558
& 0.840 & 0.880 & 3.790 & 0.680
& 0.750 & 0.843 & 4.508 & 0.660 \\

Qwen3.5-27B-FP8
& 0.805 & 0.898 & 2.055 & 0.523
& 0.795 & 0.900 & 3.630 & 0.670
& 0.713 & 0.843 & 4.408 & 0.658 \\

Qwen3-30B-A3B-Thinking-FP8
& 0.710 & 0.840 & 2.750 & 0.515
& 0.490 & 0.743 & 3.545 & 0.420
& 0.275 & 0.618 & 3.743 & 0.233 \\

Qwen3.5-35B-A3B-FP8
& 0.868 & 0.945 & 2.030 & 0.518
& 0.815 & 0.905 & 3.418 & 0.650
& 0.720 & 0.863 & 4.245 & 0.608 \\

Qwen3-Next-80B-A3B-Thinking-FP8
& 0.390 & 0.440 & 1.708 & 0.253
& 0.248 & 0.288 & 2.300 & 0.208
& 0.173 & 0.283 & 2.588 & 0.153 \\

Qwen3.5-122B-A10B-FP8
& 0.968 & 0.975 & 2.720 & 0.695
& 0.843 & 0.925 & 3.868 & 0.685
& 0.758 & 0.875 & 4.735 & 0.668 \\

gpt-oss-20B-medium
& 0.750 & 0.830 & 4.263 & 0.608
& 0.575 & 0.745 & 5.503 & 0.520
& 0.505 & 0.695 & 5.948 & 0.460 \\

gpt-oss-20B-high
& 0.935 & 0.933 & 4.110 & 0.710
& 0.840 & 0.908 & 5.490 & 0.720
& 0.818 & 0.900 & 6.470 & 0.733 \\

gpt-oss-120B-medium
& 0.913 & 0.963 & 3.610 & 0.713
& 0.795 & 0.903 & 4.778 & 0.695
& 0.688 & 0.838 & 5.465 & 0.655 \\

gpt-oss-120B-high
& 0.978 & 0.983 & 2.635 & 0.680
& 0.940 & 0.975 & 4.308 & 0.815
& 0.903 & 0.953 & 5.103 & 0.815 \\

Gemini-3.1-Pro
& 0.923 & 0.960 & 1.775 & 0.393
& 0.610 & 0.808 & 2.888 & 0.435
& 0.380 & 0.698 & 3.168 & 0.290 \\

\bottomrule
\end{tabular}
}
\end{table*}

\begin{table*}[htbp]
\centering
\scriptsize
\renewcommand{\arraystretch}{0.8}
\caption{GeneReg-MT results. Suff. denotes final sufficiency, Acc. denotes answer accuracy, \# Turn denotes the average number of turns, and OverQ denotes over-questioning rate. Gemini-3.1-Pro is evaluated on a reduced set of conditions and is therefore excluded from the cross-model analyses reported in the main text.}
\label{tab:genereg_mt_summary}
\resizebox{\textwidth}{!}{
\begin{tabular}{llcccccccc}
\toprule
\multirow{2}{*}{\textbf{Model}} & \multirow{2}{*}{$k$}
& \multicolumn{4}{c}{\textbf{Steady state}}
& \multicolumn{4}{c}{\textbf{Marker}} \\
\cmidrule(lr){3-6} \cmidrule(lr){7-10}
& & Suff. $\uparrow$ & Acc. $\uparrow$ & \# Turn & OverQ $\downarrow$
  & Suff. $\uparrow$ & Acc. $\uparrow$ & \# Turn & OverQ $\downarrow$ \\
\midrule

\multirow{4}{*}{GPT-5}
 & $k=1$ & 0.956 & 0.964 & 1.004 & 0.036 & 0.895 & 0.925 & 0.925 & 0.020 \\
 & $k=2$ & 0.887 & 0.927 & 1.981 & 0.063 & 0.760 & 0.865 & 1.765 & 0.040 \\
 & $k=3$ & 0.802 & 0.849 & 2.960 & 0.079 & 0.600 & 0.820 & 2.330 & 0.065 \\
 & $k=4$ & 0.628 & 0.769 & 3.431 & 0.081 & 0.515 & 0.790 & 2.860 & 0.045 \\

\midrule
\multirow{4}{*}{GPT-5-mini}
 & $k=1$ & 0.861 & 0.916 & 1.159 & 0.227 & 0.815 & 0.895 & 1.065 & 0.135 \\
 & $k=2$ & 0.716 & 0.770 & 2.136 & 0.230 & 0.515 & 0.735 & 1.520 & 0.065 \\
 & $k=3$ & 0.575 & 0.671 & 2.705 & 0.204 & 0.375 & 0.715 & 1.910 & 0.060 \\
 & $k=4$ & 0.388 & 0.544 & 3.391 & 0.156 & 0.335 & 0.650 & 2.285 & 0.065 \\

\midrule
\multirow{4}{*}{Gemini-3-Flash}
 & $k=1$ & 0.729 & 0.833 & 0.781 & 0.028 & 0.750 & 0.875 & 0.775 & 0.005 \\
 & $k=2$ & 0.423 & 0.599 & 1.134 & 0.028 & 0.490 & 0.765 & 1.340 & 0.040 \\
 & $k=3$ & 0.238 & 0.440 & 1.295 & 0.025 & 0.295 & 0.675 & 1.520 & 0.015 \\
 & $k=4$ & 0.106 & 0.359 & 1.194 & 0.019 & 0.225 & 0.585 & 1.845 & 0.020 \\

\midrule
\multirow{4}{*}{Qwen3-4B-Thinking}
 & $k=1$ & 0.693 & 0.546 & 2.120 & 0.434 & 0.325 & 0.625 & 1.125 & 0.040 \\
 & $k=2$ & 0.275 & 0.364 & 1.934 & 0.181 & 0.090 & 0.535 & 1.110 & 0.050 \\
 & $k=3$ & 0.166 & 0.261 & 2.325 & 0.125 & 0.010 & 0.530 & 1.040 & 0.000 \\
 & $k=4$ & 0.103 & 0.222 & 2.369 & 0.078 & 0.025 & 0.550 & 1.185 & 0.000 \\

\midrule
\multirow{4}{*}{Qwen3.5-9B}
 & $k=1$ & 0.594 & 0.693 & 1.235 & 0.279 & 0.540 & 0.760 & 0.915 & 0.170 \\
 & $k=2$ & 0.486 & 0.538 & 2.202 & 0.324 & 0.160 & 0.570 & 0.875 & 0.075 \\
 & $k=3$ & 0.321 & 0.406 & 2.541 & 0.217 & 0.075 & 0.470 & 0.820 & 0.025 \\
 & $k=4$ & 0.169 & 0.266 & 2.622 & 0.109 & 0.060 & 0.530 & 1.140 & 0.010 \\

\midrule
\multirow{4}{*}{Qwen3.5-27B-FP8}
 & $k=1$ & 0.773 & 0.821 & 1.649 & 0.382 & 0.690 & 0.795 & 0.995 & 0.130 \\
 & $k=2$ & 0.580 & 0.627 & 2.331 & 0.298 & 0.285 & 0.615 & 1.260 & 0.080 \\
 & $k=3$ & 0.361 & 0.490 & 2.645 & 0.214 & 0.085 & 0.570 & 1.225 & 0.020 \\
 & $k=4$ & 0.191 & 0.338 & 2.934 & 0.128 & 0.055 & 0.505 & 1.355 & 0.015 \\

\midrule
\multirow{4}{*}{Qwen3-30B-A3B-Thinking-FP8}
 & $k=1$ & 0.629 & 0.729 & 1.402 & 0.263 & 0.445 & 0.700 & 0.880 & 0.110 \\
 & $k=2$ & 0.390 & 0.507 & 2.066 & 0.251 & 0.205 & 0.565 & 0.940 & 0.050 \\
 & $k=3$ & 0.246 & 0.399 & 2.217 & 0.142 & 0.055 & 0.550 & 0.705 & 0.020 \\
 & $k=4$ & 0.200 & 0.319 & 2.584 & 0.103 & 0.045 & 0.520 & 0.900 & 0.005 \\

\midrule
\multirow{4}{*}{Qwen3-Next-80B-A3B-Thinking-FP8}
 & $k=1$ & 0.562 & 0.438 & 1.203 & 0.215 & 0.385 & 0.595 & 0.835 & 0.110 \\
 & $k=2$ & 0.279 & 0.305 & 1.512 & 0.141 & 0.090 & 0.550 & 0.595 & 0.010 \\
 & $k=3$ & 0.157 & 0.231 & 1.696 & 0.093 & 0.005 & 0.460 & 0.545 & 0.000 \\
 & $k=4$ & 0.094 & 0.184 & 1.856 & 0.063 & 0.005 & 0.515 & 0.695 & 0.000 \\

\midrule
\multirow{4}{*}{gpt-oss-20B-medium}
 & $k=1$ & 0.546 & 0.681 & 1.219 & 0.287 & 0.410 & 0.655 & 1.015 & 0.165 \\
 & $k=2$ & 0.401 & 0.523 & 2.056 & 0.263 & 0.140 & 0.540 & 0.850 & 0.070 \\
 & $k=3$ & 0.219 & 0.408 & 2.471 & 0.168 & 0.090 & 0.555 & 1.225 & 0.060 \\
 & $k=4$ & 0.072 & 0.250 & 2.141 & 0.059 & 0.035 & 0.500 & 1.125 & 0.010 \\

\midrule
\multirow{4}{*}{gpt-oss-20B-high}
 & $k=1$ & 0.590 & 0.705 & 1.323 & 0.315 & 0.390 & 0.635 & 0.945 & 0.175 \\
 & $k=2$ & 0.394 & 0.523 & 2.073 & 0.293 & 0.155 & 0.535 & 1.040 & 0.100 \\
 & $k=3$ & 0.206 & 0.403 & 2.353 & 0.168 & 0.075 & 0.525 & 0.990 & 0.050 \\
 & $k=4$ & 0.078 & 0.238 & 2.194 & 0.047 & 0.020 & 0.565 & 1.015 & 0.005 \\

\midrule
\multirow{4}{*}{gpt-oss-120B-medium}
 & $k=1$ & 0.709 & 0.805 & 0.920 & 0.124 & 0.540 & 0.725 & 0.705 & 0.045 \\
 & $k=2$ & 0.528 & 0.669 & 1.613 & 0.110 & 0.195 & 0.620 & 0.925 & 0.035 \\
 & $k=3$ & 0.335 & 0.514 & 2.025 & 0.098 & 0.065 & 0.530 & 0.870 & 0.000 \\
 & $k=4$ & 0.175 & 0.344 & 1.991 & 0.044 & 0.030 & 0.470 & 0.805 & 0.000 \\

\midrule
\multirow{4}{*}{gpt-oss-120B-high}
 & $k=1$ & 0.729 & 0.841 & 0.948 & 0.135 & 0.580 & 0.765 & 0.710 & 0.040 \\
 & $k=2$ & 0.519 & 0.627 & 1.500 & 0.092 & 0.230 & 0.590 & 0.925 & 0.040 \\
 & $k=3$ & 0.342 & 0.507 & 2.083 & 0.113 & 0.065 & 0.600 & 0.850 & 0.005 \\
 & $k=4$ & 0.156 & 0.406 & 1.941 & 0.038 & 0.045 & 0.505 & 0.945 & 0.005 \\

\midrule
\multirow{4}{*}{Gemini-3.1-Pro}
 & $k=1$ & 0.992 & 0.984 & 1.060 & 0.056 & 0.940 & 0.975 & 1.015 & 0.020 \\
 & $k=2$ & 0.965 & 0.969 & 2.127 & 0.080 & 0.960 & 0.970 & 2.140 & 0.130 \\
 & $k=3$ & 0.883 & 0.928 & 3.095 & 0.081 & 0.955 & 0.975 & 3.080 & 0.180 \\
 & $k=4$ & 0.906 & 0.941 & 4.038 & 0.178 & 0.905 & 0.960 & 3.925 & 0.155 \\

\midrule
\multirow{4}{*}{Qwen3.5-122B-A10B-FP8}
 & $k=1$ & 0.825 & 0.817 & 2.506 & 0.570 & 0.730 & 0.840 & 1.070 & 0.115 \\
 & $k=2$ & 0.634 & 0.615 & 3.265 & 0.472 & 0.250 & 0.605 & 1.300 & 0.100 \\
 & $k=3$ & 0.431 & 0.480 & 3.635 & 0.342 & 0.110 & 0.575 & 1.310 & 0.050 \\
 & $k=4$ & 0.250 & 0.341 & 3.678 & 0.203 & 0.060 & 0.540 & 1.465 & 0.010 \\

\midrule
\multirow{4}{*}{Qwen3.5-35B-A3B-FP8}
 & $k=1$ & 0.693 & 0.681 & 1.259 & 0.227 & 0.605 & 0.755 & 0.745 & 0.040 \\
 & $k=2$ & 0.329 & 0.444 & 1.793 & 0.176 & 0.110 & 0.565 & 0.830 & 0.020 \\
 & $k=3$ & 0.174 & 0.316 & 2.002 & 0.125 & 0.020 & 0.490 & 0.725 & 0.000 \\
 & $k=4$ & 0.084 & 0.244 & 2.191 & 0.063 & 0.000 & 0.485 & 0.825 & 0.000 \\

\midrule
\multirow{4}{*}{Qwen3.5-4B}
 & $k=1$ & 0.502 & 0.594 & 1.068 & 0.195 & 0.390 & 0.670 & 0.755 & 0.085 \\
 & $k=2$ & 0.272 & 0.394 & 1.509 & 0.150 & 0.090 & 0.510 & 0.670 & 0.020 \\
 & $k=3$ & 0.147 & 0.278 & 1.669 & 0.091 & 0.025 & 0.510 & 0.555 & 0.000 \\
 & $k=4$ & 0.084 & 0.209 & 1.900 & 0.075 & 0.020 & 0.590 & 0.820 & 0.005 \\

\bottomrule
\end{tabular}
}
\end{table*}

\begin{table*}[htbp]
\centering
\caption{
GSME-Q-MT and its extension results. 
Suff. denotes final sufficiency, Acc. denotes answer accuracy, \# Turn denotes the average number of turns, and OverQ denotes over-questioning rate.
}
\label{tab:gsme_q_mt_summary}
\resizebox{\textwidth}{!}{
\begin{tabular}{llcccccccc}
\toprule
\multirow{2}{*}{\textbf{Model}} & \multirow{2}{*}{$k$}
& \multicolumn{4}{c}{\textbf{GSME-Q-MT}}
& \multicolumn{4}{c}{\textbf{GSME-Q-MT-Ext}} \\
\cmidrule(lr){3-6} \cmidrule(lr){7-10}
& 
& Suff. $\uparrow$ & Acc. $\uparrow$ & \# Turn & OverQ $\downarrow$
& Suff. $\uparrow$ & Acc. $\uparrow$ & \# Turn & OverQ $\downarrow$ \\
\midrule

\multirow{4}{*}{Qwen3-4B-Thinking}
 & $k=1$ & 0.989 & 0.988 & $0.99{\pm}0.12$ & 0.004 & 0.813 & 0.813 & $0.84{\pm}0.42$ & 0.030 \\
 & $k=2$ & 0.986 & 0.985 & $1.97{\pm}0.22$ & 0.001 & 0.681 & 0.639 & $1.42{\pm}0.89$ & 0.028 \\
 & $k=3$ & 0.983 & 0.985 & $2.95{\pm}0.37$ & 0.001 & 0.585 & 0.520 & $1.78{\pm}1.37$ & 0.008 \\
 & $k=4$ & 0.981 & 0.980 & $3.93{\pm}0.53$ & 0.001 & 0.417 & 0.342 & $1.76{\pm}1.76$ & 0.024 \\

\midrule
\multirow{4}{*}{Qwen3-30B-A3B-Thinking-FP8}
 & $k=1$ & 0.985 & 0.984 & $0.99{\pm}0.12$ & 0.004 & 0.463 & 0.485 & $0.46{\pm}0.50$ & 0.000 \\
 & $k=2$ & 0.981 & 0.974 & $1.96{\pm}0.24$ & 0.001 & 0.368 & 0.319 & $0.74{\pm}0.91$ & 0.004 \\
 & $k=3$ & 0.985 & 0.980 & $2.96{\pm}0.29$ & 0.000 & 0.258 & 0.211 & $0.79{\pm}1.23$ & 0.005 \\
 & $k=4$ & 0.981 & 0.973 & $3.92{\pm}0.46$ & 0.000 & 0.264 & 0.195 & $1.09{\pm}1.55$ & 0.008 \\

\midrule
\multirow{4}{*}{gpt-oss-20B-medium}
 & $k=1$ & 0.993 & 0.988 & $0.99{\pm}0.09$ & 0.001 & 0.948 & 0.933 & $0.96{\pm}0.23$ & 0.015 \\
 & $k=2$ & 0.871 & 0.838 & $1.74{\pm}0.62$ & 0.000 & 0.941 & 0.896 & $1.89{\pm}0.38$ & 0.004 \\
 & $k=3$ & 0.780 & 0.697 & $2.34{\pm}1.08$ & 0.001 & 0.911 & 0.837 & $2.75{\pm}0.67$ & 0.005 \\
 & $k=4$ & 0.693 & 0.564 & $2.77{\pm}1.55$ & 0.000 & 0.821 & 0.724 & $3.33{\pm}1.24$ & 0.010 \\

\midrule
\multirow{4}{*}{gpt-oss-120B-medium}
 & $k=1$ & 0.999 & 0.998 & $1.00{\pm}0.03$ & 0.000 & 1.000 & 1.000 & $1.00{\pm}0.00$ & 0.000 \\
 & $k=2$ & 0.990 & 0.983 & $1.98{\pm}0.15$ & 0.001 & 0.993 & 0.993 & $1.99{\pm}0.17$ & 0.000 \\
 & $k=3$ & 0.860 & 0.718 & $2.58{\pm}0.72$ & 0.000 & 0.986 & 0.984 & $2.96{\pm}0.32$ & 0.000 \\
 & $k=4$ & 0.601 & 0.318 & $2.40{\pm}1.17$ & 0.000 & 0.882 & 0.813 & $3.53{\pm}1.00$ & 0.000 \\

\midrule
\multirow{4}{*}{GPT-5-mini}
 & $k=1$ & 1.000 & 0.998 & $1.00{\pm}0.04$ & 0.002 & 1.000 & 1.000 & $1.00{\pm}0.00$ & 0.000 \\
 & $k=2$ & 0.998 & 0.998 & $2.00{\pm}0.06$ & 0.001 & 1.000 & 1.000 & $2.01{\pm}0.08$ & 0.004 \\
 & $k=3$ & 0.999 & 0.999 & $3.00{\pm}0.06$ & 0.000 & 1.000 & 1.000 & $3.00{\pm}0.00$ & 0.000 \\
 & $k=4$ & 0.998 & 0.997 & $3.99{\pm}0.13$ & 0.000 & 1.000 & 1.000 & $4.00{\pm}0.00$ & 0.000 \\

\midrule
\multirow{4}{*}{Gemini-3-Flash}
 & $k=1$ & 0.999 & 0.996 & $1.00{\pm}0.07$ & 0.004 & 1.000 & 1.000 & $1.01{\pm}0.09$ & 0.008 \\
 & $k=2$ & 0.995 & 0.993 & $1.99{\pm}0.12$ & 0.002 & 0.993 & 0.993 & $1.99{\pm}0.19$ & 0.004 \\
 & $k=3$ & 0.990 & 0.982 & $2.98{\pm}0.26$ & 0.002 & 0.989 & 0.984 & $2.97{\pm}0.28$ & 0.000 \\
 & $k=4$ & 0.980 & 0.960 & $3.93{\pm}0.42$ & 0.003 & 0.988 & 0.984 & $3.96{\pm}0.39$ & 0.002 \\

\bottomrule
\end{tabular}
}
\end{table*}

\begin{table}[htbp]
\centering
\footnotesize
\setlength{\tabcolsep}{2.8pt}
\renewcommand{\arraystretch}{1.08}
\caption{ClinGuide-MT results. ``d'' represents the set of distractor variables added in each problem, all obtained from other ClinGuide-MT problems and filtered. \#d refers to the number of distracting variable sets.}
\label{tab:clinguide_summary}
\resizebox{\textwidth}{!}{
\begin{tabular}{llccccccccccccccc}
\toprule
\multirow{2}{*}{\textbf{Model}} & \multirow{2}{*}{$k$}
& \multicolumn{3}{c}{\textbf{Final acc.} $\uparrow$}
& \multicolumn{3}{c}{\textbf{Order correctness} $\uparrow$}
& \multicolumn{3}{c}{\textbf{Query correctness} $\uparrow$}
& \multicolumn{3}{c}{\textbf{Query coverage} $\uparrow$}
& \multicolumn{3}{c}{\textbf{Avg. turn used}} \\
\cmidrule(lr){3-5}
\cmidrule(lr){6-8}
\cmidrule(lr){9-11}
\cmidrule(lr){12-14}
\cmidrule(lr){15-17}
& & \#d=0 & \#d=10 & \#d=20
& \#d=0 & \#d=10 & \#d=20
& \#d=0 & \#d=10 & \#d=20
& \#d=0 & \#d=10 & \#d=20
& \#d=0 & \#d=10 & \#d=20 \\
\midrule

\multirow{4}{*}{Qwen3-4B-Thinking}
 & $k=1$ & 0.503 & 0.436 & 0.426 & 0.916 & 0.739 & 0.749 & 0.575 & 0.384 & 0.384 & 0.916 & 0.739 & 0.749 & 3.084 & 5.269 & 5.284 \\
 & $k=2$ & 0.461 & 0.350 & 0.333 & 0.917 & 0.874 & 0.883 & 0.700 & 0.493 & 0.514 & 0.743 & 0.618 & 0.594 & 3.249 & 5.159 & 5.145 \\
 & $k=3$ & 0.414 & 0.322 & 0.276 & 0.857 & 0.892 & 0.873 & 0.785 & 0.618 & 0.619 & 0.701 & 0.572 & 0.558 & 3.538 & 4.754 & 4.905 \\
 & $k=4$ & 0.339 & 0.317 & 0.304 & 0.779 & 0.819 & 0.858 & 0.828 & 0.674 & 0.684 & 0.681 & 0.532 & 0.531 & 3.868 & 4.894 & 4.758 \\

\midrule

\multirow{4}{*}{Qwen3-30B-A3B-Thinking-FP8}
 & $k=1$ & 0.560 & 0.486 & 0.451 & 0.783 & 0.616 & 0.601 & 0.546 & 0.409 & 0.397 & 0.783 & 0.616 & 0.601 & 2.296 & 2.987 & 2.950 \\
 & $k=2$ & 0.543 & 0.396 & 0.428 & 0.910 & 0.828 & 0.801 & 0.735 & 0.597 & 0.560 & 0.693 & 0.572 & 0.547 & 2.650 & 3.344 & 3.423 \\
 & $k=3$ & 0.495 & 0.381 & 0.386 & 0.850 & 0.836 & 0.830 & 0.793 & 0.708 & 0.673 & 0.662 & 0.542 & 0.545 & 3.162 & 3.459 & 3.551 \\
 & $k=4$ & 0.396 & 0.339 & 0.322 & 0.840 & 0.838 & 0.859 & 0.839 & 0.750 & 0.736 & 0.597 & 0.536 & 0.518 & 3.401 & 3.797 & 3.828 \\

\midrule

\multirow{4}{*}{Qwen3.5-27B-FP8}
 & $k=1$ & 0.749 & 0.714 & 0.712 & 0.948 & 0.906 & 0.906 & 0.578 & 0.447 & 0.454 & 0.948 & 0.906 & 0.906 & 3.203 & 4.781 & 4.987 \\
 & $k=2$ & 0.686 & 0.692 & 0.665 & 0.915 & 0.895 & 0.892 & 0.712 & 0.563 & 0.570 & 0.873 & 0.856 & 0.843 & 3.849 & 5.568 & 5.704 \\
 & $k=3$ & 0.689 & 0.622 & 0.632 & 0.812 & 0.805 & 0.805 & 0.755 & 0.641 & 0.630 & 0.852 & 0.813 & 0.824 & 4.597 & 6.346 & 6.392 \\
 & $k=4$ & 0.652 & 0.626 & 0.617 & 0.715 & 0.702 & 0.729 & 0.809 & 0.704 & 0.689 & 0.855 & 0.814 & 0.790 & 5.317 & 6.943 & 6.767 \\

\midrule

\multirow{4}{*}{GPT-5-mini}
 & $k=1$ & 0.720 & 0.666 & 0.687 & 0.929 & 0.839 & 0.852 & 0.646 & 0.497 & 0.494 & 0.929 & 0.839 & 0.852 & 3.269 & 3.793 & 3.779 \\
 & $k=2$ & 0.646 & 0.591 & 0.610 & 0.928 & 0.912 & 0.912 & 0.729 & 0.619 & 0.629 & 0.782 & 0.752 & 0.710 & 3.715 & 4.373 & 4.260 \\
 & $k=3$ & 0.584 & 0.557 & 0.568 & 0.851 & 0.856 & 0.854 & 0.798 & 0.694 & 0.707 & 0.734 & 0.691 & 0.688 & 4.092 & 4.719 & 4.700 \\
 & $k=4$ & 0.568 & 0.546 & 0.533 & 0.787 & 0.814 & 0.796 & 0.832 & 0.768 & 0.750 & 0.725 & 0.665 & 0.676 & 4.674 & 5.123 & 5.233 \\

\midrule

\multirow{4}{*}{GPT-5}
 & $k=1$ & 0.793 & 0.745 & 0.739 & 0.935 & 0.881 & 0.875 & 0.668 & 0.584 & 0.557 & 0.935 & 0.881 & 0.875 & 2.461 & 2.779 & 2.841 \\
 & $k=2$ & 0.748 & 0.721 & 0.704 & 0.930 & 0.921 & 0.936 & 0.778 & 0.709 & 0.706 & 0.848 & 0.838 & 0.812 & 3.298 & 3.753 & 3.694 \\
 & $k=3$ & 0.727 & 0.673 & 0.657 & 0.867 & 0.853 & 0.846 & 0.822 & 0.768 & 0.749 & 0.814 & 0.786 & 0.776 & 3.997 & 4.454 & 4.500 \\
 & $k=4$ & 0.696 & 0.656 & 0.656 & 0.782 & 0.775 & 0.801 & 0.849 & 0.803 & 0.802 & 0.807 & 0.789 & 0.773 & 4.749 & 5.295 & 5.269 \\

\midrule

\multirow{4}{*}{gpt-oss-20B-medium}
 & $k=1$ & 0.436 & 0.390 & 0.365 & 0.810 & 0.637 & 0.645 & 0.453 & 0.342 & 0.357 & 0.810 & 0.637 & 0.645 & 4.656 & 4.935 & 5.265 \\
 & $k=2$ & 0.396 & 0.331 & 0.323 & 0.893 & 0.813 & 0.807 & 0.653 & 0.492 & 0.487 & 0.748 & 0.615 & 0.596 & 4.727 & 5.767 & 5.671 \\
 & $k=3$ & 0.359 & 0.281 & 0.295 & 0.827 & 0.787 & 0.813 & 0.688 & 0.574 & 0.565 & 0.708 & 0.582 & 0.607 & 5.249 & 5.703 & 6.065 \\
 & $k=4$ & 0.366 & 0.295 & 0.251 & 0.730 & 0.767 & 0.741 & 0.742 & 0.634 & 0.619 & 0.693 & 0.595 & 0.602 & 5.476 & 6.463 & 6.696 \\

\midrule

\multirow{4}{*}{Gemini-3-Flash}
 & $k=1$ & 0.866 & 0.856 & 0.841 & 0.979 & 0.987 & 0.985 & 0.728 & 0.673 & 0.638 & 0.979 & 0.987 & 0.985 & 2.230 & 2.885 & 2.891 \\
 & $k=2$ & 0.813 & 0.834 & 0.822 & 0.938 & 0.931 & 0.930 & 0.793 & 0.743 & 0.735 & 0.889 & 0.914 & 0.898 & 3.086 & 3.690 & 3.537 \\
 & $k=3$ & 0.824 & 0.778 & 0.814 & 0.854 & 0.841 & 0.838 & 0.846 & 0.797 & 0.789 & 0.857 & 0.874 & 0.873 & 3.786 & 4.370 & 4.400 \\
 & $k=4$ & 0.793 & 0.775 & 0.784 & 0.757 & 0.743 & 0.736 & 0.872 & 0.816 & 0.823 & 0.852 & 0.836 & 0.840 & 4.722 & 5.115 & 5.079 \\

\bottomrule
\end{tabular}
}
\end{table}

\begin{table*}[htp]
\centering
\small
\setlength{\tabcolsep}{4pt}
\renewcommand{\arraystretch}{1.12}
\setlength{\tabcolsep}{4pt}
\caption{20 Questions results on Common and Things domains.}
\label{tab:twenty_question_summary}
\resizebox{\textwidth}{!}{
\begin{tabular}{llcccccccc}
\toprule
\multirow{2}{*}{\textbf{Examiner}} & \multirow{2}{*}{\textbf{Guesser}} & \multicolumn{4}{c}{\textbf{Common}} & \multicolumn{4}{c}{\textbf{Things}} \\
\cmidrule(lr){3-6} \cmidrule(lr){7-10}
 &  & \textbf{Acc.} $\uparrow$ & \textbf{\#avg turns} & \textbf{avg. norm IG} $\uparrow$ & \textbf{avg. final $H$} $\downarrow$ & \textbf{Acc.} $\uparrow$ & \textbf{\#avg turns} & \textbf{avg. norm IG} $\uparrow$ & \textbf{avg. final $H$} $\downarrow$ \\
\midrule
\multirow{6}{*}{GPT-5-mini} & gpt-oss-20B-medium & 0.1712 & 18.7928 & 0.1563 & 1.7203 & 0.1100 & 19.3550 & 0.1644 & 1.6599 \\
 & Qwen3-4B-Thinking & 0.2252 & 18.6847 & 0.1557 & 1.6206 & 0.1350 & 19.1400 & 0.1561 & 1.6941 \\
 & Qwen3-30B-A3B-Thinking-FP8 & 0.2793 & 18.0000 & 0.1673 & 1.7864 & 0.2650 & 18.0950 & 0.1607 & 1.8684 \\
 & Gemini-3-Flash & 0.4775 & 16.2883 & 0.1768 & 1.2130 & 0.3300 & 17.5550 & 0.2050 & 1.3496 \\
 & GPT-5-mini & 0.4234 & 17.1982 & 0.2030 & 1.4651 & 0.3800 & 17.5900 & 0.1829 & 1.6172 \\
 & GPT-5 & 0.5405 & 17.0270 & 0.2047 & 1.3696 & 0.4200 & 17.7200 & 0.2221 & 1.3044 \\
\midrule
\multirow{6}{*}{Qwen3-30B-A3B-Instruct-FP8} & gpt-oss-20B-medium & 0.2072 & 18.8288 & 0.1779 & 1.6254 & 0.1450 & 19.0800 & 0.1815 & 1.5974 \\
 & Qwen3-4B-Thinking & 0.2072 & 18.8739 & 0.1692 & 1.7097 & 0.2100 & 18.7550 & 0.1973 & 1.5362 \\
 & Qwen3-30B-A3B-Thinking-FP8 & 0.2703 & 18.2703 & 0.1535 & 1.7776 & 0.2600 & 17.9300 & 0.1545 & 1.7649 \\
 & Gemini-3-Flash & 0.4595 & 16.7207 & 0.1681 & 1.5350 & 0.3000 & 17.9350 & 0.1994 & 1.4787 \\
 & GPT-5-mini & 0.3964 & 17.4234 & 0.2249 & 1.3660 & 0.3300 & 17.8650 & 0.2488 & 1.4678 \\
 & GPT-5 & 0.3964 & 17.4775 & 0.2111 & 1.1027 & 0.3400 & 18.0150 & 0.2398 & 1.1745 \\
\bottomrule
\end{tabular}
}
\end{table*}

\clearpage

\section{Additional results and experiments}
\label{app:add_results}

\subsection{Logic-Q-MT and GeneReg-MT}

\subsubsection{Details on first-turn CoT analysis}
\label{app:turn1-cot-recovery}

The unit of analysis is a interaction with at least one interaction turn. We study two outcomes: Final sufficiency and second-turn hit (whether the second query is in the MSS, restricted to interactions whose first query misses the MSS and that continue to the second turn).
Interactions with no strict query-context variable mention are retained and assigned MSS relative mention rate $R_i=0$.

\paragraph{Feature extraction.}
Let $M_i$ denote the MSS for interaction $i$, and let $Q_i$ denote the set of queryable variables. We first detect variables that appear in strict query-planning contexts using the deterministic templates in Table~\ref{tab:turn1-cot-patterns}. 
We count a variable as appearing in a query-planning context only when it occurs in text explaining what information the model needs, intends to ask for, or considers useful for disambiguating the target. Mentions copied from the problem statement or appearing only in unrelated reasoning are not counted.
The strict query-planning templates are intentionally conservative. Annotated examples in Appendix~\ref{app:logicq_multi_example_cot} illustrate the distinction between awareness and MSS-focused planning.
The primary feature is
\begin{equation}
  R_i = \frac{|\{v \in M_i : v \text{ is mentioned in a strict query-planning context}\}|}{|\{v \in Q_i : v \text{ is mentioned in a strict query-planning context}\}|},
\end{equation}
with $R_i=0$ when the denominator is zero. We refer to $R_i$ as \emph{MSS relative mention rate}. As a contrastive feature, we also extract \emph{awareness of underspecification}, a binary indicator for generic statements that more information is still needed, without naming a specific variable.

\begin{table*}[htp]
\centering
\scriptsize
\begin{tabular}{p{0.05\textwidth} p{0.05\textwidth} p{0.05\textwidth} p{0.75\textwidth}}
\hline
Dataset & Construct & Feature & Templates used \\
\hline
Logic-Q-MT & Knows what is missing & MSS relative mention rate & \texttt{ask about \{var\}}, \texttt{ask is alice \{var\}}, \texttt{ask if alice is \{var\}}, \texttt{ask whether alice is \{var\}}, \texttt{ask if she is \{var\}}, \texttt{ask whether she is \{var\}}, \texttt{consider querying \{var\}}, \texttt{should ask about \{var\}}, \texttt{should ask if alice is \{var\}}, \texttt{should ask whether alice is \{var\}}, \texttt{need to ask about \{var\}}, \texttt{need to ask if alice is \{var\}}, \texttt{need to ask whether alice is \{var\}}, \texttt{if i ask \{var\}}, \texttt{if i ask about \{var\}}, \texttt{if i ask is alice \{var\}}, \texttt{if i ask if alice is \{var\}}, \texttt{if we ask \{var\}}, \texttt{if we ask about \{var\}}, \texttt{if we ask is alice \{var\}}, \texttt{if we ask if alice is \{var\}}, \texttt{next question is \{var\}}, \texttt{next question is is alice \{var\}}, \texttt{question is alice \{var\}}, \texttt{question is is alice \{var\}}, \texttt{question should be \{var\}}, \texttt{question should be is alice \{var\}}, \texttt{question would be \{var\}}, \texttt{question would be is alice \{var\}}, \texttt{best question is \{var\}}, \texttt{best question is is alice \{var\}}, \texttt{best question would be \{var\}}, \texttt{best question would be is alice \{var\}}, \texttt{choose \{var\} as the next question}, \texttt{choose asking whether alice is \{var\}}, \texttt{choose to ask whether alice is \{var\}}, \texttt{need to know \{var\}}, \texttt{need to know whether alice is \{var\}} \\
GeneReg-MT & Knows what is missing & MSS relative mention rate & \texttt{ask about \{var\}}, \texttt{ask \{var\}}, \texttt{ask for \{var\}}, \texttt{ask for the value of \{var\}}, \texttt{ask the value of \{var\}}, \texttt{should ask about \{var\}}, \texttt{should ask \{var\}}, \texttt{should ask for \{var\}}, \texttt{need to ask about \{var\}}, \texttt{need to ask \{var\}}, \texttt{need to ask for \{var\}}, \texttt{next question is \{var\}}, \texttt{next question should be \{var\}}, \texttt{next question \{var\}}, \texttt{question should be \{var\}}, \texttt{best question is \{var\}}, \texttt{want to ask about \{var\}}, \texttt{want to ask \{var\}}, \texttt{let me ask \{var\}}, \texttt{let me ask about \{var\}}, \texttt{lets ask \{var\}}, \texttt{lets ask about \{var\}}, \texttt{go with \{var\}}, \texttt{going with \{var\}}, \texttt{choose \{var\}}, \texttt{choosing \{var\}}, \texttt{choose to ask \{var\}}, \texttt{choose to query \{var\}}, \texttt{query \{var\}}, \texttt{query about \{var\}}, \texttt{query the value of \{var\}}, \texttt{should query \{var\}}, \texttt{need to query \{var\}} \\
Both datasets & Knows something is missing & Awareness of underspecification & \texttt{need more information}, \texttt{need more info}, \texttt{need more facts}, \texttt{need another question}, \texttt{need additional information}, \texttt{need additional facts}, \texttt{more information is needed}, \texttt{more info is needed}, \texttt{more facts are needed}, \texttt{more information is required}, \texttt{more facts are required}, \texttt{dont have enough information}, \texttt{do not have enough information}, \texttt{dont have enough info}, \texttt{do not have enough info}, \texttt{dont have enough facts}, \texttt{do not have enough facts}, \texttt{this is ambiguous}, \texttt{the situation is ambiguous}, \texttt{not yet enough}, \texttt{not yet sufficient}, \texttt{not yet determined}, \texttt{cannot yet determine}, \texttt{cannot yet conclude}, \texttt{cannot yet answer}, \texttt{can not yet determine}, \texttt{can not yet conclude}, \texttt{cant yet determine}, \texttt{cant yet conclude}, \texttt{still need to ask}, \texttt{still need to check}, \texttt{still need to determine}, \texttt{still need to find out}, \texttt{need to ask one more}, \texttt{need to ask another} \\
\hline
\end{tabular}
\caption{Pattern templates used in the first-turn CoT analysis. The GeneReg-MT query-planning templates are intentionally narrower than the Logic-Q-MT templates to reduce false positives from regulatory-rule enumeration.}
\label{tab:turn1-cot-patterns}
\end{table*}

\paragraph{Additional controls.}
The first-turn CoT controls are first-turn CoT length, first-turn chosen-query mention, and variable search breadth. The problem-difficulty controls are:
\begin{itemize}
\item \textbf{Logic-Q-MT}: \# rules, mean rule size, problem depth, \# queryable variables, $\lvert\mathcal{A}\rvert$, and fraction of rules with MSS variables.
\item \textbf{GeneReg-MT}: \# regulatory edges, dynamics depth, \# queryable variables, $\lvert\mathcal{A}\rvert$, and fraction of total regulatory in-edges into MSS genes. 
\end{itemize}

\paragraph{Pooled fixed-effect models.}
For Logic-Q-MT, the pooled model is
\begin{equation*}
  \operatorname{logit}\Pr(Y_i=1) = \alpha_{m(i)} + \gamma_{k(i)} + \beta_R R_i + \beta_A A_i + \mathbf{x}_i^\top\boldsymbol\theta,
\end{equation*}
where $\alpha_{m(i)}$ are model fixed effects, $\gamma_{k(i)}$ are underspecification-level fixed effects, $A_i$ is awareness of underspecification, and $\mathbf{x}_i$ contains the remaining CoT and problem controls. Standard errors are clustered by interaction. For GeneReg-MT we use the same specification separately within each subset $s \in \{\text{steady state},\text{ marker}\}$,
\begin{equation*}
  \operatorname{logit}\Pr(Y_i=1 \mid s(i)=s) = \alpha^{(s)}_{m(i)} + \gamma^{(s)}_{k(i)} + \beta_R^{(s)} R_i + \beta_A^{(s)} A_i + \mathbf{x}_i^\top\boldsymbol\theta^{(s)},
\end{equation*}
All continuous predictors are standardized for comparability in the coefficient plots; plotted values are standardized log-odds ratios, i.e., log odds ratios for a 1-SD increase in the predictor.

\paragraph{Stability analysis.}
To check whether the pooled coefficient is driven by one model or one underspecification level, we refit the same outcome model within each model-by-$k$ cell in Logic-Q-MT and within each model-task-$k$ cell in GeneReg-MT, omitting fixed effects inside the cell. Controls that are constant within a cell are dropped from that cell's regression. A cell is marked NA if it has fewer than 50 usable interactions, no outcome variation, no variation in MSS relative mention rate, or a separated fit; NA cells are excluded from summary counts.

\begin{figure}[htp]
    \centering
    \includegraphics[width=0.8\linewidth]{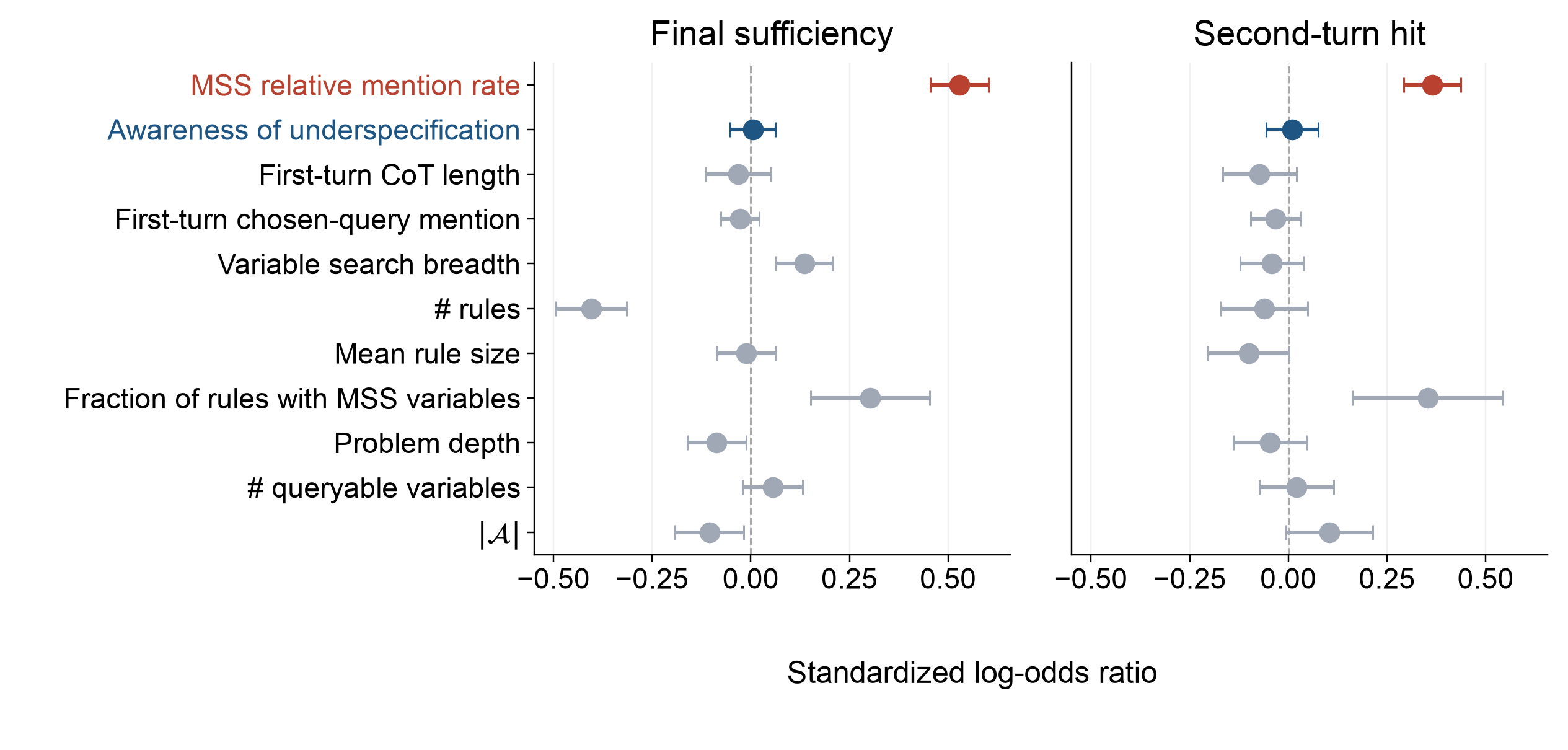}
    \caption{Turn-1 CoT predictors in Logic-Q-MT. Each panel shows standardized log-odds ratios with 95\% confidence intervals from a pooled model with model and $k$ fixed effects and standard errors clustered by interaction. All predictors are standardized before fitting, so each value is the log odds ratio for a one-standard-deviation increase in the predictor. Left: final sufficiency over all interactions with at least one turn. Right: second-turn hit among interactions whose first-turn query misses the MSS and that continue to a second turn. MSS relative mention rate is the strongest CoT predictor in both panels ($\beta_z=0.528$ for final sufficiency; $\beta_z=0.365$ for second-turn hit), while awareness of underspecification remains near zero ($\beta_z=0.005$ for final sufficiency and $\beta_z=0.010$ for second-turn hit).}
    \label{app_fig:turn1_cot_covariate_comparison_logicq_multi}
\end{figure}

\begin{figure}[htp]
    \centering
    \includegraphics[width=0.7\linewidth]{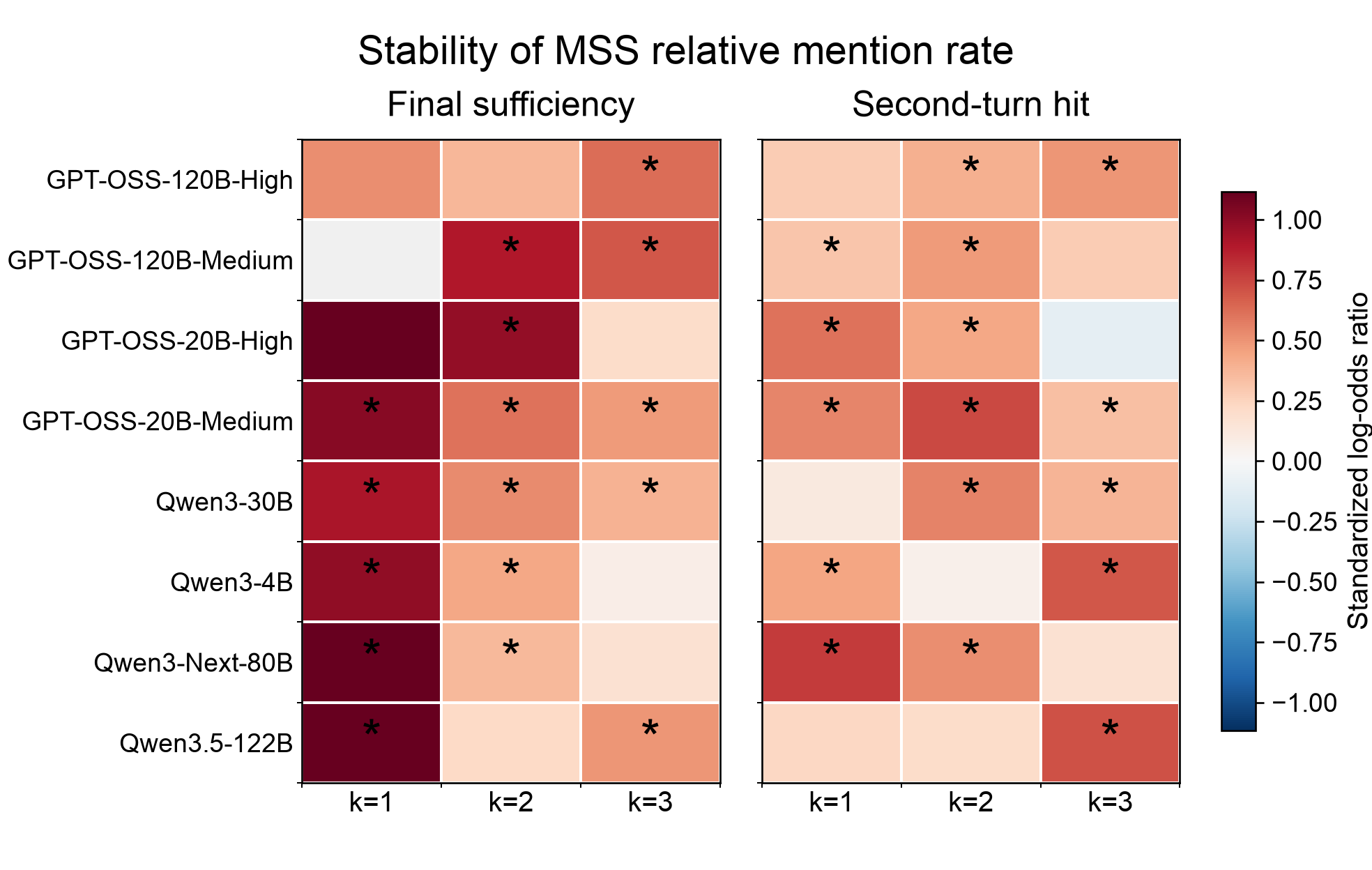}
    \caption{Stability of MSS relative mention rate in Logic-Q-MT. Each cell refits the same outcome model within one model-by-$k$ stratum, omitting fixed effects inside the stratum. Color indicates the standardized log-odds ratio for MSS relative mention rate; asterisks mark nominal $p<0.05$. Controls that are constant within a cell are dropped from that cell's regression; cells are marked NA when they have fewer than 50 usable interactions, no outcome variation, no variation in MSS relative mention rate, or a separated fit.}
    \label{app_fig:turn1_cot_model_k_stability_logicq_multi}
\end{figure}

\paragraph{Logic-Q-MT results.}
MSS relative mention rate is positively associated with final sufficiency ($\beta_z=0.528$, $p=6.71\times 10^{-45}$), whereas awareness of underspecification is near zero ($\beta_z=0.005$, $p=0.853$) (Appendix Fig.~\ref{app_fig:turn1_cot_covariate_comparison_logicq_multi}). The same pattern holds in the turn-1-miss subset for second-turn hit: MSS relative mention rate remains positive ($\beta_z=0.365$, $p=7.99\times 10^{-23}$), while awareness again stays near zero ($\beta_z=0.010$, $p=0.774$). The model-by-$k$ stability diagnostic (Appendix Fig.~\ref{app_fig:turn1_cot_model_k_stability_logicq_multi}) shows 23 positive cells out of 23 estimable cells for final sufficiency and 23 positive cells out of 24 estimable cells for second-turn hit; the corresponding nominally significant positive counts are 16 and 16.

\begin{figure}[htp]
    \centering
    \includegraphics[width=0.8\linewidth]{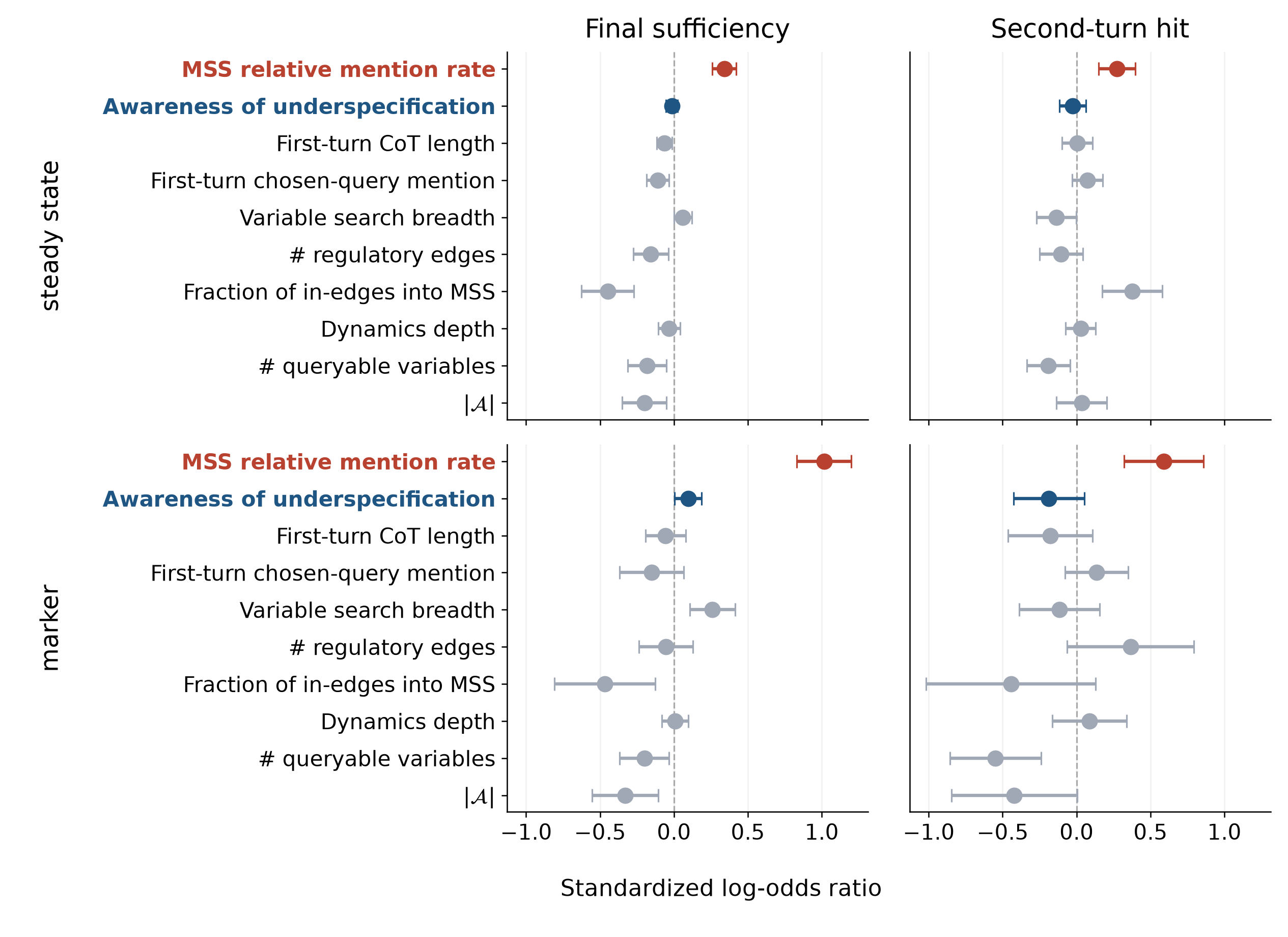}
    \caption{Turn-1 CoT predictors in GeneReg-MT. The specification matches Logic-Q-MT conceptually but is fit separately for steady-state and marker tasks and uses a narrower gene-query template set to reduce false positives from regulatory-rule enumeration. MSS relative mention rate remains positive in both task families for final sufficiency ($\beta_z=0.340$ for steady state; $\beta_z=1.016$ for marker) and for second-turn hit after a first-turn miss ($\beta_z=0.274$ for steady state; $\beta_z=0.590$ for marker).
    }
    \label{app_fig:turn1_cot_covariate_comparison_grn_multi}
\end{figure}

\begin{figure}[htp]
    \centering
    \includegraphics[width=1.\linewidth]{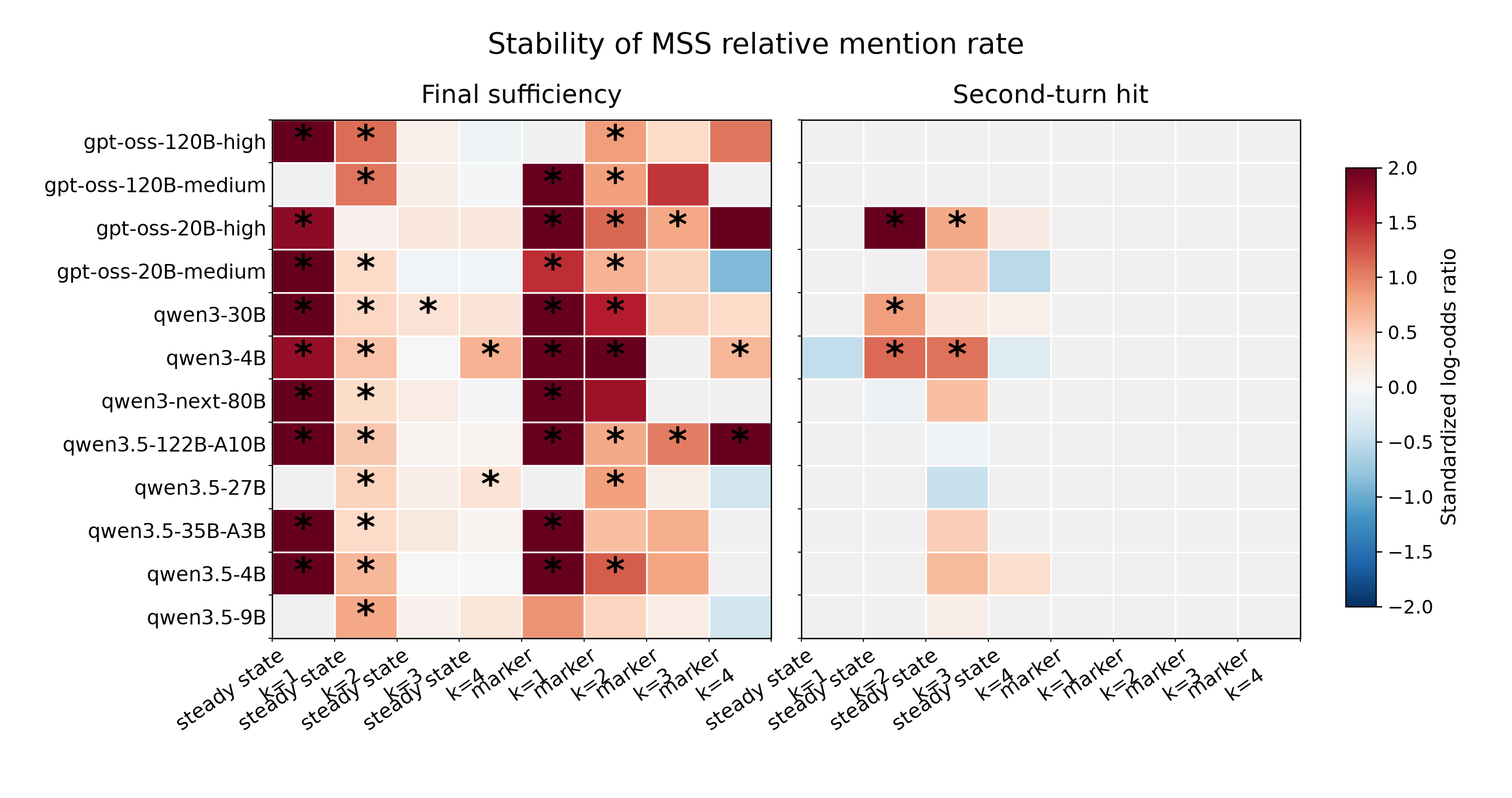}
    \caption{Stability of MSS relative mention rate in GeneReg-MT. Each cell is a separate model-task-$k$ regression with the same controls as the pooled model and heteroskedasticity-robust standard errors. Many recovery cells are NA because the outcome is defined only for interactions that both miss on turn 1 and continue to turn 2; the appendix text reports the exact NA criteria and summary counts.}
    \label{app_fig:turn1_cot_model_task_k_stability_grn_multi}
\end{figure}

\paragraph{GeneReg-MT results.}
GeneReg-MT is directionally supportive but less clean. For final sufficiency, MSS relative mention rate is positive in both steady-state ($\beta_z=0.340$, $p=5.39\times 10^{-17}$) and marker tasks ($\beta_z=1.016$, $p=3.77\times 10^{-27}$) (Appendix Fig.~\ref{app_fig:turn1_cot_covariate_comparison_grn_multi}). For second-turn hit after a first-turn miss, the same cue is again positive in steady-state ($\beta_z=0.274$, $p=1.52\times 10^{-5}$) and marker tasks ($\beta_z=0.590$, $p=1.59\times 10^{-5}$). However, the model-task-$k$ stability diagnostic is much sparser: 85 estimable cells for final sufficiency and 20 estimable cells for second-turn hit, with many non-estimable recovery cells produced by the sample restriction above (Appendix Fig.~\ref{app_fig:turn1_cot_model_task_k_stability_grn_multi}).

\paragraph{Adjustment sensitivity.}
We report the marginal effect (regression without problem factor) in addition to the fully adjusted model. The standardized log-odds ratio for MSS relative mention rate changes only modestly when problem factors are dropped: Logic-Q-MT final sufficiency 0.528 to 0.562, Logic-Q-MT second-turn hit 0.365 to 0.392, GeneReg-MT steady-state final sufficiency 0.340 to 0.334, GeneReg-MT marker final sufficiency 1.016 to 1.007, GeneReg-MT steady-state second-turn hit 0.274 to 0.342, and GeneReg-MT marker second-turn hit 0.590 to 0.620.
We also verify that in the fixed-effect design, the VIF for MSS relative mention rate ranges from 1.158 to 1.563 across all Logic-Q-MT and GeneReg-MT analysis frames. Thus it is not collinear with the MSS-incidence control.

\subsubsection{Example CoTs}
\label{app:logicq_multi_example_cot}

\newcommand{\awarenesshl}[1]{\colorbox{yellow!25}{#1}}
\newcommand{\msshl}[1]{\colorbox{cyan!20}{#1}}

\paragraph{Annotated first-turn CoT examples.}
All excerpts below are real first-turn CoTs from original Logic-Q-MT forbid-alternatives runs. We report the raw target, turn-1 query, turn-2 query, and the MSS so each excerpt can be traced back to the source result JSON.

\emph{Qwen3-4B-Thinking, target = thoughtless, turn-1 query = exuberant, turn-2 query = straightforward, MSS = \{straightforward, weary\}.}
\begin{quote}\small
``I need to ask about an attribute that is part of the antecedent for thoughtless. Let's consider querying \msshl{`weary'} because it's in the second thoughtless rule. If I ask \msshl{`Is Alice weary?'} and she is, then I need to know if she is distinct and \msshl{straightforward}.''
\end{quote}

\emph{Qwen3-30B-A3B-Thinking-FP8, target = smart, turn-1 query = proud, turn-2 query = plain, canonical MSS = \{long, plain, thoughtless\}.}
\begin{quote}\small
``...The most direct rule is `plain and proud -> smart.' So we need both \msshl{plain} and proud. We can't ask both, so we need to ask one that helps us get closer. If we ask about proud [...] Similarly, \msshl{if we ask about `plain'}, and she's plain, then we need to check if she's proud...''
\end{quote}

\emph{Qwen3.5-122B-A10B-FP8, target = gifted, turn-1 query = talented, turn-2 query = mean, MSS = \{careless, mean\}.}
\begin{quote}\small
``...Path 1: `blushing' AND `mean' [...] Forbidden to ask: blushing, courageous, cruel, cute, easy, fancy, gifted, helpful, plain, rational, shiny. Allowed to ask: calm, careless, frail, talented, [...] mean. [...] \msshl{If I ask `careless'}: No -> `gifted' False. Yes -> need `talented' AND `uptight'...''
\end{quote}

\emph{gpt-oss-20B-high, target = elegant, turn-1 query = versatile, turn-2 query = busy, MSS = \{busy, sincere, uptight\}.}
\begin{quote}\small
``...The rule that concludes elegant is rule 9: busy \& elated \& uptight => elegant. So to prove elegant, we need to prove that all three of those hold. So we could \msshl{ask about `busy'}, `elated', or \msshl{`uptight'}. But we cannot ask about elated (forbidden)...''
\end{quote}

\emph{gpt-oss-120B-high, target = tame, turn-1 query = rational, turn-2 query = light, MSS = \{light, straightforward, talkative\}.}
\begin{quote}\small
``...Rule 18: calm \& light \& talkative => adorable. [...] Thus we might want to \msshl{ask about talkative and light}. Light is allowed (not prohibited). Talkative is allowed. If we ask \msshl{`Is Alice talkative?'} [...] Alternatively, we could \msshl{ask about light} directly...''
\end{quote}

\subsubsection{Additional results}

\begin{figure}[htp]
    \centering
    \includegraphics[width=0.8\linewidth]{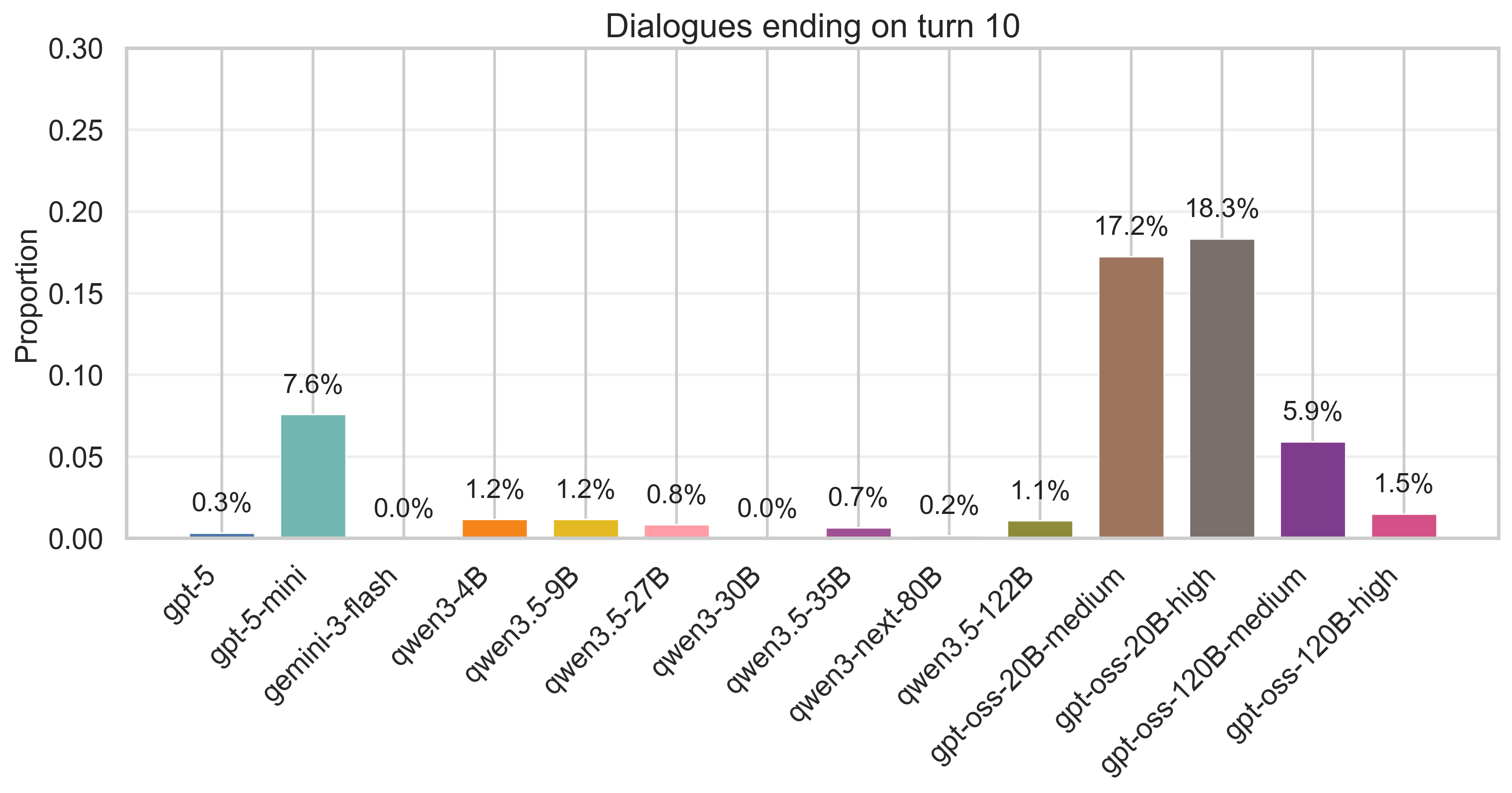}
    \caption{Proportions of interactions that end at 10 turns for models in Logic-Q-MT task-solving, with no budget in prompt but capped at 10 turns.}
    \label{app_fig:logicq_multi_multiturn_turn10_end_proportion}
\end{figure}

Fig.~\ref{app_fig:logicq_multi_multiturn_turn10_end_proportion} shows the proportion of interactions that reach the 10-turn cap; complementarily, Appendix~\ref{app:force_ask} tests an intervention that rejects answer attempts issued before $k$ queries, probing whether premature stopping is correctable under explicit feedback.

\begin{figure}[htp]
    \centering
    \includegraphics[width=0.8\linewidth]{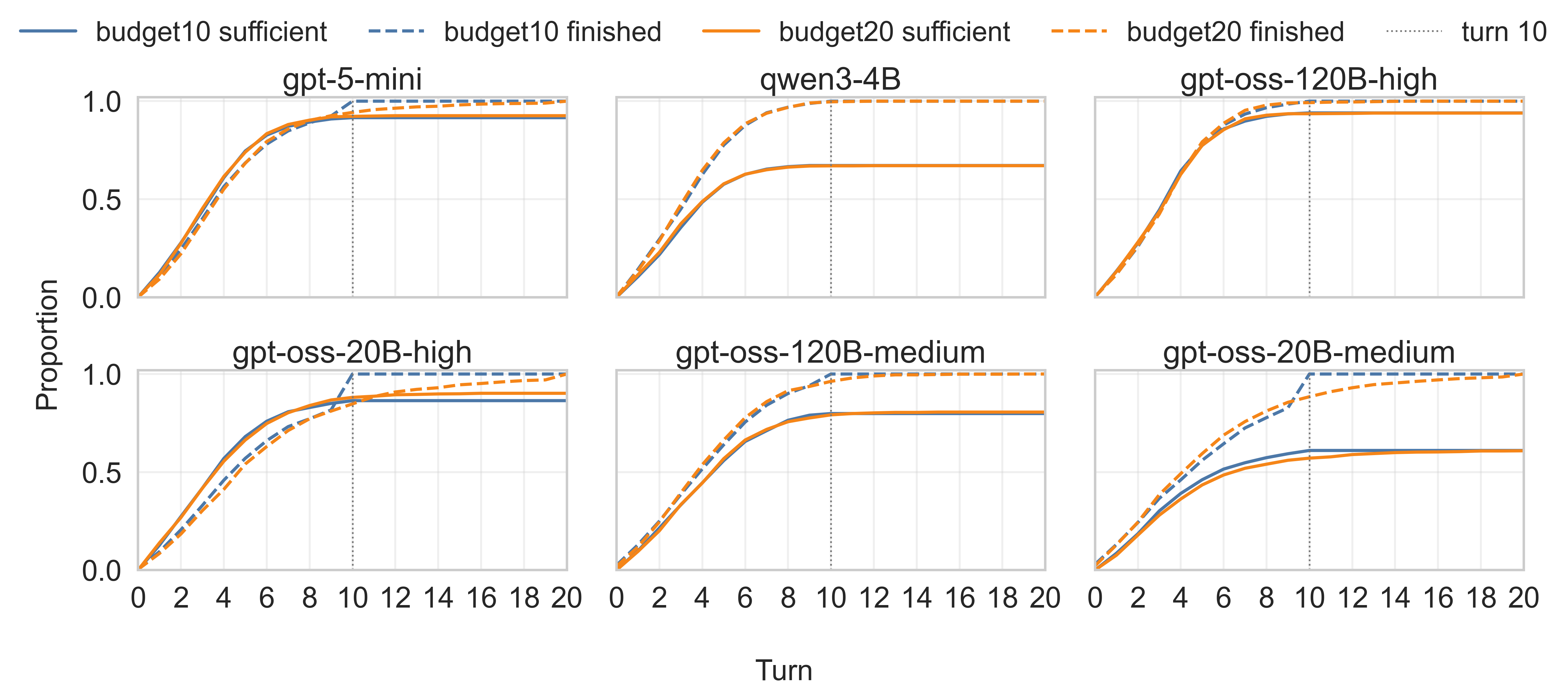}
    \caption{Performance of representative models with interactions ending on turn 10 in Logic-Q-MT task-solving, with no budget in prompt but capped at 10 or 20 turns.}
    \label{app_fig:logicq_multi_multiturn_budget20_vs_budget10}
\end{figure}

\begin{figure}[htp]
    \centering
    \includegraphics[width=0.6\linewidth]{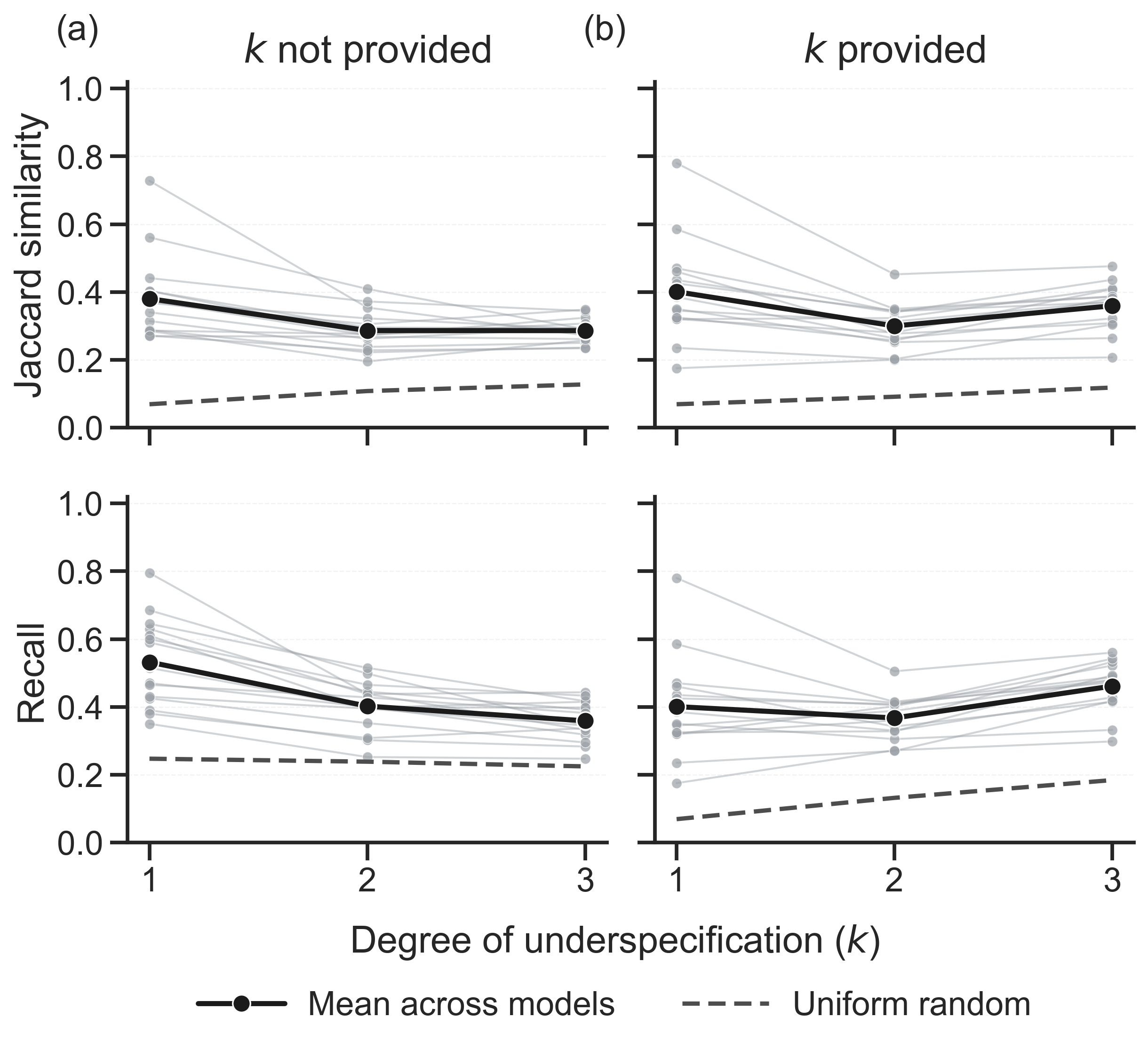}
    \caption{Recall with regard to the MSS in predicting missing variables. Same models and settings as Fig.~\ref{fig:logicq_multi_mc}; Jaccard similarity and recall are reported as a complement to accuracy. (a) $k$ not provided; (b) $k$ provided. Recall is lower when $k$ is provided for $k{=}2,3$ because forcing exactly $k$ selections decreases the chance of overlap with a $k$-MSS, compared with being allowed to predict a larger set of variables.}
    \label{app_fig:logicq_multi_mc_jaccard_recall}
\end{figure}

\begin{table}[htp]
\centering
\small
\caption{
Per-model accuracy of predicting the degree of underspecification $k$ in Logic-Q-MT (multiple-choice $k$-prediction; cf.\ Fig.~\ref{fig:logicq_multi_level_underspec}a).
Each cell reports exact-match accuracy over 200 problems per $k$; predictions of ``not ambiguous,'' ``not sure,'' and unparseable responses are counted as incorrect.
}
\label{app_tab:logic_q_k_prediction}
\setlength{\tabcolsep}{5pt}
\renewcommand{\arraystretch}{1}
\begin{tabular}{lccc}
\toprule
Model & $k=1$ & $k=2$ & $k=3$ \\
\midrule
\texttt{Qwen3-30B-A3B-Thinking-FP8}      & 0.720 & 0.505 & 0.405 \\
\texttt{Qwen3-30B-A3B-Thinking}          & 0.710 & 0.495 & 0.330 \\
\texttt{Qwen3-4B-Thinking}               & 0.770 & 0.340 & 0.125 \\
\texttt{Qwen3-Next-80B-A3B-Thinking-FP8} & 0.550 & 0.390 & 0.290 \\
\texttt{Qwen3.5-122B-A10B-FP8}           & 0.835 & 0.465 & 0.265 \\
\texttt{Qwen3.5-27B}                     & 0.785 & 0.330 & 0.200 \\
\texttt{Qwen3.5-27B-FP8}                 & 0.825 & 0.395 & 0.200 \\
\texttt{Qwen3.5-35B-A3B}                 & 0.900 & 0.365 & 0.190 \\
\texttt{Qwen3.5-35B-A3B-FP8}             & 0.900 & 0.390 & 0.235 \\
\texttt{Qwen3.5-4B}                      & 0.735 & 0.280 & 0.240 \\
\texttt{Qwen3.5-9B}                      & 0.845 & 0.355 & 0.155 \\
\texttt{GPT-5}                           & 0.935 & 0.500 & 0.255 \\
\texttt{GPT-5-mini}                      & 0.875 & 0.445 & 0.185 \\
\texttt{gpt-oss-120B-high}              & 0.875 & 0.440 & 0.185 \\
\texttt{gpt-oss-120B-medium}            & 0.765 & 0.520 & 0.250 \\
\texttt{gpt-oss-20B-high}               & 0.915 & 0.435 & 0.275 \\
\texttt{gpt-oss-20B-medium}             & 0.780 & 0.430 & 0.335 \\
\bottomrule
\end{tabular}
\end{table}

\begin{table*}[htp]
\centering
\small
\caption{
Performance of LLMs on Logic-Q-MT missing variable and full-information problem-solving. 
We report missing-variable prediction performance when $k$ is not provided or provided, and full-information problem-solving correctness as a reference upper bound for sequential task-solving.
}
\label{app_tab:logic_q_all_conditions}
\setlength{\tabcolsep}{3.5pt}
\renewcommand{\arraystretch}{1}
\begin{tabular}{ll|ccc|ccc|c}
\toprule
\multirow{2}{*}{Model} & \multirow{2}{*}{$k$}
& \multicolumn{3}{c|}{$k$ not provided}
& \multicolumn{3}{c|}{$k$ provided}
& \multirow{2}{*}{Full Info.} \\
\cmidrule(lr){3-5} \cmidrule(lr){6-8}
& & Acc. & Jac. & Rec. & Acc. & Jac. & Rec. & Acc. \\
\midrule
\multirow{3}{*}{\texttt{Qwen3-30B-A3B-Thinking-FP8}}
& 1 & 0.250 & 0.314 & 0.390 & 0.320 & 0.320 & 0.320 & 0.885 \\
& 2 & 0.070 & 0.238 & 0.302 & 0.165 & 0.323 & 0.402 & 0.948 \\
& 3 & 0.040 & 0.250 & 0.283 & 0.135 & 0.410 & 0.522 & 0.976 \\
\midrule
\multirow{3}{*}{\texttt{Qwen3-4B-Thinking}}
& 1 & 0.230 & 0.288 & 0.350 & 0.320 & 0.320 & 0.320 & 0.855 \\
& 2 & 0.065 & 0.222 & 0.252 & 0.120 & 0.280 & 0.360 & 0.959 \\
& 3 & 0.030 & 0.236 & 0.247 & 0.095 & 0.408 & 0.533 & 0.852 \\
\midrule
\multirow{3}{*}{\texttt{Qwen3-Next-80B-A3B-Thinking-FP8}}
& 1 & 0.295 & 0.372 & 0.470 & 0.345 & 0.345 & 0.345 & 0.780 \\
& 2 & 0.115 & 0.322 & 0.400 & 0.165 & 0.313 & 0.388 & 0.844 \\
& 3 & 0.055 & 0.291 & 0.318 & 0.115 & 0.371 & 0.478 & 0.808 \\
\midrule
\multirow{3}{*}{\texttt{Qwen3.5-122B-A10B-FP8}}
& 1 & 0.235 & 0.374 & 0.630 & 0.435 & 0.435 & 0.435 & -- \\
& 2 & 0.065 & 0.272 & 0.438 & 0.210 & 0.340 & 0.405 & -- \\
& 3 & 0.075 & 0.324 & 0.443 & 0.175 & 0.436 & 0.542 & -- \\
\midrule
\multirow{3}{*}{\texttt{Qwen3.5-27B-FP8}}
& 1 & 0.300 & 0.403 & 0.590 & 0.470 & 0.470 & 0.470 & 1.000 \\
& 2 & 0.060 & 0.295 & 0.445 & 0.205 & 0.343 & 0.412 & 1.000 \\
& 3 & 0.050 & 0.293 & 0.392 & 0.130 & 0.370 & 0.472 & 1.000 \\
\midrule
\multirow{3}{*}{\texttt{Qwen3.5-35B-A3B-FP8}}
& 1 & 0.235 & 0.285 & 0.380 & 0.350 & 0.350 & 0.350 & 0.990 \\
& 2 & 0.030 & 0.195 & 0.308 & 0.145 & 0.252 & 0.305 & 1.000 \\
& 3 & 0.065 & 0.257 & 0.337 & 0.100 & 0.264 & 0.332 & 1.000 \\
\midrule
\multirow{3}{*}{\texttt{Qwen3.5-9B}}
& 1 & 0.245 & 0.340 & 0.515 & 0.425 & 0.425 & 0.425 & 0.995 \\
& 2 & 0.095 & 0.262 & 0.398 & 0.220 & 0.345 & 0.408 & 1.000 \\
& 3 & 0.035 & 0.294 & 0.415 & 0.165 & 0.386 & 0.480 & 0.980 \\
\midrule
\multirow{3}{*}{\texttt{Gemini-3-Flash}}
& 1 & 0.695 & 0.728 & 0.795 & 0.780 & 0.780 & 0.780 & -- \\
& 2 & 0.220 & 0.354 & 0.438 & 0.345 & 0.452 & 0.505 & -- \\
& 3 & 0.135 & 0.272 & 0.332 & 0.285 & 0.476 & 0.560 & -- \\
\midrule
\multirow{3}{*}{\texttt{Gemini-3.1-Pro}}
& 1 & 0.475 & 0.561 & 0.685 & -- & -- & -- & -- \\
& 2 & 0.285 & 0.409 & 0.498 & -- & -- & -- & -- \\
& 3 & 0.145 & 0.290 & 0.345 & -- & -- & -- & -- \\
\midrule
\multirow{3}{*}{\texttt{GPT-5}}
& 1 & 0.310 & 0.441 & 0.645 & 0.585 & 0.585 & 0.585 & -- \\
& 2 & 0.185 & 0.372 & 0.515 & 0.220 & 0.350 & 0.415 & -- \\
& 3 & 0.175 & 0.344 & 0.417 & 0.165 & 0.387 & 0.488 & -- \\
\midrule
\multirow{3}{*}{\texttt{GPT-5-mini}}
& 1 & 0.270 & 0.403 & 0.610 & 0.460 & 0.460 & 0.460 & -- \\
& 2 & 0.110 & 0.276 & 0.398 & 0.145 & 0.277 & 0.342 & -- \\
& 3 & 0.095 & 0.307 & 0.398 & 0.075 & 0.308 & 0.415 & -- \\
\midrule
\multirow{3}{*}{\texttt{gpt-oss-120B-high}}
& 1 & 0.230 & 0.368 & 0.600 & 0.325 & 0.325 & 0.325 & 1.000 \\
& 2 & 0.090 & 0.305 & 0.465 & 0.115 & 0.257 & 0.328 & 0.990 \\
& 3 & 0.130 & 0.348 & 0.433 & 0.130 & 0.382 & 0.493 & 1.000 \\
\midrule
\multirow{3}{*}{\texttt{gpt-oss-120B-medium}}
& 1 & 0.170 & 0.271 & 0.425 & 0.175 & 0.175 & 0.175 & 0.975 \\
& 2 & 0.045 & 0.228 & 0.352 & 0.055 & 0.200 & 0.272 & 0.957 \\
& 3 & 0.040 & 0.234 & 0.295 & 0.020 & 0.207 & 0.298 & 0.955 \\
\midrule
\multirow{3}{*}{\texttt{gpt-oss-20B-high}}
& 1 & 0.175 & 0.286 & 0.465 & 0.385 & 0.385 & 0.385 & 0.985 \\
& 2 & 0.075 & 0.276 & 0.428 & 0.130 & 0.263 & 0.330 & 0.981 \\
& 3 & 0.070 & 0.287 & 0.382 & 0.080 & 0.322 & 0.428 & 0.977 \\
\midrule
\multirow{3}{*}{\texttt{gpt-oss-20B-medium}}
& 1 & 0.160 & 0.270 & 0.430 & 0.235 & 0.235 & 0.235 & 0.800 \\
& 2 & 0.085 & 0.268 & 0.395 & 0.065 & 0.202 & 0.270 & 0.830 \\
& 3 & 0.040 & 0.261 & 0.345 & 0.055 & 0.304 & 0.417 & 0.898 \\
\bottomrule
\end{tabular}
\end{table*}

\begin{figure}[htp]
    \centering
    \includegraphics[width=1.0\linewidth]{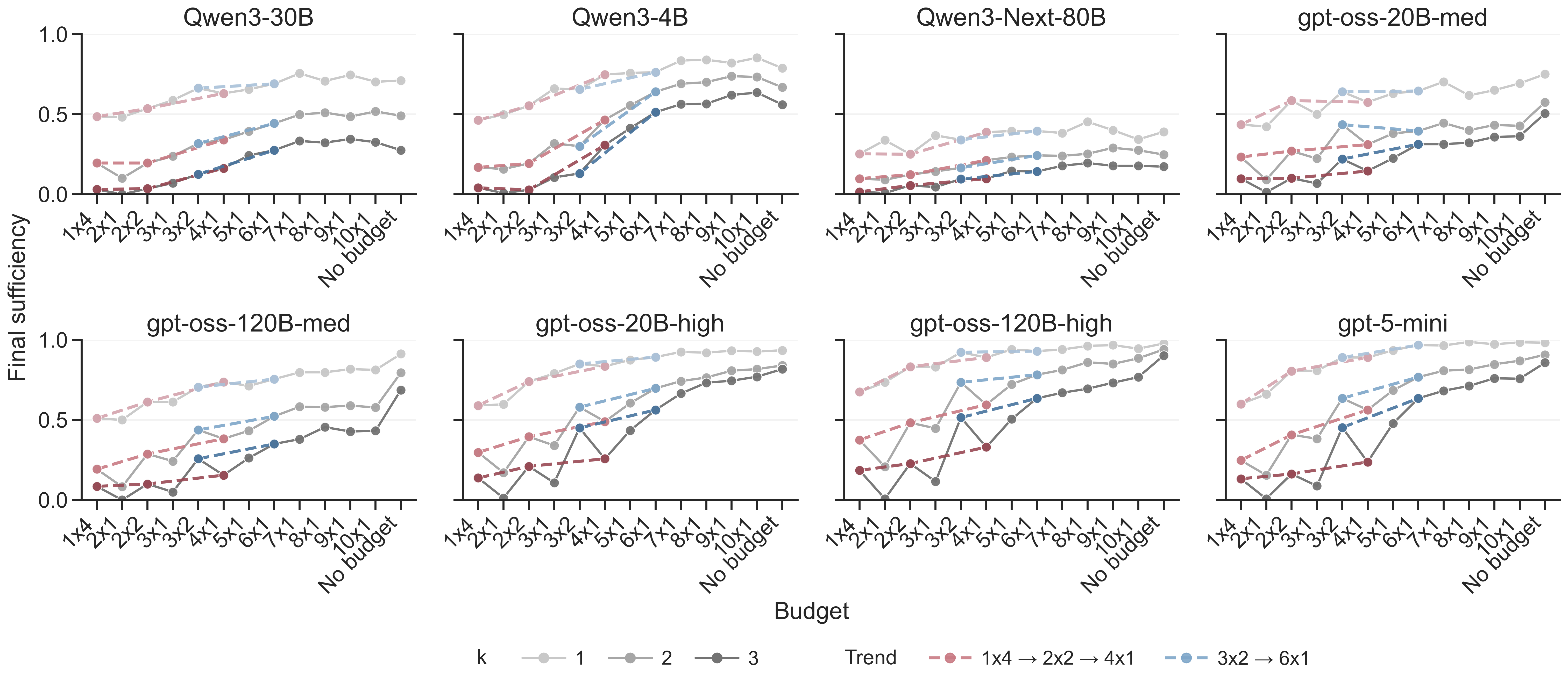}
    \caption{Performance of LLMs in Logic-Q-MT task-solving with varying budget in prompt across models. Budgets are set in the form of ``number of turns allowed'' x ``number of queries per turn allowed''. ``No budget'' refers to the setting where no budget is mentioned in the prompt. To avoid infinite loops, we cap the number of turns in the ``No budget'' setup at 10. This cap does not significantly affect final sufficiency (Fig.~\ref{app_fig:logicq_multi_multiturn_budget20_vs_budget10}). With adversarial oracles. }
    \label{app_fig:logicq_multi_multiturn_budget_exp}
\end{figure}

\begin{figure}[htp]
    \centering
    \includegraphics[width=1.0\linewidth]{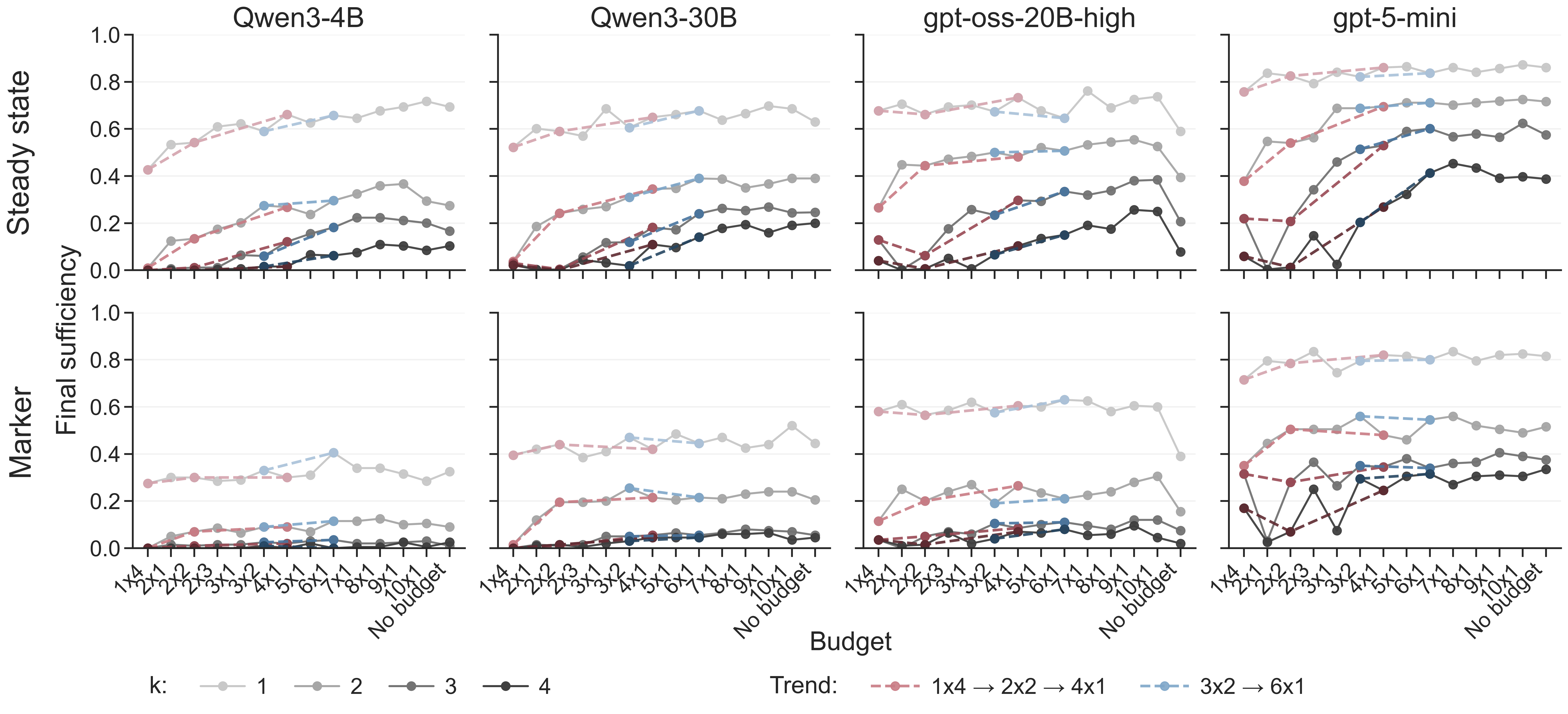}
    \caption{Performance of LLMs in GeneReg-MT marker and steady state identification task-solving with varying budget in prompt across models. Same setup as in Fig.~\ref{app_fig:logicq_multi_multiturn_budget_exp}.}
    \label{app_fig:genereg_multi_multiturn_budget_exp}
\end{figure}

\begin{figure}[htp]
    \centering
    \includegraphics[width=1.0\linewidth]{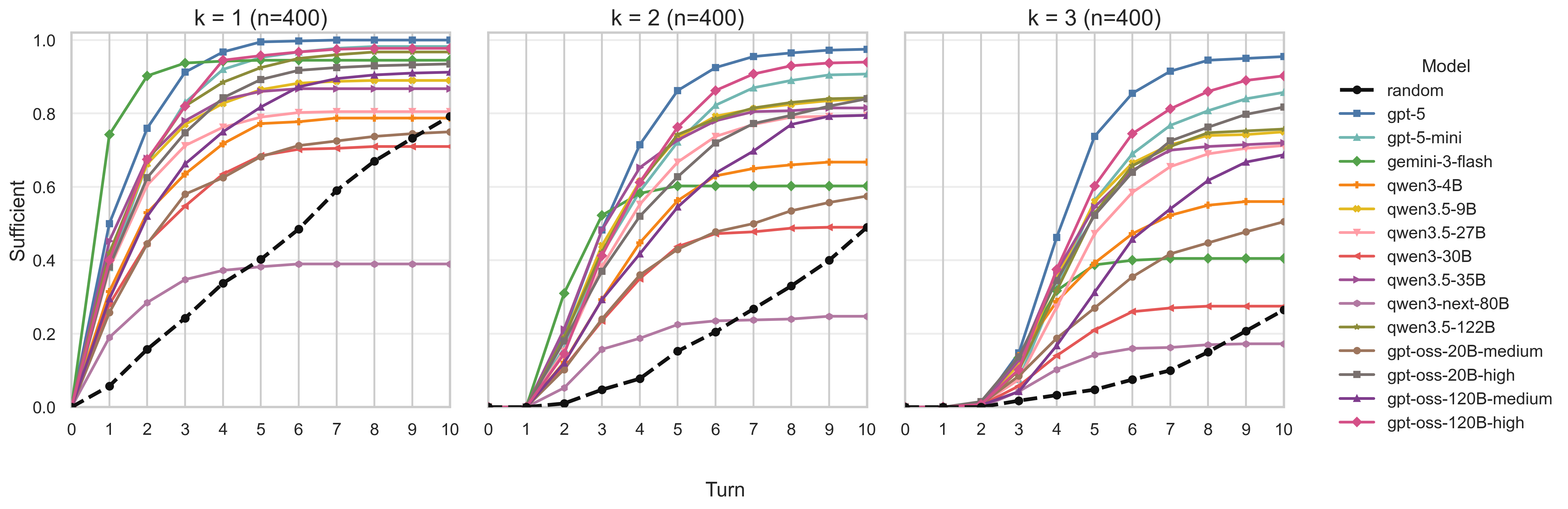}
    \caption{Proportion of tasks that have reached sufficiency at turn $t$ per model, across each k, for Logic-Q-MT. }
    \label{app_fig:logicq_multi_multiturn_sufficient_proportion_vs_turn}
\end{figure}

\begin{figure}[htp]
    \centering
    \includegraphics[width=0.8\linewidth]{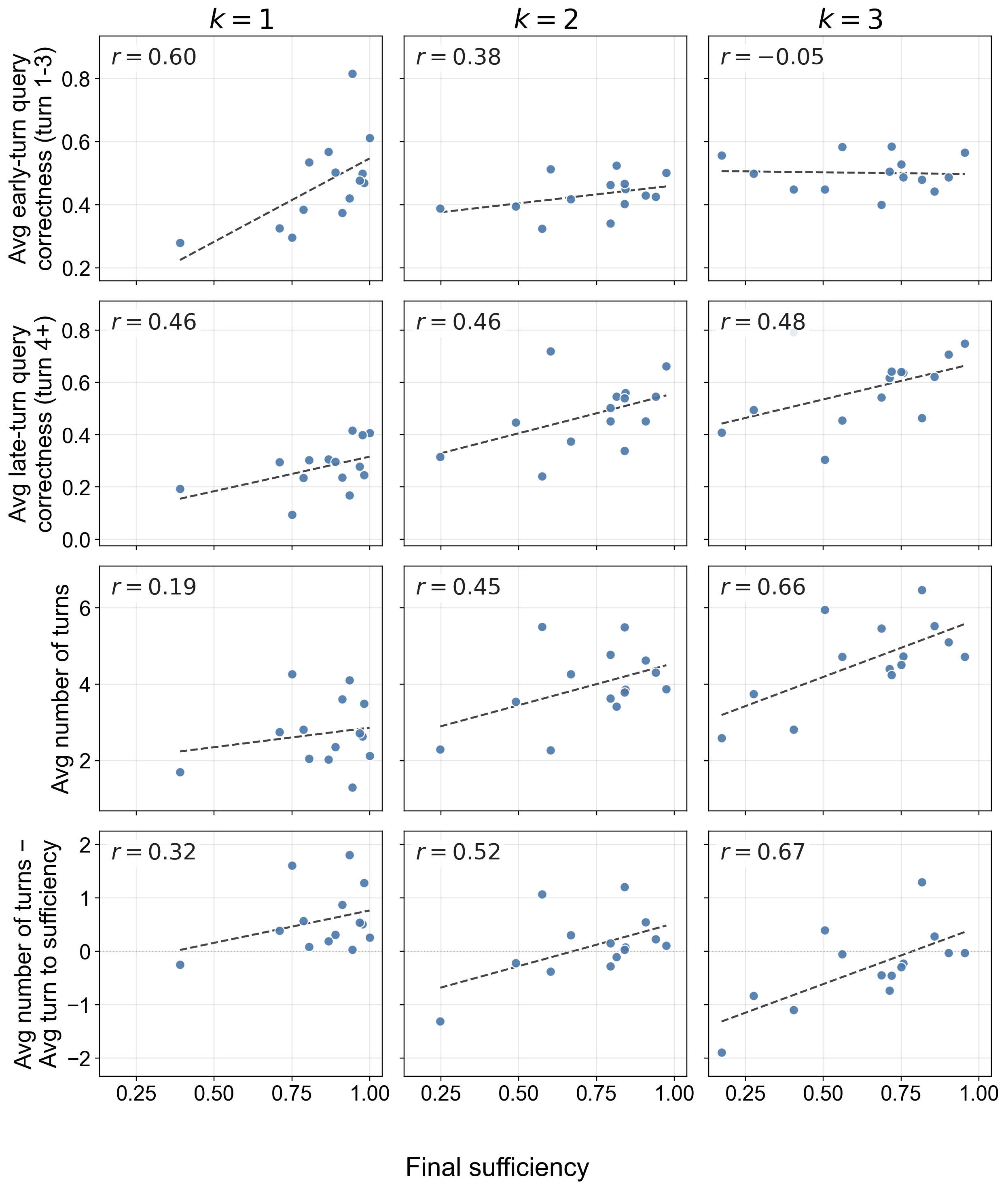}
    \caption{Scatterplot underlying every cell of Fig.~\ref{fig:logicq_multi_multiturn_corr_model_behavior_vs_model_performance_heatmap}, plus a control row, in Logic-Q-MT. Each panel plots one of four behavioral features against final sufficiency; one dot per model, columns vary the true degree of underspecification $k$, the dashed grey line is the OLS fit, and the upper-left annotation reports the Pearson $r$. The first three rows correspond directly to the cells of the heatmap. The bottom row, avg number of turns - avg turn to sufficiency, is a control: it counts the turns each model uses past its first-sufficient turn, isolating the ``keeps asking after it could stop'' component from the raw turn count. Its correlation with final sufficiency strengthens with $k$ ($r=0.32 \to 0.67$), mirroring the avg-number-of-turns row above and showing that on harder problems the benefit of longer interactions persists even after netting out when sufficiency is first reached. }
    \label{app_fig:logicq_multi_multiturn_corr_model_behavior_vs_performance_scatterplot}
\end{figure}

\begin{figure}[htp]
    \centering
    \includegraphics[width=0.6\linewidth]{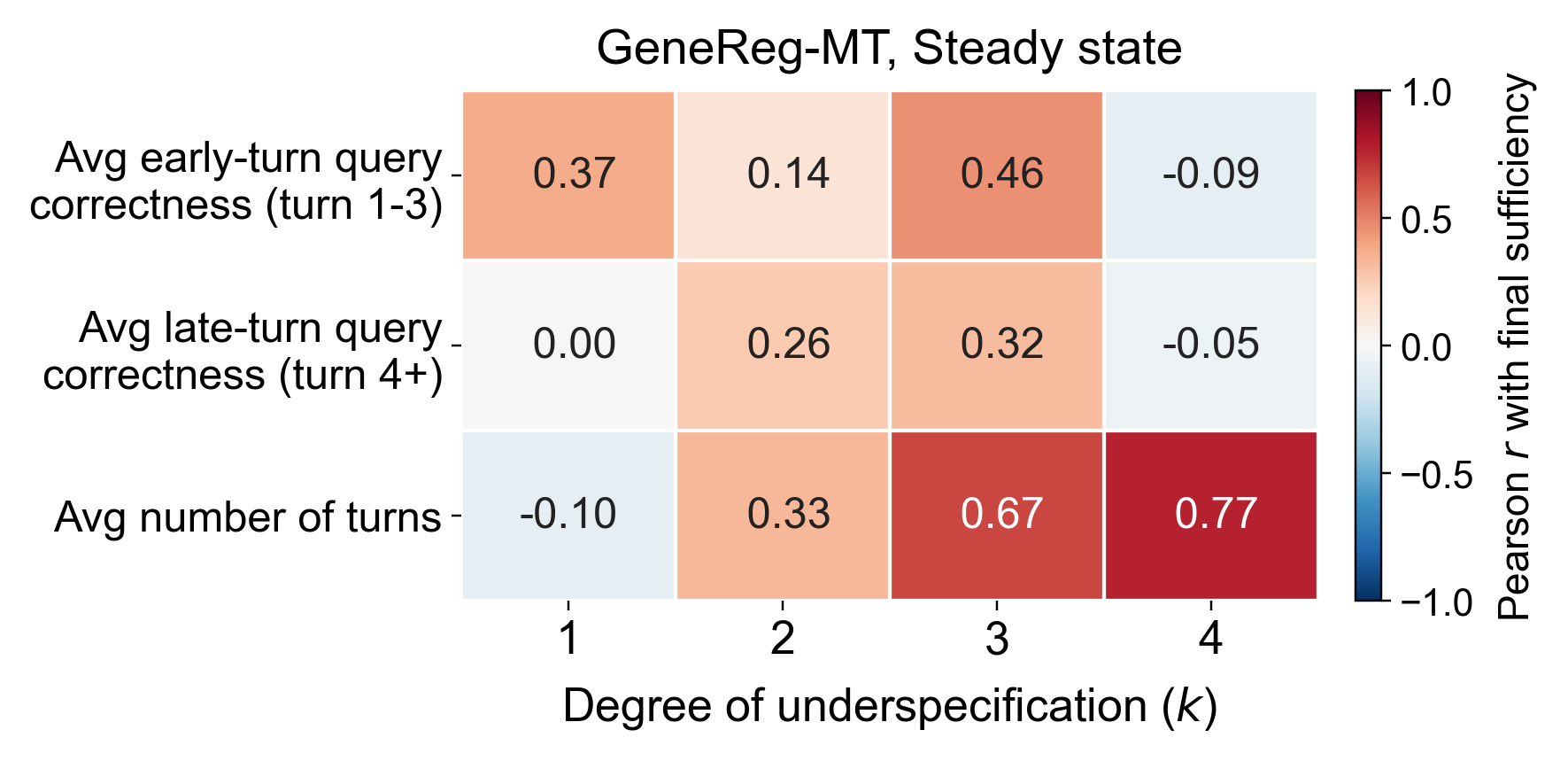}
    \caption{Heatmap of correlations between per-model behavioral features with average final sufficiency in GeneReg-MT dynamic steady state identification. Each cell is the Pearson $r$ between the behavioral feature and average final sufficiency, computed across models within a fixed $k$.}
    \label{app_fig:genereg_multi_dyn_attr_multiturn_correlation_model_behavior_vs_suff_heatmap}
\end{figure}

\begin{figure}[htp]
    \centering
    \includegraphics[width=0.6\linewidth]{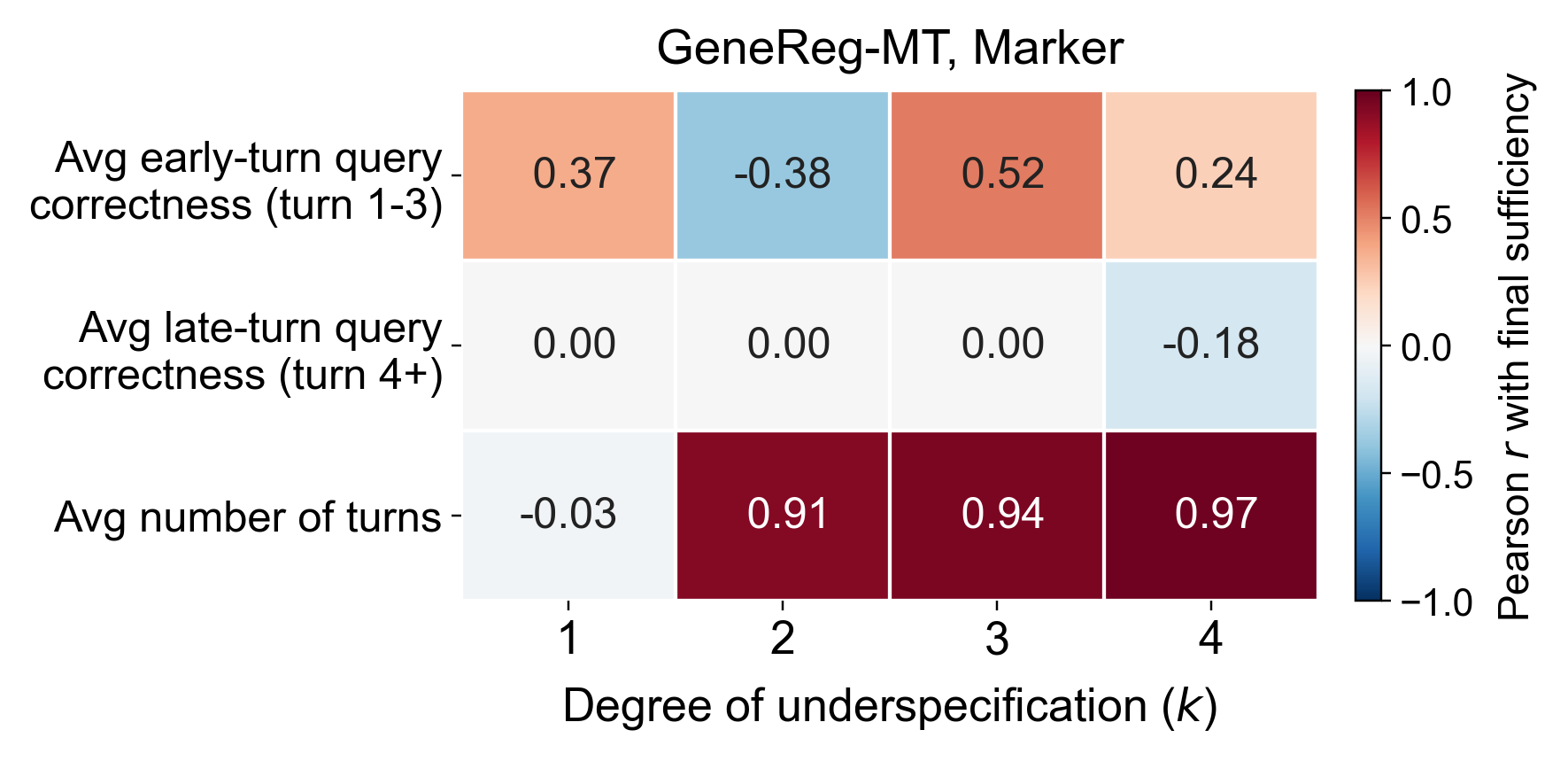}
    \caption{Heatmap of correlations between per-model behavioral features with average final sufficiency in GeneReg-MT dynamic marker identification. Each cell is the Pearson $r$ between the behavioral feature and average final sufficiency, computed across models within a fixed $k$. 
}
    \label{app_fig:genereg_multi_dyn_marker_multiturn_correlation_model_behavior_vs_suff_heatmap}
\end{figure}

\begin{figure}[htp]
    \centering
    \includegraphics[width=1.\linewidth]{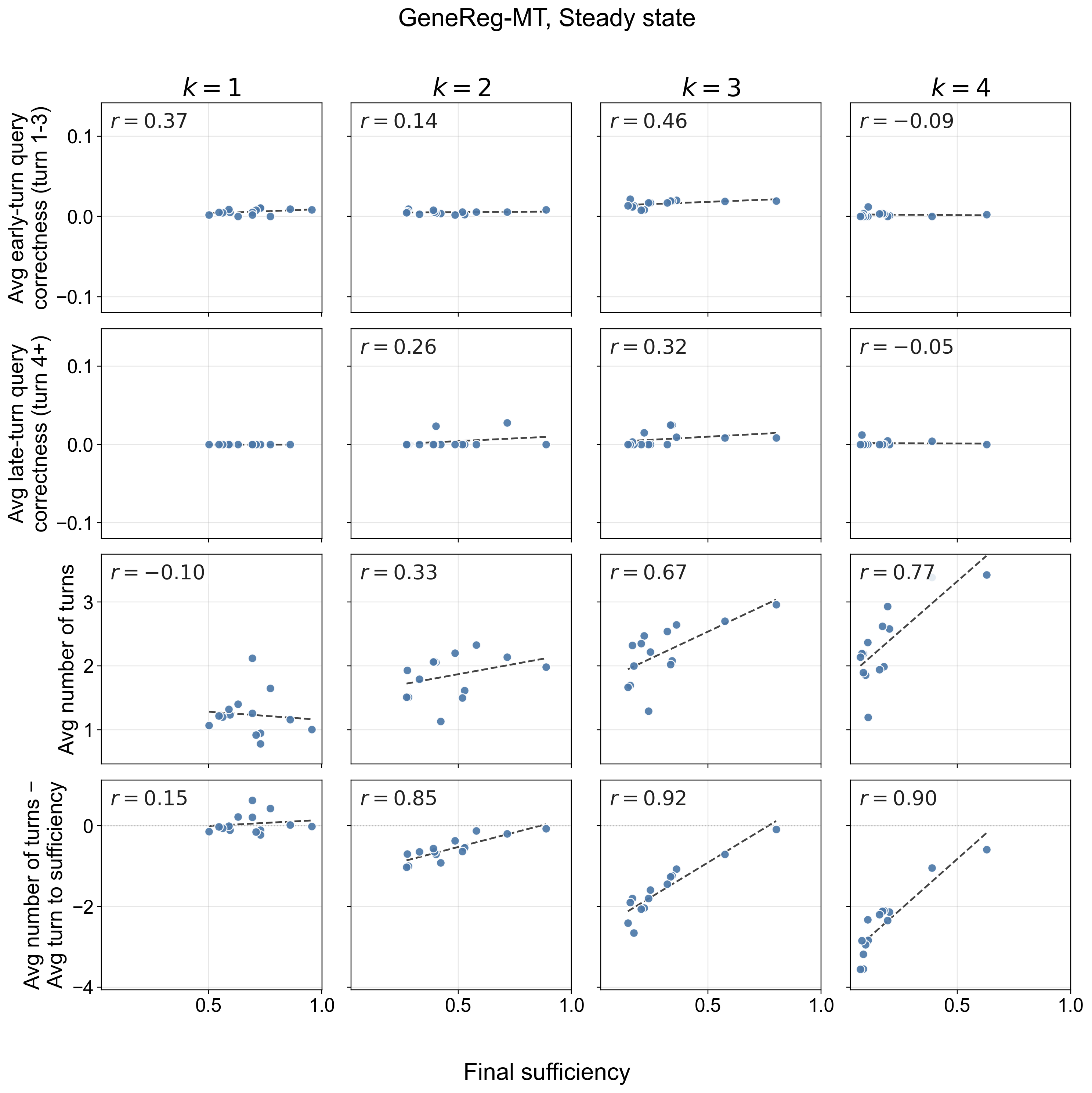}
    \caption{Scatterplot underlying every cell of Fig.~\ref{app_fig:genereg_multi_dyn_attr_multiturn_correlation_model_behavior_vs_suff_heatmap}, plus a control row, in GeneReg-MT dynamic steady state identification. Each panel plots one of four behavioral features against final sufficiency; one dot per model, columns vary the true degree of underspecification $k$, the dashed grey line is the OLS fit, and the upper-left annotation reports the Pearson $r$. The first three rows correspond directly to the cells of the heatmap. The bottom row, avg number of turns - avg turn to sufficiency, is a control: it counts the turns each model uses past its first-sufficient turn, isolating the ``keeps asking after it could stop'' component from the raw turn count. 
}
    \label{app_fig:genereg_multi_dyn_attr_multiturn_corr_model_behavior_vs_performance_scatterplot}
\end{figure}

\begin{figure}[htp]
    \centering
    \includegraphics[width=1.\linewidth]{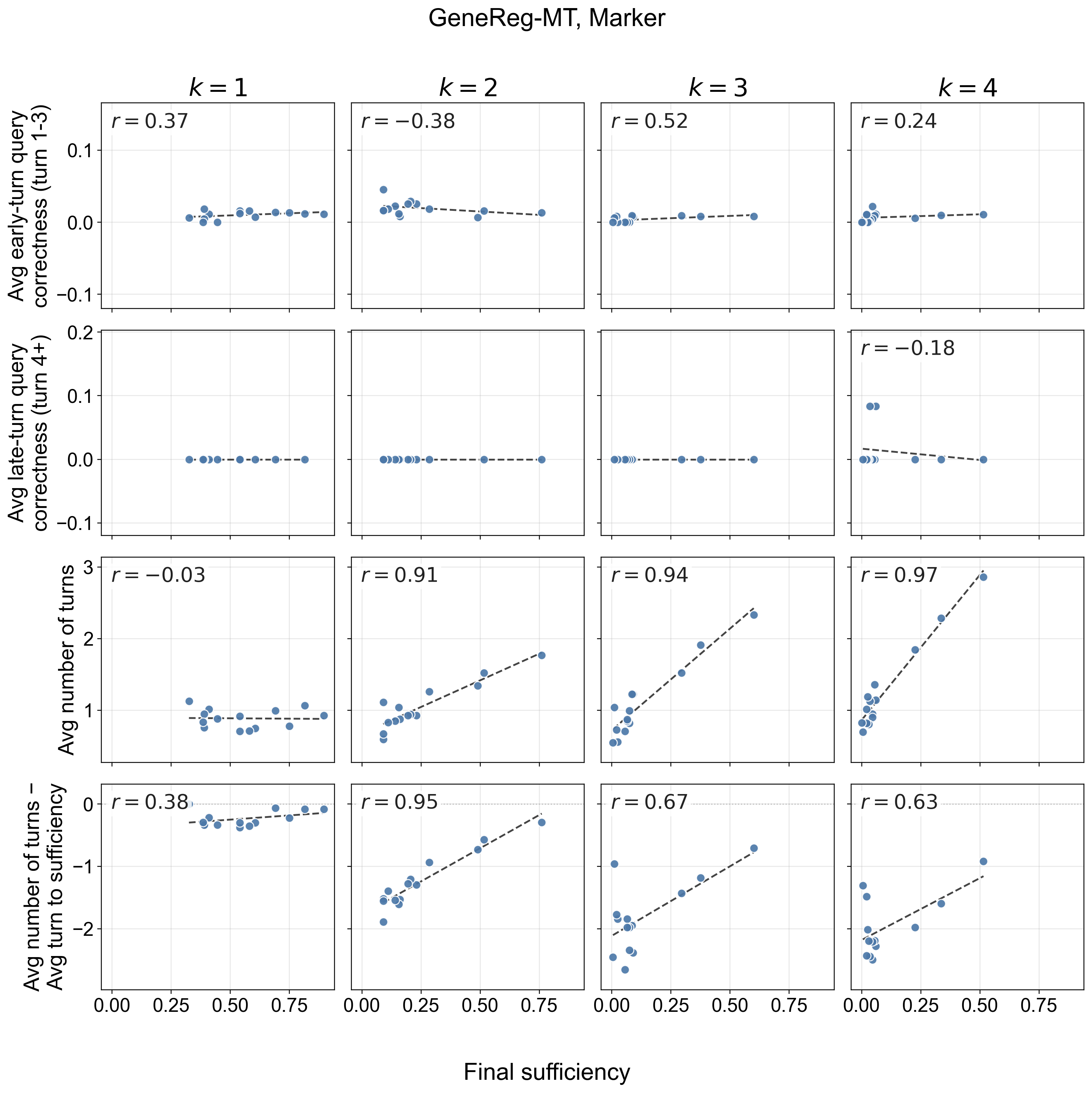}
    \caption{Scatterplot underlying every cell of Fig.~\ref{app_fig:genereg_multi_dyn_marker_multiturn_correlation_model_behavior_vs_suff_heatmap}, plus a control row, in GeneReg-MT dynamic marker identification. Each panel plots one of four behavioral features against final sufficiency; one dot per model, columns vary the true degree of underspecification $k$, the dashed grey line is the OLS fit, and the upper-left annotation reports the Pearson $r$. The first three rows correspond directly to the cells of the heatmap. The bottom row, avg number of turns - avg turn to sufficiency, is a control: it counts the turns each model uses past its first-sufficient turn, isolating the ``keeps asking after it could stop'' component from the raw turn count. 
}
    \label{app_fig:genereg_multi_dyn_marker_multiturn_corr_model_behavior_vs_performance_scatterplot}
\end{figure}

\begin{figure}[htp]
    \centering
    \includegraphics[width=1.0\linewidth]{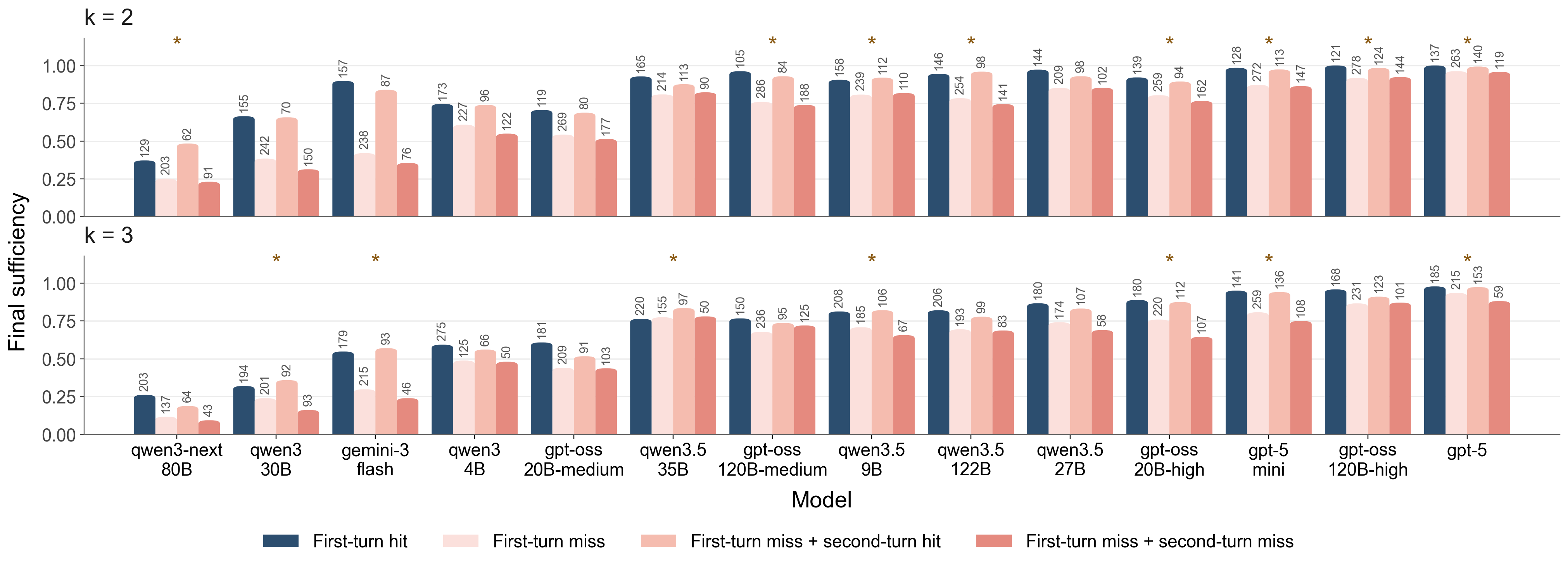}
    \vspace{-10pt}
    \caption{
    Recovery from a wrong first-turn query in $k=2,3$ Logic-Q-MT tasks. Bars show final sufficiency for each model, partitioned by whether the model queries a correct variable in the first and second turns. 
    For every model, the first-turn miss + second-turn hit (recovery) group performs within 10 percentage points of the first-turn hit group. 
    A $\star$ marks models for which this near-parity is statistically supported by a one-sided Miettinen–Nurminen non-inferiority test with a 10pp margin ($p<0.05$). 
    }
    \label{fig:logicq_multi_muliturn_early_recovery_allk_gap10pp}
\end{figure}

\begin{figure}[htp]
    \centering
    \includegraphics[width=1.0\linewidth]{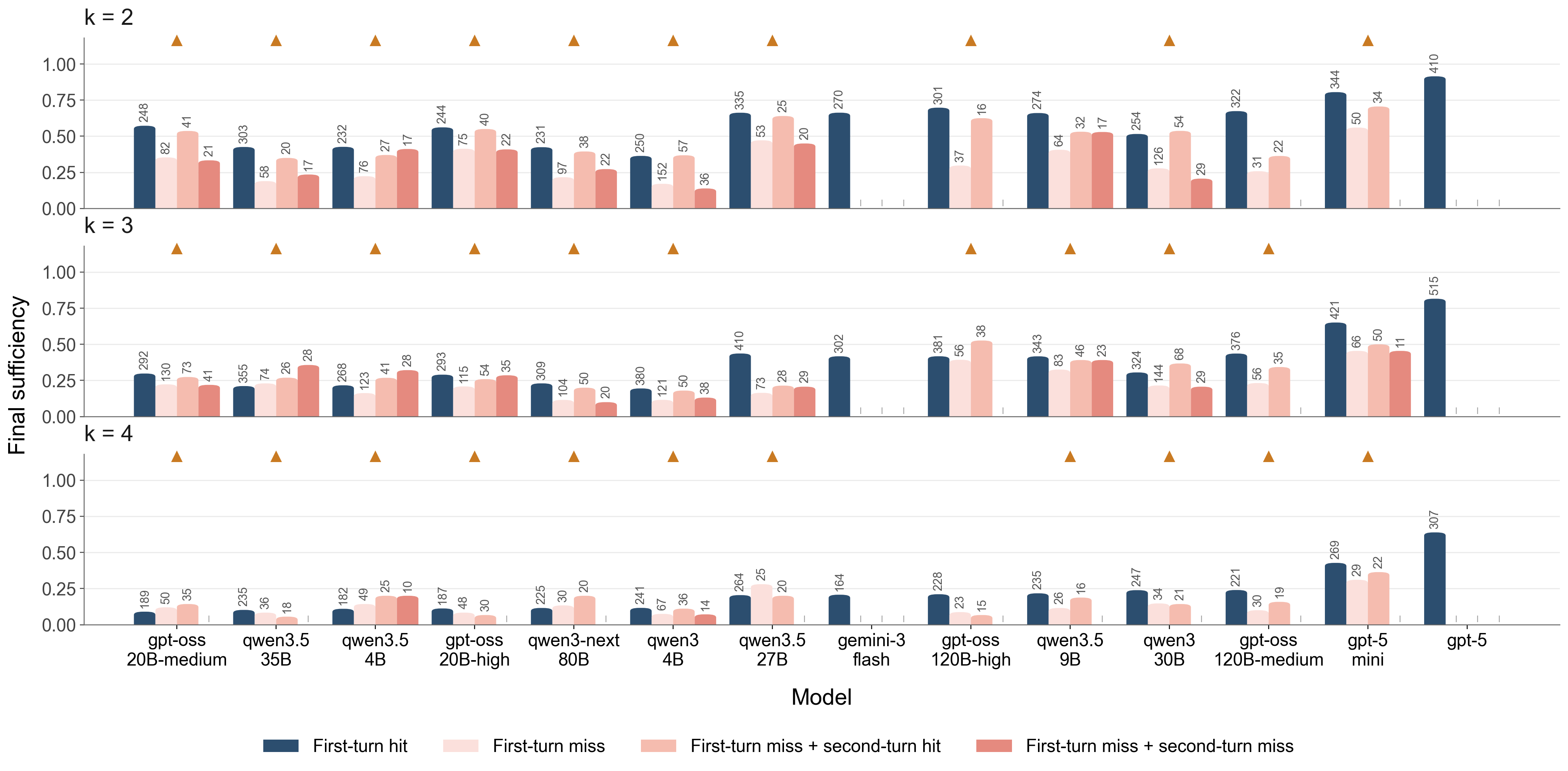}
    \vspace{-10pt}
    \caption{
    Recovery from a wrong first-turn query in $k=2,3,4$ GeneReg-MT dynamic steady state identification. Bars show final sufficiency for each model, partitioned by whether the model queries a correct variable in the first and second turns. 
    $\blacktriangle$ indicates that the first-turn miss + second-turn hit (recovery) group performs within 10 percentage points of the first-turn hit group. 
    The number on each bar represents the number of interactions in each group. Groups with fewer than 10 interactions are not shown (a dash).
    No non-inferiority tests are performed due to the lack of power. 
    Because some interactions emit zero or only one query, the miss+hit and miss+miss counts need not sum to first-turn miss; for the same reason, the totals of first-turn hit and first-turn miss vary slightly across models.
    }
    \label{fig:genereg_multi_dyn_attr_muliturn_early_recovery_allk_gap10pp}
\end{figure}

\begin{figure}[htp]
    \centering
    \includegraphics[width=1.0\linewidth]{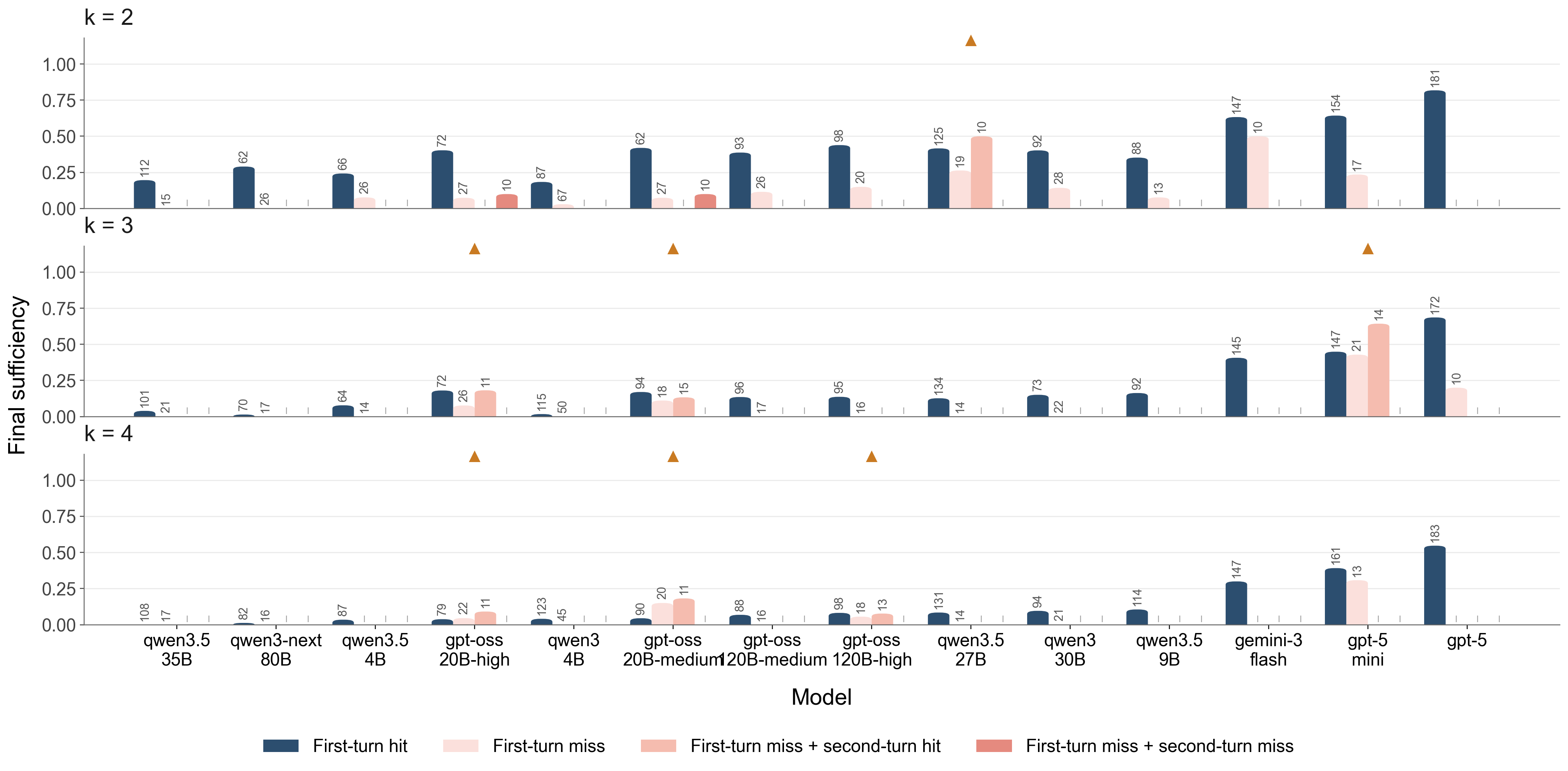}
    \vspace{-8pt}
    \caption{
    Recovery from a wrong first-turn query in $k=2,3,4$ GeneReg-MT dynamic marker identification. Bars show task-completion rate for each model, partitioned by whether the model queries a correct variable in the first and second turns. 
    $\blacktriangle$ indicates that the first-turn miss + second-turn hit (recovery) group performs within 10 percentage points of the first-turn hit group. 
    The number on each bar represents the number of interactions in each group. Groups with with fewer than 10 interactions are not shown (a dash).
    No non-inferiority tests are performed due to the lack of power. 
    Because some interactions emit zero or only one query, the miss+hit and miss+miss counts need not sum to first-turn miss; for the same reason, the totals of first-turn hit and first-turn miss vary slightly across models.
    }
    \label{fig:genereg_multi_dyn_marker_muliturn_early_recovery_allk_gap10pp}
    \vspace{-8pt}
\end{figure}

\begin{figure}[htp]
    \centering
    \includegraphics[width=0.8\linewidth]{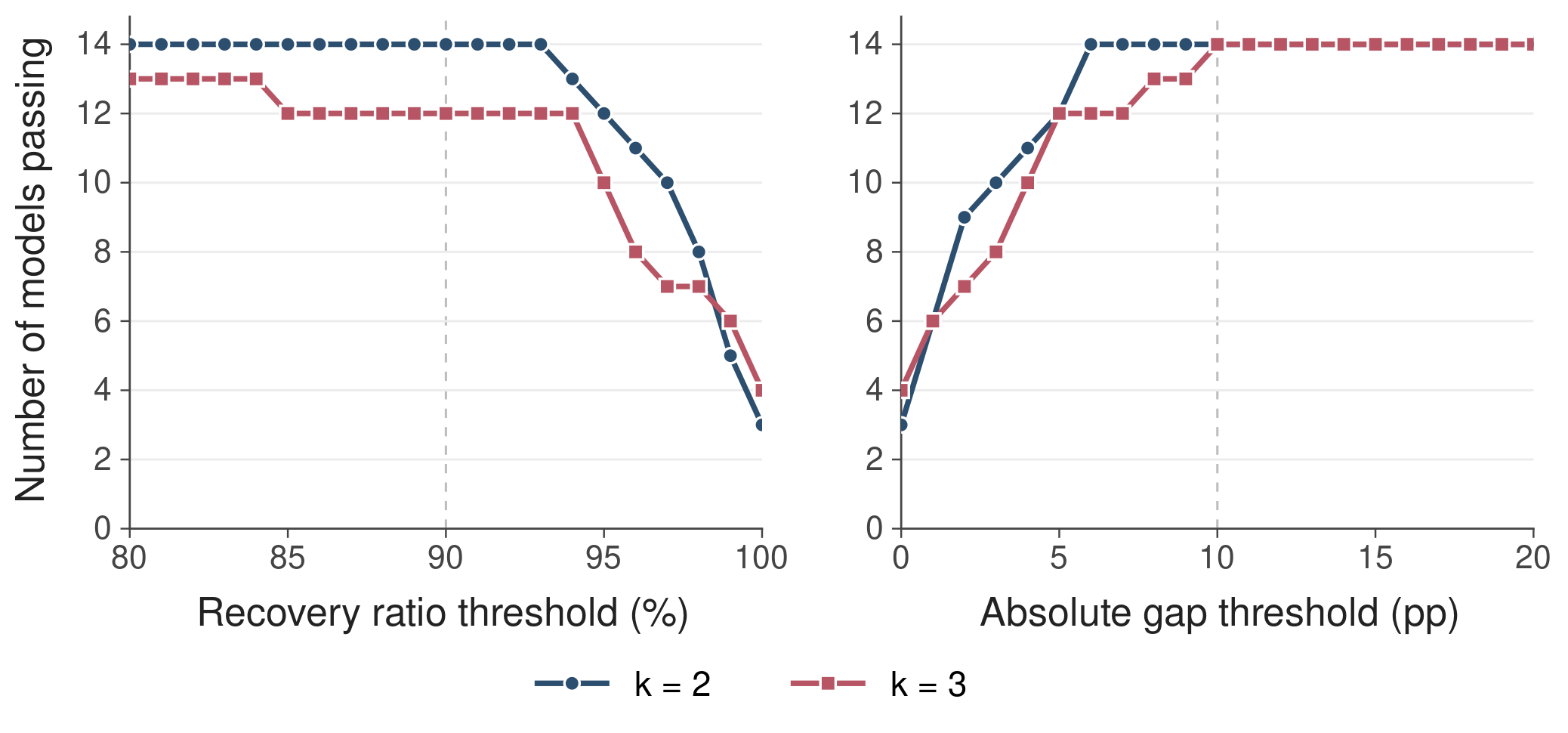}
    \caption{Sensitivity of the recovery criterion in Logic-Q-MT. For each threshold, we compute how many models satisfy the corresponding recovery criterion when comparing the ``first-turn miss + second-turn hit'' group against the ``first-turn hit'' group. The left panel uses a relative threshold based on the ratio of mean final sufficiency, while the right panel uses an absolute threshold based on the difference in mean final sufficiency. Results are shown for $k=2$ and $k=3$.
    }
    \label{fig:logicq_multi_muliturn_early_recovery_sensitivity_analysis}
\end{figure}

\begin{figure}[htp]
    \begin{subfigure}[t]{0.48\linewidth}
        \centering
        \includegraphics[width=\linewidth]{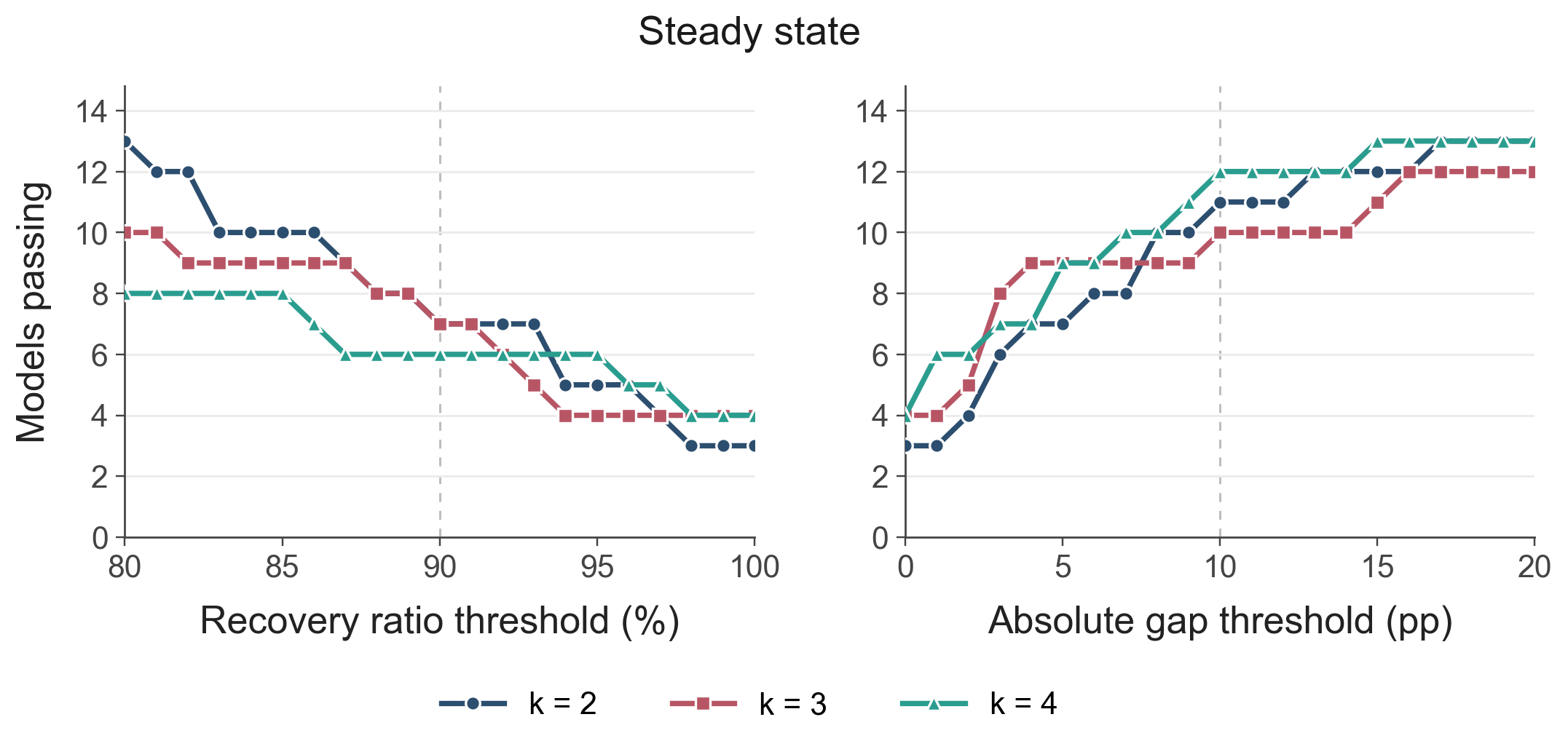}
        \caption{Steady-state identification.}
        \label{fig:genereg_multi_dyn_attr_muliturn_early_recovery_sensitivity_analysis}
    \end{subfigure}
    \begin{subfigure}[t]{0.48\linewidth}
        \centering
        \includegraphics[width=\linewidth]{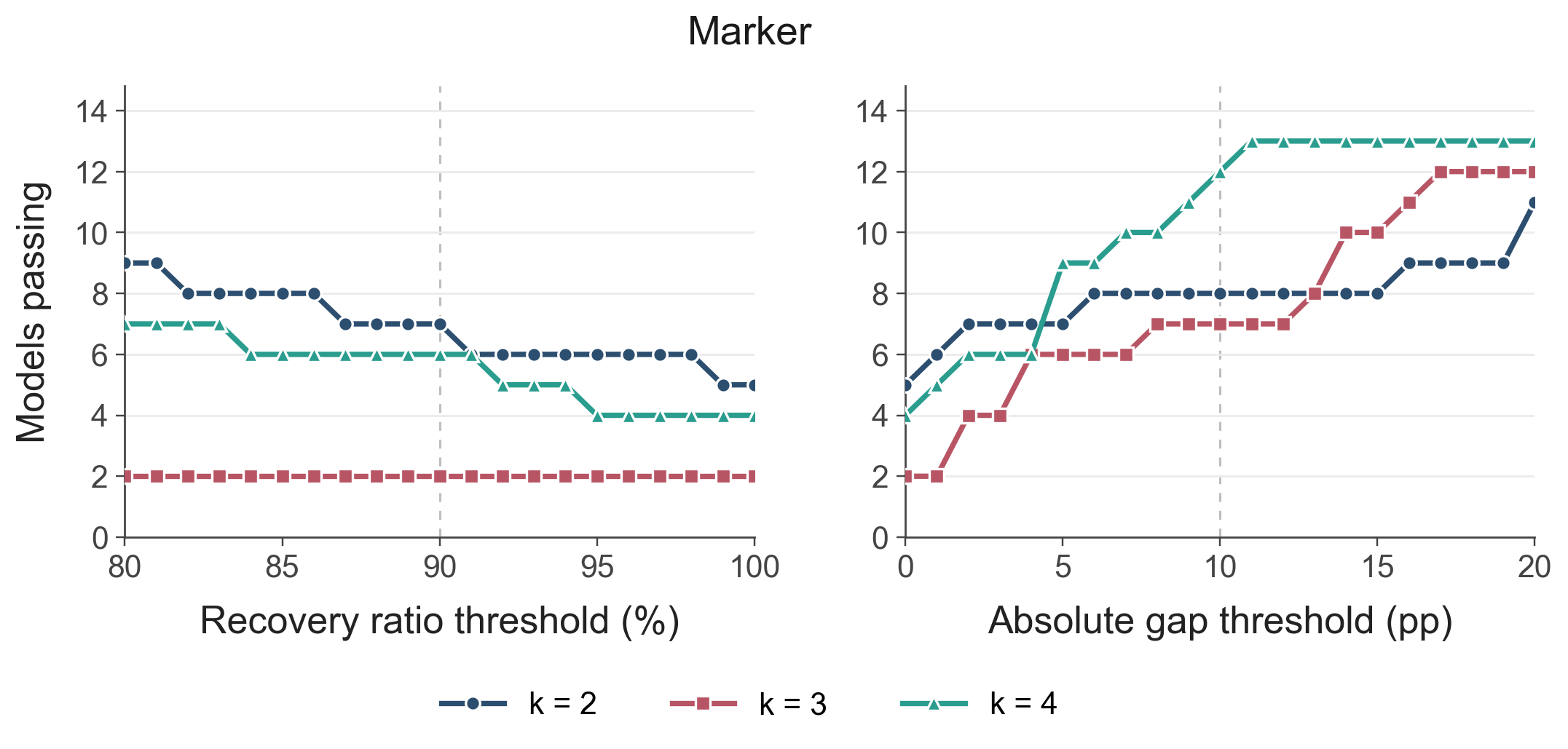}
        \caption{Marker identification.}
        \label{fig:genereg_multi_dyn_marker_muliturn_early_recovery_sensitivity_analysis}
    \end{subfigure}

    \caption{
    Sensitivity of the recovery criterion in GeneReg-MT.
    For each threshold, we count how many models satisfy the recovery criterion when comparing the ``first-turn miss + second-turn hit'' group against the ``first-turn hit'' group.
    In each subfigure, the left panel uses a relative threshold based on the ratio of mean final sufficiency, while the right panel uses an absolute threshold based on the difference in mean final sufficiency.
    Results are shown for \(k=2,3,4\).
    }
    \label{fig:genereg_multi_muliturn_early_recovery_sensitivity_analysis}
\end{figure}

\subsubsection{Prompt sensitivity}
\label{app:prompt_sensitivity}

Because LLM performance can shift substantially under semantically minor prompt reformulations~\citep{sclar2024quantifying,mizrahi2024state}, we quantify how sensitive the Logic-Q-MT and GeneReg-MT results are to the wording of the interaction prompt. We compare the default prompts (Appendix~\ref{appendix:logicq_multi_prompts} and Appendix~\ref{appendix:genereg_multi_prompts}) against three variants: (P1) reordering the presentation of the rules, facts, and forbidden attributes; (P2) toggling the query-selection guidance, where the Logic-Q-MT default already contains this guidance, so P2 removes it, whereas the GeneReg-MT default does not, so P2 adds it; and (P3) dropping the reminder that the provided information may be insufficient. We evaluate Qwen3-4B-Thinking, Qwen3-30B-A3B-Thinking-FP8, gpt-oss-20B-high, and GPT-5-mini; the Logic-Q-MT runs use the same 1{,}200 tasks (600 problems, each with two target-value variants) as the main experiments.

\paragraph{Noise calibration.}
A prompt variant should count as consequential only if it moves performance by more than run-to-run variability. We therefore rerun the original prompt under a different random seed (repl) to calibrate this noise and, for each model and variant, test whether the variant shifts final sufficiency more than the rerun does, i.e., $H_1\colon \lvert\Delta(\mathrm{orig},\mathrm{variant})\rvert > \lvert\Delta(\mathrm{orig},\mathrm{repl})\rvert$, where $\Delta$ is the signed difference in final sufficiency (variant minus original). $p$-values are obtained from a problem-level cluster bootstrap with 10{,}000 resamples and Holm-corrected over the model family. Table~\ref{app_tab:prompt_sensitivity} reports the signed shifts.

\begin{table*}[htbp]
\centering
\small
\setlength{\tabcolsep}{3pt}
\caption{Prompt sensitivity of final sufficiency in Logic-Q-MT and GeneReg-MT. Each cell reports the signed change in final sufficiency under a prompt variant, $\Delta = \mathrm{variant} - \mathrm{original}$, with the corresponding $p$-value in parentheses: a problem-level cluster bootstrap (10{,}000 resamples, Holm-corrected over the model family) tests whether the variant moves final sufficiency more than a rerun of the original prompt under a different seed. n.s.\ denotes $p > 0.05$. The Logic-Q-MT default prompt already contains the query-selection guidance, so P2 removes it, whereas the GeneReg-MT default does not, so P2 adds it.}
\label{app_tab:prompt_sensitivity}
\resizebox{\textwidth}{!}{
\begin{tabular}{lcccc}
\toprule
\textbf{Variant} & \textbf{Qwen3-4B-Thinking} & \textbf{Qwen3-30B-A3B-Thinking-FP8} & \textbf{gpt-oss-20B-high} & \textbf{GPT-5-mini} \\
\midrule
\multicolumn{5}{l}{\emph{Logic-Q-MT}} \\
P1: reorder items & $+0.019$ (n.s.) & $+0.053$ (n.s.) & $+0.010$ (n.s.) & $-0.005$ (n.s.) \\
P2: drop query-selection guidance & $+0.076$ (.002) & $-0.109$ (.006) & $-0.070$ (.011) & $-0.040$ (.011) \\
P3: drop insufficiency reminder & $-0.141$ ($<$.001) & $-0.014$ (n.s.) & $-0.027$ (n.s.) & $-0.005$ (n.s.) \\
\midrule
\multicolumn{5}{l}{\emph{GeneReg-MT}} \\
P1: reorder items & $-0.018$ (n.s.) & $-0.040$ (n.s.) & $-0.039$ (n.s.) & $-0.012$ (n.s.) \\
P2: add query-selection guidance & $-0.037$ (.046) & $-0.046$ (.046) & $-0.031$ (n.s.) & $+0.002$ (n.s.) \\
P3: drop insufficiency reminder & $-0.080$ ($<$.001) & $-0.137$ ($<$.001) & $-0.050$ (.012) & $-0.023$ (n.s.) \\
\bottomrule
\end{tabular}
}
\end{table*}

\paragraph{Results.}
The findings are consistent across both domains. P1, a meaning-preserving reordering, has no measurable effect on any model in either domain: every cell stays within run-to-run noise, so the reported numbers are not artifacts of item order or formatting. P2 is content-bearing, so some change is expected; the effect is nevertheless small and model-dependent, the strongest model (GPT-5-mini) is essentially unaffected in both domains ($-0.040$ in Logic-Q-MT, $+0.002$ in GeneReg-MT), and the direction varies (removing the guidance helps Qwen3-4B-Thinking in Logic-Q-MT, while adding it slightly hurts the two Qwen models in GeneReg-MT), indicating that the default guidance is not what drives the reported failures. P3, removing the insufficiency reminder, lowers final sufficiency, with the effect concentrated in the smaller open-weight models (Qwen3-4B-Thinking, Qwen3-30B-A3B-Thinking-FP8, gpt-oss-20B-high) and no measurable effect on GPT-5-mini in either domain, suggesting that weaker models rely more on the explicit cue that the provided information may be insufficient and are less spontaneously calibrated. Overall, GPT-5-mini is robust to all three perturbations up to noise, apart from the small $-0.040$ P2 effect in Logic-Q-MT, and the sensitivity that does exist is confined to content-bearing edits and to weaker models, so the qualitative conclusions are not prompt artifacts.

\paragraph{Prompt provenance and defaults.}
The prompts we use may not be globally optimal. The $k$-prediction and MSS-identification prompts were minimally adapted from QuestBench~\citep{li2025questbench} to support multiple jointly necessary variables, and the multi-turn evaluation prompts were fixed before the full model comparison and shared across all models, with no model-specific or test-instance tuning. We adopt the no-budget formulation as the default because explicitly stating a turn budget can change the stopping behavior being evaluated; the turn-budget ablation and the perturbations above both show that prompting shapes information-seeking behavior, and the development of better prompting strategies for multi-turn information seeking is left to future work.

\subsubsection{Preventing premature answers}
\label{app:force_ask}

\paragraph{Intervention.}
Directly forcing interaction to continue until the acquired information is sufficient would make the outcome true by construction. We therefore test a weaker intervention: in a $k$-underspecified task, any attempt to answer before issuing $k$ queries is rejected, and the model is asked to query again. Query selection remains unconstrained, and the model may stop freely once it has issued $k$ queries, whether or not those queries are actually sufficient. The intervention thus prevents clear under-querying without guaranteeing sufficiency. We focus on $k=3$, since tasks with a higher degree of underspecification show a higher rate of premature stopping.

\paragraph{Results.}
Table~\ref{app_tab:force_ask} compares final sufficiency and the average number of turns between the original and force-ask protocols on Logic-Q-MT at $k=3$. Preventing early stopping significantly improves final sufficiency for both Qwen models ($+9.7$\,pp, $p=0.003$ for Qwen3-4B-Thinking; $+25.7$\,pp, $p<5\times 10^{-4}$ for Qwen3-30B-A3B-Thinking-FP8; exact McNemar tests), whereas the effect for gpt-oss-20B-high and GPT-5-mini is insignificant. Interestingly, the intervention also increases turn counts beyond what is needed merely to satisfy the $k$-query minimum. A possible explanation is that rejecting an early answer signals to the model that its information is currently insufficient---a signal it would not otherwise receive---so that it becomes more cautious afterward.

\paragraph{Interpretation.}
This setting arguably leaks the degree of underspecification and enforces continued querying in a way that is not typical of how LLMs seek information. We therefore view the gains as evidence of a latent capability that can be elicited under explicit correction, rather than as evidence of the models' spontaneous calibration.

\begin{table*}[htbp]
\centering
\small
\setlength{\tabcolsep}{3pt}
\caption{Effect of preventing premature answers on Logic-Q-MT at $k=3$ (400 episodes per model). Any attempt to answer before issuing $k$ queries is rejected and the model is asked to query again; query selection remains unconstrained, and the model may stop freely once $k$ queries have been issued. Final sufficiency is compared with exact McNemar tests and the average number of turns with Wilcoxon signed-rank tests, both paired at the episode level.}
\label{app_tab:force_ask}
\resizebox{\textwidth}{!}{
\begin{tabular}{lcccccccc}
\toprule
\multirow{2}{*}{\textbf{Model}}
& \multicolumn{4}{c}{\textbf{Final sufficiency}}
& \multicolumn{4}{c}{\textbf{Number of turns}} \\
\cmidrule(lr){2-5} \cmidrule(lr){6-9}
& Original & Force-ask & Change (pp) & $p$ (McNemar)
& Original & Force-ask & Change & $p$ (Wilcoxon) \\
\midrule
Qwen3-4B-Thinking
& 0.560 & 0.657 & $+9.7$ & 0.003
& 4.725 & 5.192 & $+0.467$ & $<5\times 10^{-4}$ \\
Qwen3-30B-A3B-Thinking-FP8
& 0.275 & 0.532 & $+25.7$ & $<5\times 10^{-4}$
& 3.743 & 4.910 & $+1.167$ & $<5\times 10^{-4}$ \\
gpt-oss-20B-high
& 0.818 & 0.825 & $+0.7$ & 0.824
& 6.470 & 6.678 & $+0.208$ & 0.187 \\
GPT-5-mini
& 0.858 & 0.887 & $+2.9$ & 0.175
& 5.530 & 5.647 & $+0.117$ & 0.471 \\
\bottomrule
\end{tabular}
}
\end{table*}

\subsubsection{Sensitivity to the designated MSS}
\label{app:alt_mss}

\paragraph{MSS multiplicity across domains.}
GeneReg-MT, GSME-Q-MT, and ClinGuide-MT have a unique MSS by construction; for GeneReg-MT we additionally verified uniqueness exhaustively. Multiple MSSs are common in Logic-Q-MT: 96.8\% of problems have more than one, with an average of 5.98, 8.74, and 10.34 MSSs for $k=1$, $2$, and $3$, respectively (8.35 overall). Our primary sequential metric, final sufficiency, is invariant to the choice of MSS: it checks whether the acquired information uniquely determines the target, irrespective of the MSS. The set of queryable variables, however, does depend on the designated MSS, which can affect model behavior. Because evaluation outcomes can be sensitive to seemingly incidental design choices \citep{sclar2024quantifying,mizrahi2024state}, we quantify the sensitivity of Logic-Q-MT results to the designated MSS directly.

\paragraph{Resampling design.}
We independently resampled two alternative canonical MSSs (\texttt{altgt1} and \texttt{altgt2}) for every Logic-Q-MT problem; all three canonical MSSs are pairwise different for 538 of the 600 problems (89.7\%; Table~\ref{app_tab:alt_mss_classes}). We then re-evaluated four representative models on the two Logic-Q-MT variants with alternative canonical MSSs, and additionally re-ran the original tasks with an identical configuration but a different random seed (\texttt{repl}), which calibrates run-to-run noise. For every model and redraw, we test whether the redraw effect exceeds this noise, $H_1\!:\, |\Delta(\mathrm{orig}, \mathrm{alt})| > |\Delta(\mathrm{orig}, \mathrm{repl})|$, via a problem-level cluster bootstrap (10{,}000 resamples). All 8 tests per metric are run and reported, with the raw one-sided $p$-value and its Holm adjustment over the 8. We focus on $k=3$, the stratum with the largest MSS multiplicity.

\begin{table}[htp]
\centering
\small
\caption{Overlap classes of the three canonical MSS draws per Logic-Q-MT problem: the original designated MSS and two independently resampled alternatives (\texttt{altgt1}, \texttt{altgt2}).}
\label{app_tab:alt_mss_classes}
\begin{tabular}{lr}
\toprule
Class & Problems \\
\midrule
All three identical & 19 \\
Two alternatives identical, both different from the original & 43 \\
All three pairwise different & 538 (89.7\%) \\
\bottomrule
\end{tabular}
\end{table}

\begin{table*}[htbp]
\centering
\small
\setlength{\tabcolsep}{3pt}
\caption{Sensitivity of final sufficiency to the designated MSS in Logic-Q-MT at $k=3$. $\Delta$ columns are signed (run minus original): $\Delta_{\mathrm{suff}}(\mathrm{alt}-\mathrm{orig})$ compares the alternative-MSS variant with the original run, and $\Delta_{\mathrm{suff}}(\mathrm{repl}-\mathrm{orig})$ compares a rerun of the original configuration with a different random seed. The test statistic is $|\Delta_{\mathrm{alt}}| - |\Delta_{\mathrm{repl}}|$, i.e., whether the redraw effect exceeds run-to-run noise, assessed by a problem-level cluster bootstrap (10{,}000 resamples) with the 95\% percentile confidence interval shown; $p_{\mathrm{boot}}$ is the raw one-sided bootstrap $p$-value and $p_{\mathrm{Holm}}$ its Holm adjustment over the 8 tests within each metric. n.s.\ denotes $p > 0.2$. Statistics are computed from unrounded values and rounded independently, so displayed columns may disagree in the last digit.}
\label{app_tab:alt_mss_suff}
\resizebox{\textwidth}{!}{
\begin{tabular}{llcccccc}
\toprule
Model & Comparison & $\Delta_{\mathrm{suff}}(\mathrm{alt}-\mathrm{orig})$ & $\Delta_{\mathrm{suff}}(\mathrm{repl}-\mathrm{orig})$ & $|\Delta_{\mathrm{alt}}| - |\Delta_{\mathrm{repl}}|$ & 95\% CI & $p_{\mathrm{boot}}$ & $p_{\mathrm{Holm}}$ \\
\midrule
Qwen3-4B-Thinking & vs \texttt{altgt1} & $+0.037$ & $-0.015$ & $+0.022$ & $[-0.057, +0.088]$ & n.s. & n.s. \\
Qwen3-4B-Thinking & vs \texttt{altgt2} & $+0.035$ & $-0.015$ & $+0.020$ & $[-0.057, +0.088]$ & n.s. & n.s. \\
Qwen3-30B-A3B-Thinking-FP8 & vs \texttt{altgt1} & $+0.072$ & $+0.027$ & $+0.045$ & $[-0.018, +0.100]$ & 0.090 & n.s. \\
Qwen3-30B-A3B-Thinking-FP8 & vs \texttt{altgt2} & $-0.005$ & $+0.027$ & $-0.022$ & $[-0.062, +0.050]$ & n.s. & n.s. \\
gpt-oss-20B-high & vs \texttt{altgt1} & $+0.065$ & $+0.017$ & $+0.047$ & $[-0.005, +0.088]$ & 0.042 & n.s. \\
gpt-oss-20B-high & vs \texttt{altgt2} & $+0.052$ & $+0.017$ & $+0.035$ & $[-0.015, +0.077]$ & 0.117 & n.s. \\
GPT-5-mini & vs \texttt{altgt1} & $+0.027$ & $-0.005$ & $+0.022$ & $[-0.040, +0.057]$ & n.s. & n.s. \\
GPT-5-mini & vs \texttt{altgt2} & $+0.007$ & $-0.005$ & $+0.002$ & $[-0.040, +0.037]$ & n.s. & n.s. \\
\bottomrule
\end{tabular}
}
\end{table*}

\begin{table*}[htbp]
\centering
\small
\setlength{\tabcolsep}{3pt}
\caption{Sensitivity of the number of turns to the designated MSS in Logic-Q-MT at $k=3$. Same protocol as Table~\ref{app_tab:alt_mss_suff}: $\Delta$ columns are signed (run minus original), the test statistic $|\Delta_{\mathrm{alt}}| - |\Delta_{\mathrm{repl}}|$ is assessed by a problem-level cluster bootstrap (10{,}000 resamples), and Holm adjustment is over the 8 tests within each metric. n.s.\ denotes $p > 0.2$.}
\label{app_tab:alt_mss_turns}
\resizebox{\textwidth}{!}{
\begin{tabular}{llcccccc}
\toprule
Model & Comparison & $\Delta_{\mathrm{turns}}(\mathrm{alt}-\mathrm{orig})$ & $\Delta_{\mathrm{turns}}(\mathrm{repl}-\mathrm{orig})$ & $|\Delta_{\mathrm{alt}}| - |\Delta_{\mathrm{repl}}|$ & 95\% CI & $p_{\mathrm{boot}}$ & $p_{\mathrm{Holm}}$ \\
\midrule
Qwen3-4B-Thinking & vs \texttt{altgt1} & $-0.355$ & $-0.087$ & $+0.268$ & $[-0.013, +0.485]$ & 0.032 & n.s. \\
Qwen3-4B-Thinking & vs \texttt{altgt2} & $-0.557$ & $-0.087$ & $+0.470$ & $[+0.192, +0.683]$ & 0.001 & 0.006 \\
Qwen3-30B-A3B-Thinking-FP8 & vs \texttt{altgt1} & $-0.040$ & $-0.058$ & $-0.018$ & $[-0.188, +0.167]$ & n.s. & n.s. \\
Qwen3-30B-A3B-Thinking-FP8 & vs \texttt{altgt2} & $-0.048$ & $-0.058$ & $-0.010$ & $[-0.198, +0.183]$ & n.s. & n.s. \\
gpt-oss-20B-high & vs \texttt{altgt1} & $-0.345$ & $-0.007$ & $+0.338$ & $[-0.125, +0.542]$ & 0.090 & n.s. \\
gpt-oss-20B-high & vs \texttt{altgt2} & $-0.270$ & $-0.007$ & $+0.263$ & $[-0.177, +0.475]$ & 0.173 & n.s. \\
GPT-5-mini & vs \texttt{altgt1} & $-0.303$ & $+0.165$ & $+0.138$ & $[-0.337, +0.525]$ & n.s. & n.s. \\
GPT-5-mini & vs \texttt{altgt2} & $-0.293$ & $+0.165$ & $+0.128$ & $[-0.330, +0.487]$ & n.s. & n.s. \\
\bottomrule
\end{tabular}
}
\end{table*}

\paragraph{Results.}
Tables~\ref{app_tab:alt_mss_suff} and~\ref{app_tab:alt_mss_turns} report the comparisons. Alternative-MSS variants yield modest gains: final sufficiency increases by up to 7.2 percentage points at $k=3$ (positive in 7 of 8 comparisons) while requiring slightly fewer turns. Most differences fall within run-to-run noise: across all 16 noise-anchored comparisons, only Qwen3-4B-Thinking's turn reduction on \texttt{altgt2} survives Holm correction, and two further comparisons reach uncorrected $p < 0.05$ (Qwen3-4B-Thinking's turn reduction on \texttt{altgt1}, and gpt-oss-20B-high's final-sufficiency shift on \texttt{altgt1}).

\paragraph{Why alternative draws are slightly easier.}
The direction of the shift reflects a construction asymmetry: canonical MSSs had to be certified by forward chaining during dataset construction, whereas alternative MSSs could also be sufficient by refutation, making them shallower and easier to identify. Table~\ref{app_tab:alt_mss_depth} quantifies this. At $k=3$, all original canonical MSSs are forward-derivable with mean depth 4.40, versus roughly 60\% and mean depth 3.5 for the resampled draws. The original draw is therefore a restricted and systematically deeper subpopulation, and a shallower MSS is easier to identify; in this sense the designated MSS is the harder one.

\begin{table}[htp]
\centering
\small
\caption{Forward-derivability and derivation depth of the designated MSS by draw at $k=3$. \% forward-derivable is the fraction of designated MSSs whose sufficiency admits an explicit forward-chaining (unit-propagation) derivation; MSS depth is the length of that derivation. Mean and median depths are computed over forward-derivable draws only, as depth is undefined for MSSs whose sufficiency holds only by refutation; for the original draw this includes all problems.}
\label{app_tab:alt_mss_depth}
\begin{tabular}{lccc}
\toprule
Draw & \% forward-derivable & Mean MSS depth & Median MSS depth \\
\midrule
original & 100.0\% & 4.400 & 4.0 \\
\texttt{altgt1} & 59.0\% & 3.550 & 4.0 \\
\texttt{altgt2} & 60.5\% & 3.499 & 3.0 \\
\bottomrule
\end{tabular}
\end{table}

\clearpage

\subsection{GSME-Q-MT}

\begin{figure}[htp]
    \centering
    \includegraphics[width=0.6\textwidth]{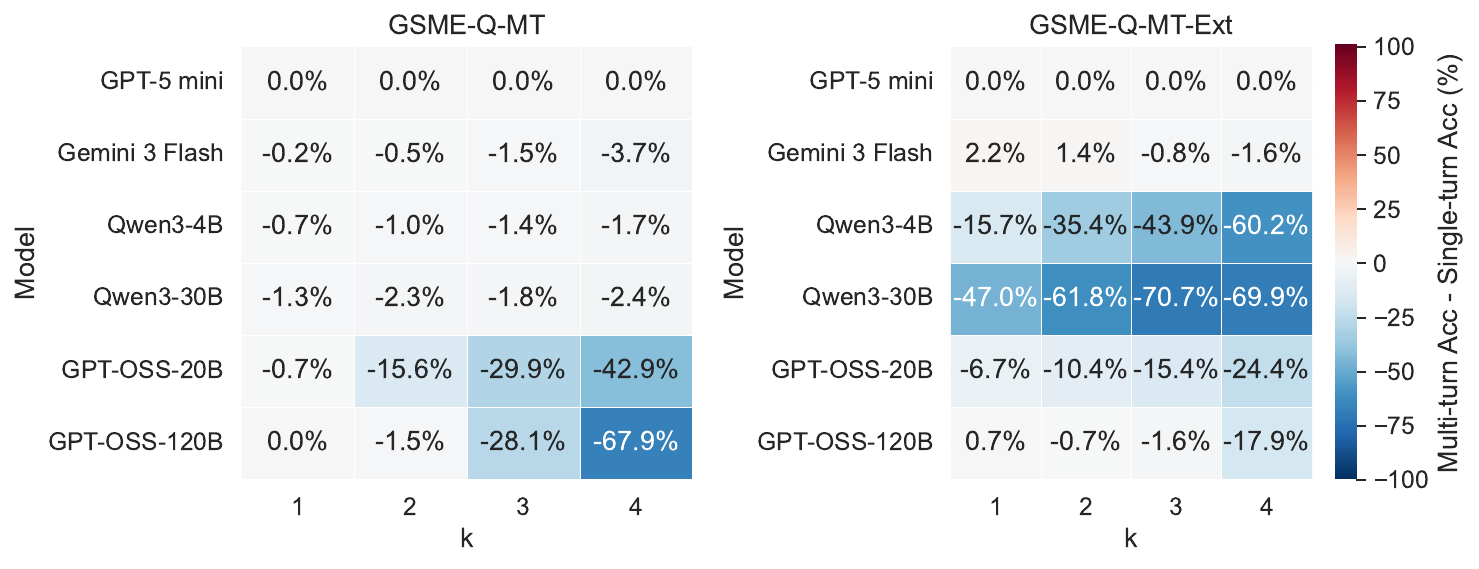}
    \caption{
    Analysis of multi-turn degradation in GSME. 
    Accuracy change from single-turn to multi-turn querying, measured as multi-turn accuracy minus single-turn accuracy. Negative values indicate degradation under multi-turn interaction. 
    }
    \label{app_fig:gsme_multi_degrad}
\end{figure}

\paragraph{Strong single-turn performance does not guarantee reliable information seeking.}
As shown in Figure~\ref{app_fig:gsme_multi_degrad}, several models perform well when the missing information is requested in a single turn, but become less reliable when the same information must be acquired through interaction. This gap is small on the GSME-Q-MT, where most instances are relatively simple and many models remain near saturation. In contrast, the GSME-Q-MT-Ext exposes clearer differences across models and values of $k$. For example, Qwen3-4B-Thinking and Qwen3-30B-A3B-Thinking-FP8 show increasingly large drops as $k$ grows, suggesting that the difficulty is not solving the arithmetic problem, but managing the information-seeking process needed to make the problem fully specified. Interestingly, GPT-5-mini remains stable across both settings, while Gemini-3-Flash shows only minor changes, indicating that some models are better able to preserve their single-turn planning ability during interaction. 

\begin{figure}[htp]
    \centering
    \includegraphics[width=1.0\linewidth]{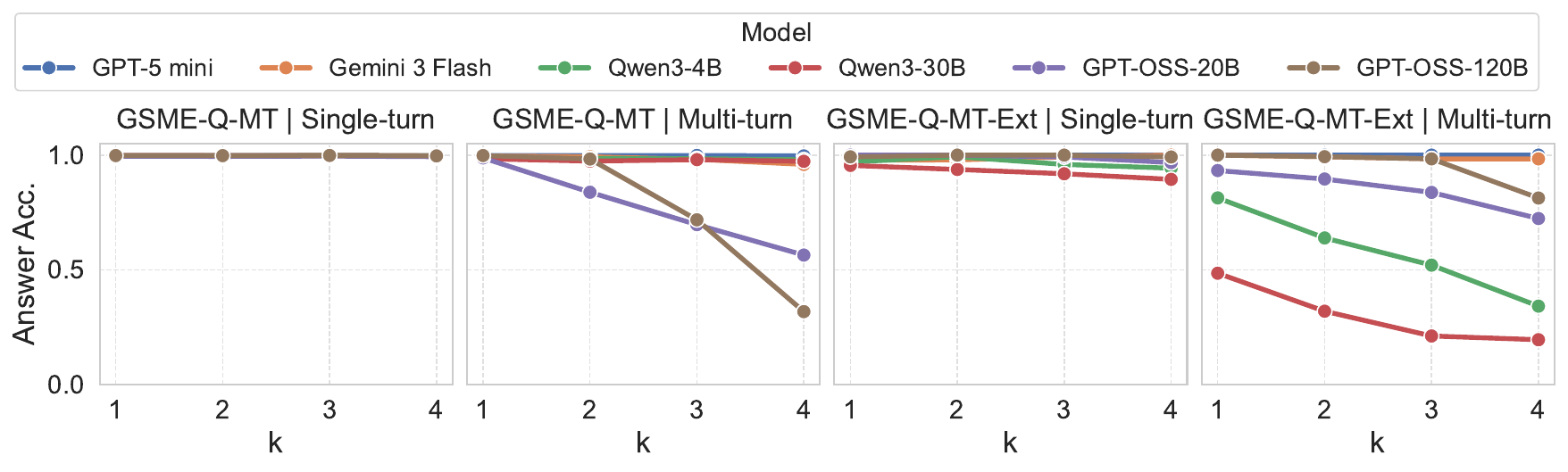}
    \caption{Answer accuracy on GSME across different values of $k$. The original GSME setting is nearly saturated across models, whereas the DAG-generated complex GSME setting reveals a sharp degradation in the multi-turn setting.}
    \label{app_fig:gsme_acc_k}
\end{figure}

\subsection{ClinGuide-MT full results}
\label{app:clinguide_fullresultstext}

\begin{figure}[htp]
    \centering
    \includegraphics[width=1\linewidth]{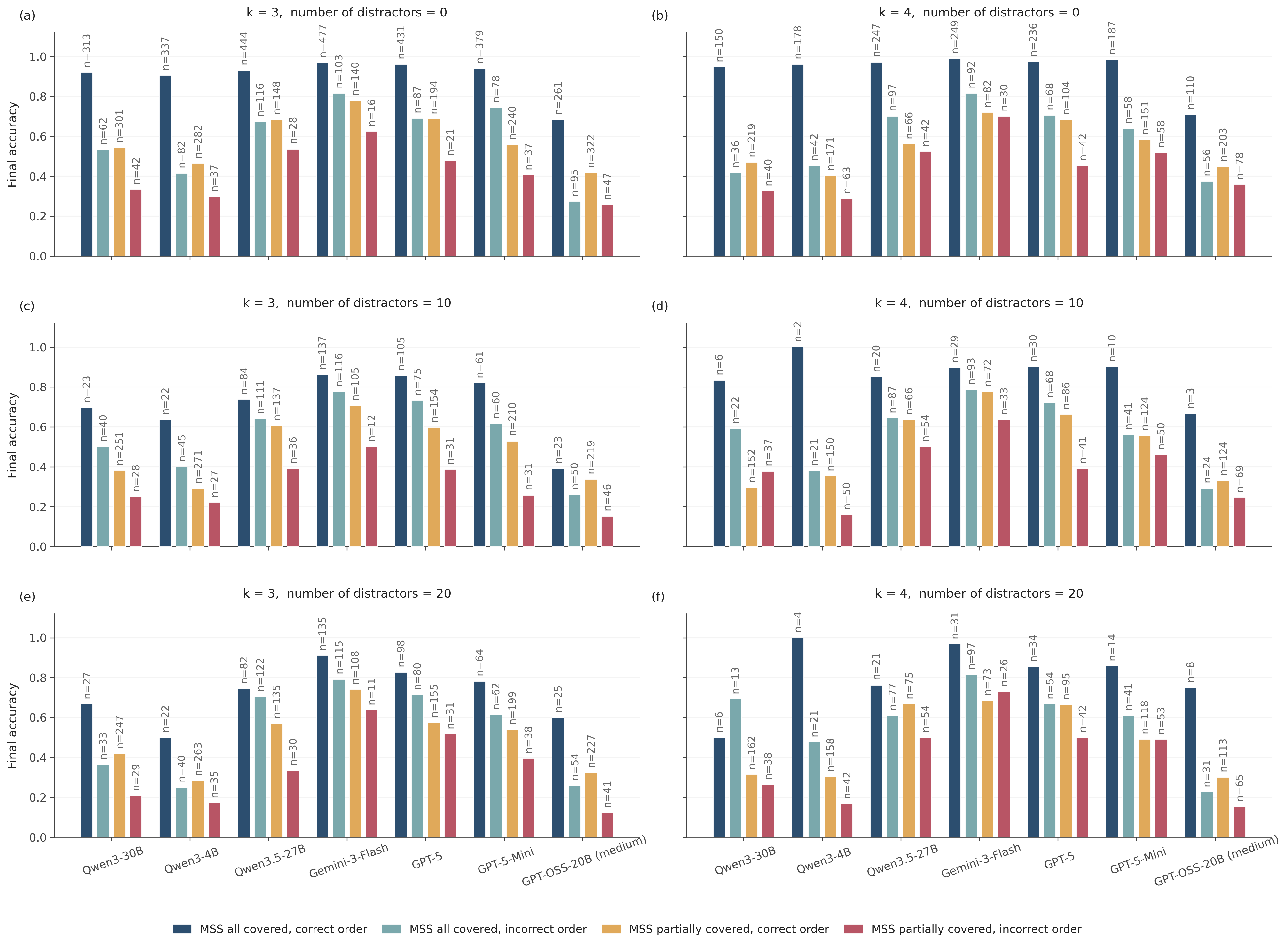}
    \caption{ClinGuide-MT final accuracy by MSS coverage and MSS ordering in the queried variables (a-f: $k=3, 4$, and $0, 10, 20$ distractors).
    }
    \label{app:clinguide-fullresult}
\end{figure}

Appendix Fig.~\ref{app:clinguide-fullresult} reports final accuracy stratified by query coverage and query order across \(k\in\{3,4\}\) and different numbers of distractors $0, 10, 20$. Across nearly all models and settings, interactions that cover MSS in full achieve substantially higher accuracy than those that only partially cover it, confirming that acquiring the necessary information is a prerequisite for accurate final prediction. Within both the full-coverage and partial-coverage groups, following the order induced by the clinical diagnostic pathways generally improves accuracy, showing that the order of information acquisition does affect downstream reasoning. This effect is especially pronounced in more difficult settings with larger \(k\) and more distractors, where incorrect ordering often leads to noticeable accuracy drops even among MSS all covered interactions.

\begin{figure}[htp]
    \centering
    \includegraphics[width=1\linewidth]{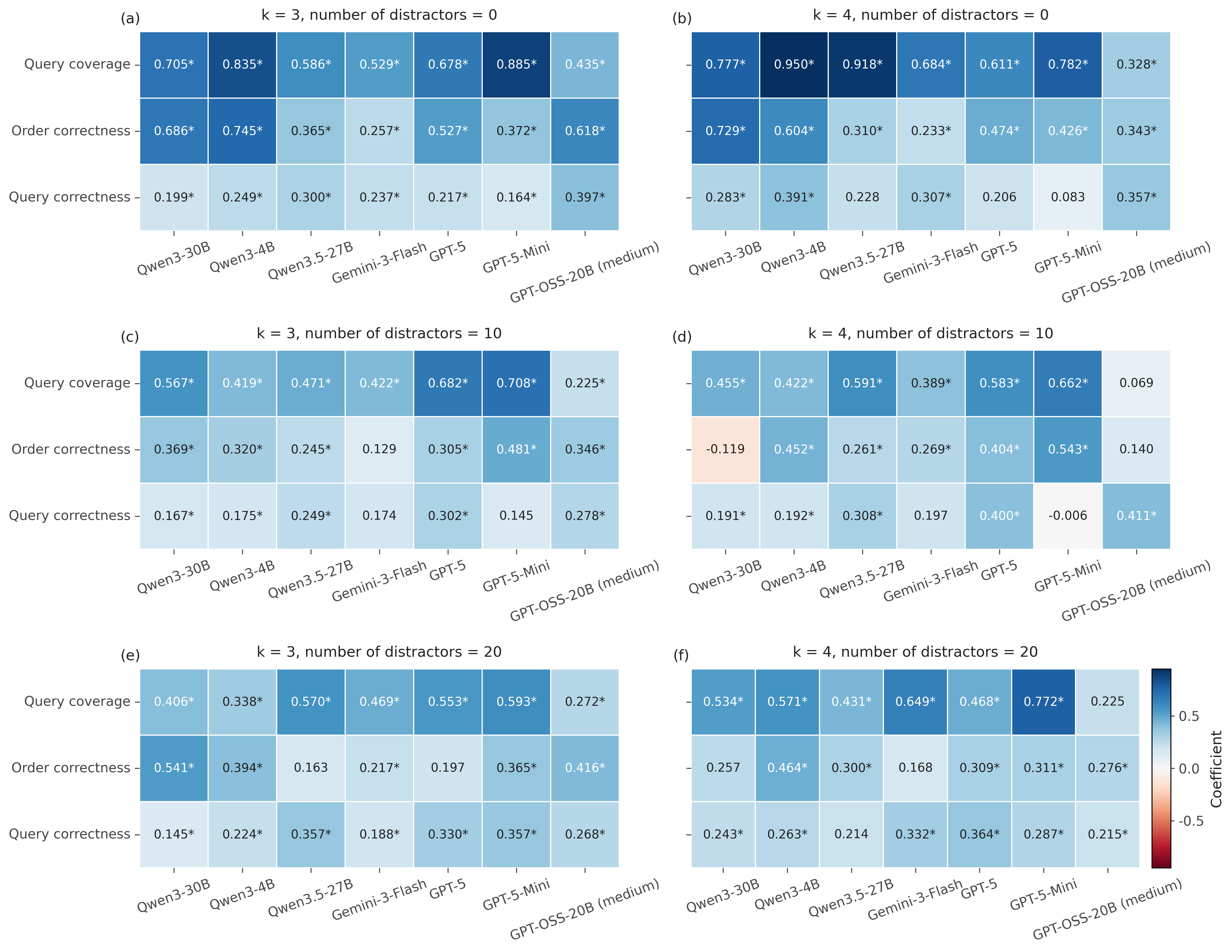}
    \caption{The effect of query coverage, order correctness, and query correctness on final accuracy for ClinGuide-MT Problem (a-f: $k=3, 4$, and $0, 10, 20$ distractors).
    }
    \label{app:clinguide_heatmap_full}
\end{figure}

Appendix Fig.~\ref{app:clinguide_heatmap_full} further investigates the impact of query coverage, order correctness, and query correctness on final accuracy using regression analysis on their associations with final accuracy \(Y_i=\alpha+\gamma_{k_i}+\beta_{\mathrm{cov}} \mathrm{QueryCov}(\hat Q_i)+\beta_{\mathrm{ord}} \mathrm{OrdCorr}(\hat Q_i)+\beta_{\mathrm{corr}} \mathrm{QueryCorr}(\hat Q_i)+\varepsilon_i\). This can be calculated with the metrics we defined in Appendix ~\ref{appendix:metrics}. Across model and setting pairs, query coverage remains the strongest and most stable positive predictor to final accuracy, as it is most frequently the top-ranked predictor. Order correctness also contributes positively in most settings, remaining the second-ranked predictor to final accuracy. Query correctness, which captures avoiding off-branch or irrelevant queries, provides an additional positive association and becomes more salient in more difficult settings. Together, these full results support the main-text conclusion that to achieve high final accuracy in ClinGuide-MT problem, the model needs to ask the correct question and in the correct order.

\subsubsection{Open-book evaluation}
\label{app:clinguide_openbook}

The default ClinGuide-MT evaluation is closed-book: no diagnostic reference material is provided, so recognizing which information is missing partly depends on the model's internal medical knowledge. To explicitly separate this from information-seeking, we additionally evaluate an open-book setting in which models are given all 59 diagnostic algorithms extracted from the source textbook \citep{patient_history}, in their original wording, including the algorithm relevant to each case (pathway provenance and validation are described in Appendix~\ref{app:clinguide_validation}). These diagnostic algorithms are in decision-tree format with an explicit chart of procedure; the patient-specific values along the pathway are to be acquired through questioning.

\paragraph{Effect on final accuracy.}
Providing the full set of diagnostic algorithms substantially improves final accuracy across all four evaluated models and both difficulty levels (Table~\ref{app_tab:clinguide_openbook_acc}), with the largest gains for Qwen3-4B-Thinking. Performance is nevertheless far from saturated even in this setting, indicating that access to the relevant diagnostic algorithm alone does not solve the task.

\begin{table}[htp]
\centering
\small
\caption{ClinGuide-MT final accuracy in the closed-book and open-book settings for $k=3$ and $k=4$ tasks. In the open-book setting, the model is given all 59 diagnostic decision trees from the source textbook \citep{patient_history} in their original wording, including the tree relevant to the case.}
\label{app_tab:clinguide_openbook_acc}
\begin{tabular}{lcccc}
\toprule
\multirow{2}{*}{\textbf{Model}}
& \multicolumn{2}{c}{$\boldsymbol{k=3}$}
& \multicolumn{2}{c}{$\boldsymbol{k=4}$} \\
\cmidrule(lr){2-3} \cmidrule(lr){4-5}
& Closed-book & Open-book & Closed-book & Open-book \\
\midrule
GPT-5-mini         & 55.7\% & 88.4\% & 54.6\% & 85.0\% \\
gpt-oss-20B-high   & 28.1\% & 56.5\% & 29.5\% & 54.2\% \\
Qwen3-30B-A3B-Thinking-FP8 & 38.1\% & 71.1\% & 33.9\% & 64.3\% \\
Qwen3-4B-Thinking  & 32.2\% & 78.4\% & 31.7\% & 80.2\% \\
\bottomrule
\end{tabular}
\end{table}

\paragraph{Coverage and ordering effects persist.}
Table~\ref{app_tab:clinguide_openbook_strat} stratifies final accuracy by MSS coverage and query order in both settings. The coverage and ordering findings established above carry over to the open-book setting wherever the strata are populated: interactions that cover the full MSS attain the highest accuracy, and querying in the order the diagnostic algorithm suggests further boosts performance.

\begin{table*}[htbp]
\centering
\scriptsize
\setlength{\tabcolsep}{3pt}
\caption{ClinGuide-MT final accuracy stratified by MSS coverage and query order for $k\in\{3,4\}$ in the closed-book setting (no diagnostic decision trees provided) and the open-book setting (all 59 diagnostic decision trees provided). Column headers abbreviate the strata: All/Partial denotes full versus partial MSS coverage, and Correct/Incorrect denotes whether the queries follow the order induced by the diagnostic pathway. Each cell reports final accuracy with the stratum size $N$ in parentheses; ``--'' marks strata with fewer than five samples, which are not reported because of insufficient statistical power.}
\label{app_tab:clinguide_openbook_strat}
\resizebox{\textwidth}{!}{
\begin{tabular}{lccccccccc}
\toprule
\multirow{2}{*}{\textbf{Model}} & \multirow{2}{*}{$k$}
& \multicolumn{4}{c}{Closed-book (decision trees $=0$)}
& \multicolumn{4}{c}{Open-book (all 59 decision trees)} \\
\cmidrule(lr){3-6} \cmidrule(lr){7-10}
& & All/Correct & All/Incorrect & Partial/Correct & Partial/Incorrect
& All/Correct & All/Incorrect & Partial/Correct & Partial/Incorrect \\
\midrule
Qwen3-30B-A3B-Thinking-FP8 & 3 & 69.6\% (N=23) & 50.0\% (N=40) & 38.2\% (N=251) & 25.0\% (N=28) & 94.0\% (N=184) & -- & 50.0\% (N=168) & -- \\
Qwen3-30B-A3B-Thinking-FP8 & 4 & 83.3\% (N=6) & 59.1\% (N=22) & 29.6\% (N=152) & 37.8\% (N=37) & 97.8\% (N=93) & -- & 41.5\% (N=123) & 28.6\% (N=7) \\
Qwen3-4B-Thinking & 3 & 63.6\% (N=22) & 40.0\% (N=45) & 29.2\% (N=271) & 22.2\% (N=27) & 94.4\% (N=195) & 88.9\% (N=18) & 60.8\% (N=143) & 40.0\% (N=5) \\
Qwen3-4B-Thinking & 4 & -- & 38.1\% (N=21) & 35.3\% (N=150) & 16.0\% (N=50) & 93.6\% (N=110) & 100.0\% (N=7) & 67.0\% (N=103) & 50.0\% (N=6) \\
GPT-5-mini & 3 & 82.0\% (N=61) & 61.7\% (N=60) & 52.9\% (N=210) & 25.8\% (N=31) & 97.9\% (N=243) & 93.3\% (N=15) & 68.9\% (N=103) & 80.0\% (N=5) \\
GPT-5-mini & 4 & 90.0\% (N=10) & 56.1\% (N=41) & 55.6\% (N=124) & 46.0\% (N=50) & 100.0\% (N=111) & 92.9\% (N=14) & 68.1\% (N=94) & 62.5\% (N=8) \\
gpt-oss-20B-high & 3 & 39.1\% (N=23) & 26.0\% (N=50) & 33.8\% (N=219) & 15.2\% (N=46) & 74.8\% (N=155) & 63.0\% (N=27) & 43.8\% (N=160) & -- \\
gpt-oss-20B-high & 4 & -- & 29.2\% (N=24) & 33.1\% (N=124) & 24.6\% (N=69) & 85.4\% (N=48) & 46.2\% (N=13) & 46.2\% (N=143) & 57.1\% (N=14) \\
\bottomrule
\end{tabular}
}
\end{table*}

Note that this is a favorable, near-upper-bound version of open-book settings. Although all 59 decision trees are provided and the model must still locate the relevant one, the trees are given in the exact wording used to construct the tasks and with a clear procedural structure, so solving largely reduces to matching the case to the right tree and node and then following the branch, rather than deciding what is relevant within a large or noisy knowledge source. The large gains are therefore expected and likely overstate real-world open-book performance. Even in this favorable setting, however, accuracy is not saturated and the coverage and ordering effects persist, so the residual errors reflect failures to identify, order, and acquire case-specific information rather than missing medical knowledge.

\subsection{20Q evaluations}

\subsubsection{Full results}
\label{app:20q_full_results}

\begin{figure}[htp]
    \centering
    \begin{subfigure}[t]{0.48\textwidth}
        \centering
        \includegraphics[width=\textwidth]{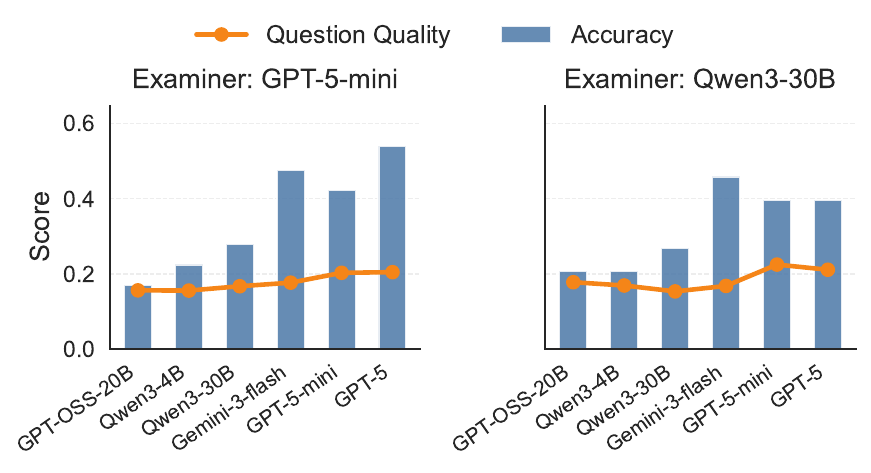}
        \caption{\textsc{Common}.}
        \label{app:common_acc_quality}
    \end{subfigure}
    \hfill
    \begin{subfigure}[t]{0.48\textwidth}
        \centering
        \includegraphics[width=\textwidth]{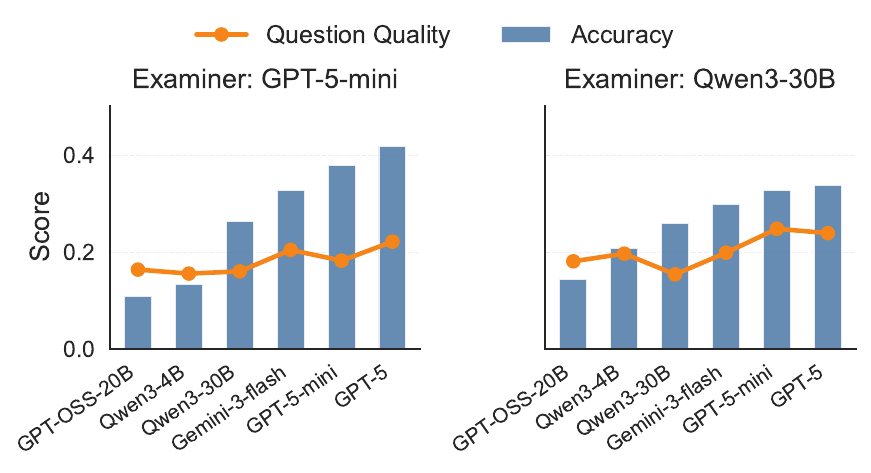}
        \caption{\textsc{Thing}.}
        \label{app:thing_acc_quality}
    \end{subfigure}
    \caption{
    Accuracy versus offline question informativeness on the 20Q tasks. 
    Bars show final guessing accuracy, and lines show the average offline question-quality score of the questions asked by each guesser model. 
    }
    \label{app:full_acc_quality}
\end{figure}

\paragraph{Better questions generally improve accuracy, but question informativeness alone does not determine success.}
Figure~\ref{app:full_acc_quality} compares final accuracy with our offline question-quality score across datasets and examiners. 
Overall, models that ask higher-quality questions tend to achieve higher final accuracy, suggesting that informative question selection is an important factor in 20Q performance. 
This trend is especially clear under the GPT-5-mini examiner, where stronger guesser models obtain both higher question-quality scores and higher final accuracy on both \textsc{Common} and \textsc{Thing}. 
However, this relationship is not deterministic.
For example, under the Qwen3-30B-A3B-Instruct-FP8 examiner, some models achieve similar or even lower question-quality scores while maintaining competitive final accuracy. 
This indicates that final success depends not only on asking informative questions, but also on whether the model can integrate the accumulated interaction history and use the remaining evidence to make a correct final guess. 
Thus, question informativeness captures an important but incomplete aspect of multi-turn information seeking.

\begin{figure}[htp]
    \centering
    \begin{subfigure}[t]{0.48\textwidth}
        \centering
        \includegraphics[width=\textwidth]{figs/20q/remaining_entropy_by_dataset_examiner_GPT-5-mini.pdf}
        \caption{Examiner: GPT-5-mini.}
        \label{app:remaining_entropy_by_dataset_gpt5}
    \end{subfigure}
    \hfill
    \begin{subfigure}[t]{0.48\textwidth}
        \centering
        \includegraphics[width=\textwidth]{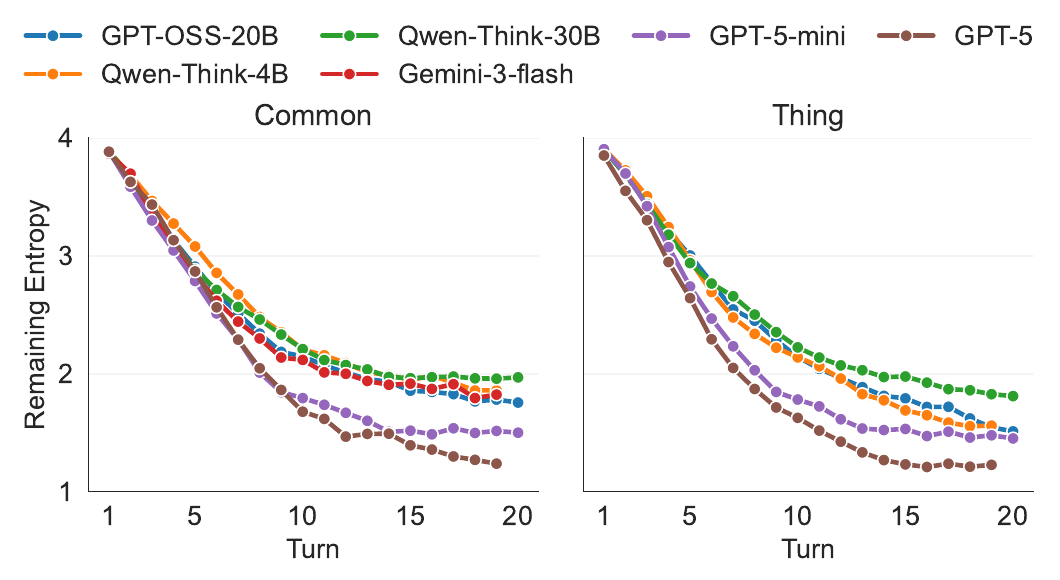}
        \caption{Examiner: Qwen3-30B-A3B-Instruct-FP8.}
        \label{app:remaining_entropy_by_dataset_qwen3}
    \end{subfigure}
    \caption{
    Remaining entropy of the candidate distribution across turns. 
    Each curve tracks the average posterior uncertainty after each turn for a given guesser model, dataset, and examiner. 
    }
    \label{app:full_remaining_entropy}
\end{figure}

\paragraph{Multi-turn interaction consistently reduces uncertainty, and later turns become less effective.}
Figure~\ref{app:full_remaining_entropy} shows the remaining entropy of the candidate distribution after each turn. 
Across both datasets, both examiners, and almost all guesser models, entropy decreases steadily over the interaction. 
This confirms that the generated questions are informative overall and that the interaction progressively narrows the candidate space. 
At the same time, most curves exhibit a clear slowdown in later turns. 
The largest entropy reduction usually occurs in the early stage, when models ask broad categorical questions that separate large groups of candidates. 
After the candidate space has been narrowed, later turns require more fine-grained follow-up questions, and many models show weaker marginal progress. 
This suggests that a central challenge in 20Q is not only initiating useful information seeking, but sustaining discriminative questioning after the broad category has already been identified.

\begin{figure}[htp]
    \centering
    \begin{subfigure}[t]{0.48\textwidth}
        \centering
        \includegraphics[width=\textwidth]{figs/20q/pass_mass_by_dataset_examiner_gpt-5-mini.pdf}
        \caption{Examiner: GPT-5-mini.}
        \label{app:pass_mass_gpt5}
    \end{subfigure}
    \hfill
    \begin{subfigure}[t]{0.48\textwidth}
        \centering
        \includegraphics[width=\textwidth]{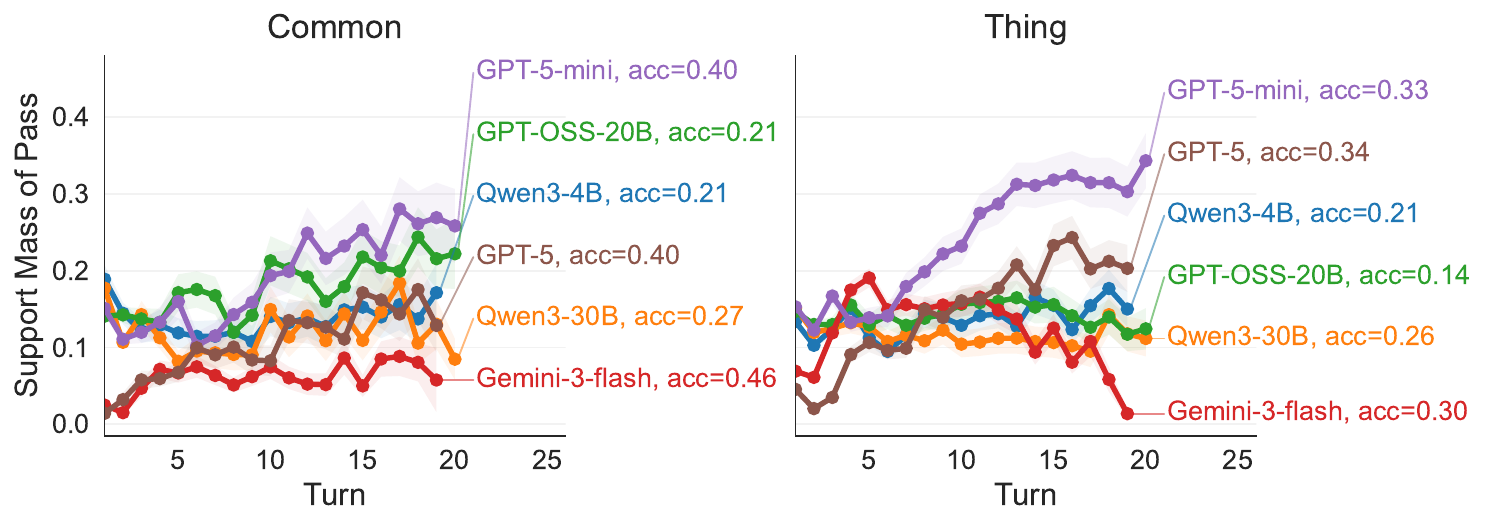}
        \caption{Examiner: Qwen3-30B-A3B-Instruct-FP8.}
        \label{app:pass_mass_qwen3}
    \end{subfigure}
    \caption{
    Pass mass across turns in the 20Q tasks. 
    Pass mass denotes the posterior support mass of candidates for which the current question would receive a \textit{pass} response, meaning that the question is ambiguous or not cleanly answerable by yes or no for those candidates. 
    }
    \label{app:entropy_passmass_analysis}
\end{figure}

\subsubsection{Case study}
\label{app:20q_case_study}

\paragraph{Case-study setup.}
To better understand why models can ask seemingly high-quality questions yet still achieve low final accuracy in the 20 Questions game, we conduct a qualitative case-study analysis of representative failure interaction. For each selected episode, we jointly visualize three elements: the QA trace, the remaining entropy of the candidate set after each turn, and the support mass assigned to the answer \textit{pass}. Remaining entropy measures how much uncertainty is left in the search state, while pass-support mass captures how much of the current candidate space is consistent with an ambiguous or non-committal answer. We select four cases that reflect distinct failure patterns repeatedly observed in the logs. In this analysis, we focus on QA traces generated by Qwen3-4B-Thinking, because this model attains relatively high question-quality scores but performs noticeably worse in recovering the final target.

\begin{figure*}[htp]
    \centering

    \begin{subfigure}[t]{0.9\textwidth}
        \centering
        \includegraphics[width=\textwidth]{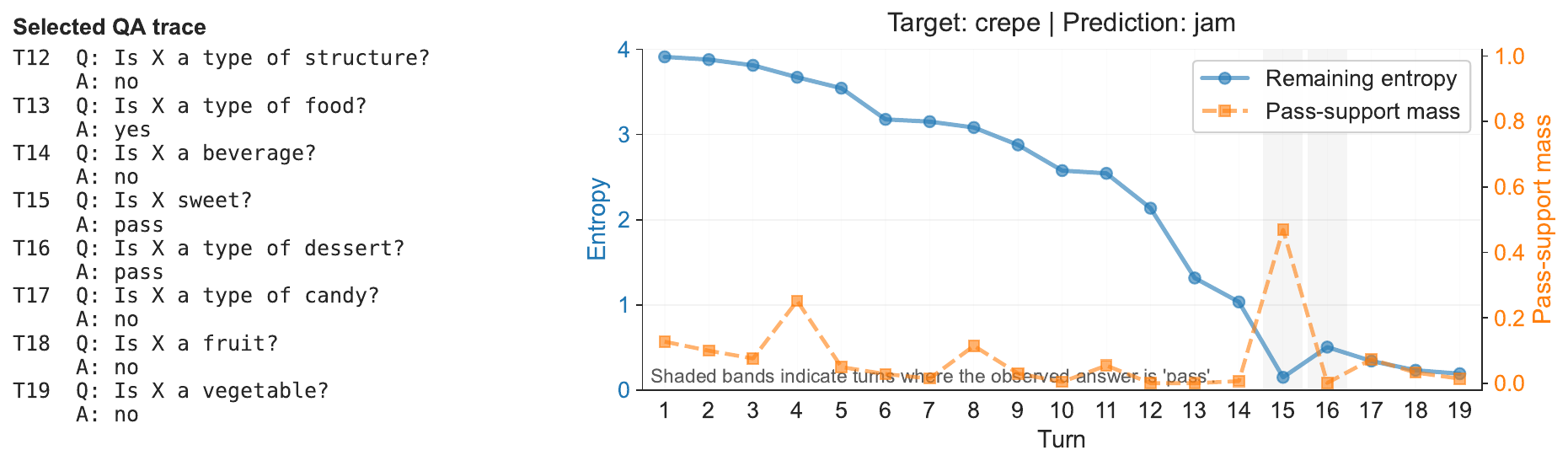}
        \caption{\textbf{Weak final disambiguation.} The model narrows the search to the correct semantic region but fails to separate the true target from nearby alternatives in the final stage. Late ambiguous questions and repeated \textit{pass} answers leave the belief insufficiently resolved, leading to a wrong final guess.}
        \label{fig:20q_case1}
    \end{subfigure}

    \vspace{0.8em}

    \begin{subfigure}[t]{0.9\textwidth}
        \centering
        \includegraphics[width=\textwidth]{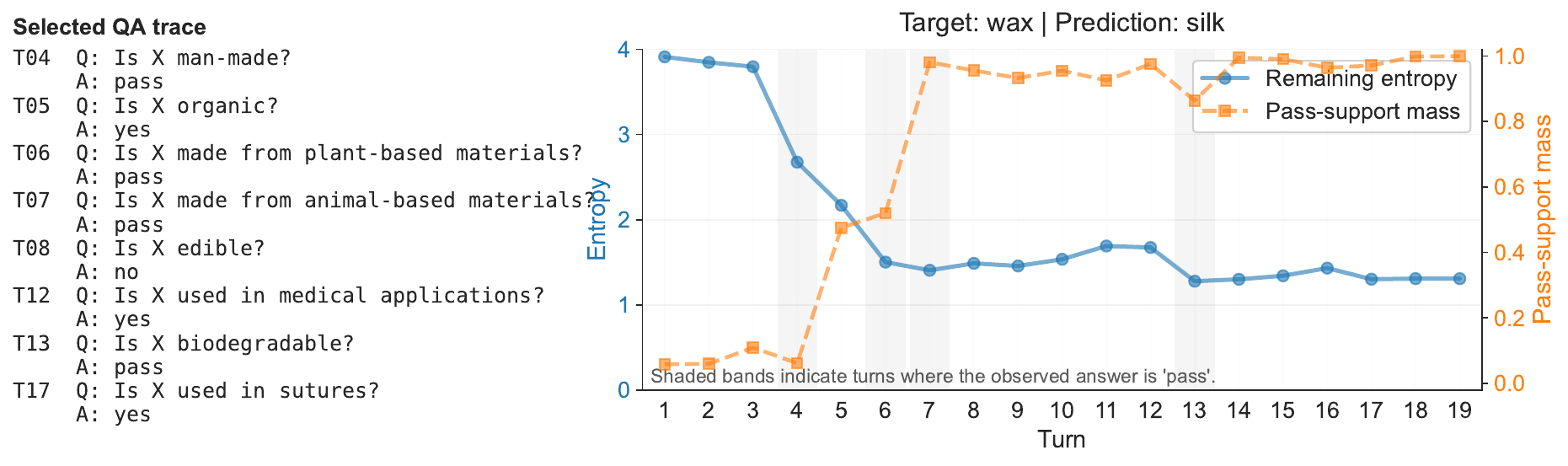}
        \caption{\textbf{Ambiguity accumulation.} Repeated \textit{pass} answers on material- and origin-related attributes destabilize the search process. Instead of clarifying the object’s core identity, the model continues refining along an ambiguous axis and eventually drifts toward an incorrect target.}
        \label{fig:20q_case2}
    \end{subfigure}

    \vspace{0.8em}

    \begin{subfigure}[t]{0.9\textwidth}
        \centering
        \includegraphics[width=\textwidth]{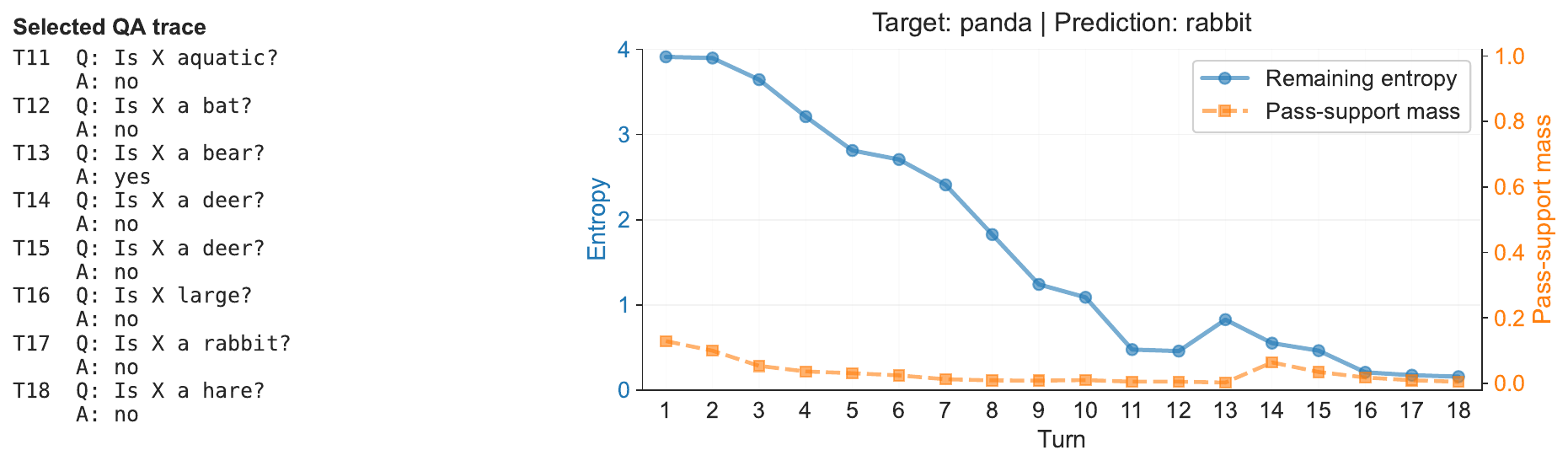}
        \caption{\textbf{Belief-state update failure.} Although the interaction provides useful evidence that the target belongs to a bear-like category, the model fails to consistently update and maintain its belief state. It revisits incompatible candidates and ends with an incorrect prediction.}
        \label{fig:20q_case3}
    \end{subfigure}

    \vspace{0.8em}

    \begin{subfigure}[t]{0.9\textwidth}
        \centering
        \includegraphics[width=\textwidth]{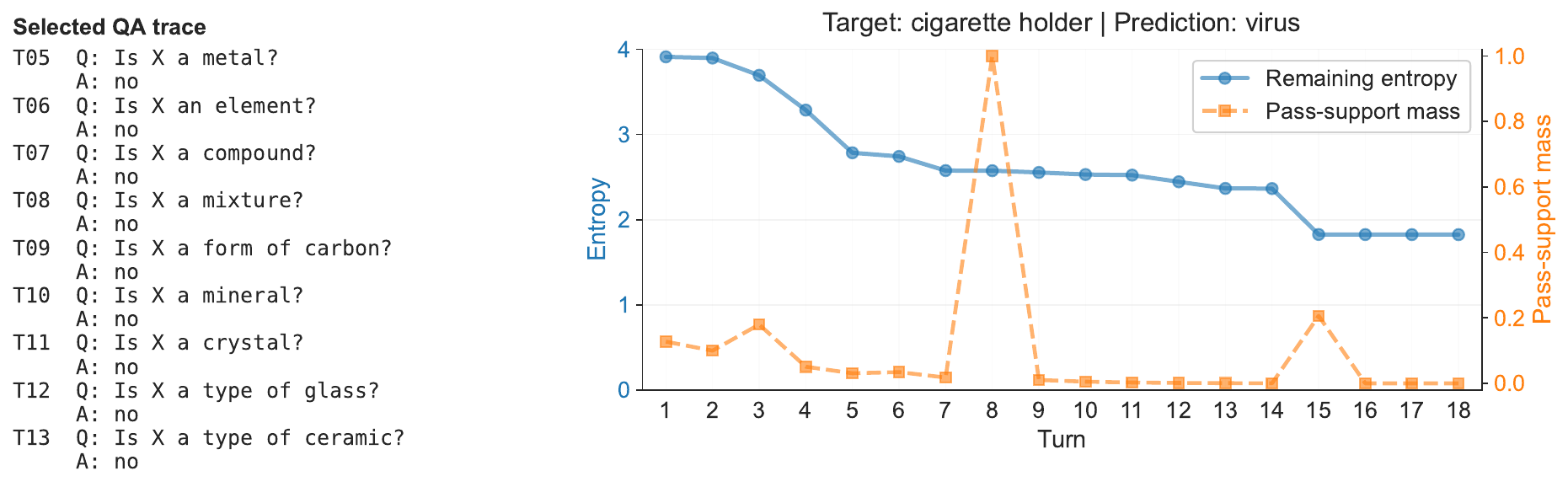}
        \caption{\textbf{Wrong abstraction level.} The target requires reasoning in a functional or object-level semantic space, but the model keeps searching through material and chemistry attributes. This misaligned abstraction level leads the interaction away from the target and produces a nonsensical final answer.}
        \label{fig:20q_case4}
    \end{subfigure}

    \caption{Representative failure modes in 20 questions. }
    \label{fig:case_studies_20q}
\end{figure*}

\paragraph{Case-study observations.}
The four examples in Figure~\ref{fig:case_studies_20q} illustrate several recurring failure patterns in 20 questions, including weak final disambiguation, ambiguity accumulation, inconsistent belief-state updates, and search at the wrong abstraction level. Despite these differences, they share a common structure: the model often makes some progress in narrowing the candidate space, but fails to reliably consolidate that progress into a correct final answer. This suggests that effective information seeking depends not only on asking high-quality questions in isolation, but also on maintaining a coherent search state, using ambiguous evidence appropriately, and refining the search at the right semantic granularity.

\end{document}